# ENGLISH-BHOJPURI SMT SYSTEM: INSIGHTS FROM THE KĀRAKA MODEL

*Thesis submitted to Jawaharlal Nehru University*
*in partial fulfillment of the requirements*
*for award of the*
*degree of*

## DOCTOR OF PHILOSOPHY

## ATUL KUMAR OJHA

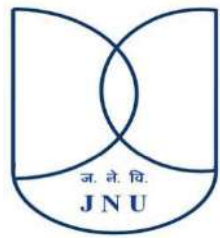

## SCHOOL OF SANSKRIT AND INDIC STUDIES

## JAWAHARLAL NEHRU UNIVERSITY,
## NEW DELHI-110067, INDIA

## 2018

संस्कृत एवं प्राच्यविद्या अध्ययन संस्थान

जवाहरलाल नेहरू विश्वविद्यालय

नई दिल्ली – ११००६७

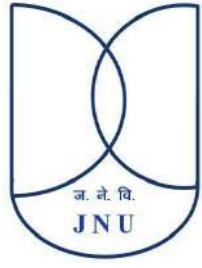

# SCHOOL OF SANSKRIT AND INDIC STUDIES
# JAWAHARLAL NEHRU UNIVERSITY
# NEW DELHI – 110067

January 3, 2019

# D E C L A R A T I O N

I declare that the thesis entitled "**English-Bhojpuri SMT System: Insights from the Kāraka Model**" submitted by me for the award of the degree of Doctor of Philosophy is an original research work and has not been previously submitted for any other degree or diploma in any other institution/university.

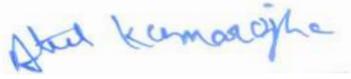

**(ATUL KUMAR OJHA)**

संस्कृत एवं प्राच्यविद्या अध्ययन संस्थान

जवाहरलाल नेहरू विश्वविद्यालय

नई दिल्ली – ११००६७

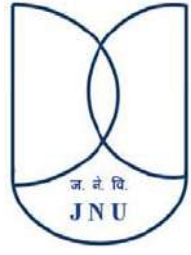

## SCHOOL OF SANSKRIT AND INDIC STUDIES
## JAWAHARLAL NEHRU UNIVERSITY
## NEW DELHI – 110067

January 3, 2019

# C E R T I F I C A T E

The thesis entitled **"English-Bhojpuri SMT System: Insights from the Kāraka Model"** submitted by **Atul Kumar Ojha** to **School of Sanskrit and Indic Studies, Jawaharlal Nehru University** for the award of degree of **Doctor of Philosophy** is an original research work and has not been submitted so far, in part or full, for any other degree or diploma in any University. This may be placed before the examiners for evaluation.

**Prof. Girish Nath Jha**　　　　　　　　　　　　**Prof. Girish Nath Jha**
　　**(Dean)**　　　　　　　　　　　　　　　　　　　**(Supervisor)**

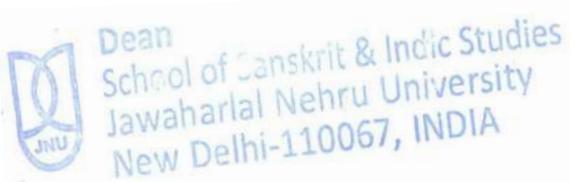

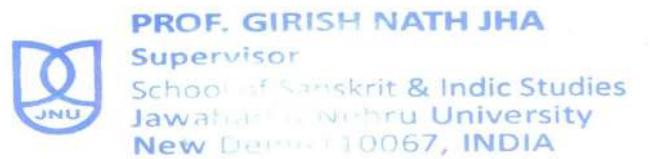

*To my grandfather*

*Late Shastri Shyam Awadh Ojha*

*&*

*To*

*My Parents*

*Sri S.N. Ojha and Smt Malti Ojha*

# Table of Contents

# List of Abbreviations

ABBREVIATIONS USED IN THE TEXT

| AdjP | Adjectival Phrase |
|---|---|
| AdvP | Adverbial Phrase |
| AGR | Agreement |
| AI | Artificial Intelligence |
| AngO | AnglaBharti Output |
| AnuO | Anusāraka Output |
| ASR | Automatic Speech Recognition |
| AV | Adjective+Verb |
| BLEU | Bilingual Evaluation Understudy |
| BO-2014 | Bing Output-2014 |
| BO-2018 | Bing Output-2018 |
| C-DAC | Centre for Development of Advanced Computing |
| CFG | Context-Free Grammar |
| CLCS | Composition of the LCS |
| CMU, USA | Carnegie Mellon University, USA |
| Corpus-based MT | Corpus-based Machine Translation |
| CP | Complementizer Phrase |
| CPG | Computational Pāṇinian Grammar |
| CRF | Conditional Random Field |
| CSR | Canonical Syntactic Realization |
| Dep-Tree-to-Str SMT | Dependency Tree-to-String Statistical Machine Translation |
| DLT | Disambiguation Language Techniques |
| DLT | Distributed Language Translation |
| D-Structure | Deep Structure |
| EBMT | Example-based Machine Translation |
| EBMT | Example-Based MT |
| EB-SMT | English-Bhojpuri Statistical Machine Translation |
| EB-SMT System-1 | PD based Dep-Tree-to-String SMT |
| EB-SMT System-2 | UD based Dep-Tree-to-String SMT |
| ECM | Exception Case Marking |
| ECP | Empty Category Principle |
| ECV | Explicator Compound Verb |
| E-ILMTS | English-Indian Language Machine Translation System |
| E-ILs | English-Indian Languages |
| EM | Expectation Maximization |
| EST | English to Sanskrit Machine Translation |
| EXERGE | Expansive Rich Generation for English |
| FBSMT | Factor-based Statistical Machine Translation |
| FT | Functional Tags |
| GATE | General Architecture for Text Engineering |
| GB | Government and Binding |
| GHMT | Generation Heavy Hybrid Machine Translation |
| GLR | Generalized Linking Routine |
| GNP | Gender, Number, Person |
| GNPH | Gender, Number, Person and Honorificity |
| GO-2014 | Google Output-2014 |
| GO-2018 | Google Output-2018 |
| GTM | General Text Matcher |
| HEBMT | Hybrid Example-Based MT |
| Hierarchical phrase-based | No linguistic syntax |
| HMM | Hidden Markov Model |
| HPBSMT | Hierarchal Phrase-based Statistical Machine Translation |
| HPSG | Head-Driven Phrase Structure Grammar |
| HRM | Hierarchical Re-ordering Model |
| HWR | Handwriting Recognition |
| Hybrid-based MT | Hybrid-based Machine Translation |
| IBM | International Business Machine |
| IHMM | Indirect Hidden Markov Model |
| IIIT-H/Hyderabad | International Institute of Information Technology, Hyderabad |
| IISC-Bangalore | Indian Institute of Science, Bangalore |
| IIT-B/Bombay | Indian Institute of Technology, Bombay |
| IIT-K/Kanpur | Indian Institute of Technology, Kanpur |
| ILCI | Indian Languages Corpora Initiative |
| IL-Crawler | Indian Languages Crawler |
| IL-IL | Indian Language-Indian Language |
| ILMT | Indian Language to Indian Language Machine Translation |
| ILs-E | Indian Languages-English |
| ILs-ILs | Indian Languages-Indian Languages |
| IMPERF | Imperfective |
| IR | Information Retrieval |
| IS | Input Sentence |
| ITrans | Indian language Transliteration |
| JNU | Jawaharlal Nehru University |
| KBMT | Knowledge-based MT |
| KN | Kneser-Ney |
| LDC | Linguistic Data Consortium |
| LDC-IL | Linguistic Data Consortium of Indian Languages |
| LFG | Lexical Functional Grammar |
| LGPL | Lesser General Public License |
| LLR | Log-Likelihood-Ratio |
| LM | Language Model |
| LP | Link Probability |
| LRMs | Lexicalized Re-ordering Models |
| LSR | Lexical Semantic Representation |
| LT | Language Technology |
| LTAG | Lexicalized Tree Adjoining Grammar |
| LTRC | Language Technology Research Centre |

| LWG | Local Word Grouping |
|---|---|
| ManO | Mantra Output |
| MatO | Matra Output |
| MERT | Minimum Error Rate Training |
| METEOR | Metric for Evaluation of Translation with Explicit Ordering |
| MLE | Maximum Likelihood Estimate |
| MT | Machine Translation |
| MTS | Machine Translation Systems |
| NE | Named Entity |
| NER | Named-entity Recognition |
| NIST | National Institute of Standard and Technology |
| NLP | Natural Language Processing |
| NLU | Natural Language Understanding |
| NMT | Neural Machine Translation |
| NN | Common Noun |
| NP | Noun Phrase |
| NPIs | Negative Polarity Items |
| NPs | Noun Phrases |
| NV | Noun+Verb |
| OCR | Optical Character Recognition |
| OOC | Out of Character |
| OOV | Out of Vocabulary |
| P&P | Principle & Parameter |
| PBSMT | Phrase-based Statistical Machine Translation |
| PD | Pāṇinian Dependency |
| PD-EB-SMT | UD based Dep-Tree-to-String SMT |
| PER | Position-independent word Error Rate |
| PERF | Perfective |
| PG | Pāṇinian Grammar |
| PLIL | Pseudo Lingua for Indian Languages |
| PNG | Person Number Gender |
| POS | Part-Of-Speech |
| PP | Prepositional Phrase |
| PP | Postpositional/Prepositional Phrase |
| PROG | Progressive |
| PSG | Phrase-Structure Grammars |
| RBMT | Rule-based Machine Translation |
| RBMT | Rule-based MT |
| RLCS | Root LCS |
| RLs | Relation Labels |
| Rule-based MT | Rule-based Machine Translation |
| SBMT | Statistical Based Machine Translation |
| SBSMT | Syntax-based Statistical Machine Translation |
| SCFG | Synchronous Context Free Grammar |
| SD | Stanford Dependency |
| SGF | Synchronous Grammar Formalisms |
| SL | Source Language |
| SMT | Statistical Machine Translation |
| SOV | Subject-Object-Verb |
| SOV | Subject Object Verb |
| SPSG | Synchronous Phrase-Structure Grammars |
| SR | Speech Recognition |
| SSF | Shakti Standard Format |
| S-structure | Surface structure |
| String-to-Tree | Linguistic syntax only in output (target) language |
| STSG | Synchronous Tree-Substitution Grammars |
| SVM | Support Vector Machine |
| SVO | Subject-Verb-Object |
| TAG | Tree-Adjoining Grammar |
| TAM | Tense Aspect & Mood |
| TDIL | Technology Development for Indian Languages |
| TG | Transfer Grammar |
| TL | Target Language |
| TM | Translation Model |
| TMs | Translation Models |
| Tree-to-String | Linguistic syntax only in input/source language |
| Tree-to-Tree | Linguistic syntax only in both (source and traget) language |
| TTS | Text-To-Speech |
| UD | Universal Dependency |
| UD-EB-SMT | PD based Dep-Tree-to-String SMT |
| ULTRA | Universal Language Translator |
| UNITRAN | Universal Translator |
| UNL | Universal Networking Language |
| UNU | United Nations University |
| UOH | University of Hyderabad |
| UPOS | Universal Part-of-Speech Tags |
| UWs | Universal Words |
| VP | Verb Phrase |
| WER | Word Error Rate |
| WMT | Workshop on Machine Translation |
| WSD | Word Sense Disambiguation |
| WWW | World Wide Web |
| XML | Extensible Markup Language |
| XPOS | Language-Specific Part-of-Speech Tag |

## ABBREVIATIONS USED IN GLOSSING OF THE EXAMPLE SENTENCES

| 1      | First person          |
|--------|-----------------------|
| 2      | Second person         |
| 3      | Third person          |
| M      | Masculine             |
| F      | Feminine              |
| S      | Singular              |
| P/pl   | Plural                |
| acc    | Accusative            |
| adj/JJ | Adjective             |
| adv/RB | Adverb                |
| caus   | Causative             |
| CP     | Conjunctive Participle|
| emph   | Emphatic              |
| fut    | Future tense          |
| gen    | Genitive              |
| impf   | Imperfective          |
| inf    | Infinitive            |
| ins    | Instrumental          |
| PR     | Present tense         |
| PRT    | Particle              |
| PST    | Past Tense            |

# List of Tables

# List of Figures

# ACKNOWLEDGEMENTS

*This thesis is a fruit of love and labour possible with the contributions made by many people, directly and indirectly. I would like to express my gratitude to all of them.*

*I would first like to thank my thesis advisor Prof. Girish Nath Jha of the School of Sanskrit and Indic Studies (SSIS) at Jawaharlal Nehru University, Delhi. The door to Prof. Jha's office was always open whenever I felt a need for academic advice and insight. He allowed this research work to be my own work but steered me in the right direction as and when needed. Frankly speaking, it was neither possible to start or to finish without him. Once again, I thank him for valuable remarks and feedback which helped me to organize the contents of this thesis in a methodical manner.*

*I wish to extend my thanks to all the faculty members of SSIS, JNU for their supports. I also would like to thanks Prof. K.K. Bhardwaj of the School of Computer and System Sciences, JNU for teaching me Machine Learning during this PhD coursework.*

*Next, I extend my sincere thanks to Prof Martin Volk of the University of Zürich who taught me Statistical and Neural Machine Translation in a fantastic way; Martin Popel of UFAL, Prague and Prof. Bogdan Baych of the University of Leeds for sharing different MT evaluation methodologies with me that tremendously enhanced the quality of my research.*

*There was an input of tremendous efforts while writing the thesis as well. My writing process would have been less lucid and even less presentable if I had not received support from my friends, seniors and colleagues. Biggest thanks must go to Akanksha Bansal and Deepak Alok for their immeasurable support. They read my manuscript to provide valuable feedback. Their last -minute efforts made this thesis presentable. I admit that their contributions need much more acknowledgement than expressed here.*

*I am extremely thankful to Mayank Jain and Rajeev Gupta for proof-reading my draft. Special thanks to Mayank Jain for being by my side for the entire writing process that kept me strong, sane and motivated. In addition, I am also thankful to Pinkey Nainwani and Esha Banerjee for proofreading and their constant support.*

*A special thanks to Prateek, Atul Mishra, Rajneesh Kumar and Rahul Tiwari for their support. Prateek and Atul have contributed in the creation of parallel corpora while Rajneesh and Rahul helped me for evaluation the developed SMT system. I cannot forget to thank the efforts put in by Saif Ur Rahman who helped me crawl the current Google and Bing MT output, thus, enriching*

*the process of evaluation of the existing Indian MT systems.*

*I also acknowledge the office staff of the SSIS Shabanam, Manju, Arun and Vikas Ji, for their cooperation and assistance on various occasions. Their prompt responses and willingness made all the administrative work a seamless process for me.*

*A heartfelt thanks is also due to all of my friends and juniors, particularly Ritesh Kumar, Sriniket Mishra, Arushi Uniyal, Devendra Singh Rajput, Abhijit Dixit, Bharat Bhandari, Bornini Lahiri, Abhishek Kumar, Ranjeet Mandal, Shiv Kaushik, Devendra Kumar and Priya Rani.*

*I would like to thank Hieu Hoang of University Edinburgh and MOSES support group members for solving the issues with SMT training.*

*I would like to thank all the ILCI Project Principal Investigators for their support to manage the project smoothly while I was completely immersed in my experiments.*

*My final thanks and regards go to all my family members, who motivated me to enhance myself academically and helped me reach the heights I've reached today. They are the anchor to my ship.*

# Chapter 1

# Introduction

*"When I look at an article in Russian, I say: 'This is really written in English, but it has been coded in some strange symbols. I will now proceed to decode'."*

*(Warren Weaver, 1947)*

**1.1 Motivation**

In the last two decades, the Statistical Machine Translation (SMT) (Brown et al., 1993) method has garnered a lot of attention as compared to the Rule-based Machine Translation (RBMT) and Interlingua-based MT or Example-based MT (EBMT) (Nago, 1984) in the field of Machine Translation (MT), especially after the availability of Moses (details provided in chapter 4) open source toolkit (Koehn et al., 2007). However, it is also imperative to note that the neural model for resolving machine related tasks has also gained a lot of momentum during the recent past after it was proposed by Kalchbrenner and Blunsom (2013), Sutskever et al (2014) and Cho et al. (2014). The neural machine translation method is different from the traditional phrase-based statistical machine translation system (see below section or follow Koehn et al., 2003 article). The latter consists of many small sub-components that are individually trained and tuned whereas the neural machine translation method attempts to build and formulate a large neural network and fine tunes it as a whole. This means the single, large neural network reads sentences and offers correct translation as an output. Presently, there are many NMT open source toolkits that can be accessed by translators such as OpenNMT (Klein et al., 2017), Neural Monkey (Helcl et al., 2017), Nematus (Sennrich et al., 2017) etc. Although, there seem to be many advantages to the NMT method, there are also challenges as it continues to underperform when it comes to low-resource languages such as the Indian languages. The SMT, on the other hand, can produce better results in English languages (Ojha et al., 2018) even on small corpus whereas, the NMT cannot. Due to its vast and complex neural network, the NMT requires a longer time to be tuned or trained. Moreover, training the NMT also depends on the system configuration. For instance, if the NMT system is trained on a GPU-based system or a cluster machine then the time taken is less than CPU which can take up to more time (may be three weeks to a month).

There have been remarkable improvements in the field of MT (machine translations) and high-quality MT can be obtained for rich-resourced language pairs such as English-German, German-English, French-English, Spanish-English. This is because these language pairs are known to have overlaps in linguistic properties, structure, and phenomena including vocabulary, cognate, and grammar. Nevertheless, these MT systems are still not near perfection and usually offer unsatisfactory outputs when it comes to English-Indian languages. This is because English-Indian Languages (E-ILs)/ Indian-English (ILs-E) or Indian-Indian languages (ILs-ILs) consist of complicated structures such as the free-word order, morphological richness, and belongs to different language families, etc. According to Ojha et al. (2014) in the SMT, most of incorrect translations occur mainly due to the following reasons: morph analysis, tokenization, and grammatical differences (including word order, agreement etc.).

During the course of my PhD research, I collected a sample of English-Hindi and Hindi-English MT translations and have presented them below. (These Hindi-English MT systems translations were taken from Ojha et al. (2014) article). These examples also show a progressive trend in the quality of Google and Bing MT's translations from 2014-2018[1].

- **English-Hindi MT**

(a) Where did you hide the can opener? (IS[2])

आपने डिब्बा ओपनर को कहाँ छिपाया (AngO)

आपने कैन खोलनेवाला कहाँ छिपाया? (AnuO)

---

[1] The output of other Indian MT systems (AnglaBharati, Anusāraka, Mantra, Matra) are not given due to their unavailability and also because they do not support Hindi-English translation.

[2] Is= Input sentence, AngO= AnglaBharti output, AnuO= Anusāraka output, ManO= Mantra output, MatO= Matra output, Go-2014= Google output in 2014, GO-2018 = Google output in 2018, BO-2014= Bing output in 2014, BO-2018 = Bing output in 2018

| | |
|---|---|
| जहाँ किया हुआ आप प्रारम्भ करने वाला छुपाते हैं | (ManO) |
| आप कैन खोलने वाला छिपाते हो | (MatO) |
| तुम खोल कर सकते हैं कहाँ छिपा था? | (GO-2014) |
| आपने सलामी बल्लेबाज को कहां छिपाया? | (GO-2018) |
| जहाँ आप सलामी बल्लेबाज कर सकते छिपा था ? | (BO-2014) |
| आप कर सकते है सलामी बल्लेबाज कहां छिपा हुआ? | (BO-2018) |

*Manual Translation*: तुमने ढक्कन खोलने वाला कहाँ छिपाया?

- **Hindi-English MT**

(b) एच.आई.वी. क्या है ?     (IS)

| | |
|---|---|
| HIV what is it? | (GO-2014) |
| HIV. What is it? | (GO-2018) |
| What is the HIV? | (BO-2014) |
| What is HIV? | (BO-2018) |

*Manual Translation*: What is the HIV?

(c) वह जाती है ।     (IS)

| | |
|---|---|
| He is. | (GO-2014) |
| He goes. | (GO-2018) |
| She goes. | (BO-2014) |
| He goes. | (BO-2018) |

*Manual Translation:* She goes.

(d) छुआरे डालकर मिलाएँ और एक मिनिट पकाएँ।                  ( IS)

Mix and cook one minute, add Cuare                           (GO-2014)

Add the spices and cook for a minute.                        (GO-2018)

One minute into the match and put chuare                     (BO-2014)

Add the Chuare and cook for a minute.                        (BO-2018)

*Manual Translation*: Put date palm, stir and cook for a minute.

The most common issues found in the above mentioned examples when analyzed were related to word-order, morph issue, gender agreement, incorrect word, etc. Consequently, the most important task at hand is to work on improving the accuracy of the already in-place and developed MT systems and to further develop MT systems for the languages that have not yet been accessed or explored using the statistical method. Improving upon an MT system poses a huge challenge because of many limitations and restrictions. So, now the question arises as how can we improve the accuracy and fluency of the available MTs?

Dependency structures, which can be utilized to tackle the afore-mentioned problems, represent a sentence as a set of dependency relations applying the principles of dependency grammar[3]. Under ordinary circumstances, dependency relations create a tree structure to connect all the words in a sentence. Dependency structures have found to have their use in several theories dwelling on semantic structures, for example, in theories dwelling on semantic relations/cases/theta roles (where arguments have defined semantic relations to the head/predicate) or in the predicate (arguments depend on the predicate). A salient feature of dependency structures is their ability to represent long distance dependency between words with local structures.

A dependency-based approach to solving the problem of word and phrase reordering weakens the requirement for long distance relations which become local in dependency tree structures. This particular property is attractive when machine translation is supposed to engage with languages with diverse word orders, such as diversity between subject-verb-object (SVO) and subject-object-verb (SOV) languages; long distance reordering becomes one of the principle features. Dependency structures target lexical items directly, which turn out to be simpler in form when compared with phrase-structure trees since constituent labels are missing. Dependencies are typically meaningful - i.e. they usually carry semantic relations and are more abstract than their surface order. Moreover, dependency relations between words model the semantic structure of a sentence directly. As such, dependency trees are desirable prior models for the process of preserving semantic structures from source to target language through translation. Dependency structures have been known to be a promising direction for several components of SMT (Ma et al., 2008; Shen et al., 2010; Mi and Liu, 2010; Venkatpathy, 2010, Bach, 2012) such as word alignment, translation models and language models.

---

[3] a type of grammar formalism

Therefore in this work, I had proposed research on English-Indian language SMT with special reference to English-Bhojpuri language pair using the Kāraka model based dependency (known as Pāṇinian Dependency) parsing. The Pāṇinian Dependency is more suitable for Indian languages to parse at syntactico-semantic levels as compared to other models like phrase structure and Government of Binding theory (GB) (Kiparsky et al., 1969). Many researchers have also reported that the Pāṇinian Dependency (PD) is helpful for MT system and NLP applications (Bharati et al., 1995).

**1.2 Methodology**

For the present research, firstly Bhojpuri corpus was created both monolingual (Bhojpuri) and parallel (English-Bhojpuri). After the corpus creation, the corpora were annotated at the POS level (for both SL and TL) and at the dependency level (only SL). For the dependency annotation both PD and UD frameworks were used. Then, the MT system was trained using the statistical methods. Finally, evaluation methods (automatic and human) were followed to evaluate the developed EB-SMT systems. Furthermore, the comparative study of PD and UD based EB-SMT systems was also conducted. These processes have been briefly described below:

- **Corpus Creation**: There is a big challenge to collect data for the corpus. For this research work, 65,000 English-Bhojpuri parallel sentences and 100000 sentences for monolingual corpora have been created.

- **Annotation**: After corpus collection, these corpora have been annotated and validated. In this process, Karaka and Universal models have been used for dependency parsing annotation.

- **System Development**: The Moses toolkit has been used to train the EB-SMT systems.

- **Evaluation**: After training, the EB-SMT systems have been evaluated. The problems of EB-SMT systems have been listed in the research.

## 1.3 Thesis Contribution

There are five main contributions of this thesis:

- The thesis studies the available English-Indian Language Machine Translation System (E-ILMTS) (given in the below section).

- It presents a feasibility study of Kāraka model for using the SMT between English-Indian languages with special reference to the English-Bhojpuri pair (see chapter 2 and 4).

- Creation of LT resources for Bhojpuri (see chapter 3).

- An R&D method has been initiated towards developing an English-Bhojpuri SMT (EB-SMT) system using the Kāraka and the Universal dependency model for dependency based tree-to-string SMT model (see chapter 4).

- A documentation of the problems has been secured that enlists the challenges faced during the EB-SMT system and another list of current and future challenges for E-ILMTS with reference of the English-Bhojpuri pair has been curated. (see chapter 5 and 6).

## 1.4 Bhojpuri Language: An Overview

Bhojpuri is an Eastern Indo-Aryan language, spoken by approximately 5,05,79,447 (Census of India Report, 2011) people, primarily in northern India which consist of the Purvanchal region of Uttar Pradesh, western part of Bihar, and north-western part of Jharkhand. It also has significant diaspora outside India, e.g. in Mauritius, Nepal, Guyana, Suriname, and Fiji. Verma (2003) recognises four distinct varieties of Bhojpuri spoken in India (shown in Figure 1.1, it has adopted from Verma, 2003):

1. <u>Standard Bhojpuri (also referred to as Southern Standard)</u>: spoken in, Rohtas, Saran, and some part of Champaran in Bihar, and Ballia and eastern Ghazipur in Uttar Pradesh (UP).

2. <u>Northern Bhojpuri</u>: spoken in Deoria, Gorakhpur, and Basti in Uttar Pradesh, and some parts of Champaran in Bihar.

3. <u>Western Bhojpuri</u>: spoken in the following areas of UP: Azamgarh, Ghazipur, Mirzapur and Varanasi.

4. <u>Nagpuria</u>: spoken in the south of the river Son, in the Palamu and Ranchi districts in Bihar.

Verma (2003) mentions there could be a fifth variety namely 'Thāru' Bhojpuri which is spoken in the Nepal Terai and the adjoining areas in the upper strips of Uttar Pradesh and Bihar, from Baharaich to Champaran.

Bhojpuri is an inflecting language and is almost suffixing. Nouns are inflected for case, number, gender and person while verbs can be inflected for mood, aspect, tense and phi-agreement. Some adjectives are also inflected for phi-agreement. Unlike Hindi but like other Eastern Indo-Aryan languages, Bhojpuri uses numeral classifiers such as *Tho, go, The, kho* etc. which vary depending on the dialect.

Syntactically, Bhojpuri is SOV with quite free word-order, and generally head final, wh-in-situ language. It allows pro drop of all arguments and shows person, number and gender agreement in the verbal domain. An interesting feature of the language is that it also marks honorificity of the subject in the verbal domain. Unlike Hindi, Bhojpuri has nominative-accusative case system with differential object marking. Nominative can be considered unmarked case in Bhojpuri while other cases (in total six or seven) are marked through postpositions. Unlike Hindi, Bhojpuri does not have oblique case. However, like Hindi, Bhojpuri has series of complex verb constructions such as conjunct verb constructions and serial verb constructions.

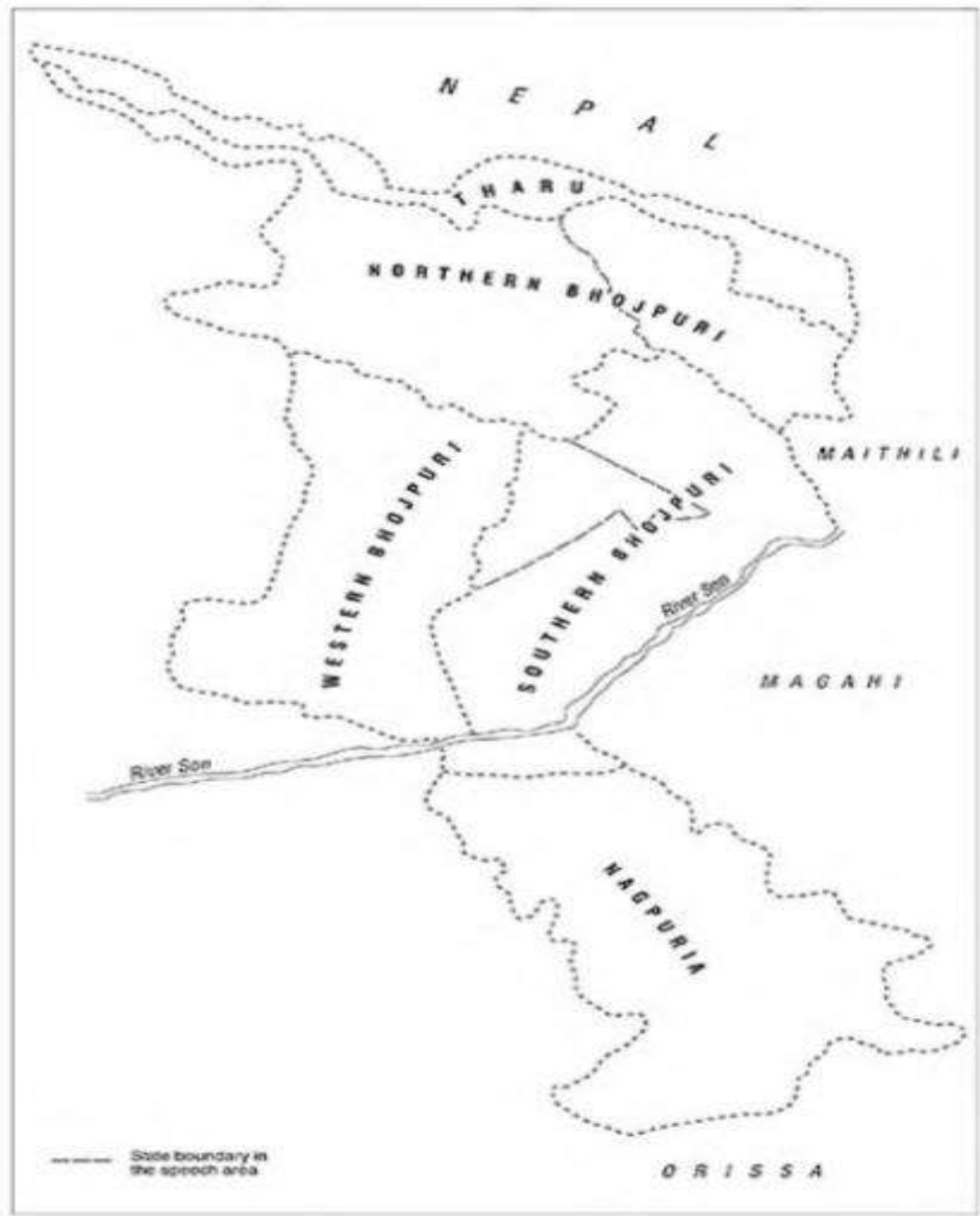

Figure 1. 1: Areas showing Different Bhojpuri Varieties

On the other hand, Hindi and English are very popular languages. Hindi is one of the scheduled languages of India. While English is spoken worldwide and now considered as an international language; Hindi and English are official languages of India.

## 1.5 Machine Translation (MT)

A MT is an application that translates the source language (SL) into target-language (TL) with the help of a computer. MT is one of the most important NLP applications. Previously, the MTs were only used for text translations but currently, they are also employed for image-to-text translation as well as speech-to-speech translations. A machine translation system usually operates on three broad types of approaches: rule-based, corpus-based, and hybrid-based. The author has explained these approaches very briefly, below. Details of each approach can also be found in the following sources – Hutchins and Somers, 1992, 'An Introduction to Machine Translation' and Bhattacharyya, 2015 'Machine Translation'; Poibeau, 2017 'Machine Translation').

- **Rule-based MT:** Rule-based MT techniques are linguistically oriented as the method requires dictionary and grammar to understand syntactic, semantic, and morphological aspects of both languages. The primary objective of this approach is to derive the shortest path from one language to another, using rules of grammar. The RBMT approach is further classified into three categories: (a) Direct MT, (b) Transfer based MT, and (c) Interlingua based MT. All these categories require an intensive and in-depth knowledge of linguistics and the method becomes progressively complex to employ when one moves from one category to the other.

- **Corpus-based MT:** This method uses and relies on previous translations collected over time to propose translations of new sentences using the statistical/neural model. This method is divided into three main subgroups: EBMT, SMT, and NMT. In these subgroups, EBMT (Nagao, 1984; Carl and Way, 2003) presents translation by analogy. A parallel corpus is required for this, but instead of assigning probabilities to words, it tries to learn by example or using previous data as sample.

- **Hybrid-based MT:** As the name suggests, this method employs techniques of both rule-based and statistical/corpus based methods to devise a more accurate translation technique. First the rule-based MT is used to present a translation and then statistical method is used to offer correct translation.

## 1.6 An Overview of SMT

The SMT system is a probabilistic model automatically inducing data from corpora. The core SMT methods (Brown et al., 1990, 1993; Koehn et al., 2003) - emerged in the 1990s and matured in the 2000s to become a commonly used method. The SMT learnt direct correspondence between surface forms in the two languages, without requiring abstract linguistic representations. The main advantages of SMT are versatility and cost-effectiveness: in principle, the same modeling framework can be applied to any pair of language with minimal human effort or over the top technical modifications. The SMT has three basic components: translation model, language model, and decoder (shown in Figure 1.2). Classic SMT systems implement the noisy channel model (Guthmann, 2006): given a sentence in the source language 'e' (denotes to English), we try to choose the translation in language 'b' (denotes to Bhojpuri) that maximizes $(b|e)$. According to Bayes rule (Koehn, 2010), this can be rewritten as:

$$\text{argmax}_e \, p(b|e) = \text{argmax}_e \, p(e|b)p(b) \qquad (1.1)$$

$p(b)$ is materialized with a language model – typically, a smoothed n-gram language model in the target language – and $p(e|b)$ with a translation model – a model induced from the parallel corpora - aligned documents which are, basically, the translation of each other. There are several different methods that have been used to implement the translation model, and other models such as fertility and reordering models have also been employed, since the first translation schemes proposed by the IBM Models[4] were used 1 through 5 in the late 1980s (Brown et al, 1993). Finally, it comes down to the decoder that is an algorithm which calculates and selects the most probable and appropriate translation out of several possibilities derived from the models at hand.

The paradigms of SMT have emerged from word-based translations (Brown et al., 1993) and also from phrase-based translations (Zens et al., 2002; Koehn et al., 2003; Koehn, 2004). , Hierarchical Phrase-based translation (Chiang, 2005; Li et al, 2012), Factor-based translation (Koehn et al., 2007; Axelrod, 2006; Hoang, 2011), and Syntax-based translation (Yamada and Knight, 2001; Galley et al., 2004; Quirk et al., 2005; Liu et al.,

---

[4] See chapter 4 for detailed information

2006; Zollmann and Venugopal, 2006; Williams et al., 2014; Williams et al., 2016). All these have been explained briefly in the sub-sections below.

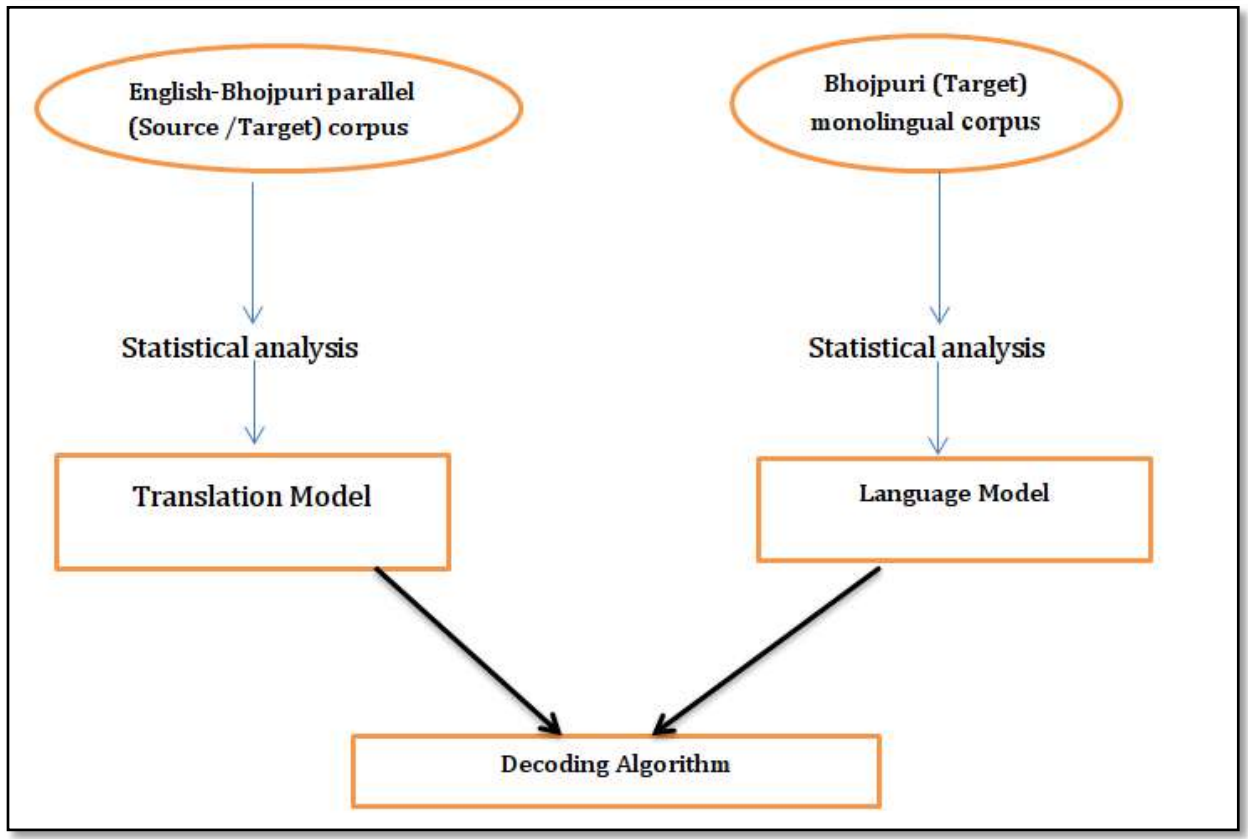

Figure 1. 2: Architecture of SMT System

**1.6.1 Phrase-based SMT (PBSMT)**

Word-based translation (Brown et al., 1993) models are based on the independent assumption that translation probabilities can be estimated at a word-level while ignoring the context that word occurs in. This assumption usually falters when it comes to natural languages. The translation of a word may depend on its context for morpho-syntactic reasons (e.g. agreement within noun phrases), or because it is part of an idiomatic expression that cannot be translated literally or compositionally in another language which may not bear the same structures. Also, some (but not all) translation ambiguities can be disambiguated in a larger context.

Phrase-based SMT (PBSMT) is driven by a phrase-based translation model, which connects, relates, and picks phrases (contiguous segments of words) in the source to match with those in the target language. (Och and Ney, 2004). A generative tale of PBSMT systems goes on in the following manner:

- source sentence is segmented into phrases

- each phrase-based unit represented on phrase tables is translated

- translated phrases are permuted in their final order

Koehn et al. (2003) examines various methods by which phrase translation pairs can be extracted from any parallel corpus in order to offer phrase translation probabilities and other features that match the target language accurately. Phrase pair extraction is based on the symmetrical results of the IBM word alignment algorithms (Brown et al., 1993). After that all phrase pairs, consistent with word alignment (Och et al., 1999), are extracted that only intravenous word alignment has taken place which means that words from the source phrase and target phrase are aligned with each other only and not with any words outside each other's domain. Relative frequency is used to arrive at an estimate about the phrase translation probabilities$(e|b)$.

While using the phrase-based models, one has to be mindful of the fact that a sequence of words can be treated as a single translation unit by the MT system. And, increasing the length of the unit may not yield accurate translation as the longer phrase units will be limited due to data scarcity. Long phrases are not as frequent and many are specific to the module developed during training. Such low frequencies bear no result and the relative frequencies result in unreliable probability estimates. Thus, Koehn et al. (2003) proposes that lexical weights may be added to phrase pairs as an extra feature. These lexical weights are obtained from the IBM word alignment probabilities. They are preferred over directly estimated probabilities as they are less prone to data sparseness. Foster et al. (2006) introduced more smoothing methods for phrase tables (sample are shown in the chapter 4), all aimed at penalizing probability distributions that are unfit for translation and overqualified for the training data because of data sparseness. The search in phrase-based machine translation is done using heuristic scoring functions based on beam search.

A beam search phrase-based decoder (Vogel, 2003; Koehn et al., 2007) employs a process that consists of two stages. The first stage builds a translation lattice based on its existing corpus. The second stage searches for the best path available through the lattice.

This translation lattice is created by obtaining all available translation pairs from the translation models for a given source sentence, which are then inserted into a lattice to deduce a suitable output. These translation pairs include words/ phrases from the source sentence. The decoder inserts an extra edge for every phrase pair and pastes the target sentence and translation scores to this edge. The translation lattice then holds a large number of probable paths covering each source word exactly once (a combination of partial translations of words or phrases). These translation hypotheses will greatly vary in quality and the decoder will make use of various knowledge sources and scores to find the best possible path to arrive at a translation hypothesis. It is that this step that one can also perform limited reordering within the found translation hypotheses. To supervise the search process, each state in the translation lattice is associated with two costs, current and future translation costs. The future cost is an assessment made for translation of the remaining words in any source sentence. The current cost refers to the total cost of those phrases that have been translated in the current partial hypothesis that is the sum of features' costs.

### 1.6.2 Factor-based SMT (FBSMT)

The idea of factored translation models was proposed by Koehn & Hoang (Koehn and Hoang, 2007). In this methodology, the basic unit of language is a vector, annotated with multiple levels of information for the words of the phrase, such as lemma, morphology, part-of-speech (POS) etc. This information extends to the generation step too, i.e. this system also allows translation without any surface form information, making use of instead the abstract levels which are first translated and the surface form is then generated from these using only the target-side operations (shown in Figure 1.3, taken from Koehn, 2010). Thus, it is preferable to model translation between morphologically rich languages at the level of lemmas, and thus collect the proof for different word forms that derive from a common lemma.

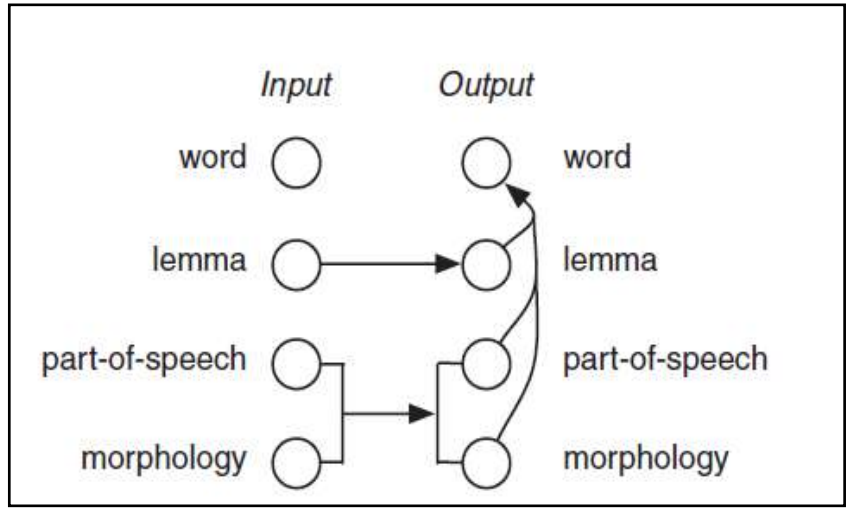

Figure 1. 3: Workflow of Decomposition of Factored Translation Model

Koehn & Hoang's experiments indicate that translation improves in the event of data sparseness. They also exhibit an effect that wears off while moving towards larger amounts of training data. It is this approach implemented in the open source decoder, Moses (Koehn et al., 2007).

### 1.6.3   Hierarchal Phrase-based SMT (HPBSMT)

Hierarchical phrase-based models (Chiang, 2005) come across as a better method to design discontinuous phrase pairs and re-orderings in a translation model than crafting a separate distortion model. The model permits hierarchical phrases that consist of words (terminals) and sub-phrases (non-terminals), in this case English to Bhojpuri. For example:

$$X \rightarrow \langle \text{ is } X_1 \text{ going}, \text{जात हऽ} X_1 \rangle$$

This makes the model a weighted synchronous context-free grammar (CFG), and CYK parsing helps perform decoding. The model does not require any linguistically-motivated set of rules. In fact, the hierarchical phrases are learned using the technique of similar phrase extraction heuristics similar to the process in phrase-based models. However, the formalism can be applied to rules learned through a syntactic parser. Chiang (2010) provides a summary of all the approaches that utilize syntactic information either on the side of the source, the target, or both.

Hierarchical models perform better than phrase-based models in a few settings but not so much in others. Birch et al. (2009) compared the performances of phrase-based with hierarchical models, only to conclude that their respective performance is dependent on the kind of re-orderings necessary for the language pair.

Except phrase-based models, hierarchical models are the only kind of translation models which the author uses in this work with experiment details discussed in chapter 4. While phrase-based, hierarchical and syntax-based models employ different types of translation units, model estimation is mathematically similar.

### 1.6.4 Syntax-based SMT (SBSMT)

Modeling syntactic information in machine translation systems is not a novelty. A syntactic translation framework was proposed by Yngve (1958) who understood the act of translation as a 3-stage process, namely: Analysis of source sentence in the form of phrase structure representations; Transferring the phrase structure representation into equivalent target phrase structures; Application of target grammar rules with the objective generating output translation

While the models mentioned above make use of structures beyond mere word-pairs, namely phrases and hierarchical rules, they do not require linguistic syntax. Syntax-based translation models date to Yamada and Knight (2001, 2002), who designed a model and a decoder for translating a source-language string into a target-language string along with its phrase structure parse. The research community added significant improvements to syntax-based statistical machine translation (SBSMT) systems in recent years. The breakthrough point came when the combination of syntax with statistics was made possible along with the availability of a large-sized training data, and synchronous grammar formalisms.

Phrase-structure grammar is credited to extend its fundamental tenets to furnish Synchronous grammar formalisms. Phrase-structure rules, for instance, NP → DET JJ NN, are the principle features of phrase-structure grammar. These rules were a product of the observation that words complement the increasing hierarchical orders in trees and can be labeled with phrase labels such as verb phrase (VP), noun phrase (NP), prepositional phrase (PP) and sentence (S). Using these principles, leaf nodes are normally labeled with the aid of part-of-speech tags. The Chomsky Hierarchy (Chomsky, 1957) classifies phrase-structure grammars in accordance with the form of their productions.

The first class of SBSMT explicitly models the translation process. It utilizes the string-to-tree approach in the form of synchronous phrase-structure grammars (SPSG). SPSGs generate two simultaneous trees, each representing the source and targets sentence, of a machine translation application. For instance, an English noun phrase 'a good boy' with Bhojpuri translation एगो नीक लइका will manifest synchronous rules as

NP → DET$_1$ JJ$_2$ NN$_3$ | DET$_1$ JJ$_2$ NN$_3$

NP → a good boy | एगो नीक लइका

Each rule will find itself associated with a set of features including PBSMT features. A translation hypothesis is measured as a product of all derivation rules associated with language models. Wu (1997) proposed bilingual bracketing grammar where only binary is used. This grammar performed well in several cases of word alignments and for word reordering constraints in decoding algorithms. Chiang (2005, 2007) presented hierarchical phrase model (Hiero) which is an amalgamation of the principles behind phrase-based models and tree structure. He proposed a resourceful decoding method based on chart parsing. This method did not use any linguistic syntactic rules/information (already explained in the previous section).

Tree-to-tree and tree-to-string models constitute the second category. This category makes use of synchronous tree-substitution grammars (STSG). The SPSG formalism gets extended to include trees on the right-hand side of rules along with non-terminal and terminal symbols. There are either non-terminal or terminal symbols at the leaves of the trees. All non-terminal symbols on the right-hand side are mapped on one-to-one basis between the two languages.

STSGs allow the generation of non-isomorphic trees. They also allow overcoming the child node reordering constraint of flat context-free grammars (Eisner, 2003). The application of STSG rules is similar to SPSG rules except for the introduction of an additional structure. If this additional structure remains unhandled, then flattening STSG rules is the way to obtain SPSG rules. Galley et al. (2004, 2006) presented the GHKM rule extraction which is a process similar to that of phrase-based extraction. The similarity lies in the fact that both extract rules which are consistent with given word alignments. However, there are differences as well of which the primary one is the application of syntax trees on the target side, instead of words sequence on. Since STSGs, consider only 1-best tree, they become vulnerable to parsing errors and rule coverage. As a result, models lose a larger amount of linguistically-unmotivated mappings. In this vein, Liu et al. (2009) propose a solution to replace the 1-best tree with a packed forest.

Cubic time probabilistic bottom-up chart parsing algorithms, such as CKY or Earley, are often applied, to locate the best derivation in SBMT models. The left-hand side, of both SPSG and STSG rules, holds only one non-terminal node. This node employs efficient dynamic programming decoding algorithms equipped with strategies of recombination and pruning (Huang and Chiang, 2007; Koehn, 2010). Probabilistic CKY/Earley decoding method has to frequently deal with binary-branching grammar so that the number of chart entries, extracted rules, and stack combinations can be brought down (Huang et al., 2009). Furthermore, incorporation of n-gram language models in decoding causes a significant rise in the computational complexity. Venugopal et al. (2007) proposed to conduct a first pass translation without using any language model. His suggestion included to then proceed with the scoring of the pruned search hyper graph in the second pass using the language model. Zollmann et al. (2008) presented a methodical comparison between PBSMT and SBSMT systems functioning in different language pairs and system scales. They demonstrated that for language pairs with sufficiently non-monotonic linguistic properties, SBMT approaches yield substantial benefits. Apart from the tree-to-string, string-to-tree, and tree-to-tree systems, researchers have added features derived from linguistic syntax to phrase-based and hierarchical phrase-based systems. In the present work, string-to-tree or tree-to-tree are not included. Only the tree-to-string method using the dependency parses of source language sentences is implemented.

## 1.7 An Overview of Existing Indian Languages MT System

This section has been divided into two subsections. First sub-section gives an overview of the MT systems developed for Indian languages while the second sub-section reports current evaluation status of English-Indian languages (Hindi, Bengali, Urdu, Tamil, and Telugu) MT systems.

## 1.7.1 Existing Research MT Systems

MT is a very composite process which requires all NLP applications or tools. A number of MT systems have been already developed for E-ILs or ILs-E, IL-ILs, English-Hindi or Hindi-English such as AnglaBharati (Sinha et al., 1995), Anusāraka (Bharti et al., 1995; Bharti et al., 2000; Kulkarni, 2003), UNL MTS (Dave et al., 2001), Mantra (Darbari, 1999), Anuvadak (Bandyopadhyay, 2004), Sampark (Goyal et al., 2009; Ahmad et al, 2011; Pathak et al, 2012; Antony, 2013).), Shakti and Shiva (Bharti et al.; 2003 and 2009), Punjabi-Hindi (Goyal and Lehal, 2009; Goyal, 2010), Bing Microsoft Translator (Chowdhary and Greenwood, 2017), Google Translate (Johnson et al., 2017), SaHit (Pandey, 2016; Pandey and Jha, 2016; Pandey et al., 2018), Sanskrit-English (Soni, 2016), English-Sindhi (Nainwani, 2015), Sanskrit-Bhojpuri (Sinha, 2017; Sinha and Jha, 2018) etc. The brief overview of Indian MT systems from 1991 to till present is listed below with the explanations of approaches followed, domain information, language-pairs and development of years:

| Sr. No. | Name of the System | Year | Language Pairs for Translation | Approaches | Domain |
|---|---|---|---|---|---|
| 1. | AnglaBharti-1(IIT K) | 1991 | Eng-ILs | Pseudo- Interlingua | General |
| 2. | Anusāraka (IIT-Kanpur, UOH and IIIT-Hyderabad) | 1995 | IL-ILs | Pāṇinian Grammar framework | General |
| 3. | Mantra (C-DAC- Pune) | 1999 | Eng-ILs Hindi- Emb | TAG | Administration, Office Orders |
| 4. | Vaasaanubaada (A U) | 2002 | Bengali- Assamese | EBMT | News |
| 5. | UNL MTS (IIT-B) | 2003 | Eng-Hindi | Interlingua | General |
| 6. | Anglabharti-II (IIT-K) | 2004 | English-ILs | GEBMT | General |
| 7. | Anubharti-II(IIT-K) | 2004 | Hindi-ILs | GEBMT | General |
| 8. | Apertium | - | Hindi-Urdu | RBMT | - |
| 9. | MaTra (CDAC-Mumbai) | 2004 | English- Hindi | Transfer Based | General |
| 10. | Shiva & Shakti (IIIT-H, IISC-Bangalore and CMU, USA) | 2004 | English- ILs | EBMT and RBMT | General |
| 11. | Anubad (Jadavapur University, Kolkata) | 2004 | English-Bengali | RBMT and SMT | News |
| 12. | HINGLISH (IIT-Kanpur) | 2004 | Hindi-English | Interlingua | General |
| 13. | OMTrans | 2004 | English-Oriya | Object oriented concept | - |
| 14. | English-Hindi EBMT system | 2005 | English-Hindi | EBMT | - |
| 15. | Anuvaadak (Super Infosoft) | | English-ILs | Not- Available | - |
| 16. | Anuvadaksh (C-DAC- Pune and other EILMT members) | 2007 and 2013 | English-ILs | SMT and Rule-based | - |
| 17. | PUNJABI-HINDI (Punjab University, Patiala) | 2007 | Punjabi-Hindi | Direct word to word | General |
| 18. | Systran | 2009 | English-Bengali, Hindi and Urdu | Hybrid-based | |
| 19. | Sampark | 2009 | IL-ILs | Hybrid-based | - |
| 20. | IBM MT System | 2006 | English-Hindi | EBMT & SMT | - |
| 21. | Google Translate | 2006 | English-ILs, IL-ILs and Other Lgs (more than 101 Lgs) | SMT & NMT | - |
| 22. | Bing Microsoft Translator | Between 1999-2000 | English-ILs, IL-ILs and Others Lgs (more than 60 Lgs) | EBMT, SMT and NMT | - |
| 23. | Sata-Anuvādak (IIT-Bombay) | 2014 | English-IL and IL-English | SMT | ILCI Corpus |
| 24. | Sanskrit-Hindi MT System (UOH, JNU, IIIT-Hyderabad, IIT-Bombay, JRRSU, KSU, | 2009 | Sanskrit-Hindi | Rule-based | Stories |
|     | BHU, RSKS-Allahabad, RSVP Triputi and Sanskrit Academy) | | | | |
| 25. | English-Malayalam SMT | 2009 | English-Malayalam | Rule-based reordering | - |
| 26. | Bidirectional Manipuri-English MT | 2010 | Manipuri-English and English-Manipuri | EBMT | News |
| 27. | English-Sindhi MT system (JNU, New Delhi) | 2015 | English-Sindhi | SMT | General Stories and Essays |
| 28. | Sanskrit-Hindi (SaHiT) SMT system (JNU, New Delhi) | 2016 | Sanskrit-Hindi | SMT | News and Stories |
| 29. | Sanskrit-English SMT system (JNU, New Delhi-RSU) | 2016 | Sanskrit-English | SMT | General Stories |
| 30. | Sanskrit-Bhojpuri MT (JNU, New Delhi) | 2017 | Sanskrit-Bhojpuri | SMT | Stories |

Table 1. 1: Indian Machine Translation Systems

**1.7.2 Evaluation of E-ILs MT Systems**

During the research, the available E-ILs MT systems have been studied (Table 1). To know current status of E-ILMT systems (based on the SMT and NMT models), five languages were chosen whose numbers of speakers and on-line-contents/web-resources availabilities are comparatively higher than other Indian languages. Census of India Report (2011), ethnologue and W3Tech reports were used to select five Indian languages (Hindi, Bengali, Tamil, Telugu and Urdu). Ojha et al. (2018) has conducted PBSMT and NMT experiments on seven Indian languages including these five languages using low-data. These experiments supported that SMT model gives better results compare to NMT model on the low-data for E-ILs. Even the Google and Bing (which have best MT systems and rich-resources) E-ILMTs performance is very low compare to PBSMT (Ojha et al., 2018) systems. Figure 1.4 demonstrates these results.

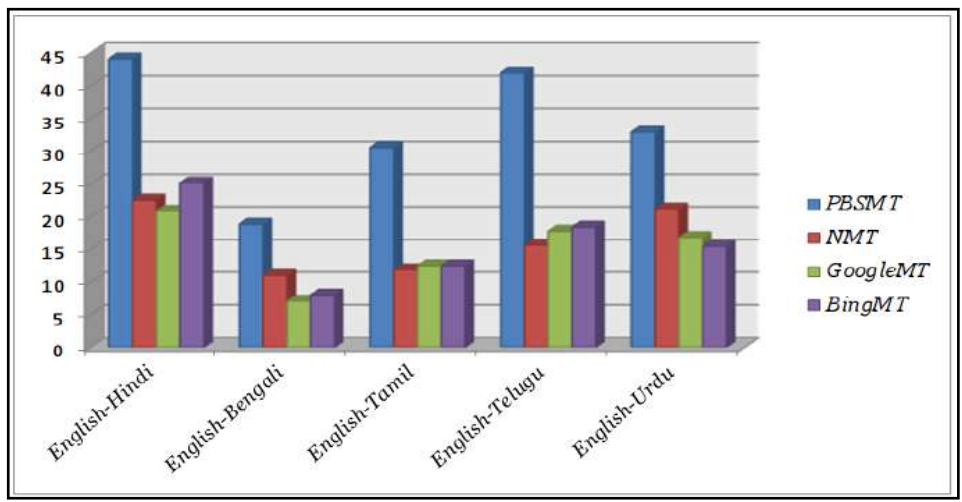

Figure 1. 4: BLEU Score of E-ILs MT Systems

**1.8 Overview of the thesis**

This thesis has been divided into six chapters namely: 'Introduction', 'Kāraka Model and it impact on Dependency Parsing', 'LT Resources for Bhojpuri', 'English-Bhojpuri SMT System: Experiment', 'Evaluation of EB-SMT System', and 'Conclusion'.

Chapter 2 talks of theoretical background of Kāraka and Kāraka model. Along with this, it talks about previous related work. It also discusses impacts of the Kāraka model in NLP and on dependency parsing. It compares Kāraka (which is also known as Pāṇinian dependency) dependency and Universal dependency. It also presents a brief idea of implementation of these models in the SMT system for English-Bhojpuri language pair.

Chapter 3 discusses the creation of language technological (LT) resources for Bhojpuri language such as monolingual, parallel (English-Bhojpuri), and annotated corpus etc. It talks about the methodology of creating LT resources for less-resourced languages. Along with these discussions, this Chapter presents already existing resources for Bhojpuri language and their current status. Finally, it provides the discussion on statistics of LT resources created and highlights issues and challenges for developing resources for less-resourced languages like Bhojpuri.

Chapter 4 explains the experiments conducted to create EB-SMT systems using various translation models such as PBSMT, FBSMT, HBSMT and Dep-Tree-to-Str (PD and UD based). It also illustrates the LM and IBM models with the example. Finally, it briefly mentions evaluation reports of trained SMT systems on the BLEU metric.

Chapter 5 discusses automatic evaluation reports of the developed PBSMT, HBSMT, FBSMT, PD based Dep-Tree-to-Str and UD based Dep-Tree-to-Str SMT systems. It also presents Human Evaluation report for only the PD and UD based Dep-Tree-to-Str SMT systems. Finally, it reports comparative error analysis of the PD and UD based SMT systems.

Chapter 6 concludes the thesis and proposes the idea of future works to improve developed EB-SMT system accuracy such as pre-editing, post-editing, and transliteration methods etc.

# Chapter 2

# Kāraka Model and its Impact on Dependency Parsing

*"Dependency grammar is rooted in a long tradition, possibly going back all the way to Pāṇini's grammar of Sanskrit several centuries before the Common Era, and has largely developed as a form for syntactic representation used by traditional grammarians, in particular in Europe, and especially for Classical and Slavic languages."*

*Sandra Kübler, Ryan McDonald, and Joakim Nivre (2009)*

## 2.1 Introduction

Sanskrit grammar is an inevitable component of many Indian languages. This is evident from the fact that many features of the Sanskrit grammar can be traced as subsets within the syntactic structure of a variety of languages such as Hindi, Telugu, Kannada, Marathi, Gujarati, Malayalam, Odia, Bhojpuri, and Maithili and so on. Some of the key features like morphological structures, subject/object and verb correlatives, free word-order, case marking and case or *kāraka* used in the Sanskrit language form the bases of many dialects and languages (Comrie, 1989; Masica, 1993; Mohanan, 1994). More importantly, it has been found that Sanskrit grammar is potent to be used in Interlingua approach for building the multilingual MT system. The features of the grammar structures are such that they prove to be a set of construction tools for the MT system (Sinha, 1989; Jha and Mishra, 2009; Goyal and Sinha, 2009). Working along those lines, the Sanskrit grammar module also has a flexibility to deal with the AI and NLP systems (Briggs, 1985; Kak, 1987; Sinha, 1989; Ramanujan, 1992; Jha , 2004; Goyal and Sinha, 2009). Here, it is worth to be emphasized that Pāṇinian grammatical (Pāṇini was an Indian grammarian who is credited with writing a comprehensive grammar of Sanskrit namely Aṣṭādhyāyī) model is efficient not only in providing a syntactic grounding but also an enhanced semantic understanding of the language through syntax (Kiparsky et al., 1969; Shastri, 1973).

It has been observed that accuracy of MT system for the Indian languages is very low. The reasons being that majority of the Indian languages are morphologically richer, free word-order etc. The Indian languages comprise free-word orders as compared to the European languages. On the parameters of linguistic models, it can be said that Indian Languages and English have divergent features both in grammatical and the syntactico-semantic structures. This difference leads to a need for a system that can fill the gaps among the antipodal languages of the Indian subcontinent and the European languages. Indian researchers have thus resorted to the use of computational Pāṇinian grammar framework. This computational model acts as filler for the evident gaps among dissimilar language structures. The concepts of the Pāṇini Grammar have been used for the computational processing purpose. This computational process of a natural language text is known as Computational Pāṇinian Grammar (CPG). Not only has the CPG framework been implemented among the Indian languages, but also has been successfully applied to English language (Bharati et al., 1995) in NLP/language technology applications. For instance, the uses of systems such as Morphological Analyzer and Generator, Parser, MT, Anaphora resolution have proven the dexterity of the Computational Pāṇinian Grammar (CPG).

In NLP, parsing is one efficient method to scrutinize a sentence at the level of syntax or semantics. There are two kinds of famous parsing methods are used for this purpose, namely constituency parse, and dependency parse. A constituency parse is used to syntactically understand the sentence structure at the level of syntax. In this process there is an allotment of the structure to the words in a sentence in terms of syntactic units. The constituency parse as is suggested by its name, is used to organize the words into close-knit nested constituents. In other words, it can be said that the word divisions of a sentence are formulated by the constituent parse into subunits called phrases. Whereas, the dependency parse is useful to analyse sentences at the level of semantics. The role of the dependency structure is to represent the words in a sentence in a head modifier structure. The dependency parse also undertakes the attestation of the relation labels to the structures.

Hence in order to comprehend the structures of morphologically richer and free word-order language the dependency parse is preferred over constituency parse. This preference is made as it is more suitable for a wide range of NLP tasks such as machine translation, information extraction, question answering. Parsing dependency trees are simple and fast.

The dependency model provides two popular annotation schemes (1) Pāṇinian Dependency (PD) and (2) Universal Dependency (UD).

The PD is developed on the module of Pāṇini's Kāraka theory (Bharati et al., 1995, Begum et al., 2008). There are several projects that have been based on this scheme. The PD offers efficient results for Indian languages (Bharati et al., 1996; Bharati et al., 2002; Begum et al., 2008; Bharati et al., 2009; Bhat et al., 2017). The UD has been acknowledged rapidly as an emerging framework for cross-linguistically consistent grammatical annotation. The efforts to promote the Universal Dependency are on the rise. For instance, an open community attempt with over two hundred distributors producing more than one hundred TreeBanks in more than seventy languages has generated a mammoth database (as per the latest release of UD-v2)[1]. Dependency tag-set of the UD is prepared on the Stanford Dependencies representation system (Marneffe et al., 2014). Detailed analysis of the description of the respective dependencies frameworks would be undertaken in section 2.4.

The dependency modal is consistently being used for improving, developing or encoding the linguistic information as given in the Statistical and Neural MT systems (Bach, 2012, Williams et al., 2016; Li et al., 2017; Chen et al., 2017). However, to the best of my knowledge, both of the PD and UD models have not been compared to check their suitability for the SMT system. Even, due to the above importance, there is no attempt to develop SMT system based on the Pāṇinian Kāraka dependency model for English-Low-resourced Indian languages (ILs) either in string-tree, tree-string, tree-tree or dependency-string approaches. The objective of the study is to undertake a palimpsest research for improving accuracy of low-resourced Indian languages SMT system using the Kāraka model. Hence in order to improve accuracy and to find suitable framework, both the Pāṇinian and Universal dependency models have been used for developing the English-Bhojpuri SMT system.

This chapter is divided into five (including introduction) subsections. An overview of the Kāraka and Kāraka model given in section 2.2. This segment also deals with the uses of the model in Indian languages and in the computational framework. The section 2.3 elaborates on literature review related to the Kāraka model. It also scrutinizes the CPG framework in the Indian language technology. The section 2.4 emphasizes the models of

---

[1] http://universaldependencies.org/#language-

Dependency Parsing, PD and UD annotation schemes as well as their comparisons. The final section 2.5 concludes this chapter.

## 2.2 Kāraka Model

The etymology of Kāraka can be traced back to the Sanskrit roots. The word Kāraka refers to 'that which brings about' or 'doer' (Joshi et al., 1975, Mishra 2007). The Kāraka in Sanskrit grammar traces the relation between a noun and a verb in a sentence structure. Pāṇini neologized the term Kāraka in the sūtra Kārake (1.4.23, Astadhyayi). Pāṇini has used the term Kāraka for a syntactico-semantic relation. It is used as an intermediary step to express the semantic relations through the usage of *vibhaktis*. As per the doctrine of Pāṇini, the rules pertaining to Kāraka explain a situation in terms of action (kriyā) and factors (kārakas). Both the action (kriyā) and factors (kārakas) play an important function to denote the accomplishment of the action (Jha, 2004; Mishra, 2007). Most of the scholars and critics agree in dividing d Pāṇini's Kāraka into six types:

- **Kartā (Doer, Subject, Agent):** "one who is independent; the agent" (स्वतंत्र: कर्ता (svataMtra: kartā), 1.4.54 Aṣṭādhyāyī). This is equivalent to the case of the subject or the nominative notion.
- **Karma (Accusative, Object, Goal):** "what the agent seeks most to attain"; deed, object (कर्तुरीप्सिततमं कर्म (karturIpsitatamaM karma), 1.4.49 Aṣṭādhyāyī). This is equivalent to the accusative notion.
- **Karaṇa (Instrumental):** "the main cause of the effect; instrument" (साधकतमं करणम् (sAdhakatamaM karaNam), 1.4.42 Aṣṭādhyāyī). This is equivalent to the instrumental notion.
- **Saṃpradāna (Dative, Recipient):** "the recipient of the object" (कर्मणायमभिप्रेति स संप्रदानम् (karmaNAyamabhipreti sa saMpradAnam), 1.4.32 Aṣṭādhyāyī). This is equivalent to the dative notion which signifies a recipient in an act of giving or similar acts.
- **Apādāna (Ablative, Source):** "that which is firm when departure takes place" (ध्रुवमपायेऽपादानम् (dhruvamapAyeSpAdAnam), 1.4.24 Aṣṭādhyāyī). This is the equivalent of the ablative notion which signifies a stationary object from which a movement proceeds.

- **Adhikaraṇa (Locative):** "the basis, location" (आधारोऽधिकरणम् (AdhAroSdhikaraNam), 1.4.45 Aṣṭādhyāyī). This is equivalent to the locative notion.

But he assigns *sambandha* (genitive) with another type of *vibhakti* (case ending). It aids in expressing the relation of a noun to another. According to Pāṇini, case endings recur to express relations between *kāraka* and kartā. Such relations are known as *prathamā* (nominative endings). In the Sanskrit language, these seven types of case endings are based on 21 sub vibhaktis/case markers that are bound to change according to the language.

Since ancient times Kāraka theory has been used to analyze the Sanskrit language, but due to its efficiency and flexibility the Pāṇinian grammatical model was adopted as an inevitable choice for the formal representation in the other Indian languages as well. The application of the Pāṇinian grammatical model to other Indian languages led to the consolidation of the *Kāraka* model. This model helps to extract the syntactico-semantic relations between lexical items. The extraction process provides two trajectories which are classified as *Kāraka* and *Non-kāraka* (Bharati et al, 1995; Begum et al., 2008; Bhat, 2017).

(a) **Kāraka:** These units are semantically related to a verb. They are direct participants in the action denoted by a verb root. The grammatical model depicts all six 'kārakas', namely the *Kartā*, the *Karma*, the *Karaṇa*, the *Sampradāna,* the *Apādāna* and the Adhikaraṇa. These relations provide crucial information about the main action stated in a sentence.

(b) **Non-kāraka:** The *Non-kāraka dependency relation* includes purpose, possession, adjectival or adverbial modifications. It also consists of cause, associative, genitives, modification by relative clause, noun complements (appositive), verb modifier, and noun modifier information. The relations are marked and become visible through '*vibhaktis*'. The term '*vibhakti*' can approximately be translated as inflections. The *vibhaktis* for both nouns (number, gender, person and case) and verbs (tense, aspect and modality (TAM)) are used in the sentence structure.

Initially, the model was applied and chosen for Hindi language. The idea was to parse the sentences in the dependency framework which is known as the PD (shown in the Figure 2.1). But an effort was made to extend the model for other Indian languages including English (see the section 2.4 for detail information of the PD model).

| **(I)** | दीपक ने अयान को लाल गेंद दी। | (Hindi sentence) |
| | dIpaka ne ayAna ko lAla geMda dI | | (ITrans) |
| | deepak ne-ERG ayan acc red ball give-PST . | (Gloss) |
| | Deepak gave a red ball to Ayan. | (English Translation) |

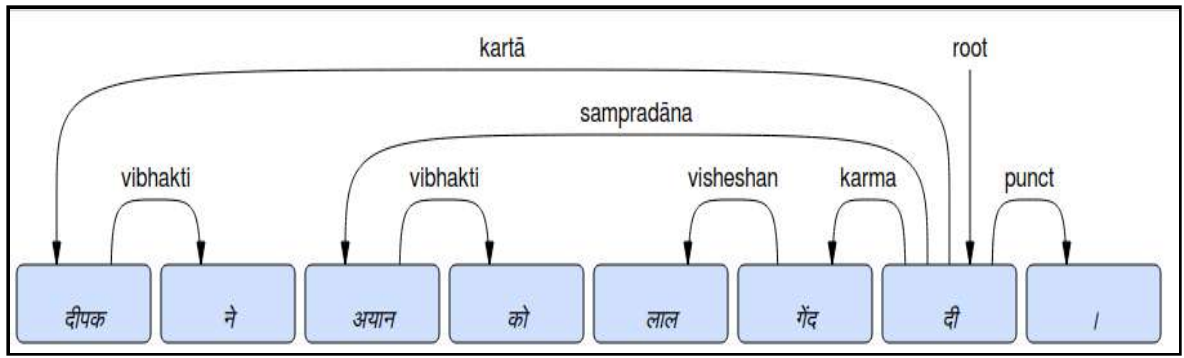

Figure 2. 1: Dependency structure of Hindi sentence on the Pāṇinian Dependency

The above figure depicts dependency relations of the example (I) sentence on the Kāraka model. In the dependency tree, verb is normally presented as the root node. The example (I) dependency relation represents that दीपक is the 'kartā' (doer marked as kartā) of the action. This is denoted by the verb दी. The word अयान is the '*saṃpradāna*' (recipient marked as *saṃpradāna*) and गेंद is the 'karma' (object marked as karma) of the verb, and दी is the root node.

**2.3 Literature Review**

There have been several language technology tools that have developed on the basis of Kāraka or computational Pāṇinian grammar model. Following is a brief summary of the linguistic tools:

- **MT (Machine Translation) System:** Machine Translation systems have been built specifically keeping in mind the Indian Language syntactic structures. Systems such as Anusāraka, Sampark, Shakti MT systems endorse Pāṇinian framework in which either full or partial framework is put to use.

**(a) Anusāraka:** The Anusāraka MT was developed in 1995. It was created by the Language Technology Research Centre (LTRC) at IIIT-Hyderabad (initially it was started at IIT-Kanpur). The funding for the project came from TDIL, Govt of India. Anusāraka is adept in using principles of Pāṇinian Grammar (PG). It also projects a close similarity with Indian languages. Through this structure, the Anusāraka essentially maps local word groups between the source and target languages. In case of deep parsing it uses *Kāraka* models to parse the Indian languages (Bharti et al., 1995; Bharti et al., 2000; Kulkarni, 2003; Sukhda, 2017). Language Accessors for this programming have been developed from indigenous languages such as Punjabi, Bengali, Telugu, Kannada and Marathi. The Language Accessors aid in accessing a plethora of languages and providing reliable Hindi and English-Indian language readings. The approach and lexicon is generalized, but the system has mainly been applied on children's literature. The primary purpose is to provide a usable and reliable English-Hindi language accessor for the masses.

**(b) Shakti:** Shakti is a form of English-Hindi, Marathi and Telugu MT system. It has the ability to combine rule-based approach with statistical approaches and follow Shakti Standard Format (SSF). This system is a product of the joint efforts by IISC-Bangalore, and International Institute of Information Technology, Hyderabad, in collaboration with Carnegie Mellon University USA. The Shakti system aids in using kāraka model in dependency parsing for extracting dependency relations of the sentences (Bharti et al., 2003; Bharti et al.; 2009; Bhadra, 2012).

**(c) Sampark:** Sampark is a form of an Indian Language to Indian Language Machine Translation System (ILMT). The Government of India funded this project where in eleven Indian institutions under the consortium of ILMT project came forwards to produce the system. The consortium has adopted the Shakti Standard Format (SSF). This format is utilized for in-memory data structure of the blackboard. The systems are based on a hybrid MT approach. The Sampark system constitutes the Computational Pāṇinian Grammar (CPG) approach for language analysis along with the statistical machine learning process (Goyal et al., 2009; Ahmad et al, 2011; Pathak et al, 2012; Antony, 2013). The system has proven beneficial as it has successfully developed language translation technology for nine Indian languages. In the process it has resulted in building MT for 18 language pairs. These are: 14 bi-directional pairs between Hindi and Urdu/Marathi/Punjabi /Bengali/Tamil/Telugu/Kannada and 4 bi-directional pairs between Tamil and Telugu/ Malayalam.

**(d) English-Sanskrit and Sanskrit-English MT System:** The AnglaBharati system is being scrutinized by linguist to formulate a system-design to translate English to Indian languages. This would be a blueprint for further developing a system that could be adapted for translations to Sanskrit (Goyal and Sinha, 2009). The researchers, Goyal and Sinha have demonstrated that the machine translation of English to Sanskrit for simple sentences could be accomplished. The simple sentence structures translated were based on PLIL generated by AnglaBharati and Aṣṭādhyāyī rules. In this experiment, the scholars have used *Kāraka* theory to decode the meaning from sentence.

Sreedeepa and Idicula (2017) reported that an Interlingua based Sanskrit-English MT system has developed using Pāṇinian framework (Sreedeepa and Idicula, 2017). They used kāraka analysis system for the semantic analysis. But the drawback of this paper, there is no evaluation report.

- **Sanskrit Kāraka Analyzer:** The Sanskrit kāraka analyzer (Mishra, 2007; Jha and Mishra, 2009) has been developed by Sudhir Mishra in 2007 at JNU, New Delhi during his doctoral research. The project was undertaken to create a translation tool for Sanskrit language. This Kāraka analyzer was efficient in the syntactico-semantic relations at the sentence level following with the rule based approach. But it is limited in a way that it is unable to perform on deep semantic structural analysis of Sanskrit sentences.

- **Constraint based Parser for Sanskrit language:** A Constrained Based Parser for Sanskrit Language has been developed by University of Hyderabad in 2010 (Kulkarni et al., 2010). The system was concretized through the principles of generative grammar. The designing features of the generative grammars helped the parser for finding the directed trees. In the tree pattern graphs, the nodes represent words, and edges depict the relations between the words and edges. To combat dead-ends and overcome the scenario of non-solutions the linguists used mimāmsā constraints of ākānksā and sannidhi. The current system at work allows only limited and simple sentences to be parsed.

- **Frame-Based system for Dravidian Language Processing:** Idicula (1999) in his doctoral thesis, 'Design and Development of an Adaptable Frame-Based system for Dravidian Language Processing' has used kāraka model. The kāraka relation has been used to extract meaning. Under this system, the author has used the modem of vibhakti-kāraka mapping for analyzing the kāraka relation.

- **Shallow Parser Tool:** The Shallow Parser Tools[2] for Indian languages (SPT-IL) project was started in 2012 in the consortium mode (University of Hyderabad, Jawaharlal Nehru University, University of Jammu, Gujarat University, University of Kashmir, Goa University, Visva-Bharati University, Guwahati University, Manipur University and North Bengal University). The project was supported and funded by TDIL, Govt. of India. The project aims to build Shallow Parser Tools for 12 Indian languages constitution Hindi, Assamese, Bodo, Dogri, Gujarati, Kashmiri, Konkani, Maithili, Manipuri, Nepali, Odia, and Santhali. The basic components that are required for this task comprise Morphological Analyzer, a POS Tagger and a Chunker. In the development of the Morph Analyzer, the Pāṇinian paradigm model was used. The Pāṇinian paradigm was already in use in diverse range of MT projects before it was used as the building block for Shallow Parser tool. The Pāṇinian paradigm has already been implemented for building the morph analyzer tools like Indian-Indian Languages Machine Translation (ILMT), Anusāraka, Sampark, AnglaBharati.

- **Dependency Treebank and Parser:** The Development of Dependency Tree Bank for Indian Languages project[3] was started in 2013 in consortium mode. The project was formulated under the IIIT- Hyderabad consortium leader and sponsored by TDIL, Govt. of India. The fundamental objective of this project was to resurrect a monolingual and parallel Treebank for languages such as Hindi, Marathi, Bengali, Kannada, and Malayalam. To accomplish this Treebank model, Pāṇinian Kāraka Dependency annotation scheme was followed. As an offshoot of the project, the annotation scheme was also utilized to annotate the data for Telugu, Urdu and Kashmiri languages (Begum et al., 2008; Hussain, 2009; Hussain et al., 2010; Bhat, 2017). The dependency parser was thus created for the Hindi, Telugu, Bengali, Urdu and Kashmiri languages on the basis of PD framework (Hussain et al., 2010, Dhar et al., 2012, Bhat, 2017).

---

[2] see following link for details information: "http://sanskrit.jnu.ac.in/projects/sptools.jsp?proj=sptools"
[3] http://meity.gov.in/content/language-computing-group-vi

## 2.4 Pāṇinian and Universal Dependency Framework

The following subsections provide an idea of the workings of both the PD and UD annotation scheme. But this description is only accomplished at the level of dependency and did not focus other levels such as morphology, POS, chunk.[4] In the section 2.4.3, the differentiation between both these schemes has been amplified illustrated examples.

### 2.4.1 The PD Annotation

The PD scheme is conceptualized on a modifier-modified relationship (Bharati et al., 1995). For the PD model, I have followed the 'Ancorra: TreeBanks for Indian Languages' guidelines (Bharati et al., 2012) which is prepared by the Govt. sponsored projects ( the IL-ILMT and Dependency TreeBank for Indian languages). In the scheme, all the kāraka relations having tags starting with k are enlisted first. This pattern is followed by non-kāraka relation labels which begin with 'r'. As per the guideline, the scheme has 40 relation tags which are mentioned in the Table1. These labels or tags are constructed from the main five labels at the coarsest level. The labels that aid in the construction process are namely- verb modifier (vmod), noun modifier (nmod), adjective modifier (jjmod), adverbial modifier (advmod) and conjunct of (ccof). Among these labels, ccof is not strictly a dependency relation (Begum et al., 2008). In the Figure 2.2, a hierarchy of relations used in the scheme becomes visible. A hierarchical set-up of dependency relations is thus established on the basis of this categorization.

---

[4] To know the PD annotation and other levels procedures see the Akshar Bharati et al., 2012 'Ancorra: TreeBanks for Indian Languages'. And, for others tagsets/information of the UD, see the UD website (http://universaldependencies.org) or follow these articles: De Marneffe et al., 2014 'Universal Stanford Dependencies: A cross-linguistic typology', Daniel Zeman ' Reusable Tagset Conversion Using Tagset Drivers', Joakim Nivre et al., 'Universal Dependencies v1: A Multilingual Treebank Collection'.

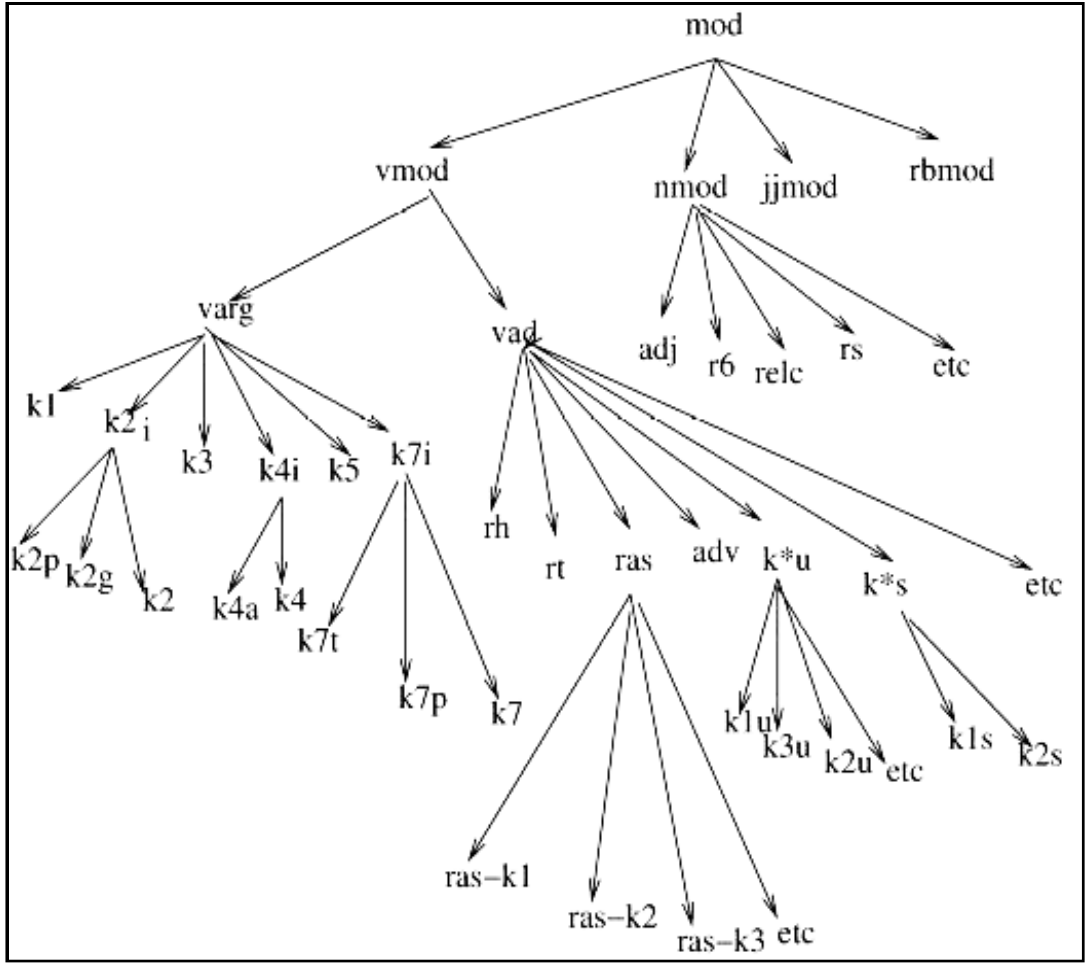

Figure 2. 2: Screen-Shot of the PD Relation Types at the Hierarchy Level

The respective part of speech category becomes the basis for the classification of the modified group (mod). For instance, all those relations whose parent (mod) is a verb fall under the verb modifier (vmod) category. These relations are further classified through the use of subsequent levels. The next level comprises of verb argument (varg), verb adjunct (vad) and vmod_1 labels under the umbrella term of vmod. One can observe the categories of adjective (jj), genitive (r6) classified under the nmod_adj label. At the most finely attuned level, varg and vad further branch into trajectories like kartā (k1), karma (k2), karan (k3), sampradāna (k4), apādāna (k5), adhikaran (k7) and hetu 'reason' (rh), tadarthya 'purpose' (rt), relation prati 'direction' (rd), vidheya kartā-kartā (k1s), sādrishya 'similarit/comparison' (k*u), kriyāvishesaṇa 'adverb' (adv) etc. The relations labeled under *varg* are the six *Kārakas* that are the most essential participants in an action sequence (the Figure1 shows its sample). On the other hand, all the other dependency tags are termed as the non-kārakas.

| Description (Relations) | Gloss | Tags/labels |
|---|---|---|
| Kartā | Doer/agent/subject | k1 |
| vidheya kartā - kartā samānādhikarana | Noun complement of kartā | k1s |
| prayojaka kartā | Causer | pk1 |
| prayojya kartā | Cause | jk1 |
| madhyastha kartā | Mediator-causer | mk1 |
| Karma | Object/patient | k2 |
| subtype of karma | Goal, Destination | k2p |
| secondary karma | Secondary karma | k2g |
| karma samānādhikarana | Object complement | k2s |
| karaṇa | Instrument | k3 |
| sampradāna | Recipient | k4 |
| anubhava kartā | Experiencer | k4a |
| Apādāna | A point of separation/departure from source | k5 |
| prakruti apādāna | Source material | K5prk |
| kAlAdhikarana | Location in time | k7t |
| Deshadhikarana | Location in space | k7p |
| vishayadhikarana | Location elsewhere | k7 |
| noun chunks with vibhaktis | According to | k7a |
| sAdrishya | Similarity/comparison | k*u |
| Shashthi | Genitive/possessive | r6 |
| Kartā of a conjunct verb | conjunct verb (complex predicate) | r6-k1 |
| Karma of a conjunct verb | conjunct verb (complex predicate) | r6-k2 |
| kA | Relation between a noun and a verb | r6v |
| kriyAvisheSaNa | Adverbs - ONLY 'manner adverbs' have to be taken here | adv |
| kriyAvisheSaNa yukta vaakya | Sentential Adverbs | sent-adv |
| relation prati | Direction | rd |
| Hetu | Reason | rh |
| Tadarthya | Purpose | rt |
| upapada_ sahakArakatwa | Associative | ras-k* |
| niShedha | Negation in Associative | ras-neg |
| relation samānādhikarana | noun elaboration/complement | rs |
| relation for duratives | Relation between the starting point and the end point of a durative expression | rsp |
| address terms | Address terms | rad |
| relative construction modifying a noun | Relative clauses, jo-vo constructions | nmod__relc |
| relative construction modifying an adjective | Relative clauses, jo-vo constructions | jjmod__relc |
| relative construction modifying an adverb | Relative clauses, jo-vo constructions | rbmod__relc |
| noun modifier | Participles etc modifying nouns | nmod |
| emphatic marker | noun modifier of the type emphatic marker | nmod emph |
| verb modifier | Verb modifier | vmod |
| modifiers of the adjectives | Modifiers of the adjectives | jjmod |
| part-of relation | Part of units such as conjunct verbs | pof |
| phrasal verb | Part of units in phrasal verb constructions | pof-phrv |
| conjunct-of | Coordination and subordination | ccof |
| fragment of | Relation to link elements of a fragmented chunk | fragof |
| enumerator | Enumerator | Enm |
| label for a symbol/ full stop | Tag for a symbol | rsym |
| the relation marked between a clause and the postposition | the relation marked between a clause and the postposition | psp__cl |

Table 2. 1: Details of the PD Annotation Tags

**2.4.2 The UD Annotation**

The UD project is developing cross-linguistically consistent Treebank annotation for many languages. The primary aim of this project is to facilitate multilingual parser development. The system will also take into account cross-lingual learning and perform parsing research from the perspective of language typology. The evolution of (universal) Stanford dependencies is the platform for the annotation scheme structures (Marneffe et al., 2006, 2008, 2014), Google universal part-of-speech tags (Petrov et al., 2012), and the Interset Interlingua (Zeman, 2008) for morpho-syntactic tagsets. As mentioned by (Nivre et al., 2016) and also elaborated by (Johannsen et al., 2015), the driving principles behind UD formalism are as follows:

- **Content over function:** In the binary relations, the content words are the heads of function words. For instance the lexical verbs form the head of periphrastic verb constructions. Whereas the nouns are the heads of prepositional phrases. In copula constructions, attributes take the head positions.
- **Head-first:** There are cases wherein the head positions are not clear at the first instance. For instance, in spans it is not immediately clear which element is the head because there is no direct application of the content-over-function rule. In such situations, the UD takes a head-first approach. The first element in the span takes the head position. The rest of the span elements attach to the head. This justly applies to the format of coordinations, multiword expressions, and proper names.
- **Single root attachment:** The root-dominated token performs an important function. Each dependency tree has exactly one token. This token is directly dominated by the artificial root node. Other candidates that seek direct root attachment are instead attached to this root-dominated token.

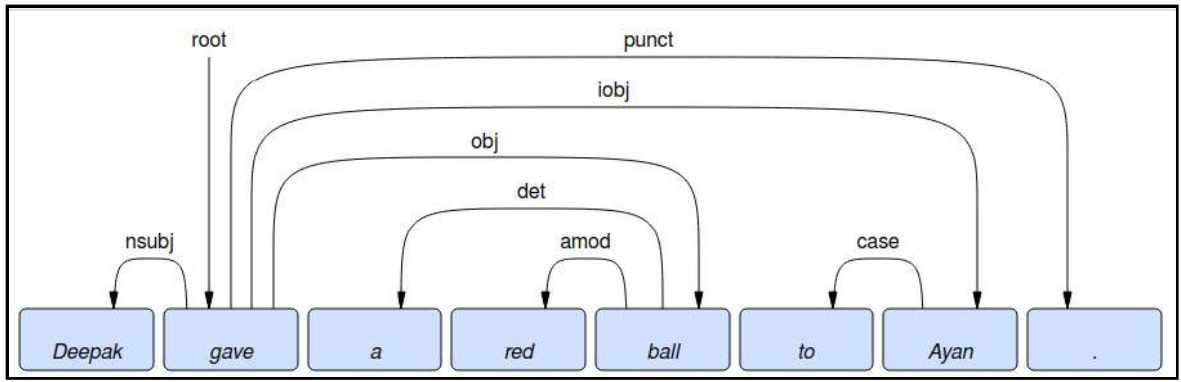

Figure 2. 3: Dependency annotation of English sentence on the UD scheme

Currently in the UD, 37 universal syntactic relations are in use (listed in the Table 2 and taken from the UD website). Some of these relations are demonstrated as a sample of the above English translated sentences (I) in Figure 2.3. The figure also showcases the above mentioned UD principles. The organizing principles of the UD taxonomy can be studied through the upper segment of the following table. The given rows correspond to functional categories in relation to the head (core arguments of clausal predicates, non-core dependents of clausal predicates, and dependents of nominals). The columns depict the structural categories of the dependent (nominals, clauses, modifier words, function words). The lower segment of the table lists relations that cannot be categorized as dependency relations in a narrow sense.

|  | Nominals | | Clauses | | Modifier Words | | Function Words | |
| --- | --- | --- | --- | --- | --- | --- | --- | --- |
|  | Tags | Description | Tags | Description | Tags | Description | Tags | Description |
| **Core Arguments** | nsubj | nominal subject | csubj | clausal subject | | | | |
|  | obj | object | ccomp | clausal complement | | | | |
|  | iobj | indirect object | xcomp | open clausal complement | | | | |
| **Non-core Arguments** | obl | oblique nominal | advcl | adverbial clause modifier | advmod | adverbial modifier | aux | auxiliary |
|  | vocative | vocative | | | discourse | discourse element | cop | copula |
| | | expl | expletive | | | | | mark | marker |
| | | disolated | dislocated elements | | | | | | |
| Nominal dependents | | nmod | nominal modifier | acl | clausal modifier of noun | amod | adjectival modifier | det | determiner |
| | | appos | appositional modifier | | | | | clf | classifier |
| | | nummod | numeric modifier | | | | | case | case marking |
| Coordination | | MWE | | Loose | | Special | | Other | |
| conj | conjunct | fixed | fixed multiword expression | list | List | orphan | orphan | punct | punctuation |
| cc | coordinating conjunction | flat | flat multiword expression | parataxis | Parataxis | goeswith | goes with | root | root |
| | | Compound | compound | | | reparandum | overridden disfluency | dep | unspecified dependency |

Table 2. 2: Details of the UD Annotation Tags

### 2.4.3 A Comparative Study of PD and UD

We have seen some differences in comparison of the annotation schemes of PD and UD in terms of correspondences between tags. There is no one-to-one correspondence in most of the tags which have the two annotation schemes. It was either many-to-one or one-to-many mappings between their tags. We will discuss these differences in detail in the coming sections.

### 2.4.3.1 Part of Speech (POS) Tags

In UD POS tagset, the total number of tags is 17 and it is less than as compared to the 32 tags in the BIS based POS tagset (TDIL, 2010; Kumar et al., 2011) developed for the Indian languages. We can see mapping in two annotation schemes: There may be several tags in UD POS tagset which correspond to a single tag in the BIS Indian languages tagset and vice versa. For instance, BIS POS tags RB (Adverb), WQ (question words),

NN (noun), INTF (intensifier), NST (spatial-temporal), etc. map to a single UD POS tag ADV. For example:

    WQ: कहाँ, कब

    NN: काल्ह, आजु

    INT: अतना, बहुते, तबहीं

    NST: के उपरे, ओहिजा

Above all tags correspond to the same POS tag 'ADV' in the UD. The reason being more granularities of BIS POS as compared to UD. In the same way, PRON (pronoun), DET (determiner) and ADV of UD POS tagset are annotated with the same BIS POS tag WQ.

**2.4.3.2 Compounding**

Bhojpuri has compound conjunctions like 'aur to aur' bhojpuri examples अउर त अउर (all the more) and जइसे कि (like/as) etc. In BIS POS schema, these are tagged as follows: अउर\CC_CCD त\RP_RPD अउर\CC_CCD.

In the UD compounding is however marked at the level of dependency relations by three tags: compound, mwe and name.

**2.4.3.3 Differences between PD and UD Dependency labels**

The PD and UD relations also show asymmetry in many cases as reflected in the POS tagsets (Tandon et al., 20116). The correspondence between these two dependencies relations are not always one to one. Such mapping can be of two types - one to many or many to one. For example, PD relation k2 corresponds to more than one relation (ccomp, dobj, xcomp) in UD. The nmod relation of UD corresponds similarly to k3, k7p, k7t and r6 relations of PD. We can see another example where the nsubj relation of UD corresponds to k1, k4a and pk1 relations of PD.

**2.4.3.4 Dependency Structure**

It is a difficult to find correspondences between the PD and UD schemas. They do not have always similar relationship. We have shown this with the help of some specific types of constructions given below.

    **(a) Conjunctions:** According to the PD framework, conjunction (subordinate or coordinate) is the head of the clause (shown Figure 2.4 and 2.6). The other

elements of clause are dependent on the head. On the other hand, the head is the first element of the coordinated construction in UD framework while the conjunction and other coordinated elements are considered as the dependents of the first element (shown Figure 2.5 and 2.7). In case of subordinating conjunction which is a dependent of the head of the subordinate clause, it is annotated as mark. We can see this in the following illustrated examples:

(II)  Anita and Ravi came yesterday.                    (English sentence)

अनिता अउर रवि काल्हि आइल रहलन ।                         (Bhojpuri sentence)

anita   aur   ravi   kalhi   Ail   rahalan.            (ITrans of Bhojpuri sentence)

Anita   and   Ravi yesterday come be-PST               (Gloss of Bhojpuri)

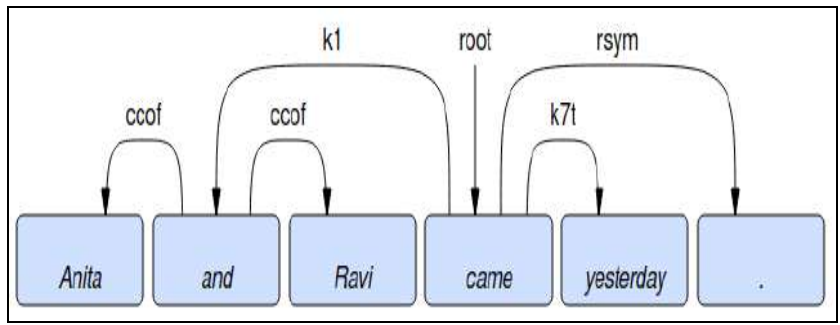

Figure 2. 4: PD Tree of English Example-II

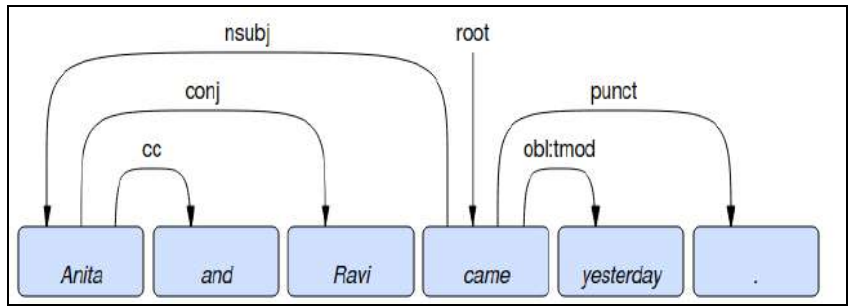

Figure 2. 5: UD Tree of English Example-II

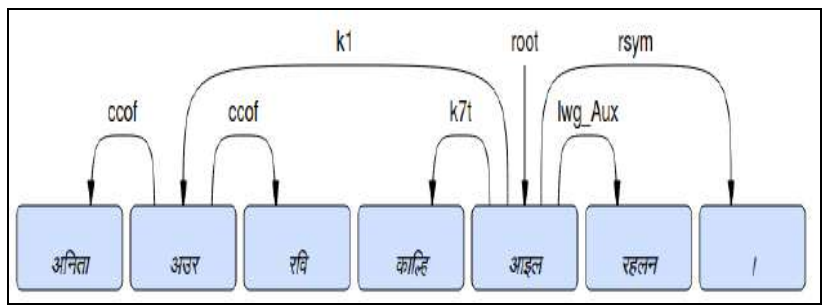

Figure 2. 6: PD Tree of Bhojpuri Example-II

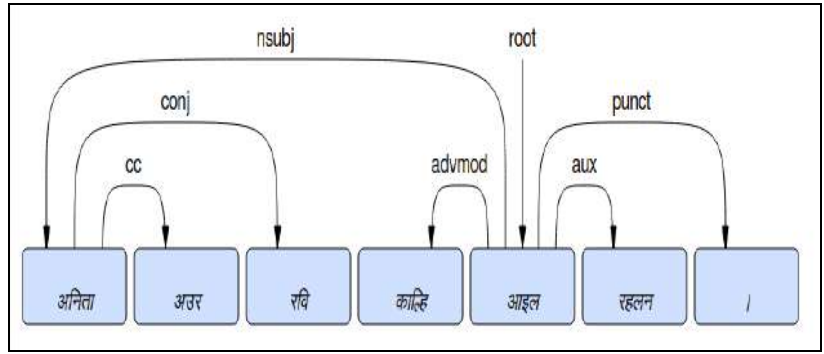

Figure 2. 7: UD Tree of Bhojpuri Example-II

**(b) Copula:** According to the PD framework (Figure 2.9 and 2.11), a copular verb is considered as the head of a copula construction whereas in UD (Figure 2.8 and 2.10), the 'be' verb becomes a cop dependent and predicative nominal in the copula construction is marked as head.

(III)  The Tajmahal is beautiful.                (English sentence)

ताजमहल सुन्नर हऽ.                                 (Bhojpuri sentence)

 tAjamahal sunnar hS.                            (ITrans of Bhojpuri sentence)

 Tajmahal beautiful be-PRS                       (Gloss of Bhojpuri)

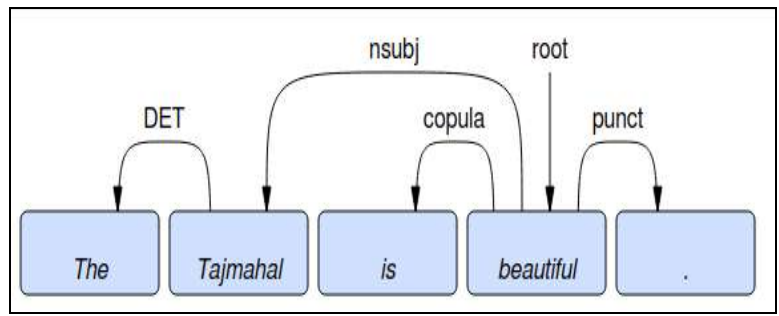

Figure 2. 8: UD Tree of English Example-III

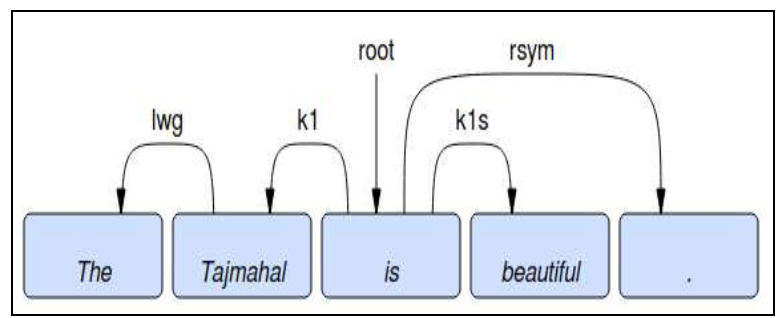

Figure 2. 9: PD Tree of English Example-III

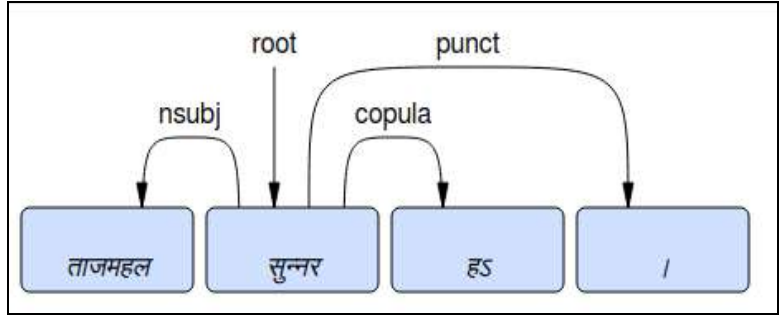

Figure 2. 10: UD Tree of Bhojpuri Example-III

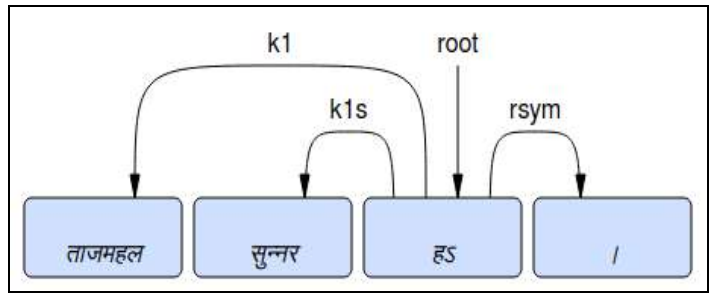

Figure 2. 11: PD Tree of Bhojpuri Example-III

'सुन्दर'/'beautiful is treated as the heads in the UD, while 'ताजमहल'/'Tajmahal' and the 'be' verb 'हS'/'is' and its dependents of the type nsubject and cop.

- **(c) Multiword names:** In the UD framework, (Johannsen et al., 2015), "In spans where it is not immediately clear which element is the head, UD takes a head-first approach: the first element in the span becomes the head, and the rest of the span elements attach to it. This applies mostly to coordinations, multiword expressions, and proper names." For instance, in a name such as 'Sachin Ramesh Tendulkar', in UD framework, the first word in a compound name 'Sachin', is considered the head and the rest becomes its dependents. On the other hand in PD framework, 'Tendulkar' is regarded as the head and 'Sachin' and 'Ramesh' are its dependents.
- **(d) Active and Passive:** One of the anomalies of the Kāraka system according to Panini shows that constructions as active and passive are the realizations of the same structure except for certain morphological distinctions (Vaidya et al. 2009).

  We also aim at the same principle to handle the case of passives in the annotation scheme (Figure 2.13). While (Figure 2.12) demonstrates the analysis of an active sentence, the same dependency tree is drawn for the passive, only marking the verb's TAM (Tense, Aspect & Modality) as passive. The feature structure that marks the verb morphology as passive will show that the agreement and positional information in the tree is applicable to k2 and not k1 (see Figure 2.13). The distinction between the two constructions is lexical (and morphological) rather than syntactic in this framework.

(IV) Deepak hit the ball.              (Active voice example of English)

(V) The ball was hit by Deepak.        (Passive vocie example of English)

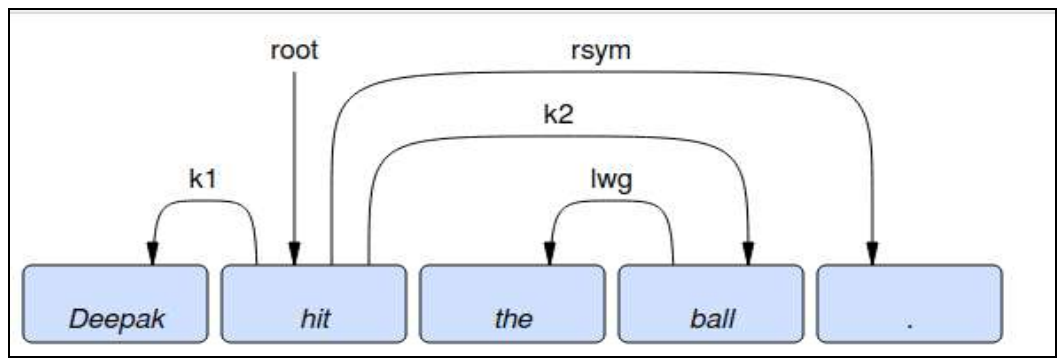

Figure 2. 12: PD Tree of Active Voice Example-IV

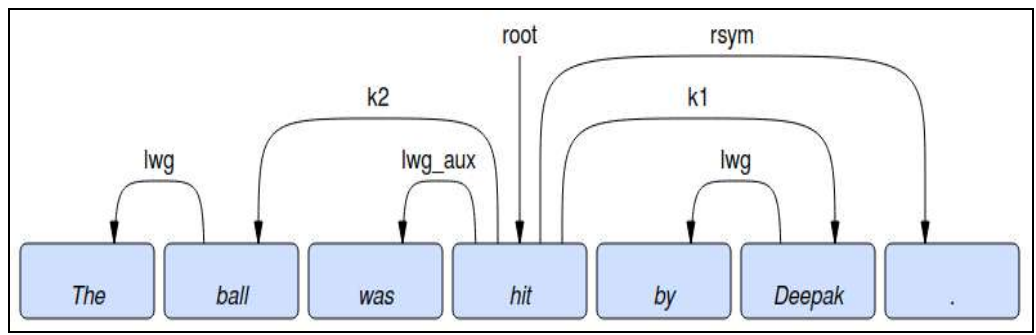

Figure 2. 13: PD Tree of Passive Voice Example-V

(e) **Yes-No Questions:** In English, we have seen interrogative sentences of two types: i) yes-no type or ii) Wh-type. In both cases, we consider the displaced element without the use of traces. The moved constituent is instead analyzed in situ. In the case of yes-no type questions, (Figure 2.14) demonstrate the dependency tree. We append the information that the sentence is a yes-no type of interrogative sentence. The moved TAM marker is given the label 'fragof' to convey that it is related to the verb chunk that is its head. We eventually mark the remaining arguments of the verb with Kāraka relations.

(VI) Did Deepak hit the ball?          (Yes-No Interrogative example of English)

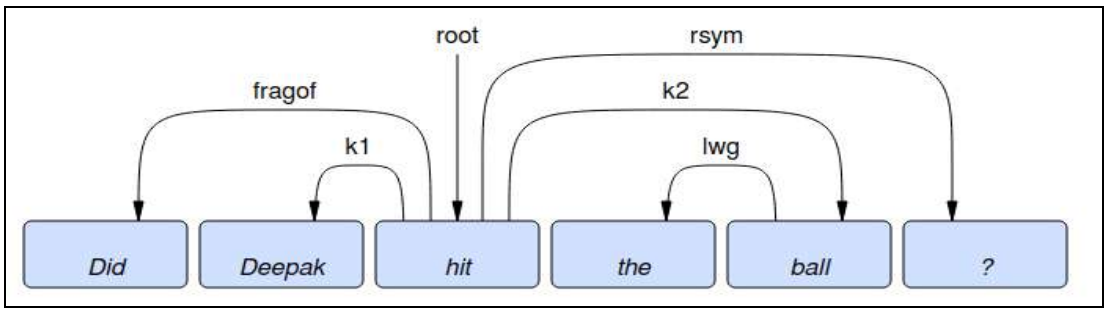

Figure 2. 14: PD Tree of English Yes-No Example-VI

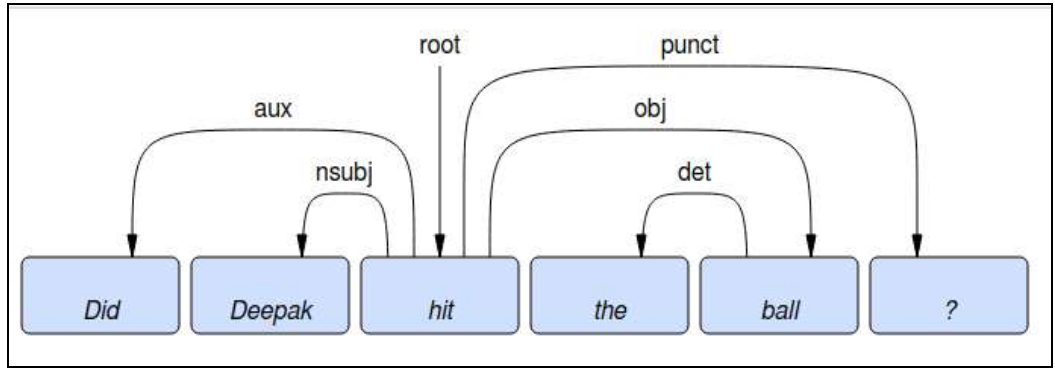

Figure 2. 15: UD Tree of English Yes-No Example-VI

**(f) Expletive Subjects:** Expletives are semantically vacuous words meant to fill the syntactic vacancy in a sentence. 'it' and 'there' are the two commonest expletives of English. Though Expletives are subjects syntactically, they cannot be kartā in a sentence since they are semantically vacuous, and kartā, though syntactically grounded, entails some semantics too. Since expletives are not found in Hindi, a new label 'dummy-sub' was formalized to facilitate annotation of the expletives of English.

(VII) It rained yesterday.           (Expletive example of English)

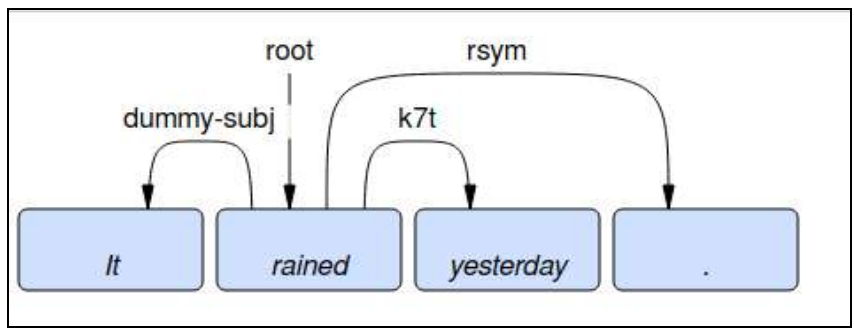

Figure 2. 16: PD Tree of English Expletive Subjects Example-VII

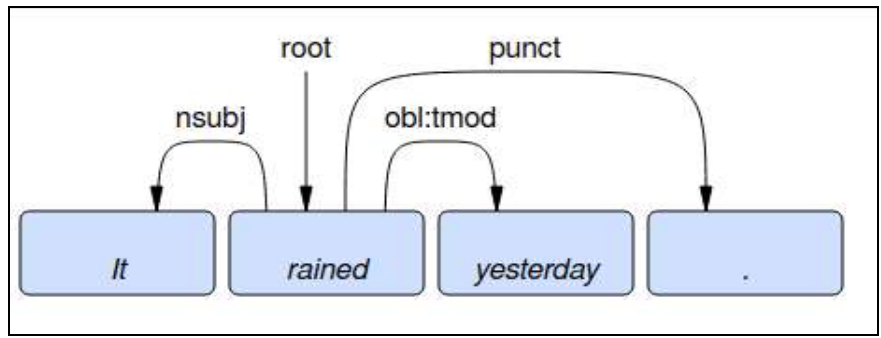

Figure 2. 17: UD Tree of English Expletive Subjects Example-VII

As seen in Figure 2.16, 'it' is a semantically vacuous element that serves to fill the empty subject position in the sentence. The fact that 'it' is not the locus of the action in the sentence substantiates that though it fills the subject position, it fails to function as kartā. Therefore, we mark it with a special relation 'dummy-sub', which reflects the fact that 'it' is a dummy element in the sentence.

**(g) Subordinate Clauses:** Verbs such as want that take subordinate clauses can be represented where the subordinate clause is related with the relation k2 as karma. In Figure 2.18 for example, 'expects' takes 'to study' as its immediate child with a k2 relation and 'students' is shown attached to 'to study' with a relation 'k1'. Figure 2.18 reflects the predicate argument of 'expects' and 'study'. It is important to note that Kāraka relations are **not equivalent to theta roles** (although they are mapped sometimes, for the sake of elucidation).

(VIII) Deepak expects students to study more.     (Subordinate Clauses of English)

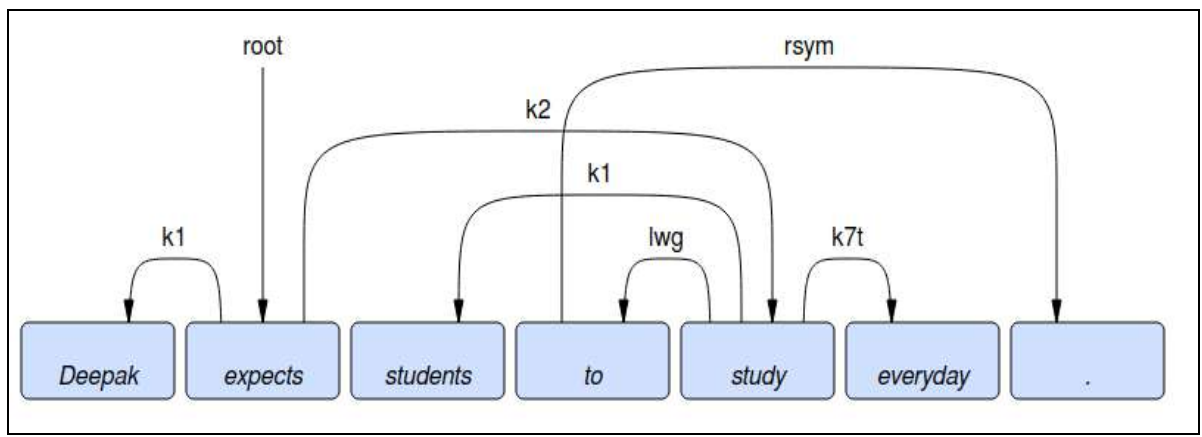

Figure 2. 18: Tree of English Subordinate Clause Example-VIII

While thematic roles are *purely semantic* in nature Kāraka relations are *syntactico-semantic*.

Extending the annotation scheme based on the CPG model to English, helps capture semantic information along with providing a syntactic analysis. The level of semantics they capture is reflected in the surface form of the sentences, and is important syntactically. Such a level of annotation makes available a syntactico-semantic interface that can be easy to exploit computationally, for linguistic investigations and experimentation. This includes facilitating mappings between semantic arguments and syntactic dependents.

> **(h) Reflexive Pronouns:** Sometimes, a reflexive pronoun is used in an appositive way to indicate that the person who realizes the action of the verb is also the person who receives the action, whereas at other times reflexives are used to emphasize the subject, and are called emphatic pronouns e.g. himself, itself etc. Our dependency analysis of the above two cases would differ per their role in the sentence, i.e. according to the relation the pronoun has with the other entities in the sentence. In the case of normally occurring reflexive pronouns such as the below in the example:

(IX) Deepak saw himself in the mirror.

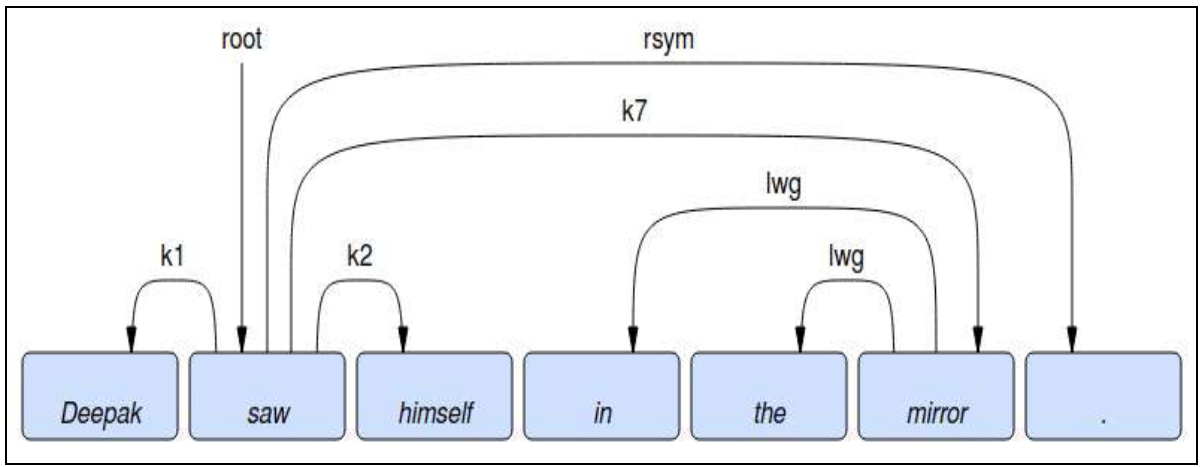

Figure 2. 19: PD Tree of English Reflexive Pronoun Example-IX

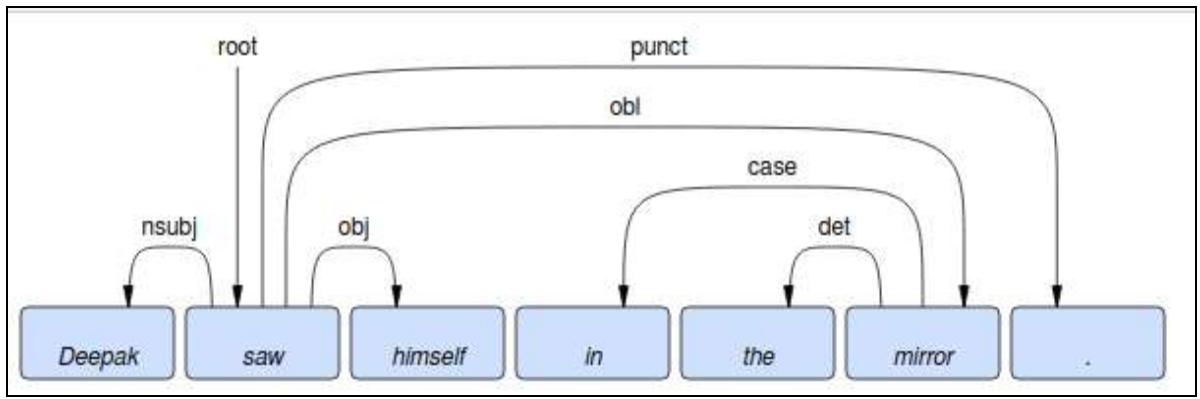

Figure 2. 20: UD Tree of English Reflexive Pronoun Example-IX

The reflexive pronoun 'himself' in this sentence will be labeled 'k2' of the verb 'saw', since it is the 'karma' of the verb. Whereas, in case of emphatic pronouns the pronoun isn't the karma of the verb, but a modifier of the noun that it goes back to. Thus, to handle emphatic pronouns, I use the dependency label nmod_emph, that makes its role in the sentence lucid. The label 'nmod_emph stands for nmod of the type emph. An example sentence for emphatic pronouns would be:

(X) The news had come out in the report of the commission itself.

As seen in Figure 2.21, the emphatic marker 'itself' is annotated nmod_emph of the noun 'commission', within the PP 'of the commission'.

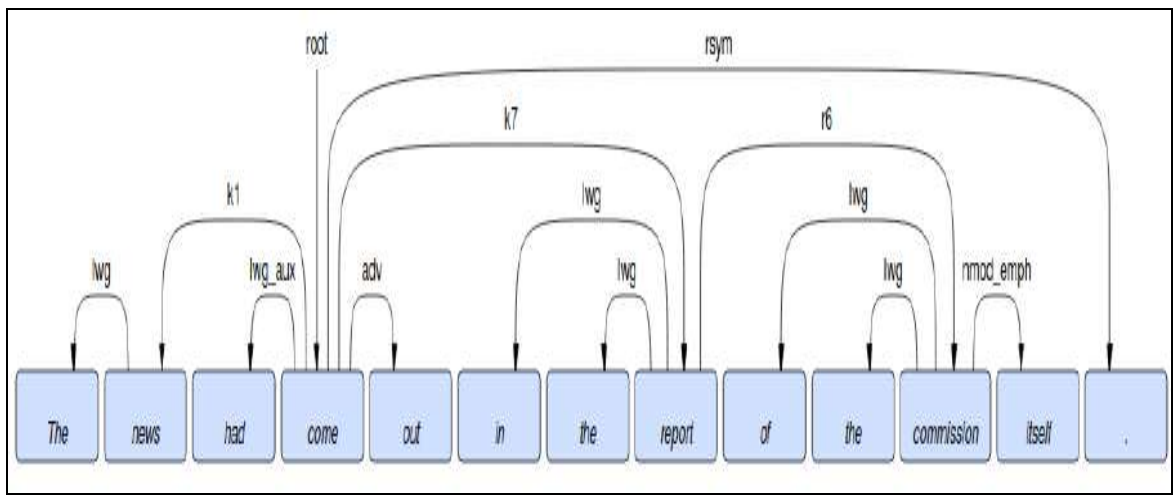

Figure 2. 21: PD Tree of English Emphatic Marker Example-X

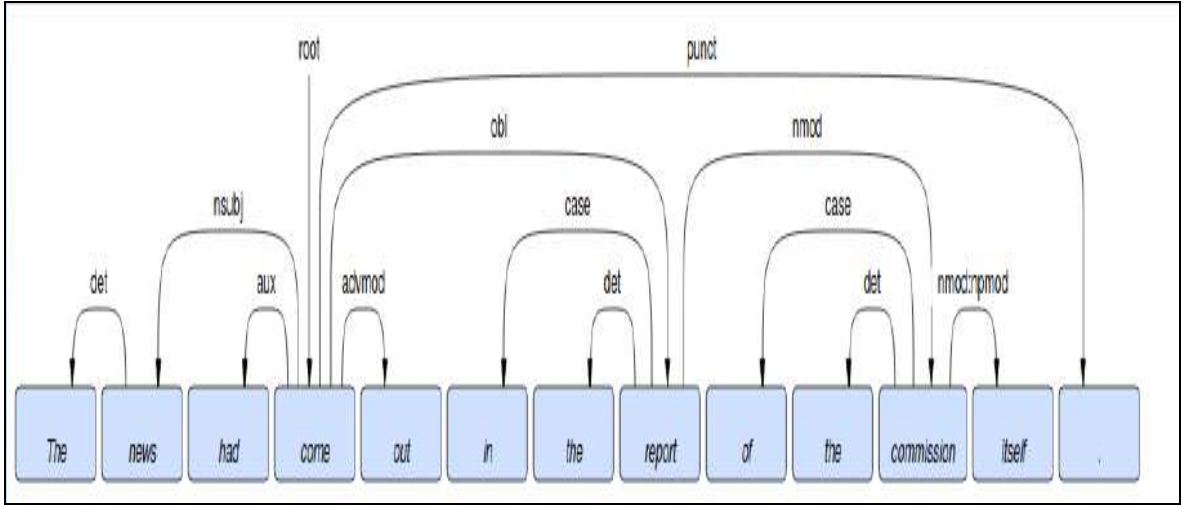

Figure 2. 22: UD Tree of English Emphatic Marker Example-X

## 2.5 Conclusion

In this chapter, I have tried to present theoretical framework of the Pāṇini's Kāraka, Kāraka based dependency model and its importance in the MT and parsing of the sentences. Another dependency model, UD, has been also introduced. The UD model follows universal cross-lingual annotation schema. Along with their description, we have also discussed a comparison between the two. In course of comparison of these above-mentioned dependency models, we observe the PD model is established/formed on the basis of syntatico-semantic level, while the UD is based on syntactic level. When we parse parallel sentences typically in MT, then PD framework is found to be more accommodating as compare to the UD to capture semantic information and does not affect its accuracy if source language and target language belong two different families or follow free-word order.

In chapter 3, these models will be used to annotate source language (English) data. The PD and UD based annotated data will be integrated into the English-Bhojpuri SMT system for building tree-to-string, syntax-directed etc. models. Detailed processes of these rules and model creations will be discussed in chapter 4. In Chapter 5, the comparative study of PD and UD based SMT systems will be discussed.

# Chapter 3

# LT Resources for Bhojpuri

Linguistic resources are crucial elements in the development of NLP applications. A corpus, thus, is a valuable linguistic resource for any language and more so if the language is endangered or less-resourced/lesser known; with the availability of a corpus a language's utility increases. A corpus can be built in any format, either text, speech, image or multimodal. It assists in creating other resources such as language technology tools, for instance, POS Tagger, Chunker, MT, TTS, ASR, Information extraction, Ontology etc. It also helps in creating resources for linguistics analysis that aids us in substantiated study of all aspects of language and linguistics e.g. grammar/syntax, morphology, phonology. It is also useful for a comparative study of languages.

Hence, in the digital era, it is vital to create text, speech and multimodal corpora and language technology tools for all the languages. With this consideration, the Ministry of Communication and Information Technology (MCIT), Government of India (now Ministry of Electronics and Information Technology) started Technology Development for the Indian Languages (TDIL) program in 1991(Das, 2008; Jha, 2010) with the aim of building linguistic resources and language technology tools for Indian scheduled languages. The TDIL has funded several projects for developing language technology tools and corpus (including text, speech and image corpus) for scheduled languages, such as, IL-MT by IIIT-Hyderabad, E-ILMT by CDAC-Pune, Gyan-Nidhi by CDAC-Noida, development of Text-to-Speech (TTS) synthesis systems for Indian languages by IIT-Madras, development of Robust Document Analysis & Recognition System for Indian Languages by IIT-Delhi, development of On-line handwriting recognition system by I.I.Sc, Bangalore, development of Cross-lingual Information Access by IIT-Bombay, and Indian Languages Corpora Initiative (ILCI) by JNU etc. Another government organization - Central Institute of Indian Languages (CIIL), Mysore, under the Ministry of Human Resource Development (MHRD) has been working to develop a corpus for scheduled languages. There are several programs for promoting and preserving Indian languages - National Translation Mission, Linguistic Data Consortium of Indian Languages (LDC-IL), Bharatavani, National Testing Mission etc.

In the above-mentioned programs, Gyan-Nidhi, LDC-IL and ILCI are some of the popular projects with primary focus on corpus creation while the remaining projects created corpus as a byproduct of a separate objective.

Gyan-Nidhi developed parallel text corpus in English and 12 Indian languages using million of pages (Shukla, 2004) from the National Book Trust, Sahitya Academy, Pustak Mahal etc.

The LDC-IL[1] was established in 2003 especially for building linguistic resources in all Indian languages. But till now it has been able to build only text and speech corpus for the scheduled languages with sources for the former ranging from printed books, magazine, newspapers, government documents etc.

The ILCI[2] (Jha, 2010; Choudhary and Jha, 2011) is the first Indian text corpora project which developed a POS annotated corpus based on the Bureau of Indian Standard (BIS) scheme and national standard format (Kumar et al., 2011). Under this project, 1,00,000 sentences (including parallel and monolingual text corpora) are created with POS and Chunk annotation (Banerjee et al., 2014 Ojha et al., 2016) for 17 languages (including 16 scheduled and English languages).

Apart from these initiatives, several universities/institutions and industries are also working to build the linguistics resources and language technological tools such as IIT-Bombay, JNU New Delhi, IIIT-Hydearbad, IIT-Kharagpur, UOH Hyderabad, IIT-BHU, Jadavpur University, Linguistic Data Consortium Pennsylvania, European Language Resources Association (ELRA), Google, Microsoft, Amazon, Samsung, Nuance, and SwiftKey etc. But, as per the author's best knowledge, there is no plan or support to create corpus for non-scheduled or closely-related languages.

This chapter further discusses corpus creation methodology, statistics and issues in text corpora for a non-scheduled language: Bhojpuri and English-Bhojpuri languages including both monolingual and parallel types of the corpus. Furthermore, it gives details of annotation for these corpora based on the Universal Dependency (UD) framework.

---

[1] See following link to know in detail : " http://www.ldcil.org/ "
[2] See following link to know in detail : " http://sanskrit.jnu.ac.in/projects/ilci.jsp?proj=ilci "

## 3.1 Related work

Despite having a substantial number of speakers and a significant amount of literature in Bhojpuri, there is little digital content available over the Internet. Hence, it is a daunting yet an essential task to create a corpus in this language. There have been a few attempts to build Bhojpuri corpus in the developments of language technological tools. However, these are not accessible by public. These works are:

- **Automatics POS Tagger:** the existing Bhojpuri POS taggers were developed on statistical approach based on SVM and CRF algorithm. The SVM-based and CRF++-based POS tagger yields 88.6% and 86.7% accuracy respectively (Ojha et al., 2015; Singh and Jha., 2015).

- **Machine-Readable dictionary**: A Bhojpuri-Hindi-English Machine-Readable dictionary was developed with 7,650 words (Ojha 2016).

- **Sanskrit-Bhojpuri Machine Translation**: A Sanskrit-Bhojpuri Machine Translation (SBMT) system was developed to translate conversational Sanskrit texts. It was trained on statistical approach using Microsoft Translator Hub and gives 37.28 BLEU score (Sinha and Jha, 2018). The authors have reported use of 10,000 parallel sentences in the development of this system.

- **Monolingual Bhojpuri corpus:** few works have mentioned that they have collected monolingual corpus used for developing LT tools such as language identification tool (Kumar et al. 2018), POS tagger (Singh and Jha, 2015) and SBMT (Sinha, 2017). Of these, only 15,000 sentences are available on the web.

However, English does not face this kind of problem. There are several open source resources available for free use for research purpose. For example: Brown, Kolhapur, Europarl Parallel corpus, WMT, OPUS subtitles, UD Tree bank, ELRA, LDC etc.

## 3.2 Corpus Building

This section elaborates on corpus building methodology for Bhojpuri and its statistics. It is divided into three subsections. Sections (3.2.1) and (3.2.2) discuss monolingual and parallel corpora creation methodology, data sources and domains and its statistics. Section (3.2.3), discusses monolingual and parallel annotated corpus.

**3.2.1 Monolingual (Bhojpuri) Corpus Creation**

To create monolingual corpus, the following approaches are followed:

**(i) Manually Created Monolingual Corpus**

Less than 3,000 sentences were collected through manual typing. Standalone version of Indian Languages Corpora Initiative Corpora Creation Tool (ILCICCT) was utilized for this purpose. This tool was developed by ILCI, JNU group under the ILCI project sponsored by DeiTY, Government of India (Bansal et al., 2013). Its process can be easily understood through Figure 3.1. In this process, each sentence was assigned a unique ID saved in UTF-8 format using the naming convention "languagename_domainname_setnumber". It allows storing detailed and accurate metadata information. For example: title, name of the book/magazine, blog/web portal (name of website), name of article, name of the author, date/year of publication, place of publishing, website URL, date of retrieved online data etc. After following these processes, data is created. Its output can be seen in the Figure 3.2.

The metadata information helps other researchers to use the corpus in better way. Hence, the basic idea of this project to maintain standard format in corpus creation and the metadata has been followed.

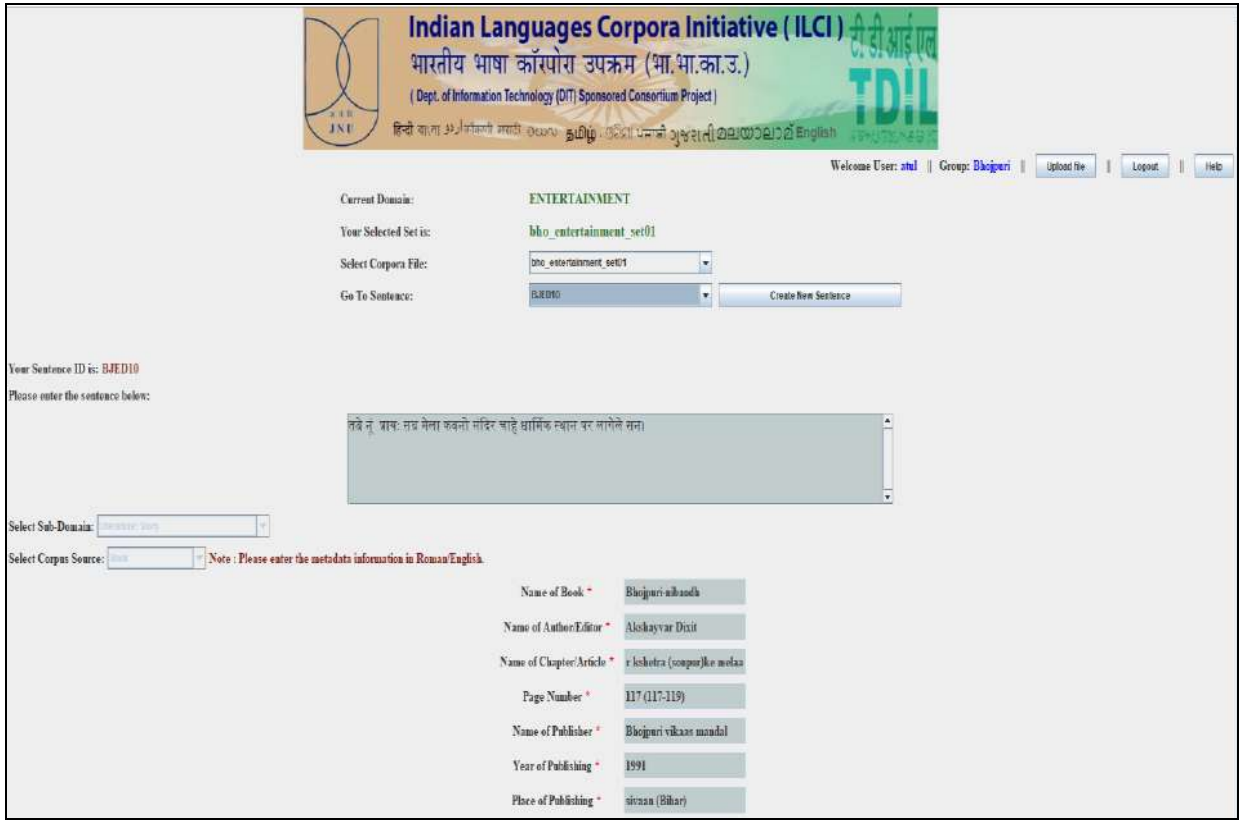

Figure 3. 1: Snapshot of ILCICCT

Figure 3. 2: Snapshot of Manually Collected Monolingual Corpus

**(ii) Semi-Automated Corpus**

Approx. 57,000 sentences were extracted from printed and digital version of several books & magazines available in Bhojpuri language. To extract these texts, semi-automated method was followed. At first, all the documents were printed and then scanned. Next, the scanned documents were run on the Tesseract[3] Hindi OCR. It gives good results in case of Hindi texts but with Bhojpuri texts its results lies between 80-85%. Its accuracy depends upon used fonts in the images as well as image quality. Finally, retrieved texts from the OCR were validated manually. Its final output can be seen in Figure 3.3 and 3.4.

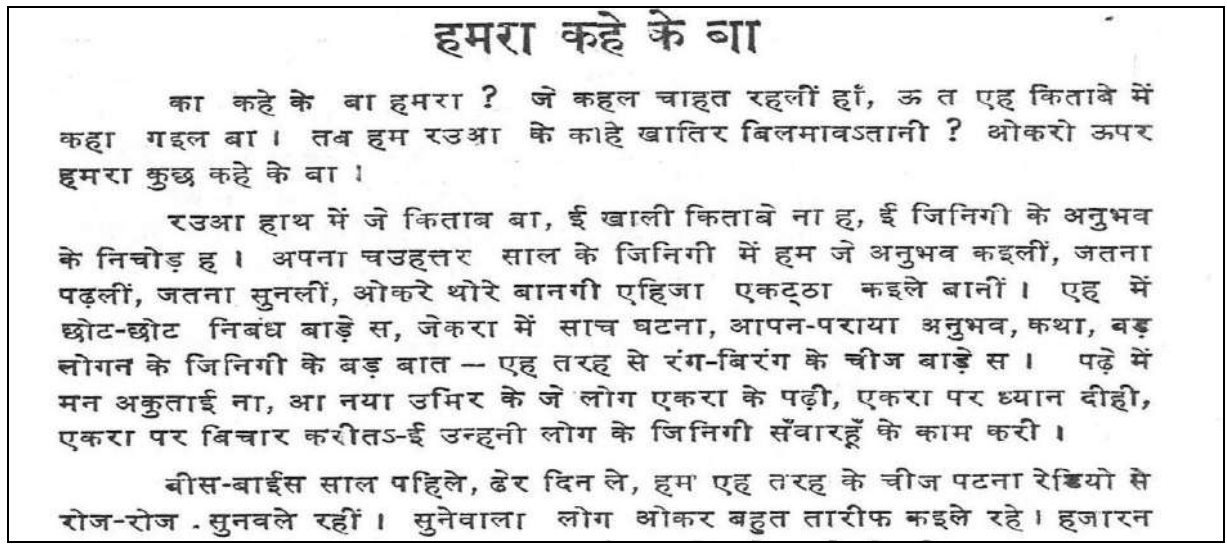

Figure 3. 3: Screen-shot of Scanned Image for OCR

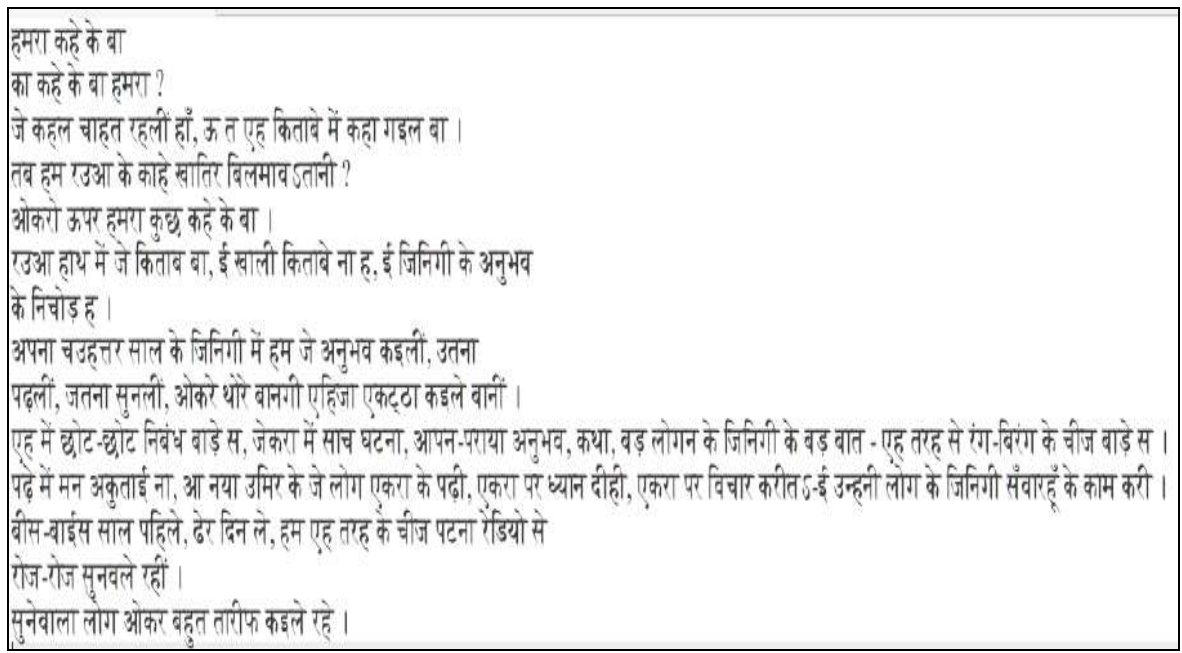

Figure 3. 4: Output of Semi-automated based Collected Monolingual Corpus

---

[3] An open source OCRengine available at https://github.com/tesseract-ocr/tesseract. It works on more than 100 Indian languages including Hindi.

### (iii) Automatically Crawled corpus

Publicly available Bhojpuri texts were collected automatically using the IL-Crawler[4] from different websites. The IL-Crawler has developed by JNU. It has built on script-based recognizer on Java using the JSOUP library. Along with the corpus, I have also saved meta-data information of corpus resources (shown in Figure 3.5). After collecting the corpus, duplicated and noisy data was deleted. Finally, the corpus was validated manually and all the data belonging to other languages was deleted. Through this method more than 40,000 sentences were created.

Figure 3. 5: Screen-Shot of Automatic Crawled Corpus

### 3.2.1.1 Monolingual Corpus: Source, Domain Information and Statistics

With the use of above methodologies, a corpus of 1,00,000 Bhojpuri sentences were developed which contains 16,16,080 words. This corpus was collected from various domains such as literature, politics, entertainment and sports using various sources. For example: Online newspaper, Blogs, printed books, magazines etc.

Its detailed statistics and source information are elaborated in the Table 3.1. These sources provide an exhaustive coverage of all kinds of language use found in Bhojpuri. For example, newspapers provide a more formal use of the language while books, blogs, and magazines provide a more colloquial use.

---

[4] http://sanskrit.jnu.ac.in/download/ILCrawler.zip

## 3.2.2 English-Bhojpuri Parallel Corpus Creation

To build this parallel corpus, English data was collected as a source language using the following:

- 30,000 sentences were taken from English grammar, English learning e-books and story books etc. (Nainwani, 2015). This source was created for development of English-Sindhi SMT system by Pinkey Nainwani at JNU, New Delhi for her PhD work.
- 33,000 sentences were used from the OpenSubtitles[5]

| Corpus source | Corpus source information | Sentences | Words | Characters |
|---|---|---|---|---|
| Books | bhojpuri nibandh | 60,000 | 10,38,202 | 50,78,916 |
| | tin nAtak | | | |
| | jial sikhiM | | | |
| | rAvan UvAch | | | |
| | bhojpuri vyakarana | | | |
| Magazines | pAti | | | |
| | Parikshan | | | |
| | Aakhar | | | |
| | samkAlin bhojpuri sAhitya | | | |
| Web-sources | Anjoria | 40,029 | 5,77,878 | 28,06,191 |
| | tatkaa Khabar | | | |
| | bhojpuria BlogSpot | | | |
| | Dailyhunt | | | |
| | Jogira | | | |
| | pandjiblogspot | | | |
| | manojbhawuk.com | | | |
| **Total number of sentences, words and characters** | | 1,00,029 | 16,16,080 | 78,85,107 |

Table 3. 1: Details of Monolingual Bhojpuri Corpus

---

[5] OPUS subtitles - http://opus.nlpl.eu/

These sentences belong to entertainment and literature domains. Once the source corpus was compiled, it was manually translated to Bhojpuri (target language) following the ILCI project translation guidelines (samples are translated sentences in Figure-3.6). According to the guidelines "The Structural and Aesthetic is considered to be the default type, since the translation has to be a balance of both" (ILCI Phase-2 Translation Guideline, 2012). During the parallel corpus creation, an attempt was made to capture and maintain different varieties of Bhojpuri to maintain a generic system.

```
At which position do you work ?          कवने पद पर तु काम करल ?
Which book do you not know to whom to give ?   कवन किताब तु नाई जानलअ जवन देवल ?
Which book will you take ?    कवन किताब तु लेबअ ?
Which book are you reading ?   कवन कितब तु पढत हउअ ?
What book do you think you can give to John ?  का तु किताब जान के दे सकल ?
What book have you read recently ?   कवन किताब तु हाल ही मे पढले हउअ ?
Which don't you know who bought ?   का तु नाइ जानत हउअ कि के खरिदले ?
Which songs do you like best?   कवन गीत तोहके बढिया लागल ?
Which graph are you going to use?   कवन गराफ तु काम मे लावे जात हउअ ?
Which juice do you prefer, Orange and Apple?   कवन जुस तु पसन्द करत हउअ, सन्तरा आकि सेव ?
Which fruit do you like?    कवन फल तु पसन्द करत हउअ ?
What colour do you like?    कवन रन्गा तु पसन्द हउअ ?
Which colour do you prefer, red or blue?   कवन कलर तोहके पसन्द हउए, लाल कि नीला ?
Which will you take?    तु का लेब ?
Which one will you buy ?  तु एकगो क खरीदब ?
What subject do you not like?   तु कवन विषय नाइ पसन्द करल ?
Which subject have you chosen ?    तु कवन विषय चुन भइल ?
Which subject do you teach ?   तु कवन विषय पढावतड ?
Which subject does you like better, Art or Music?    तु कवन विषय बढिया से पसन्द करल कला कि म्युजिक ?
```

Figure 3. 6: Sample of English-Bhojpuri Parallel Corpus

For this purpose, Bhojpuri speakers from various regions (Purvanchal region of Uttar Pradesh and western part of Bihar) have been chosen to translate the source sentences. This exercise was done because there is no standard grammar or writing style in this language. Also, there are various varieties at spoken level as well as written texts.

Out of the manual translation, 2,000 parallel sentences were collected through printed books and Bhojpuri Wikipedia page. Approx. 1,100 aligned parallel sentences were extracted from the *Discover the Diamond in You* written by Arindam Chaudhari and translated in Bhojpuri by Onkareshwar Pandey. Approx. 800 sentences were taken from *Sahaj Bhojpuri Vyakaran: An Easy Approach to Bhojpuri Grammar* and *Bhojpuri-Hindi-English lok shabdkosh dictionary* written or edited by Sarita Boodhoo and Kumar Arvind respectively. Rest of the data was collected from the भोजपुरी and *Bhojpuri language* Wikipedia pages.

| Types of the Corpus | Sentences | Words | Characters |
|---|---|---|---|
| **English-Bhojpuri** | 65,000 | 4,40,609 | 23,29,093 |
| | | 4,58,484 | 21,17,577 |

Table 3. 2: Statistics of English-Bhojpuri Parallel Corpus

As previously mentioned, there is no English-Bhojpuri parallel corpus available publicly or on the Internet. This is the first work in this language pair where 65,000 parallel sentences have been created which containing 4,40,609 and 4,58,484 words in English and Bhojpuri respectively.

**3.2.3 Annotated Corpus**

Both, monolingual corpus (Bhojpuri) consisting of 1,00,000 sentences and parallel corpus (English-Bhojpuri) consisting of 65,000 sentences, have been POS annotated. English data, constituent of parallel data, has also been annotated as per the UD and PD framework. UD[6] is a project to build Treebank annotation for many languages which is cross-linguistically consistent. Its annotation scheme is formed of universal Stanford dependencies (de Marneffe et al., 2006, 2014), the Interest Interlingua for morpho-syntactic tagset (Zeman, 2008) and Google universal part-of-speech tags (Petrov et al., 2012). The Treebank includes the following information in CONLLU format: lemma, universal part-of-speech tags (UPOS), language-specific part-of-speech tag (XPOS), morphological features, syntactic dependency relations etc. But in these lemmas and XPOS are optional features. If it does not exist in any language, then annotator will mark that field with an underscore (_) symbol. As mentioned in the previous chapter, it has 17 tags for UPOS, 23 tags for morphological features based on lexical and inflectional categories and 37 tags for syntactic dependency relations (Zeman et al., 2018).

The Bhojpuri UD tagset[7] has been prepared on the basis of UD annotation guidelines (Zeman et al., 2017) and the English UD tagset has been followed for English language. In Bhojpuri, X-POS tags were annotated using the BIS tagset[8]. 5,000 Bhojpuri monolingual sentences were annotated manually except the XPOS tags. The XPOS tag was automatically annotated by CRF++ based Bhojpuri POS tagger (Ojha et al., 2015).

---

[6] http://universaldependencies.org/
[7] see the following link: https://github.com/UniversalDependencies/UD_Bhojpuri-BHTB
[8] This BIS tagset was released in 2010 as generic tagset for annotating corpus in Indian languages. The related document can be accessed at http://tdil-dc.in/tdildcMain/articles/134692Draft%20POS%20Tag%20standard.pdf

During the process of manual annotation, XPOS tags were validated (shown in the Figure 3.7) using Webanno tool (Eckart, 2016).

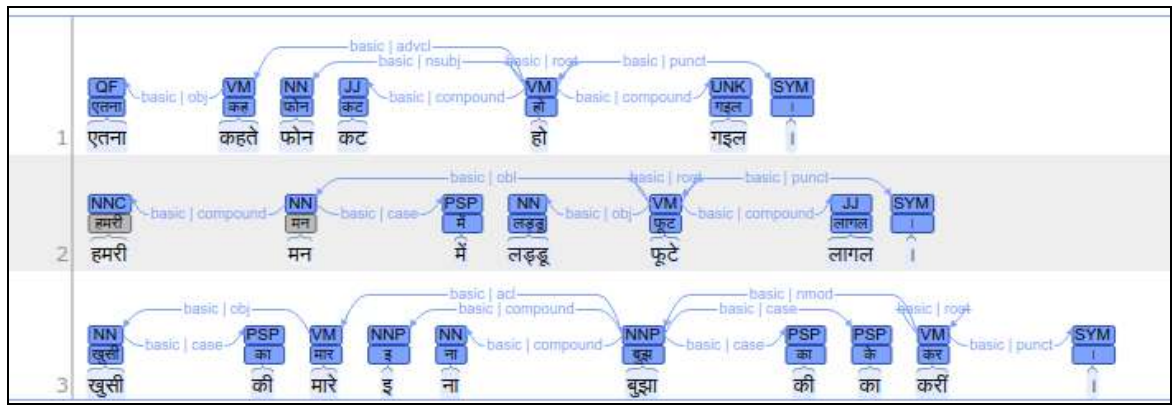

Figure 3.7: Screen-Shot of Dependency Annotation using the Webanno

However, the English-Bhojpuri parallel corpus (5000 sentences) was annotated (shown in the Figure 3.8 and 3.9). 50,000 English sentences were tagged on UD model using UDPipe (Straka et al., 2016) pipeline through English language model trained under the *CoNLL 2017 Shared Task* (Straka 2017). The annotated data was duly validated. The same data was also annotated on PD model.

So, in the project, I have built a total of 1,00,000 and 5,000 annotated sentences in monolingual and parallel corpus respectively, while 50,000 English sentences were annotated at UD and PD levels (see chapter 2 for detailed information).

```
# sent_id = 135
# text = Anita and Ravi came yesterday
1   Anita     Anita     NOUN   NN   Number=Sing   4   nsubj   _   _
2   and       and       CCONJ  CC   _             3   cc      _   _
3   Ravi      Ravi      PROPN  NNP  Number=Sing   1   conj    _   _
4   came      come      VERB   VBD  Mood=Ind|Tense=Past|VerbForm=Fin   0   root   _   _
5   yesterday yesterday NOUN   NN   Number=Sing   4   obl:tmod   _   SpaceAfter=No
6   .         .         PUNCT  .    _             4   punct   _   SpaceAfter=No
```

```
# sent_id = 135
# text = अनिता अउर रवि काल्हि आइल रहलन ।
1   अनिता    अनिता   PROPN  N_NNP  Case=Nom|Gender=Fem|Number=Sing|Person=3   5   nsubj   _   _
2   अउर      अउर     CCONJ  CC_CCD  _                                          3   cc      _   _
3   रवि      रवि     PROPN  N_NP   Case=Nom|Gender=Fem|Number=Sing|Person=3   1   conj    _   _
4   काल्हि   काल्हि  NOUN   N_NST  Case=Nom|Gender=Masc|Number=Sing|Person=3  5   obl     _   _
5   आइल      आ       VERB   V_VM   Aspect=Perf|Gender=Masc|Number=Plur|VerbForm=Part|Voice=Act   0   root   _   _
6   रहलन     रहल     AUX    V_AUX  Gender=Masc|Mood=Ind|Number=Plur|Tense=Past|VerbForm=Fin   5   aux   _   _
7   ।        ।       PUNCT  SYM    _                                          5   punct   _   SpaceAfter=No
```

Figure 3.8: Snapshot of Dependency Annotated of English-Bhojpuri Parallel Corpus

```
# sent_id = 135
# text = Anita and Ravi came yesterday
1   Anita    Anita     PROPN  NNP  Number=Sing                         4   nsubj       _   _
2   and      and       CCONJ  CC   _                                   3   cc          _   _
3   Ravi     Ravi      PROPN  NNP  Number=Sing                         1   conj        _   _
4   came     come      VERB   VBD  Mood=Ind|Tense=Past|VerbForm=Fin    0   root        _   _
5   yesterday  yesterday  NOUN  NN  Number=Sing                        4   obl:tmod    _   SpaceAfter=No
```

Figure 3.9: Snapshot of after the Validation of Dependency Annotated of English Sentence

## 3.3 Issues and Challenges in the Corpora building for a Low-Resourced language

Several problems were encountered while build these resources. For example, lack of digital content in Bhojpuri, vast variation in Bhojpuri language in both spoken and writing styles, capturing structural variation in building parallel corpus, problems with Bhojpuri corpus annotation etc. But here, only major issues will be discussed especially with the creation of Bhojpuri monolingual, parallel and annotation of the corpus. These major problems are:

a) **OCR-based extracted text:** The Hindi OCR tool generated several errors during the extraction of Bhojpuri texts which took much more time in data validation. Most of the times, the tool failed to produce correct output under these circumstances: words and numbers written in bold/italic styles, characters in large font size, word not belonging to Hindi language, the use of 'ऽ' (Avagrha) symbol and similar looking characters. The Figure 3.10 (extracted from the Figure 3.3 image) and Table 3.3 demonstrate these errors. First two rows of the Table3 are examples of bold/italic or font size error. Next four rows of the table are examples of similar looking characters that don't belong to Hindi. Next and last row of the table are examples of inability to recognize 'ऽ' (Avagrha) symbol. These problems were faced in all texts generated with the use of OCR[d].

Such issues may have cropped up because the OCR was trained on Hindi corpus. Hence, Bhojpuri words or character appeared Out of Vocabulary (OOV) or Character (OOC). It could be solved by mapping with a Bhojpuri dictionary or large size of monolingual corpus. But as mentioned earlier there is no Bhojpuri dictionary of a good size nor a large corpus is available in the language.

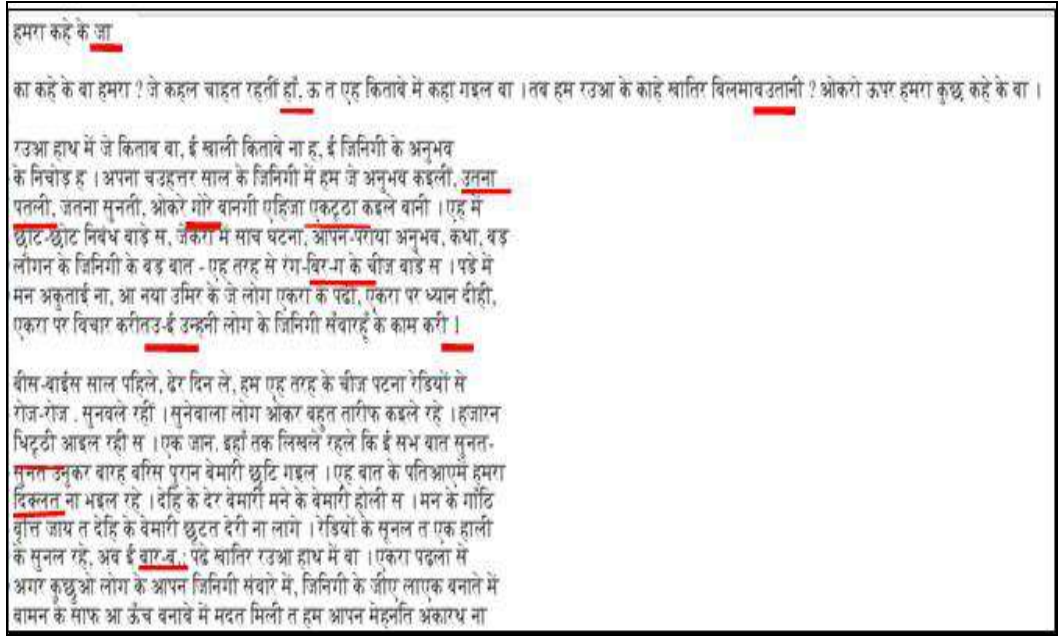

Figure 3.10: Sample of OCR Errors

| Actual styles of text or text in the Image | OCR Output | Desired output |
|---|---|---|
| *भोजपुि‍रया समाज* | *जपुि‍रया समाज* | *भोजपुि‍रया समाज* |
| *बाा ।* | *वाा 1* | *बाा ।* |
| *एकट्टा । लोगन* | *एकट्ठा लौगन* | *एकट्टा । लोगन* |
| *थोरे* | *गोरे* | *थोरे* |
| *पढ़ी /पढ़े /बाड़े* | *पढी पडो बाडे* | *पढ़ी पढ़े बाड़े* |
| *रहलीं सुनलीं* | *रहतीं सुनती* | *रहलीं सुनलीं* |
| *ि‍बलमावऽतानी। करीतऽ– ईं हऽ* | *ि‍बलमावउतानी । करीतउ-ई हइ* or *ह* | *ि‍बलमावऽतानी । करीतऽ–ई । हऽ* |

Table 3.3: Error Analysis of OCR based created corpus

b) **Issues with Automatic Crawled Bhojpuri Corpus**: Two major problems were faced during the cleaning and validation of automatically crawled Bhojpuri corpus:

**(i) Variation in Character Styles**: Variation in the use of sentence end markers was a challenge. Some texts used Roman full stop ("."), while some made use of a vertical bar "|" and others used the Devanagri full stop "।" symbol. Varieties in the use of Avagrha ('ऽ') was also found. Some content writers or authors were used roman "S" character instead of Devanagri 'ऽ' character. For example: करS, हS, जइबS etc. should have been written as करऽ, हऽ, जइबऽ. This happened because there are no standard guidelines for writing styles of Bhojpuri texts either for websites or printed material. Hence, content writers or bloggers relied on the available resource and the ease of typing.

एकर कवनो विकल्प नइखे ।
पढ़ल कबो बेकार ना जाला ।
नाजायज तरीका से चोरी से नम्बर लेआ देला से कवनो फायदा नइखे ।
उ त अपने आप के ठगल बा ।
सोचे के बात बा कि अइसे कइला से आखिर के ठगाता ।
डा० अमरेन्द्र मिश्र एक संक्षिप्त परिचय जुलाई के जन्म ।
गाँव सोनवर्षा थाना शाहपुर जिला भोजपुर बिहार ।
बेयालिस बरिस सांख्यिकी में शिक्षण कइला के बाद पटना विश्वविद्यालय से में सेवानिवृत ।
शिक्षण के अतिरिक्त पटना विश्वविद्यालय में निदेशक हैं ।
दूर शिक्षा निदेशालय अध्यक्ष सांख्यिकी विभाग निदेशक जनसंख्या शोध केन्द्र निदेशक एकेडेमिक स्टाफ कालेज संकायाध्यक्ष विज्ञान संकाय आदि अनेक पदन पर योगदान ।
आकाशवाणी पटना से अनेके भोजपुरी कहानी के प्रसारण ।
कविता संग्रह अंजुरी भर मेरी भी अवरू एक भोजपुरी कहानी संग्रह आखिर के ठगाता प्रकाशित ।
वर्तमान पता प्रेमचन्द पथ राजेन्द्र नगर पटना मोबाइल नम्बर ईमेल ।
डा० अमरेन्द्र मिश्र के धन्यवाद ।
बहुत ही शिक्षापुरद ।
अभिव्यक्ति बेमिसाल ।
भाषा काफी सटीक ।
लेखक को मेरी बधाई ।
भोजपुरी में ऐसी कहानियाँ कम पढ़ने को मिलती हैं ।
इ कहानी से हमहन के जरूर सिख लेवे के चाही ।
परणाम गुरू जी ।

Figure 3.11: Automatic Crawled Bhojpuri Sentences with Other language

**(ii) Identification and cleaning of Other Language Sentences:** During the automatic crawling of Bhojpuri corpus, Hindi and Sanskrit language sentences were also crawled (shown in the Figure 3.11). This is because the crawler crawls data on the basis of Character encoding (Charset) of the language which for the present purpose is Bhojpuri. Thus, the data validation process (which includes identification and removal of non-Bhojpuri sentences) got further complicated.

**c) The Problems of Equivalences in English-Bhojpuri Parallel Corpus**

In building the parallel corpus, some sentences were given to translators to translate from source language sentences. But several variations were found in the translated sentences owing to the judgment used by different translators. As previously mentioned, for capturing and studying of these variations, the same source sentences were assigned to different translators from different regions. Several equivalences issues were found in the target (Bhojpuri) language i.e. writing styles, lexical/word and phrase levels.

    i. **Issues in Writing Style:** The translators have followed their own writing styles. For example: English word "*is*" and "*come*" should be translated to '*हऽ*' "*आइब*" in Bhojpuri respectively. However, it was found that sometimes translator translates as '*ह*' and "*आइब*" word and sometimes as '*हऽ*' and "*आइब*". As previously mentioned, it happened due to lack of standard writing style and awareness between translator with written texts, thus creating more ambiguities to align parallel words and sentences.

    ii. **Issues with Word/lexical and Phrase levels:** Several varieties in Bhojpuri translation were found as Bhojpuri is spoken differently in different regions which were clearly reflected in the translated sentences. In the Figure 3.12, the Translator-1 belongs to Gorakhpur (Uttar Pradesh) and the Translator-2 belongs to Aara (Bihar).

| Source | I do this work . |
|---|---|
| Translator-1 | हम इह काम करेनी |
| Translator-2 | इ काम हम करींऽल |

| Source | I can do this work . |
|---|---|
| Translator-1 | हम इह काम कइ सकेनी |
| Translator-2 | हम इ कइ सकींऽल |

| Source | I did it ! |
|---|---|
| Translator-1 | हम इह कइ लेहनी |
| Translator-2 | हम कइलीं |

| Source | I did it because he told me to . |
|---|---|
| Translator-1 | हम इह कइ लेहनी काहे कि उ हमके बतवले रहने |
| Translator-2 | हम कइलीं काहे से उ हमसे बतइले रहऽल |

| Source | I did it because I wanted to . |
|---|---|
| Translator-1 | हम इह कइ लेहनी काहे कि हम चाहत हइ |
| Translator-2 | हम कइलीं काहे से हम चाहत रहलीं |

| Source | I bought it approximately a week ago . |
|---|---|
| Translator-1 | हम शायद एक हप्ता पहिले खरिदले हई |
| Translator-2 | एके लगभग एक हप्ता पहले हम खरीदले रहलीं |

| Source | Whom shall I this give to ? |
|---|---|
| Translator-1 | हम केकरे खरतिन इ छोड देब |
| Translator-2 | इ हम केके देब |

| Source | I suggest that this rule be changed . |
|---|---|
| Translator-1 | हमार सलाह हउए कि इ नियम के बदल देहल जाय |
| Translator-2 | इ हमार सुझाव ह कि नियम बदल सकऽल |

Figure 3.12: Comparison of Variation in Translated Sentences

In (1) and (2), the translated sentences by two translators demonstrate variations at the word level. Here "do" and "this" of source sentences were given different translation equivalents in the target language as translator-1 "करेनीं" and "इह" translator-2 "करीऽल" and "इ". In (8), we can see the verb source sentence, was translated as a group of three words (बदल देहल जाय) by translator-1 and two words (बदल सकऽल) by translator-2. This created problem in maintaining equivalences and building a standard parallel corpus.

The issues presented in a), b) and c) will affect the development of the LT tools and their accuracy. Hence it is required to reduce these problems further.

# Chapter 4

# English-Bhojpuri SMT System: Experiments

The chapter describes in detail the various methods and rationale behind experiments conducted for the development of an English-Bhojpuri SMT (EB-SMT) system where English is the source language while Bhojpuri, the target language. As mentioned before, there has been no such research conducted on English-Bhojpuri based SMT.

Several experiments were conducted to build a robust SMT system for the English-Bhojpuri language pair, designed using various training models and parameters, namely, Phrase-based (Koehn, 2003), Hierarchical Phrase-based (Chiang, 2005; Chiang, 2007), Factor-based (Koehn and Hoang, 2007; Hoang, 2011), Dependency Tree-to-String (Liu and Glidea, 2008; Venkatapathy, 2010; Graham, 2013; Li, 2013; Williams et al., 2016), Phrase-based reordering, Lexicalized reordering, and Hierarchical-reordering, etc. All the training models, except Factor-based, were trained on three different Language Model (LM) toolkits: IRSTLM (Federico et al., 2007), SRILM (Stolcke, 2002) and KenLM (Heafield, 2011). The Factor-based SMT (FBSMT) was trained on KenLM and SRILM only. The reason behind using different LM toolkits was to validate their suitability for a low-resourced Indian language, in this case, Bhojpuri.

As previously iterated, one of the main objectives of this research is to check compatibility between the PD and UD framework, and the SMT model for English which belongs to a language family different than Bhojpuri. To achieve this objective, other SMT experiments were conducted using the PD and UD-based English annotated data, trained on the Dependency Tree-to-String (Dep-Tree-to-Str) model. The Dep-Tree-to-Str based EB-SMT system has been built on the KenLM toolkit. The Moses toolkit (Koehn et al., 2007) has been used for experiments of these SMT systems.

This chapter is divided into six sections to enable clarity while the process of the experiment is explained. The first section elaborates the Moses toolkit and its workflow and the second section provides information of the training and development data statistics. The second section also discusses the pre-processing procedures that have been used for experiments. The third section deals with the development of EB-SMT systems, further divided into sections elaborating different steps and models used in building of the systems. The fourth section discusses the results of experiments on the scale of BLEU evaluation metrics. The fifth section discusses the outline of EB-SMT system's web interface. The sixth and last section sums up the chapter by providing concluding remarks.

## 4.1 Moses

An open source SMT toolkit, the Moses enables automatic training of statistical machine translation models and boasts of being able to work for all natural language pairs (Koehn et al., 2007). Its inception took place at the University of Edinburgh, but it was further enhanced during a summer workshop held at Johns Hopkins University (Koehn, 2010). Moses succeeded Pharaoh, another SMT toolkit developed at the University of Southern California (Koehn, 2004), and is now the most popular Phrase-based SMT framework. There are two main components in the Moses, training pipeline and decoder with training, tuning, and pre-processing tools as additions to the decoding tool. The procedures involved in training Moses[1] can be easily understood through a workflow provided in figure 4.1.

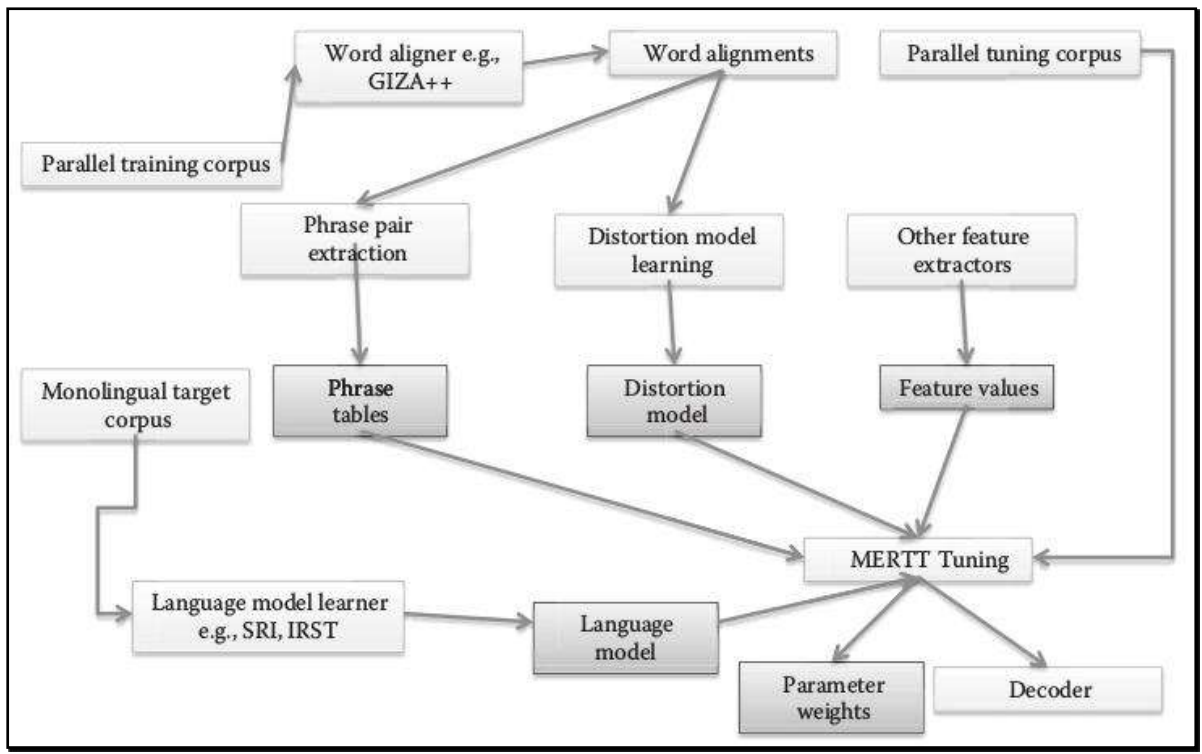

Figure 4. 1: Work-flow of Moses Toolkit (adopted from Bhattacharya, 2015)

---

[1] detailed information of this tool can be found at "http://www.statmt.org/moses/"

Fundamentally, the training pipeline is an amalgamation of various tools[2] which picks up the raw data (both, parallel and monolingual) and transforms it into a model[3] for machine translation. For example:

- KenLM, SRILM for language model estimation
- GIZA++ (Och and Ney, 2003) for word alignments of the parallel corpus
- Tokenizer, true-caser, lower-casing and de-tokenizer for pre-processing and post-processing of the input and output data.

After a system is trained, a well-crafted search algorithm of trained model immediately identifies the translation with the highest probability among an exponential number of choices available. The decoder, on the other hand, with the assistance of trained machine translation model, provides a translation of source sentence into the target language.

A great feature of Moses is that in addition to supporting Phrase-based approach it also supports Factored, Syntax-based or Hierarchical phrase-based approaches.

## 4.2 Experimental Setup

- **Data Size**

For the purpose of experimentation, different sizes of data have been used in the development of the PBSMT, HPBSMT, FBSMT and Dep-Tree-to-Str based EB-SMT systems. The translation model of all the systems, except Dep-Tree-to-Str System, is trained on 65000 English-Bhojpuri parallel sentences while the translation model of Dep-Tree-to-Str system is trained on 50000 parallel sentences. In the set with 50000 parallel sentences, only source sentences are annotated with the PD and UD dependency framework which integrates in the Dep-Tree-to-Str model. The parallel data is further divided into training, tuning and testing sets (detailed statistics are described in Table 4.1). For all the EB-SMT systems, 100000 monolingual sentences of the target language have been used to build the required LM model.

---

[2] Scripted in "Perl and C++"
[3] The code can be accessed from: "https://github.com/moses-smt/mosesdecoder"

| Types of Data | Training | Tuning/ Dev | Testing |
|---|---|---|---|
| English-Bhojpuri parallel sentences (both raw and POS annotated ) | 63000 | 1000 | 1000 |
| PD and UD based annotated English sentences | 48000 | 1000 | 1000 |
| Monolingual sentences (both raw and POS annotated) | 100000 | - | - |

Table 4. 1: Statistics of Data Size for the EB-SMT Systems

- **Preprocessing for Training the EB-SMT System**

For the scope of this work, several pre-processing steps were performed. Firstly, both types of corpora were tokenized and cleaned (removing sentences of length over 80 words). The true-casing and CoNLL-based dependency format data into Moses XML format of the English representation of the corpora was done next. Both these processes were performed using Moses scripts. The next step was tokenization of Bhojpuri data. For the pre-processing of the target language data, conversion script was used to convert POS annotated data into Moses format and tokenizer was used to ensure the canonical Unicode representation. The conversion script and tokenizer were developed by the author and the SSIS, JNU team.

## 4.3 System Development of the EB-SMT Systems

This section is arranged in five sub-sections with the objective of enumerating all the steps included in conducting experiments to build the EB-SMT systems. The first three sub-sections provide a detailed description of training parts in developing the systems. The remaining sub-sections explain procedures of testing and post-processing steps.

### 4.3.1 Building of the Language Models (LMs)

The LM is an essential model in the process of building any SMT system. It is used to capture fluency of the output similar to the native speaker's output in the target language. In short, it helps to understand syntax of the output (target) language. Therefore, it is built using the monolingual data of the target language. In the case of English-Bhojpuri language pair-based SMT system the LM is created for Bhojpuri.

In LM, a probabilistic language model $p_{LM}$ should be able to prefer the correct word order over an incorrect word order. For instance,

$$p_{LM} \text{ (इ घर छोट हऽ)} > p_{LM} \text{ (छोट घर हऽ इ)} \tag{4.1}$$

The above example reveals that the likelihood of a Bhojpuri speaker uttering the sentence इ घर छोट हऽ (i choTa ghara haS) is more than the sentence छोट घर हऽ इ (choTa ghara haS i). Hence, a good LM, $p_{LM}$ will assign a higher probability to the first sentence. Formally, an LM is a function that picks a Bhojpuri sentence and returns the higher probability to the sentence which should be produced by a Bhojpuri speaker.

LM also aids an SMT system to deduce the right word-order in the context. If an English word 'spoke' refers to multiple translations in Bhojpuri such as बतवनीं (batavanIM), बोललस (bolalasa), and कहलीं (kahalIM); an LM always assigns a higher probability to the more natural word choice in accordance with the context. For example:

$$p_{LM} \text{ (हम उनके माई के बतवनीं)} > p_{LM} \text{ (तू समझलू ना हम का कहलीं)}$$

The above example shows that कहलीं co-occurs most with the conjunction as compared to the word बतवनीं.

The language modeling methodology consists of n-gram (briefly explained below) language models (LMs), smoothing and back off methods that address the issues of data sparseness, and lastly the size of LMs with limited monolingual data.

- **N-Gram LMs**

N-gram language modeling is a crucial method for language modeling. N-gram LMs measure how likely is it for words to follow each other. In language modeling, the probability of a string is computed using the equation,

$$W = w_1\, w_2\, w_3\, w_4\, w_5 \ldots\ldots\ldots w_n \tag{4.2}$$

The statistical method to compute p(W) is to count how often W occurs in a monolingual data-set, however, most long sequences of words do not occur so often in the corpus. Therefore, the computation of p(W) needs to break down in several smaller steps aiding in collection of sufficient statistics and estimation of probability distributions. In sum, n-gram language modeling breaks up the process of predicting a word sequence *W* into predicting one word at a time. Hence, the actual number of words in the history is chosen based on the amount of monolingual data in the corpus. Larger monolingual data allows for longer histories. For instance, trigram language models consider a two-word history to predict the third word; 4-grams and 5-grams language models are used in a similar vein. This type of model steps through a sequence of words and consider for the transition only a limited history is known as a Markov Chain. The estimation of trigram language model probability is computed as follows:

$$p(w3|w1, w2) = \frac{count\ (w1,w2,w3)}{\sum w\ count\ (w1,w2,w3)} \qquad (4.3)$$

Hence, it counts how often in a corpus, the sequence $w_1$, $w_2$, $w_3$ is followed by the word $w_4$ in comparison to other words (there are several related works available to understand the LM modeling in details like Och and Ney, 2003; Koehn, 2010; Jurafsky and Martin, 2018).

- **Training of the LMs**

Three LM toolkits (SRILM, IRSTLM and KenLM) based on the following methodologies have been used to build LMs.

(a) SRILM, based on 'TRIE', used in several decoders (Stolcke, 2002).
(b) IRSTLM, a sorted 'TRIE', implementation designed for lower memory consumption (Federico et al., 2008; Heafield, 2011).
(c) KenLM, uses probing and TRIEs which renders the system faster (Heafield, 2011).

100000 tokenized monolingual sentences have been used to train the LMs which train on 3, 4 and 5-gram orders with the IRSTLM, SRILM and KenLM respectively using modified Kneser-Ney (Kneser and Ney, 1995) smoothing and interpolate method (Chen and Goodman, 1998). The above-mentioned n-grams orders were chosen because of their better performance in comparison with other numbers in the previous experiments. While creating models, the number of unigram tokens remained the same irrespective of n-grams orders except for the SRILM (detailed statistics of tokens are demonstrated in the Table 4.2). The KenLM model makes higher numbers of tokens on 3-gram orders as compared to other two while the SRILM makes lowest number. The LMs are then converted into binary language model (BLM) to improve speed.

| N-gram (order) | IRSTLM | SRILM | KenLM |
|---|---|---|---|
| **N-gram-1** | 70318 | 70317 | 70318 |
| **N-gram-2** | 474786 | 474784 | 474785 |
| **N-gram-3** | 182943 | 171200 | 841129 |
| **N-gram-4** | - | 153717 | 933978 |
| **N-gram-5** | - | - | 918330 |

Table 4. 2: Statistics of the Bhojpuri LMs

More details of integration are discussed in sections describing translation models and decoding.

### 4.3.2 Building of Translation Models (TMs)

In most instances, alignment is the first task of any SMT building process meant to identify translation relationship among the words or multiword units in a bitext (bilingual text). Word alignment is usually the first step because creating word-based models helps in phrase extraction, building of phrase-based[4] models etc. The word-based model could be lexical translation models (Koehn, 2010).

- **Methodology of Word Alignment and Word-Based Models**

This method can be better understood by looking at the examples provided below. For instance, in a large-sized corpus of English-Bhojpuri, one could count how often 'good' is translated into each of the given choices.

| Translation of English in Bhojpuri | Total occurrence |
|---|---|
| बढ़िया | 138 |
| नीक | 145 |
| अच्छा | 172 |
| निम्मन | 73 |
| ठीक | 7 |

Table 4. 3: Statistics of 'good' words Translation in English-Bhojpuri Parallel Corpus

---
[4] The meaning of *phrase* here is not similar to that in linguistics. Here, it refers to a combination of more than one word, in short, any multiword units.

The frequency of occurrence for the word 'good' is 555 in English-Bhojpuri corpus. This word is translated 172 times into अच्छा (acchA), 145 times into नीक (nIka), 138 times into बढ़िया (ba.DhiyA) and so on. These counts help in making an estimate of a lexical translation probability distribution which will further help in translating new English text. To put this function formally:

$$p_e : b \rightarrow p_e(b) \tag{4.4}$$

| Translation of English in Bhojpuri | Probability Distribution |
|---|---|
| बढ़िया | 0.249 |
| नीक | 0.261 |
| अच्छा | 0.309 |
| नीमन | 0.131 |
| ठीक | 0.012 |

Table 4. 4: Statistics of Probability Distribution of 'good' word in English-Bhojpuri Corpus

The above (4.4) function also explains that for any English word $e$ (good), a probability for each choice of Bhojpuri translation $b$ is returned. This probability indicates how likely the next translation would be. The function of probability distribution is computed on each of the above provided translation choice which is shown in the Table 4.4.

The table above represents the source and target word possibilities in a parallel corpus. Using the example of the source word 'good', we find that there is a higher possibility of it being translated as अच्छा (0.309) than ठीक (0.012).

Word position is another aspect of alignment. A Word-Based Model proposes to translate sentences word by word. The following diagram illustrates the alignment between input and output words.

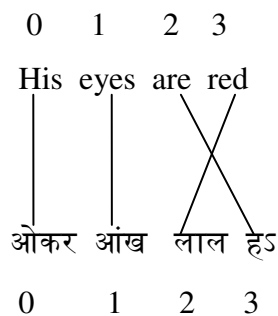

$a$: {0-0, 1-1, 2-3, 3-2}

The above example is a sentence translated from English to Bhojpuri where only an adjective alters its position in the output. A function of an alignment can be formalized using an alignment function *a*. This function maps Bhojpuri word position at *b* to an English input word at position *e*:

$$a: b \longrightarrow e \quad\quad\quad (4.5)$$

But there are structures where English requires two Bhojpuri words to capture the same meaning such as in the example provided below where the English word 'oldest' holds two words of Bhojpuri:

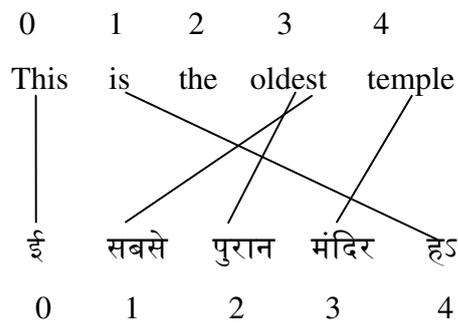

*a*: {0-0, 1-3, 2-3, 3-4, 4-1}

In the above example illustrating alignment, there is a lack of clear translation for the English word 'the' into Bhojpuri or equivalent words which should be actually dropped during the process of translation. To capture such scenarios, Moses alignment model introduces NULL token. This token should be treated like any other word in the output. This token is required to align each Bhojpuri word to an English one, to define the alignment function completely.

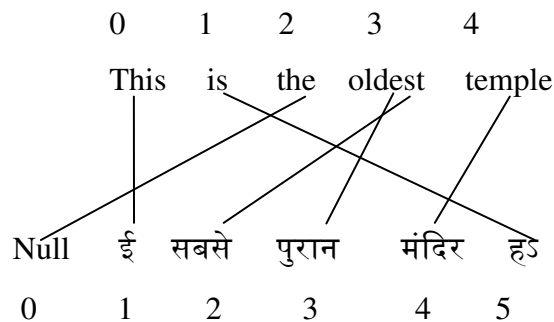

*a*: {0-2, 1-0, 2-3, 3-3, 4-4, 5-1}

At this stage, during translation, alignment model captures three kinds of word mapping behavior:
- One-to-one word mapping
- One-to two word mapping
- One-to NULL token mapping

At the second stage, Moses captures the idea of IBM Model 1 into its alignment model. IBM Model 1 allows a model that generates several different translations for a sentence. Each translation possesses a different probability. This model is fundamentally a generative model which disintegrates the process of generating the data into further smaller steps. These smaller steps are further modeled with probability distributions, and then combined into a coherent sentence.

At the third stage, Moses allows Expectation Maximization (EM) algorithm to learn translation probabilities from sentence-aligned parallel text. The EM algorithm works on sentence-aligned corpus addressing the issues caused due to incomplete data. It is an iterative learning method which helps in filling the gaps in data and in training a model in alternating steps.

The EM algorithm works as follows (Koehn, 2010):
- Initialize the model with uniform distributions
- Apply the model to the data (expectation step)
- Learn the model from the data (maximization step)
- Iterate steps 2 and 3 until convergence takes place

The first step is to initialize the model sans any prior knowledge. This means for each input, word *e* may be translated with equal probability into any output word *s*. In the expectation step, the English word 'good' is aligned to its most likely translation अच्छा. In the maximization step, the model learns from the data. Based on the learnings gains, the best guess would be determined. But it is better to consider all possible guesses and weigh them along with their corresponding probabilities. Since it is arduous to compute all of them, therefore, the model uses the technique of sampling, and higher probabilities counts. This process iterates through steps 2 and 3 until convergence takes place.

As discussed above, IBM Model 1 is only capable of handling lexical translation and is extremely weak in terms of reordering. Four more models are proposed in the original work on SMT at IBM:

IBM Model 1: lexical translation;
IBM Model 2: lexical translation with absolute alignment model
IBM Model 3: addition of fertility model
IBM Model 4: addition of relative alignment model
IBM Model 5: fixing deficiency

IBM Model 2 is successful in addressing the issue of alignment, based on the positions of both, the input and output words. An English input word's translation in position *e* to Bhojpuri word in position *b* is modeled using an alignment probability distribution

$$a(e|b, l_b, l_e) \qquad (4.6)$$

Therefore, the translation done under IBM Model 2 becomes a two-step process including a lexical translation step and an alignment step:

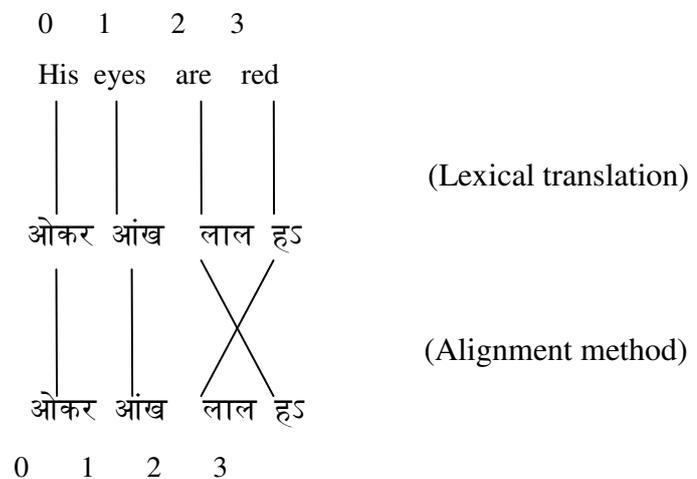

Lexical translation step is the first step which is modeled by the translation probability $t(b|e)$. The alignment step comes next which is modeled by an alignment probability $a(2|3, 4, 4)$. In the above example, the 2$^{nd}$ Bhojpuri word लाल is aligned to 3$^{rd}$ English word 'red'.

IBM Model 3 appends fertility to its model which handles the number of output words which are generated from each input word. For instance, few Bhojpuri words do not correspond to English words. These dropping or adding words are generated in IBM Model 3 with the use of the special NULL token. The fertility and NULL Token insertion enhances the translation process to four steps in IBM Model 3.

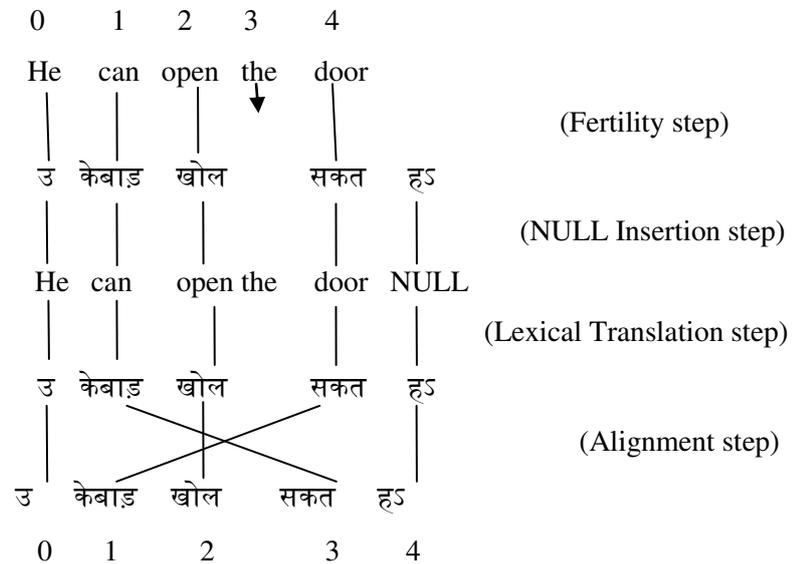

IBM Model 4 is an improved version of Model 3 with the introduction of a relative distortion model. According to this enhancement, the placement of the translation of an input word is now based on the placement of the translation of the preceding input word. For instance, some words get reordered during the process of translation, while others maintain the order. An example of this verb-object inversion that can take place during the translation process from English to Bhojpuri is when the token 'is good' of English converts into the token अच्छा हऽ in Bhojpuri. Here, Model 4 simultaneously introduces word class and vocabulary of a language mostly grouped into 50 or more classes. For each class, the probabilities are computed and translations are generated accordingly.

While Model 3 and 4 may display the possibility of placing the multiple output words in the same position, Model 5 fixes such deficiency by places words only into vacant word positions. For Model 5, the alignment process that steps through one foreign input word at a time, places words into Bhojpuri output positions while keeping track of vacant Bhojpuri positions. The fertility of English input words gets selected right at the beginning. This means that the number of Bhojpuri output positions is fixed. These IBM Models, from simple (IBM Model 1) to sophisticated (IBM Model 5), are still state-of-the-art when it comes to alignment models of SMT.

A fundamental problem with IBM Models is that each Bhojpuri word can be traced back to exactly one English word (or the NULL token). This means that each Bhojpuri token is aligned to (at most) one English token. It is impossible that one Bhojpuri word is aligned with multiple English words. To address such an alignment issue, Moses alignment model algorithm carries out IBM models training in both directions. This results in two-word alignments which can then be merged with the support of intersection or the union of alignment points of each alignment. This process is called symmetrization of word alignments. For better performance goals, Moses symmetrizes word alignments after every iteration of IBM model training.

The GIZA++ toolkit has been used for the word alignment to extract words (phrases) from its corresponding parallel corpora. The GIZA++ implements all IBM (1-5) models and HMM algorithms to align words and sequences of words (Och and Ney, 2003). The HMM word alignment model also consists of a source for the MKCLS tool which assists in generating the word classes crucial to train some of the alignment models. This aligner is used to build all kinds of EB-SMT systems.

The forthcoming section explains the newly created and trained translation models (the sections below explain only about the training and decoding process in the experiments, for theory refer the chapter - Introduction).

**4.3.2.1 Phrase-based Translation Models**

Phrase-based translation gains a two-fold edge over word-based translation:
a) phrases manage to resolve several translation ambiguities
b) phrase-based modeling is also able to resolve one-to-many mappings; case of large parallel corpus, longer and complex phrases and sometimes an entire sentence can be learnt.

Apart from these benefits, it is also able to handle local reordering captured through the neighboring phrase pairs (Galley and Manning, 2008).

In the phrase-based models, input (English as *e*) sentence is first segmented into phrases, each of which are then translated into a phrase of output (Bhojpuri as *b*) sentence (process is illustrated in an example and Figure 4.2). The phrase may be reordered and its probability is computed with the use of the following formula:

$$\emptyset(e_i|b_i) \qquad (4.7)$$

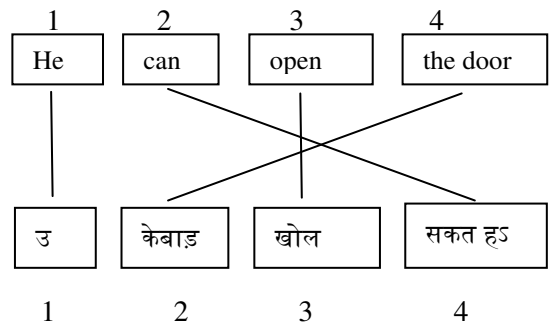

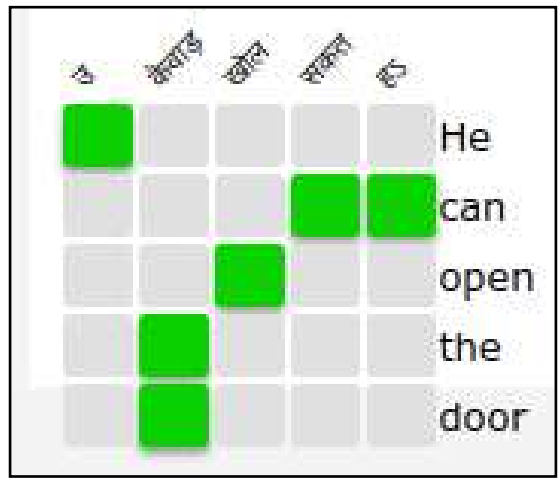

Figure 4. 2: English-Bhojpuri Phrase Alignment

After pre-processing is complete and LM is built, the next step involves creation of phrase-based translation model. The first experiment was trained using the IRSTLM-based language model wherein 63000 parallel training data-set was used (shown in table-4.1). To build the PBSMT system, following parameters were followed:

- 'grow-diag-final-and heuristic search' method was used for word alignment which helps in phrase extraction
- up to 4 n-gram order was used

After the model was built, two files were created: (a) phrase table and (b) moses.ini Based on stored phrase pairs, which is a part of the phrase table, a translation table is generated (sample of phrase table can be seen in Figure 4.3).

```
! " exclaimed the ||| क दावा ||| 0.333333 4.35515e-09 0.333333 0.0483928 ||| 2-0 3-0 2-1 ||| 3 3 1 ||| |||
! " exclaimed the ||| क दावा ह ||| 0.333333 4.35515e-09 0.333333 0.000726018 ||| 2-0 3-0 2-1 ||| 3 3 1 ||| |||
! " exclaimed the ||| क दावा ह कि ||| 0.333333 4.35515e-09 0.333333 9.70279e-06 ||| 2-0 3-0 2-1 ||| 3 3 1 ||| |||
$ 23 million movie . ||| तेईस मिलियन डालर के पिक्चर । ||| 0.5 0.0539836 1 9.65767e-05 ||| 1-0 1-1 2-1 0-2 2-3 3-4 4-5 ||| 2 1 1 ||| |||
$ 23 million movie ||| तेईस मिलियन डालर के पिक्चर ||| 0.5 0.0635271 1 0.000386793 ||| 1-0 1-1 2-1 0-2 2-3 3-4 ||| 2 1 1 ||| |||
$ 23 million ||| तेईस मिलियन डालर के ||| 0.5 0.222345 1 0.00323939 ||| 1-0 1-1 2-1 0-2 2-3 ||| 2 1 1 ||| |||
$ ||| $ ||| 1 1 0.5 0.428571 ||| 0-0 ||| 2 4 2 ||| |||
$ ||| डालर ||| 0.5 1 0.25 0.285714 ||| 0-0 ||| 2 4 1 ||| |||
$ ||| होला डालर ||| 1 0.517857 0.25 0.0408163 ||| 0-0 0-1 ||| 1 4 1 ||| |||
% is levied on Local Calls ||| परसेंट आ रोमिंग कॉल पर २५ ||| 1 0.00281294 1 6.47028e-07 ||| 5-0 0-2 2-2 1-3 3-4 4-5 ||| 1 1 1 ||| |||
```

Figure 4. 3: Snapshot of Phrase-Table of English-Bhojpuri PBSMT System

A distinctive feature of Moses phrase-based model is that the corpus can be trained in both translation directions (shown in Figure 4) $\emptyset\ (e_i|b_i)$ and $\emptyset\ (e_i|b_i)$ because the feature functions with proper weight, outperforms a model that makes use of one translation direction only. As far as the number of phrase pairs are concerned, it is always a conundrum to decide whether one should rely on longer (fewer) phrase pairs or the shorter (more) ones. To overcome this dilemma uses this test: if p <1, longer (fewer) phrase pairs is preferred, and if p >1, shorter (more) phrase pairs is preferred.

```
# Sentence pair (1) source length 6 target length 8 alignment score : 2.98015e-09
A stone , said the young one .
NULL ({ }) एगो ({ 1 }) पत्थर ({ 2 }) , ({ 3 }) कहलस ({ 4 }) छोटन ({ 5 6 7 }) । ({ 8 })
# Sentence pair (2) source length 7 target length 14 alignment score : 7.95033e-22
The ' h ' in honest is silent ' as ' in hour .
NULL ({ }) होनेस्ट ({ 1 2 }) में ({ 5 }) एच ({ 3 4 6 8 9 10 11 13 }) शान्त ({ }) होवे ({ 7 12 }) ले ({ }) । ({ 14 })
# Sentence pair (3) source length 5 target length 6 alignment score : 1.8496e-07
Who wrote ' Gitanjali ' ?
NULL ({ }) गीतांजली ({ 1 3 4 5 }) के ({ }) लिखले ({ 2 }) बा ({ }) ? ({ 6 })
# Sentence pair (4) source length 7 target length 9 alignment score : 3.92907e-13
When is ' Cheti Chand ' got celebrated ?

# Sentence pair (1) source length 8 target length 6 alignment score : 1.23773e-10
एगो पत्थर , कहलस छोटन ।
NULL ({ }) A ({ 1 }) stone ({ 2 5 }) , ({ 3 }) said ({ 4 }) the ({ }) young ({ }) one ({ }) . ({ 6 })
# Sentence pair (2) source length 14 target length 7 alignment score : 2.47908e-12
होनेस्ट में एच शान्त होवे ले ।
NULL ({ }) The ({ }) ' ({ }) h ({ 1 3 4 5 6 }) ' ({ }) in ({ 2 }) honest ({ }) is ({ }) silent ({ }) ' ({ }) a
({ }) ' ({ }) in ({ }) hour ({ }) . ({ 7 })
# Sentence pair (3) source length 6 target length 5 alignment score : 4.50532e-06
गीतांजली के लिखले बा ?
NULL ({ }) Who ({ 2 }) wrote ({ 3 }) ' ({ }) Gitanjali ({ 1 4 }) ' ({ }) ? ({ 5 })
# Sentence pair (4) source length 9 target length 7 alignment score : 1.23985e-12
चेति चान्द कब मानावल जावे ला ?
NULL ({ }) When ({ 3 }) is ({ }) ' ({ }) Cheti ({ 1 5 6 }) Chand ({ 2 4 }) ' ({ }) got ({ }) celebrated ({ }) ?
```

Figure 4. 4: Snapshot of English-Bhojpuri and Bhojpuri-English Phrase-based Translation model

To check the suitability of different LMs with the English-Bhojpuri language pair in the PBSMT system and also to improve the accuracy level, SRILM and KenLM-based LMs have been used. With the support of these two LMs, other two PBSMT systems were trained with the same parameters and training data. Out of these three systems, SRILM and KenLM-based PBSMT systems gave slightly better results as compared to IRSTLM (results are shown in the section 4.4).

English and Bhojpuri have different word-order, therefore, to solve the word ordering issue and improve performance, the experiments were conducted with reordering model (Koehn et al., 2005; Galley and Manning, 2008; Koehn, 2010) and the PBSMT system was trained. There have been several experiments done for Indian languages with reordering models and have reported improvement in the performance (Gupta et al, 2012; Patel et al., 2013; Chatterjee et al; 2014; Kunchukuttan et al., 2014; Pal et al., 2014; Shukla and Mehta et al., 2018). Three different (including lexicalized, phrase-based and hierarchal) reordering-based translation models were implemented which were extended to the '–reordering' flag in the syntax before following models:

- msd-bidirectional-fe (msd = use of three different orientations: monotone[5] (m), swap[6] (s) and discontinuous[7] (d); bidirectional = represents directionality of both backward and forward models; fe = used for both source and target language)
- phrase-msd-bidirectional-fe (phrase = phrase-based model type)
- hier-mslr-bidirectional-fe (hier = used for hierarchical based model type; mslr = uses four different orientations: monotone, swap, discontinuous-left, discontinuous-right)

The above mentioned reordering models are in addition to the other methods used in the previous experiments (phrase-based translation models including IRSTLM, SRILM, KenLM). A PBSMT system developed with these models performed the worst in comparison to the previous three PBSMT systems (results are reported in section 4.4).

A distance-based reordering model handles the issue of reordering in Moses. If we consider $start_i$ as the place of the initial word in an English input phrase that translates to the *ith* Bhojpuri phrase, and $end_i$ as the place of the final word of the same English phrase (Nainwani, 2016), then reordering will be computed using the equation, $start_i - end_{i-1} - 1$. The reordering distance is actually a measure of words skipped (either forward or backward) in the event of taking input words out of sequence.

A key problem with phrase-based SMT translation model is the loss of larger context during the process of making translation predictions. The model's functionality is also restricted to the mapping of only short chunks without any direct support of linguistic information (morphological, syntactic, or semantic). Such additional information aids in enhancing statistical performance and resolving the problems of data sparseness caused due to limited size of training data. But in PBSMT, linguistic information has been proved to be valuable through its integration in pre-processing/post-processing steps.

**4.3.2.2 Hierarchical Phrase-Based Translation Models**

We understand that a phrase is an atomic unit in the phrase-based translation models (PBSMT system), A Hierarchical model creates sub-phrases in order to weed out several problems associated with a PBSMT system especially the one with long distance reordering. HPBSMT is a tree-based model that specializes in automatic extraction of SCFG through a parallel corpus without the aid of labeled data, annotated syntactically in the form of hierarchical rules (Chiang, 2005). This model takes into consideration a formally syntax-based SMT where a grammar (hierarchical rules) sans underlying linguistic interpretation is created. In the hierarchical phrase-based model phrases, hierarchical rules extracted through HPBSMT are the fundamental units of translation. These rules are extracted in accordance with the phrase-based translation model (Koehn et al., 2003). Therefore, hierarchical rules possess the power of statistically-extracted continuous sets in addition to the ability to translate interrupted sentences as well and learn sentence reordering without a separate reordering model. The HPBSMT model has two kinds of rules: hierarchical rules and glue grammar rules. This model expands highly lexicalized-based models of sentence translation systems, lexicalized rearrangement model and disjoint sets[8] (Chiang, 2005; Chiang, 2007).

---

[5] There is evidence for monotone orientation when a word alignment point exists towards the top-left.
[6] There is evidence for a swap with the previous phrase when a word alignment point exists towards the top-right.
[7] There is evidence for discontinuous orientation when no word alignment point exists towards either top-left or top-right and there is neither monotone order nor a swap.

Both, Chiang model and Moses toolkit are widely accepted and followed for SMT. Hence, the experiments have been conducted with Hierarchical phrase-based translation model using the SRILM and KenLM-based LMs. To train the HPBSMT, the author has used the same training data size and followed identical training steps as used for the baseline of PBSMT systems. Three additional parameters[9] have also been included: '-hierarchical and -glue-grammar (for creating glue-grammar and by default rule-table)', '--extract-options (for extraction of rules)' and '--score-options (for scoring of rules)'. These parameters create 'rule-table' instead of phrase-table which is a part of the model (the 'rule-table' is an extension of the Pharaoh or Moses 'phrase-table'). See Figure 4.5 for an illustration of the rule-table of HPBSMT.

---

[8] A detailed description is included in Chapter 1, Introduction.
[9] Further details can be accessed from : *"http://www.statmt.org/moses/?n=Moses.SyntaxTutorial"*

```
[X][X] our fast bowler . [X] ||| [X][X] हमन क तेज गेंदबाज ह । [X] ||| 0.5 3.34919e-05 1 0.000631171 ||| 0-0 2-3 3-1 3-2 3-4 3-5 4-5 ||| 0.285714 0.142857 0.142857 ||| |||
[X][X] our flag [X] ||| [X][X] हमहन क झन्डा [X] ||| 1 0.0218876 1 0.00134698 ||| 0-0 1-1 2-2 2-3 ||| 0.369841 0.369841 0.369841 ||| |||
[X][X] our flag [X][X] ! [X] ||| [X][X] हमहन क झन्डा [X][X] [X] ||| 0.110113 0.000136137 1 0.00134698 ||| 0-0 1-1 2-2 2-3 3-4 ||| 0.394851 0.0434783 0.0434783 ||| |||
[X][X] our flag [X][X] [X] ||| [X][X] हमहन क झन्डा [X][X] [X] ||| 0.672173 0.0218876 1 0.00134698 ||| 0-0 1-1 2-2 2-3 3-4 ||| 0.394851 0.265408 0.265408 ||| |||
[X][X] our flag [X][X] there [X] ||| [X][X] हमहन क झन्डा [X][X] [X] ||| 0.217715 0.000172829 1 0.00134698 ||| 0-0 1-1 2-2 2-3 3-4 ||| 0.394851 0.0859649 0.0859649 ||| |||
[X][X] our flag was [X][X] [X] ||| [X][X] हमहन क झन्डा [X][X] रहे । [X] ||| 1 0.00293266 1 2.95803e-05 ||| 0-0 1-1 2-2 2-3 3-5 4-4 ||| 0.129443 0.129443 0.129443 ||| |||
[X][X] our flag was still [X] ||| [X][X] हमहन क झन्डा ओहिजे रहे । [X] ||| 1 0.00293266 1 1.4019e-07 ||| 0-0 1-1 2-2 2-3 3-5 4-4 ||| 0.135965 0.135965 0.135965 ||| |||
```

Figure 4. 5: Snapshot of Rule Table from the English-Bhojpuri HPBSMT

A major drawback of this approach, when compared to set-based systems, is that the total number of rules learned is larger in several orders of magnitude than standard phrase-based translation model. This leads to an over-generation rate and help search error, further resulting in a much longer decoding time, requiring more space and high memory in comparison to phrase-based and factor-based translation models.

### 4.3.2.3 Factor-Based Translation Models

Factor-based translation model operates on the phrase-based model, extending its basis to model variables representing linguistic information (Koehn et al., 2003; Koehn and Hoang, 2007; Koehn, 2010; Hoang, 2011). It also allows integration of additional linguistic information through annotation at word level, like labels indicating POS and lemma. Each type of additional word-level information is termed as a factor. In short, to develop a statistical translation model which is factored-based, the training data from parallel corpus should be annotated with additional factors. The remaining standard steps of training are followed with the same methods as used in the previous translation model.

From the perspective of training, there are two crucial features or steps in factored models. The first feature is translation and the second one is called generation. They originate from a word-aligned parallel corpus and determine scoring methods, thus, helping in making an accurate choice from multiple ambiguous mappings (Koehn and Hoang, 2007).

As we've learnt before, Phrase-based translation models are obtained from a parallel corpus that is word-aligned. The process for acquiring these models is to extract all phrase pairs which are consistent with word alignment. Several feature functions can be estimated if a set of extracted phrase pair are made available with counts. These feature functions can be conditional phrase translation probabilities. They stand on relative frequency estimation/lexical translation probabilities which are established on the basis of words constituting the phrases.

Similarly, the translation steps models can also be obtained with the support of a word-aligned parallel corpus. For a designated number of factors in both, the input and output, a set of phrase mappings (now over-factored representations) are extracted which are determined on the basis of relative counts and the probabilities of word-based translation (as shown in above figure).

The generation steps are actually probability distributions which are estimated only on the side of the output. It is acquired on a word-for-word basis. For example, a tabular representation of entries such as (अच्छा, JJ) is built for every generation step that is successful in mapping surface forms to part-of-speech. Conditional probability distributions, obtained by maximum likelihood estimation can be used as feature functions, e.g., p(JJ|अच्छा).

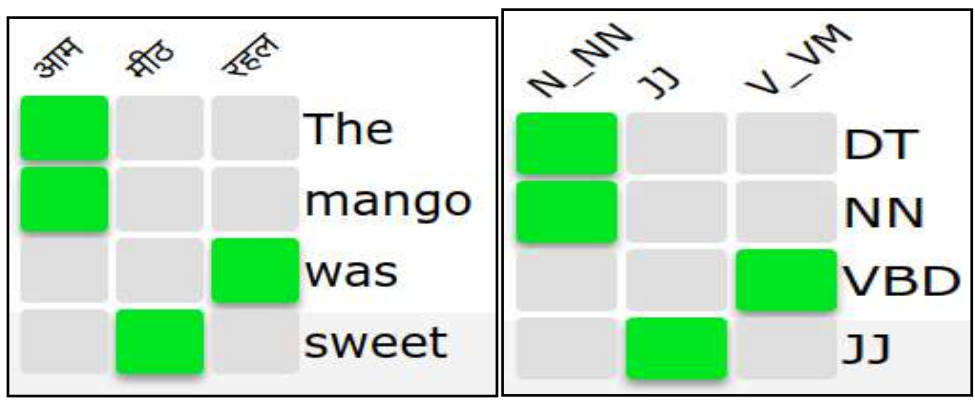

Figure 4. 6: Extraction of the translation models for any factors follows the phrase extraction method for phrase-based models

To reiterate, the LM is a crucial constituent of any SMT system which frequently creates over-surface forms of words. Such sequence models can be defined over any singular or a set of factor(s) in the framework of factored translation models. Building is straightforward for factors like part-of-speech labels.

Hence, to successfully conduct experiments using this model, the training data (including both parallel and monolingual corpus) was annotated at POS level only (shown in the Table 4.1). At first, surface factor-based LMs were developed using the monolingual sentences based on SRILM and KenLM toolkit only. The next step involved, preprocessing and then training FBSMT systems using the translation steps. After the creation of SRILM and KenLM-based two factor translation models, the author has trained other six factor-based translation models using the above-mentioned three reordering models (sample of model is shown in Figure 4.7).

```
! Come here ||| इहां|PR_PRP आवज|N_NN ||| 1 0.0193508 1 0.00172865 ||| 2-0 0-1 1-1 ||| 1 1 1 ||| |||
! Come ||| आवज|N_NN ||| 0.5 0.0277778 1 0.0066358 ||| 0-0 1-0 ||| 2 1 1 ||| |||
! I was very little when my ||| !|
RD_SYM हम|PR_PRP बहुत|RP_INTF छोट|JJ रहली|V_VM जब|N_NST हमार|PR_PRP ||| 1
0.00016405 1 1.18322e-05 ||| 0-0 1-1 3-2 4-3 2-4 5-5 6-6 ||| 1 1 1 ||| |||
! I was very little when ||| !|RD_SYM हम|PR_PRP बहुत|RP_INTF छोट|JJ रहली|V_VM जब|N_NST
||| 1 0.000553444 1 3.86604e-05 ||| 0-0 1-1 3-2 4-3 2-4 5-5 ||| 1 1 1 ||| |||
! I was very little ||| !|RD_SYM हम|PR_PRP बहुत|RP_INTF छोट|JJ रहली|V_VM ||| 1 0.00106472 1
7.29025e-05 ||| 0-0 1-1 3-2 4-3 2-4 ||| 1 1 1 ||| |||
! I ||| !|RD_SYM हम|PR_PRP ||| 1 0.121909 0.25 0.112032 ||| 0-0 1-1 ||| 1 4 1 ||| |||
! I ||| !|RD_SYM हमके|PR_PRP ||| 1 0.0812713 0.25 0.0288667 ||| 0-0 1-1 ||| 1 4 1 ||| |||
! I ||| त|RP_RPD हम|PR_PRP ||| 0.0285714 0.00360973 0.25 0.00800229 ||| 0-0 1-1 ||| 35 4 1 ||| |||
! I ||| महरानी|N_NN हम|PR_PRP ||| 1 0.91867 0.25 0.00200059 ||| 0-0 1-1 ||| 1 4 1 ||| |||
! If ||| !|RD_SYM अगर|CC_CCD ||| 1 0.060611 0.5 0.0801231 ||| 0-0 1-1 ||| 1 2 1 ||| |||
! If ||| अगर|N_NN ||| 1 0.25 0.5 0.00330055 ||| 0-0 1-0 ||| 1 2 1 ||| |||
! Just jump . ||| !|RD_SYM कूद|N_NN जा|V_VM ||| 1 0.000326308 1 0.000482381 ||| 0-0 3-0 1-1 2-1
```

Figure 4. 7: Snapshot of Phrase-Table based on the Factor-based Translation Model

**4.3.2.4 PD and UD based Dependency Tree-to-String Models**

The last two experiments were administered using the 'Tree-to-String' method relying on dependency data which acts as a constituent of syntax-based SMT models. The Tree-to-String models, as mentioned before, use a rich input language representation (source-side language information derived trees resemble the linguistic parse trees observed in the data) to translate into word sequences in the output language.

To develop the Deep-to-Tree-Str based SMT systems, the task of dependency annotation was performed using the PD and UD models (see chapter 2 to know more about PD and UD models). This annotation practice is used to annotate English source sentences. Statistics of such labeled data are placed in Table 1. Before system training, pre-processing and CoNLL format-based annotation of dependency data was transformed into XML format. This was done in order to suit the requisites of input format for Moses (shown in Figure 4.8). Because XML format demands substructure nesting, the input can be provided only in the form of projective dependency structures to the tool. Such is the case because non-projectivity is known to break nesting (Graham, 2013).

```
<tree label="sent"><tree label="root"><tree label="k1"><tree label="lwg"><tree label="WDT">What</tree></tree><tree
label="NN">idea</tree></tree><tree label="lwg_aux"><tree label="VBZ">does</tree></tree><tree label="lwg"><tree
label="DT">the</tree></tree><tree label="NN">story</tree></tree><tree label="r6"><tree label="case"><tree label="IN">of</tree></tree></tree><tree label="rsym"><tree label="`">'</tree></tree><tree label="NNP">Beauty</tree><tree
label="conj"><tree label="compound"><tree label="cc"><tree label="CC">and</tree></tree></tree><tree label="lwg"><tree
label="DT">the</tree></tree><tree label="NNP">Beast</tree></tree><tree label="case"><tree label="POS">'</tree></tree></tree><tree label="NN">convey</tree></tree></tree><tree label="rsym"><tree label=".">?</tree></tree></tree></tree>
<tree label="sent"><tree label="root"><tree label="expl"><tree label="EX">There</tree></tree><tree label="VBP">are</tree><tree label="k1"><tree label="amod"><tree label="JJ">total</tree></tree><tree label="nummod"><tree
label="CD">66</tree></tree><tree label="NNS">gates</tree></tree><tree label="r6"><tree label="case"><tree label="IN">in</tree></tree></tree><tree label="compound"><tree label="NNS">Loyds</tree></tree><tree label="NN">barrage</tree></tree><tree label="rsym"><tree label=".">.</tree></tree></tree></tree>
<tree label="sent"><tree label="root"><tree label="advmod"><tree label="WRB">How</tree></tree><tree label="k1"><tree
label="amod"><tree label="JJ">many</tree></tree><tree label="amod"><tree label="JJ">total</tree></tree><tree
label="NNS">gates</tree></tree><tree label="cop"><tree label="VBP">are</tree></tree><tree label="RB">there</tree><tree label="obl"><tree label="case"><tree label="IN">in</tree></tree><tree label="rsym"><tree
label="``">'</tree></tree><tree label="compound"><tree label="NNP">Loyds</tree></tree><tree
label="NNP">Barrage</tree></tree><tree label="rsym"><tree label="''">'</tree></tree><tree label="rsym"><tree
label=".">?</tree></tree></tree></tree>

<tree label="sent"><tree label="root"><tree label="nsubj"><tree label="det"><tree label="DT">A</tree></tree><tree
label="NN">stone</tree></tree><tree label="punct"><tree label=",">,</tree></tree><tree label="VBD">said</tree><tree
label="obj"><tree label="det"><tree label="DT">the</tree></tree><tree label="amod"><tree label="JJ">young</tree></tree><tree label="NN">one</tree></tree><tree label="punct"><tree label=".">.</tree></tree></tree></tree>
<tree label="sent"><tree label="root"><tree label="det"><tree label="DT">The</tree></tree><tree label="amod"><tree
label="JJ">'h</tree></tree><tree label="punct"><tree label="``">'</tree></tree><tree label="obl"><tree
label="case"><tree label="IN">in</tree></tree><tree label="JJ">honest</tree></tree><tree label="cop"><tree
label="VBZ">is</tree></tree><tree label="JJ">silent</tree><tree label="punct"><tree label="''">'</tree></tree><tree label="obl"><tree label="case"><tree label="IN">as</tree></tree><tree label="punct"><tree
label="``">'</tree></tree><tree label="case"><tree label="IN">in</tree></tree><tree label="NN">hour</tree></tree><tree label="punct"><tree label=".">.</tree></tree></tree></tree>
<tree label="sent"><tree label="root"><tree label="nsubj"><tree label="WP">Who</tree></tree><tree label="VBD">wrote</tree><tree label="obj"><tree label="punct"><tree label="``">'</tree></tree><tree label="NNPS">Gitanjali</tree><tree label="punct"><tree label="''">'</tree></tree></tree><tree label="punct"><tree label=".">?</tree></tree></tree></tree>
<tree label="sent"><tree label="root"><tree label="mark"><tree label="WRB">When</tree></tree><tree
label="nsubj:pass"><tree label="cop"><tree label="VBZ">is</tree></tree><tree label="punct"><tree label="``">'</tree></tree><tree label="compound"><tree label="NNP">Cheti</tree></tree><tree label="NNP">Chand</tree><tree
label="case"><tree label="POS">'</tree></tree></tree><tree label="aux:pass"><tree label="VBD">got</tree></tree><tree label="VBN">celebrated</tree><tree label="punct"><tree label=".">?</tree></tree></tree></tree>
```

Figure 4. 8: Snapshot of Converted Tree data of PD & UD based to train of Deep-to-Tree-Str SMT Systems

```
<tree label="sent"><tree label="root"><tree label="discourse"><tree label="UH">No</tree></tree><tree
label="punct"><tree label=",">,</tree></tree></tree><tree ||| में आप सांस लेवे खातिन होई ||| 1 1 0.0102041 0.0257132 ||| 5-1 0-2 1-2 2-2 3-2 4-2 5-2 6-2
5-3 5-4 5-5 ||| 1 98 1 ||| |||
<tree label="sent"><tree label="root"><tree label="discourse"><tree label="UH">No</tree></tree><tree
label="punct"><tree label=",">,</tree></tree></tree></tree><tree ||| रूप में आप सांस लेवे खातिन होई ||| 1 1 0.0102041 2.88785e-05 ||| 5-2 0-3 1-3
2-3 3-3 4-3 5-3 6-3 5-4 5-5 5-6 ||| 1 98 1 ||| |||
<tree label="sent"><tree label="root"><tree label="discourse"><tree label="UH">No</tree></tree><tree
label="punct"><tree label=",">,</tree></tree></tree></tree><tree ||| वरना निसकासित हो जईबा ||| 0.5 1 0.0102041 1 ||| 4-0 0-1 1-1 2-1 3-1 4-1 4-2 5-2 4-3 ||| 2
98 1 ||| |||
<tree label="sent"><tree label="root"><tree label="discourse"><tree label="UH">No</tree></tree><tree
label="punct"><tree label=",">,</tree></tree></tree></tree><tree ||| सइकील समूह क 18 सदस्स गिरफ्तार भईलें ||| 1 1 0.0102041 1.51168e-10 ||| 2-3 2-4
2-5 0-6 1-6 2-6 3-6 4-6 5-6 6-6 ||| 1 98 1 ||| |||
<tree label="sent"><tree label="root"><tree label="discourse"><tree label="UH">No</tree></tree><tree
label="punct"><tree label=",">,</tree></tree></tree></tree><tree ||| सब ठीक हव , ले ला ||| 0.5 1 0.0102041 1 ||| 3-0 4-1 4-2 4-3 3-4 0-5 1-5 2-5 3-5 4-5 5-5 |||
2 98 1 ||| |||
<tree label="sent"><tree label="root"><tree label="discourse"><tree label="UH">No</tree></tree><tree
label="punct"><tree label=",">,</tree></tree></tree></tree><tree ||| सबसे दुरी बात ना हव की ||| 0.2 1 0.0102041 1 ||| 2-0 0-1 1-1 2-1 2-2 1-3 1-4 1-5 ||| 5 98 1 |||
|||
<tree label="sent"><tree label="root"><tree label="discourse"><tree label="UH">No</tree></tree><tree
label="punct"><tree label=",">,</tree></tree></tree></tree><tree ||| समे कइदिन ताला बन्दी खातिन रपट करा । ||| 1 1 0.0102041 1 ||| 5-0 0-1 1-1 2-1 3-1 4-1 5-1
6-1 5-2 5-3 5-4 5-5 5-6 ||| 1 98 1 ||| |||
<tree label="sent"><tree label="root"><tree label="discourse"><tree label="UH">No</tree></tree><tree
label="punct"><tree label=",">,</tree></tree></tree></tree><tree ||| समझलू ? ||| 0.5 1 0.0102041 1 ||| 0-0 1-0 2-0 3-0 4-0 5-1 ||| 2 98 1 ||| |||
<tree label="sent"><tree label="root"><tree label="discourse"><tree label="UH">No</tree></tree><tree
label="punct"><tree label=",">,</tree></tree></tree></tree><tree ||| समूह क 18 सदस्स गिरफ्तार भईलें ||| 1 1 0.0102041 2.25624e-06 ||| 2-2 2-3 2-4
0-5 1-5 2-5 3-5 4-5 5-5 6-5 ||| 1 98 1 ||| |||
```

Figure 4. 9: Screenshot of Phrase-Table of the Dep-Tree-to-Str based EB-SMT System

In order to build the Dep-Tree-to-Str systems, there is a need to extract rules. To meet this requirement, Moses (Williams and Koehn, 2012) implements GHKM (Galley et al., 2004;

Galley et al., 2006), which otherwise is used in the process of syntax-augmented SCFG extraction from phrase-structure parses (Zollmann and Venugopal, 2006). It is the same rule extraction process that has been applied to dependency parses in a way so that there is no mandatory restriction to a particular set of node labels. The remaining part of system training (including PD and UD based) steps are the same as followed in the PBSMT/HPBSMT system. A sample of rule-table is provided in Figure 4.9 substantiating similarity to the HPBSMT.

### 4.3.3 Tuning

Tuning is the last step in the process of creating SM system. Different machine-produced translations (known as *n-best* hypotheses) are weighed against each other in order to deduce a group of best possible translations. Minimum Error Rate Training (Och et al., 2003) is an in-built tuning algorithm residing in the MOSES decoder[10] with the task objective to tune a separate set of parallel corpus. To complete this process, 1000 parallel sentences (see tuning/dev set under Table 4.1) were used which was implemented with all of the above developed systems using the 'mert-moses.pl' script. There is an additional parameter "--inputtype 3" which was included with the 'mert-moses.pl' script to tune the Dep-Tree-to-Str based EB-SMT systems. The MERT performs three tasks with the aid of tuning set: (a) Combining features set using the algorithm combines a set of features (b) determining contribution of each feature weight to the overall translation score and (c) Optimizing the value of feature weights to maximize the translation quality.

---

[10] The task of decoder is to identify the best-scoring translation from an exponential number of options available for any given source language sentence. Examining an exhaustive list of possible translations, score each of them, and pick the best one out the scored list is computationally expensive for even a sentence of modest length. Moses provides a solution in the form of a set of efficient techniques called heuristic search methods whose task is to find the best possible translations (from the LM, phrase-table/rule-table and reordering models, plus word, phrase and rule counts). These methods are able to prevent two kind of errors: a) Search error – it refers to the failure to locate the translation with highest-probability, and b) Model error - when the translation with highest probability is not a good translation according to the model. Therefore, the end goal of MT is to provide translations which is able to deliver the meaning of source sentence and is also fluent in target language (Further details can be accessed from: http://www.statmt.org/moses/?n=Advanced.Search or follow Neubig and Watanabe, 2016's article "Optimization for Statistical Machine Translation: A Survey").

### 4.3.4 Testing

As discussed in section 4.2, 1000 sentences were used to evaluate the results of the fully-developed English-Bhojpuri SMT (including EB-PBSMT, EB-HPBSMT, EB-FBSMT and Dep-Tree-to-Str) systems. The BLEU scores are reported in section 4.4.

### 4.3.5 Post-processing

After obtaining output from the SMT systems, the tasks of de-tokenization and transliteration of the names are performed in a bid to further enhance the accuracy rate of the EB-SMT outputs.

## 4.4 Experimental Results

In the current research, 24 EB-SMT systems were trained using the Moses on English-Bhojpuri with different phrase-based, hierarchical phrase-based, factor-based and dependency based tree-to-string translation models on various LMs. Figure 4.10, 4.11, 4.12 and 4.13 demonstrate results of these systems at the Precision, Recall, F-measure and BLEU metric.

Out of these systems, PD and UD-based EB-SMT systems achieves highest BLEU, Precision, Recall and F-Measure score compare with others system while HPBSMT systems lowest. But in the PD and UD, PD's BLEU score were increased +0.24. A perspective of evaluation of the reordering based model 'lexicalized reordering' based EB-SMT systems performance is very low compared with others reordering model while hierarchical reordering results are better.

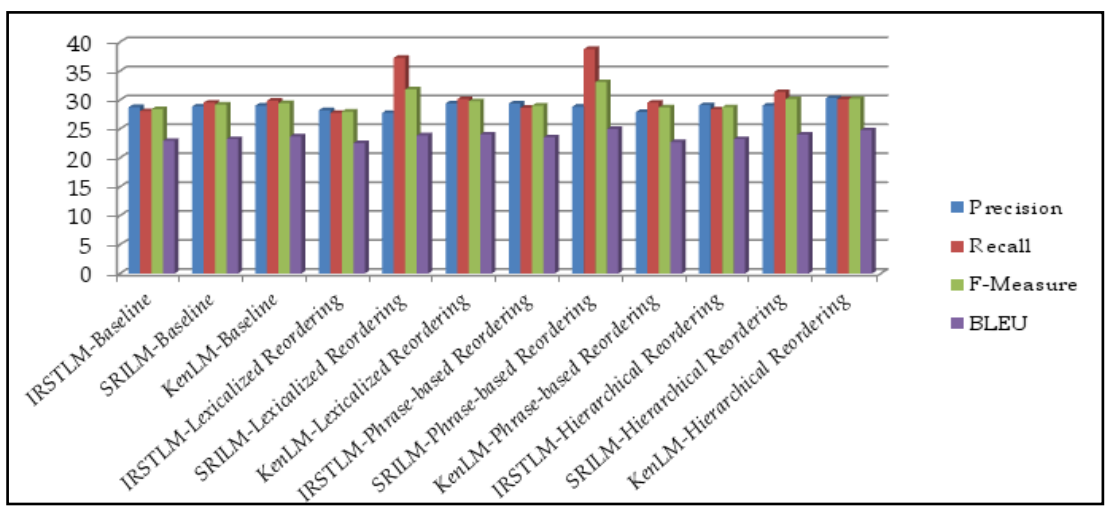

Figure 4. 10: Results of Phrase based EB-SMT Systems

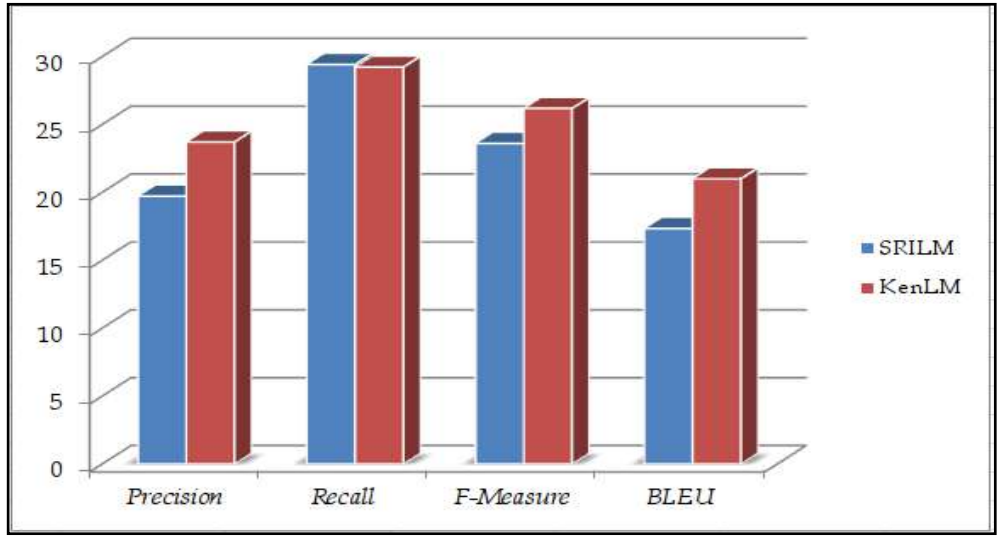

Figure 4. 11: Results of Hierarchical based EB-SMT Systems

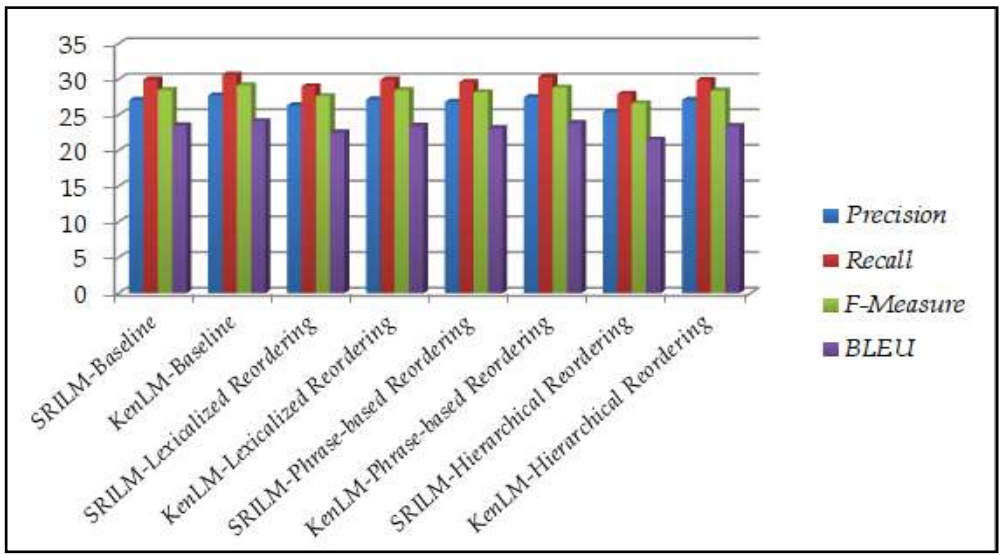

Figure 4. 12: Results of Factor based EB-SMT Systems

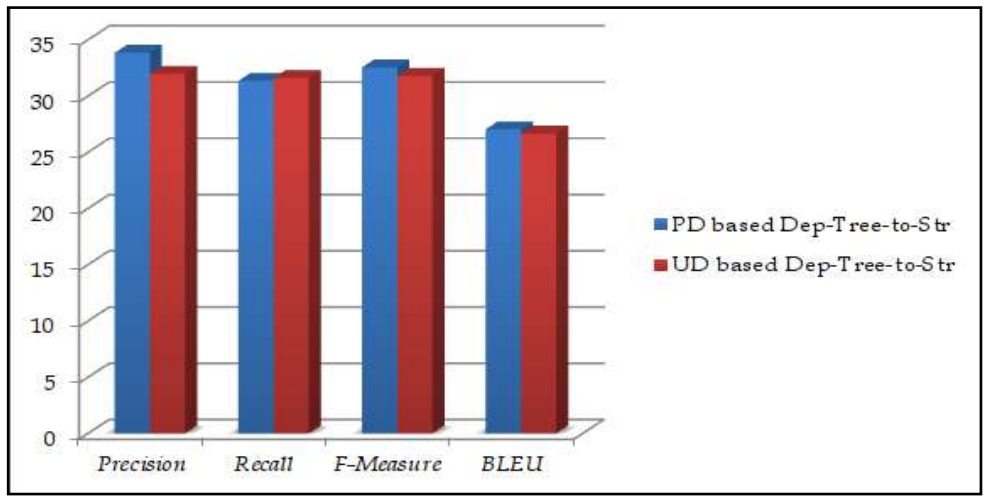

Figure 4. 13: Results of PD and UD based Dep-Tree-to-Str EB-SMT Systems

## 4. 5 GUI of EB-SMT System based on MOSES

Figure 14 demonstrates user interface of the EB-SMT system which is accessible at: http://sanskrit.jnu.ac.in/eb-smt/index.jsp to test the system. Initially, user gives input text or uploads English texts file. Once it has entered or uploaded, it goes for preprocessing such as tokenization, lowercasing of source language sentences. When it is finished, input text goes to developed tuned model file where the "moses.ini" using the related decoder generates best translation of the input sentence. After that translated sentences go for post-processing and that will be displayed on web interface.

Figure 4. 14: Online Interface of the EB-SMT System

## 4.6 Conclusion

This chapter has discussed in detail of experimental part of the developed different SMT systems for English-Bhojpuri language pair which was followed on various translation model: phrase-based, hierarchical phrase-based, factor-based and dependency-based tree-to-string. Hierarchical and Factor based SMT were implemented with two different the KenLM and SRILM LM toolkits while Phrase-based was implemented on three LMs including IRSTLM LM toolkits. The KenLM based LMs have been used to develop dependency-based tree-to-string SMT (PD and UD based).

Out of these discussions, the chapter has also explained language model, word-based, and word-alignment models with the Bhojpuri examples.

Finally, the BLEU score of 24 EB-SMT systems (including 12 PBSMT, 2 HPBSMT, 8 FBSMT and 2 Dep-Tree-to-Str) have been presented. The last section gives a brief idea of online-interface (GUI) of the English-Bhojpuri SMT systems with the web-link. Detailed evaluation and linguistic analysis of the best EB-SMT systems will be discussed in chapter 5.

# Chapter 5

# Evaluation of the EB-SMT System

## 5.1 Introduction

Evaluation is inevitable in the development of an MT system. Without credible evaluation of the final generation text, on the parameters of accuracy, fluency, and acceptability, no claims can be made on the success of the MT system in question. An evaluation task validates how the results of MT systems are inaccurate or insufficient. It is, in fact, a mandatory task for all NLP applications. In comparison to other NLP tasks, MT evaluation is complicated and tougher in whose development several unexpected issues crop up. One of the reasons is the lack of a single method to determine the perfect human translation. For instance, an English sentence can generate multiple sentences in Bhojpuri when translated with the aid of various human translators (as discussed in chapter 3).

There is no consistency in human translations which makes it very difficult to create universally accepted methods for the evaluation of MT systems. Hence, there are no globally agreed and trustworthy methods available (AMTA, 1992; Arnold et al., 1993; Falkedal, 1994). However, there is a common hypothesis and agreement on the basic structures (Hutchins & Somers, 1992; Arnold et al., 1994) of the same. The two most common practices for the MT evaluation are Automatic and Manual/Human evaluation.

In the last chapter, experiments conducted for development of the EB-SMT systems were discussed. Results of 24 EB-SMT systems were reported on Precision, Recall, F-Measure and BLEU metrics. The present chapter evaluates only top two EB-SMT systems based on their performance: PD-based Dep-Tree-to-Str and UD-based Dep-Tree-to-Str EB-SMT systems. These two EB-SMT systems are compared with each other on the basis of error analysis (Automatic and Human evaluation) and linguistic perspectives (Human evaluation).

This chapter is divided into five sections. The second section briefly discusses the methodology of automatic evaluation. It reports the results of PD and UD-based Dep-Tree-to-Str EB-SMT (based on the automatic evaluations) with a focus on error analysis. The third section gives a brief idea of Human evaluation and the methodologies followed to evaluate the systems. This section also reports and discusses the comparative results of the EB-SMT systems based on two human evaluators. The fourth section presents comparative error analysis of the best two EB-SMT systems i.e. PD and UD-based EB-SMT. The fifth and final section concludes the chapter.

## 5.2 Automatic Evaluation

This method evaluates quality of the MT system through a computer program. The primary objective behind this method is to rapidly capture performance of the developed MT systems and being less expensive in comparison to Human evaluation. There are several tested MT evaluation measures frequently used such as Precision (Melamed et al., 2003), Recall (Melamed et al., 2003), F-Measure (Melamed et al., 2003), BLEU (Papineni et al., 2002), WER (Tomás et al., 2003), METEOR (Denkowski and Lavie, 2014), and NIST (Doddington, 2002) etc. These methods which have been used to evaluate EB-SMT systems using the reference corpus[1] (the Precision, Recall, F-Measure, and BLEU metrics scores are already reported for all 24 EB-SMT systems in chapter 4) are briefly explained below.

- **Precision, Recall and F-Measure**

Precision and Recall metrics are widely used in NLP applications such as MT, POS tagger, Chunker, search engine, and speech processing etc. The precision metrics compute correct translated words from the MT output by dividing it with the output-length of the system while the recall metric divides the correct words by the length of reference translation or reference-length. The F-measure metrics is a harmonic mean of the precision and recall metrics or to put it simply, it is a combination of precision and recall (Koehn, 2010). In the MT application, precision metric is more important than the recall metric. A notable drawback of the precision metric is its sole focus on word-matches while ignoring the word-order. WER has to be borrowed from speech recognition to account for word-order. The F-measure is formed to reduce the double counting done by n-gram based metrics such as BLEU and NIST. These metrics can be computed on the following formulas[2] (taken from Koehn, 2010):

$$Precision = \frac{correct}{output-length} \quad (5.1)$$

$$Recall = \frac{correct}{reference-length} \quad (5.2)$$

$$F-measure = \frac{precision*recall}{(precision*recall)/2} \quad (5.3)$$

or

$$F-measure = \frac{correct}{(output-length+reference-length)/2} \quad (5.4)$$

---

1 Reference corpus is known as gold corpus. This type of corpus is prepared by Human. In the case of MT, reference sentences are translated by human instead of computer.

To understand concept of the precision and recall, let us take an example below:

Reference Translation: हम बहरे खेले नाई जाब ।
System A: हमें के भी बहरे ना जाईं ।
System B: हम ना बाहर जाब के चाही ।
System C: हम बहरे खेले नाई जाब ।

In the example illustrated above, System C's output exactly matches with the reference translation which shares all six tokens, while the outputs of System A and B matches one and two tokens (out of six tokens) respectively. The precision results then become: A's:14.28%, B's: 28.57% and C's: 100%. While in the recall, these scores would be: A's:16.6%, B's: 33.33% and C's: 100%.

- **BLEU (Bilingual Evaluation Understudy)**

BLEU is an n-gram based metric, popularly used in the automatic evaluation of an MT system. For each *n*, where *n* ranges from 1 to 4, the BLEU score counts the number of occurrences of n-grams in the candidate translation (MT output). It should display an exact match in the corresponding set of reference translations (human translation) for the same input. BLEU score is commonly computed on an entire test corpus and not on the sentence level. The number of matching n-grams is directly proportional to a higher BLEU score. This score ranges from 0 to 1, where the higher the score the closer the match between reference and candidate translations. A key anomaly of using BLEU is that it assigns a low score despite a candidate translation being able to express the source meaning fluently and precisely but using different lexical and syntactic choices which, although, are perfectly legitimate but are absent in one of the reference translations. Due to these problems, it fails to measure the recall metric.[3] It is also erroneous in the evaluation of the English-Hindi language pair (Ananthakrishnan et al., 2007; Gupta et al., 2010).

---

[2] to know details of these metrics see following articles or book: Tomás et al., 2003 "A Quantitative Method for Machine Translation Evaluation"; Melamed et al., 2003 "Precision and recall of machine translation"; Koehn, 2010 "Statistical Machine Translation"

- **WER (Word Error Rate)**

WER is the first automatic evaluation metric which is applied to SMT (Koehn, 2010). It is adapted from speech recognition systems and considers word order for evaluation. WER works on the 'Levenshtein distance' (Koehn, 2010), and can be defined as the minimum number of editing steps (including insertions, deletions and substitutions). In short, it is the ratio of words, which should be inserted, substituted or deleted in a translation to achieve the reference sentence (Tomás et al., 2003). The WER is computed using the formula,

$$\text{WER} = \frac{substitutions\ + insertions\ + deletions}{reference\ -length\ or\ length\ of\ reference\ translation} \quad (5.5)$$

- **METEOR**

METEOR is another extension of BLEU metrics used to extract the best evaluation report of the MT system. It incorporates the recall metric in the evaluation of the MT system which BLEU doesn't consider. According to Denkowski and Lavie (2014), it evaluates "translation hypotheses by aligning them to reference translations and calculating sentence-level similarity scores". The major advantage of this metric is that it matches MT output to reference translation on the stemming, synonyms or semantically-related word levels which help to retrieve more accurate score as compared to BLEU metric. It also allows the use of Wordnet to disambiguate similar forms or synonyms of the target words. But a drawback of this method is that its formula and method are more complicated than BLEU.

---

3 To know the drawbacks and BLEU methodology see following articles: Ananthakrishnan R et al., 2007 " Some Issues in Automatic Evaluation of English-Hindi MT: More Blues for BLEU", Papineni et al., 2002, "BLEU: a method for automatic evaluation of machine translation").

## 5.2.1 PD and UD-based EB-SMT Systems: Automatic Evaluation Results

In this section WER and METEOR automatic evaluation results of PD and UD-based EB-SMT systems are reported. This section also compares these systems on sentence level based on the scores achieved from the automatic evaluation metrics.

(a) **WER and METEOR Results of the PD and UD-based EB-SMT Systems**

As Figure 5.1 demonstrates, WER results show that the UD-based EB-SMT system generates more error at the word level as compared to PD-based EB-SMT system. At the overall METEOR accuracy, the PD-based EB-SMT system performance is +0.00359 units higher than the UD-based EB-SMT system.

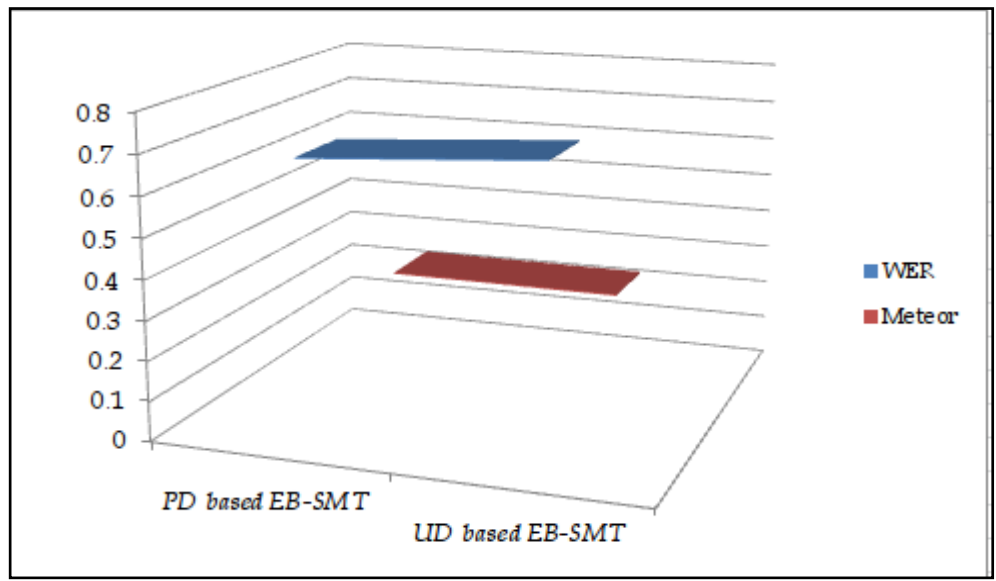

Figure 5. 1: Results of PD and UD based EB-SMT Systems at the WER and Meteor

(b) **A Comparative Analysis of PD and UD based EB-SMT Systems on the METEOR metric**

When we compare the MT output of test data of these systems at the sentence level (Figure 5.2) on the METEOR metric, PD-based EB-SMT's performance is much better reported than UD-based EB-SMT which is analyzed next.

If we analyse Score (Figure 5.2), Precision, and Recall performance of the two systems, then PD-based EB-SMT system has outperformed the UD-based EB-SMT at most instances. There is only one exception seen in the segment range of 150 to 200, in which UD's performance is better than PD. Similarly, the analysis of fragmentation-penalty (Frag) also shows that PD performance is higher as compared the UD (the only exception is in the segment range of 350-400).

When we anlayse these systems at the sentence-length level, the performances of PD and UD-based EB-SMT systems show some variation in the performance at different word-length levels which is described below (shown in figure 5.3):

- **(i)** **1-10 words length level**: the PD-based EB-SMT system has reported highest score at the segment range of 7-10 while the UD-based EB-SMT system reports better performance at the segment range of 6-8.

- **(ii)** **11-25 words length level**: the UD-based EB-SMT system has reported highest score at the segment range of 80-140 while the PD-based EB-SMT system reports better performance at the segment range of 20-60 and above 140.

- **(iii)** **25-50 words length level:** At this level, the UD-based EB-SMT system has reported the highest score at the segment range of 60-75 and 100-120 while the PD-based EB-SMT system reports better performance at the segment range of 80-100.

- **(iv)** **51+ words length level:** At this level, the UD-based EB-SMT system has reported highest score at the segment range of 6-9 and 16-18 while the PD-based EB-SMT system reported better performance at the segment range of 2-4 and 12-16.

**(c) A Comparative Analysis of PD and UD-based EB-SMT Systems at the sentence levels based on the BLEU, Precision, Recall and F-Measure metrics**

Figures 5.4 and 5.5 demonstrate some comparative examples using the above metrics score, out of the 100 best sentences (for (c) and (d), MT-CompareEval toolkit has been used (Klejch et al., 2015)). One analyzing these figures we find most of the time PD achieves 70-100% scores on all metrics (except brevity penalty) as compared to the UD-based EB-SMT system. Even there is a huge difference on the UD-based EB-SMT system output except in Figure 5.5's first and last examples.

### (d) A Comparative Analysis of PD and UD based EB-SMT Systems at the Confirmed and Un-confirmed N-gram levels

A confirmed N-gram refers to the correct match of system output with respect to the reference translation, while an unconfirmed N-gram refers to the incorrect match of the system output with respect to the reference translation.

Figure 5.6 presents the statistics of confirmed N-grams (1 to 4 gram) of PD and UD-based EB-SMT systems (Dep-Tree-to-Str). For each N-gram level (1-4 the top ten N-gram are provided, for both the systems. If we consider top ten 1-gram, then UD has more function words as compared to PD.

The statistics of unconfirmed top ten N-grams (1 to 4 gram) of the PD and UD-based EB-SMT systems are displayed in Figure 5.7. As observed in the confirmed N-gram, the top ten unconfirmed 1-gram have more function words in UD output as compared to PD output. One interesting observation is that in case of PD n-gram statistics (Figures 5.6 and 5.7), the punctuation - question mark '?' - appears only in unconfirmed n-gram.

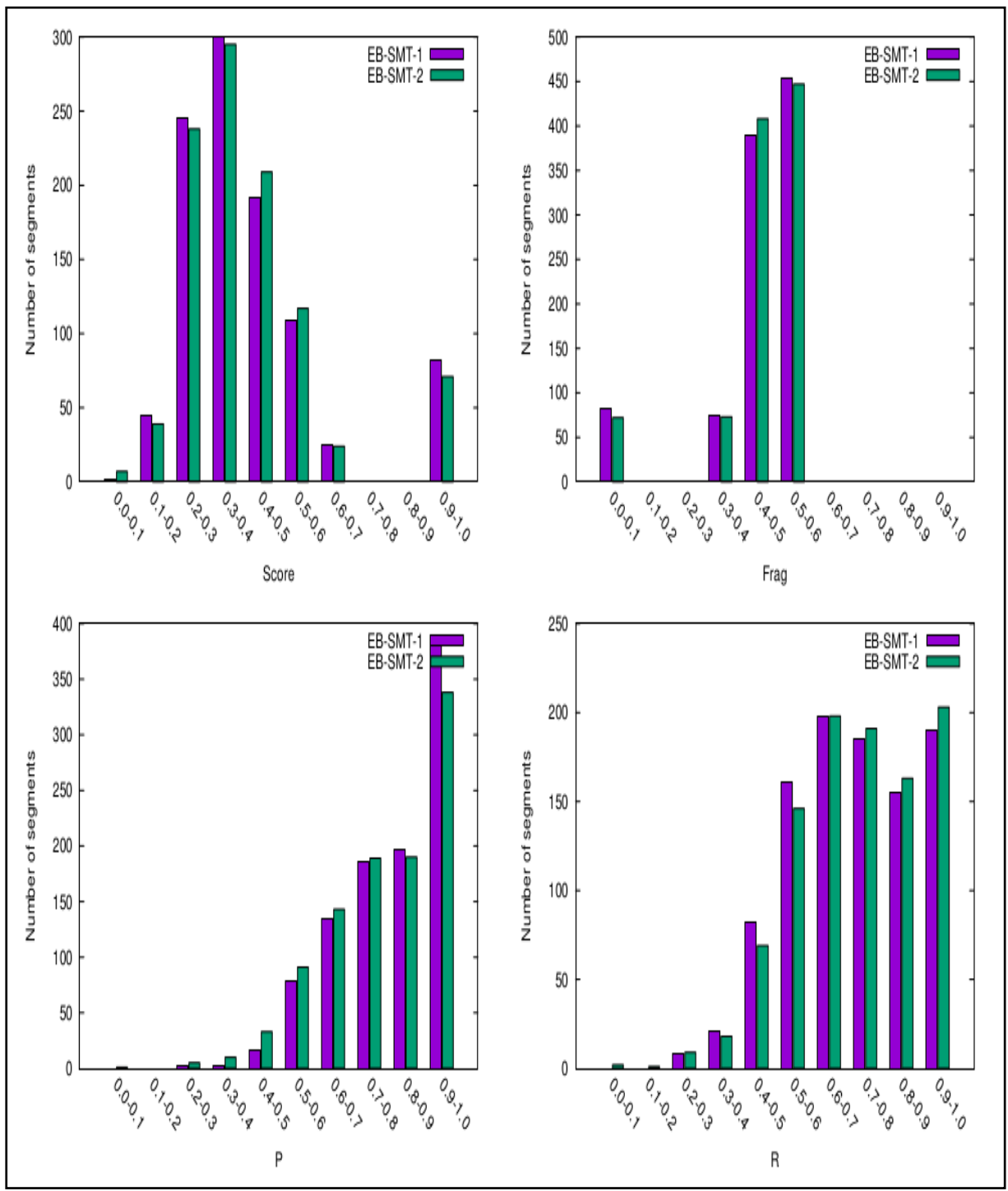

Figure 5. 2: METEOR Statistics for all sentences of EB-SMT System 1 (denotes to PD based) and 2 (denotes UD based)

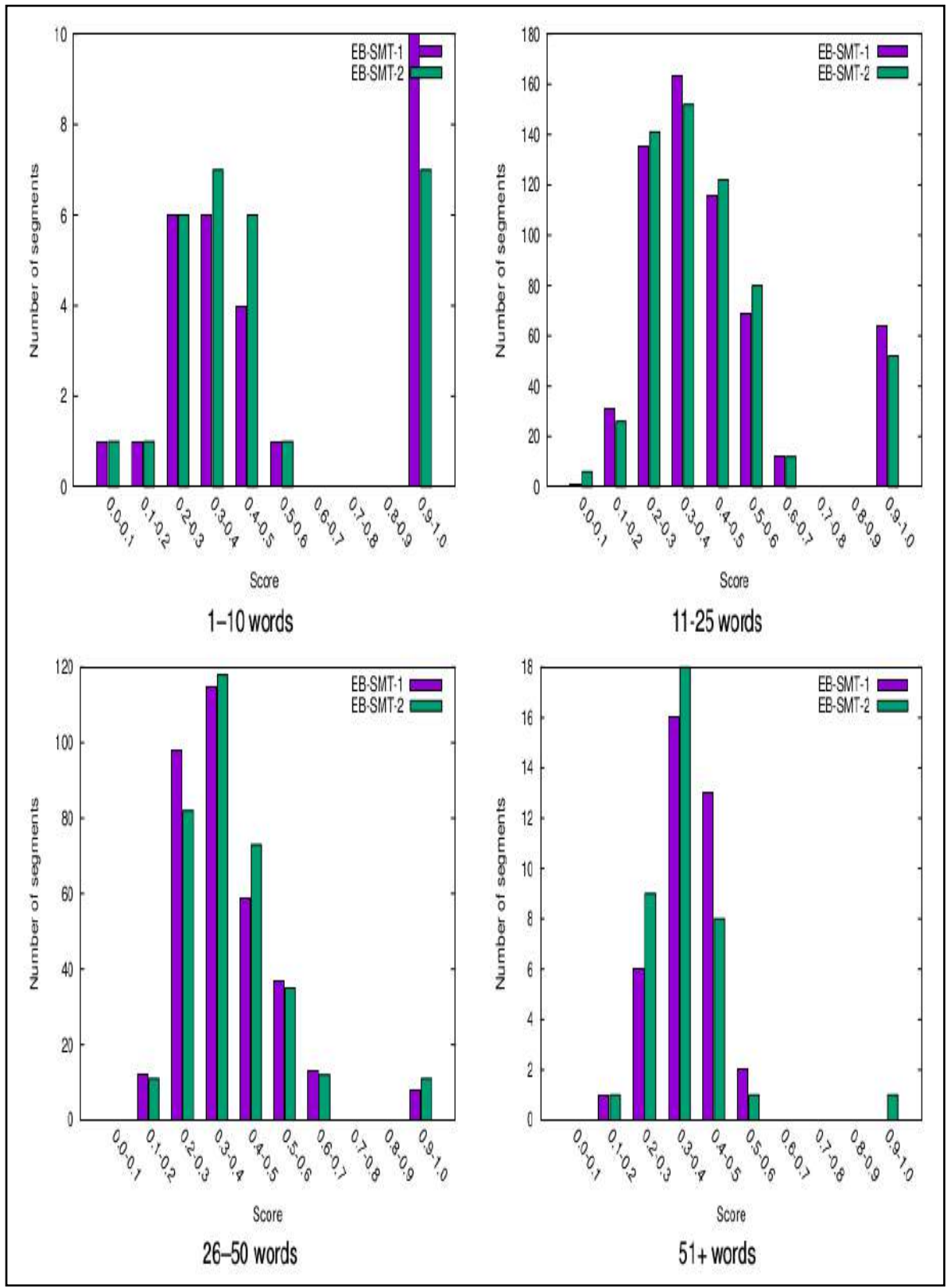

Figure 5. 3: METEOR Scores by sentence length of EB-SMT System 1 (denotes to PD based) and 2 (denotes to UD based)

| | | | | | | |
|---|---|---|---|---|---|---|
| Source | I had to visit an institution . | | | | | |
| Reference | हमके कौनो संस्थान घूमेऽक रहेऽल । | | | | | |
| UD based Dep-Tree-to-Str | हम एगो घूमे संस्था के परल । | | | | | |
| PD based Dep-Tree-to-Str | हम कौनो संस्थान घूमेऽक रहेऽल । | | | | | |
| | BREVITY-PENALTY | BLEU | BLEU-cased | PRECISION | RECALL | F-MEASURE |
| UD based Dep-Tree-to-Str | 1 | 6.57 | 6.57 | 3.57 | 4.17 | 3.85 |
| PD based Dep-Tree-to-Str | 1 | 75.98 | 75.98 | 76.25 | 76.25 | 76.25 |
| Diff | 0.0000 | -69.4100 | -69.4100 | -72.6800 | -72.0800 | -72.4000 |
| Source | I shall return within an hour . | | | | | |
| Reference | हमके एक घंटा में लौटऽक पड़ी । | | | | | |
| UD based Dep-Tree-to-Str | हम एक घंटा के भीतर वापस आ के जाई । | | | | | |
| PD based Dep-Tree-to-Str | हम एक घंटा में लौटऽक पड़ी । | | | | | |
| | BREVITY-PENALTY | BLEU | BLEU-cased | PRECISION | RECALL | F-MEASURE |
| UD based Dep-Tree-to-Str | 1 | 9.29 | 9.29 | 10.28 | 14.88 | 12.16 |
| PD based Dep-Tree-to-Str | 1 | 80.91 | 80.91 | 81.01 | 81.01 | 81.01 |
| Diff | 0.0000 | -71.6200 | -71.6200 | -70.7300 | -66.1300 | -68.8500 |
| Source | I needn ' t get up early tomorrow . | | | | | |
| Reference | हमके बिहान जल्दी उठे क जरूरत नाई ह । | | | | | |
| UD based Dep-Tree-to-Str | हम ना क आवश्यकता उठ बिहान जल्दी । | | | | | |
| PD based Dep-Tree-to-Str | हम बिहान जल्दी उठे क जरूरत नाई ह । | | | | | |
| | BREVITY-PENALTY | BLEU | BLEU-cased | PRECISION | RECALL | F-MEASURE |
| UD based Dep-Tree-to-Str | 0.8824969026 | 11.59 | 11.59 | 16.07 | 14.24 | 15.1 |
| PD based Dep-Tree-to-Str | 1 | 86.33 | 86.33 | 86.36 | 86.36 | 86.36 |
| Diff | -0.1175 | -74.7400 | -74.7400 | -70.2900 | -72.1200 | -71.2600 |

Figure 5. 4: Example-1 of Sentence Level Analysis of PD and UD EB-SMT System

| | BREVITY-PENALTY | BLEU | BLEU-cased | PRECISION | RECALL | F-MEASURE |
|---|---|---|---|---|---|---|
| Source | How can you make a request ? | | | | | |
| Reference | तु कइसे एक विनती कइ सकल ? | | | | | |
| UD based Dep-Tree-to-Str | एक विनती कइ सकल तु कइसे ? | | | | | |
| PD based Dep-Tree-to-Str | तु कइसे एक विनती कइ सकल ? | | | | | |
| UD based Dep-Tree-to-Str | 1 | 50.81 | 50.81 | 57.92 | 57.92 | 57.92 |
| PD based Dep-Tree-to-Str | 1 | 100 | 100 | 100 | 100 | 100 |
| Diff | 0.0000 | -49.1900 | -49.1900 | -42.0800 | -42.0800 | -42.0800 |
| Source | Mary sold the car for susan yesterday . | | | | | |
| Reference | मैरी सुसन खातिर कलिहां कार बेच देहलस । | | | | | |
| UD based Dep-Tree-to-Str | कार बेच मैरी सुसन खातिर कल । | | | | | |
| PD based Dep-Tree-to-Str | मैरी सुसन खातिर कालिहां कार बेच । | | | | | |
| UD based Dep-Tree-to-Str | 0.8668778998 | 27.89 | 27.89 | 38.93 | 33.63 | 36.09 |
| PD based Dep-Tree-to-Str | 0.8668778998 | 72.9 | 72.9 | 84.58 | 71.4 | 77.43 |
| Diff | 0.0000 | -45.0100 | -45.0100 | -45.6500 | -37.7700 | -41.3400 |
| Source | Why are you weeping ? | | | | | |
| Reference | तु काहे रोअत हउअ ? | | | | | |
| UD based Dep-Tree-to-Str | का तु काहे रोअत हउअ ? | | | | | |
| PD based Dep-Tree-to-Str | तु काहे रोअत हउअ ? | | | | | |
| UD based Dep-Tree-to-Str | 1 | 50.81 | 50.81 | 52.5 | 67.92 | 59.22 |
| PD based Dep-Tree-to-Str | 1 | 100 | 100 | 100 | 100 | 100 |
| Diff | 0.0000 | -49.1900 | -49.1900 | -47.5000 | -32.0800 | -40.7800 |

Figure 5. 5: Example-2 of Sentence level Analysis of PD and UD EB-SMT System

## Figure 5.6: Details of PD and UD at the Confirmed N-grams Level

### 1-gram

| PD based Dep-Tree-to-Str wins | | UD based Dep-Tree-to-Str wins | |
|---|---|---|---|
| तु | 176 - 69 = 107 | से | 81 - 56 = 25 |
| नाई | 34 - 24 = 10 | ख्याल | 23 - 2 = 21 |
| लगऽल | 13 - 6 = 7 | हउअ | 71 - 58 = 13 |
| उ | 41 - 35 = 6 | के | 195 - 183 = 12 |
| केतना | 84 - 78 = 6 | कि | 74 - 67 = 7 |
| सम्मान | 10 - 4 = 6 | काहे | 22 - 16 = 6 |
| तोहार | 38 - 33 = 5 | चुकल | 18 - 13 = 5 |
| एगो | 27 - 22 = 5 | आपन | 33 - 28 = 5 |
| बहरे | 10 - 5 = 5 | हो | 13 - 8 = 5 |
| करेंऽल | 5 - 0 = 5 | सकल | 12 - 7 = 5 |

### 2-gram

| PD based Dep-Tree-to-Str wins | | UD based Dep-Tree-to-Str wins | |
|---|---|---|---|
| तु आपन | 13 - 4 = 9 | ख्याल से | 23 - 2 = 21 |
| तु का | 12 - 3 = 9 | हउअ ? | 43 - 30 = 13 |
| के लगऽल | 7 - 0 = 7 | सकल । | 6 - 2 = 4 |
| लगऽल कि | 7 - 0 = 7 | चाहत हउअ | 9 - 5 = 4 |
| क सम्मान | 8 - 1 = 7 | चुकल हउअ | 12 - 9 = 3 |
| तु एगो | 5 - 0 = 5 | तु अपने | 5 - 2 = 3 |
| माहेर के | 5 - 0 = 5 | देख चुकल | 3 - 0 = 3 |
| करेंऽल , | 5 - 0 = 5 | से जॉन | 3 - 0 = 3 |
| सम्मान करेंऽल | 5 - 0 = 5 | से तु | 3 - 0 = 3 |
| एगो बक्सा | 6 - 2 = 4 | करे के | 4 - 1 = 3 |

### 3-gram

| PD based Dep-Tree-to-Str wins | | UD based Dep-Tree-to-Str wins | |
|---|---|---|---|
| के लगऽल कि | 7 - 0 = 7 | चुकल हउअ ? | 11 - 7 = 4 |
| तु आपन झोला | 6 - 0 = 6 | चाहत हउअ ? | 9 - 5 = 4 |
| तु एगो बक्सा | 5 - 0 = 5 | ख्याल से तु | 3 - 0 = 3 |
| सम्मान करेंऽल , | 5 - 0 = 5 | ख्याल से उ | 3 - 0 = 3 |
| क सम्मान करेंऽल | 5 - 0 = 5 | तोहके कवन पसन्द | 3 - 0 = 3 |
| माहेर के लगऽल | 5 - 0 = 5 | देख चुकल हउअ | 3 - 0 = 3 |
| के पसंद करऽल | 4 - 0 = 4 | ख्याल से जॉन | 3 - 0 = 3 |
| टीचर के पसंद | 4 - 1 = 3 | गा सकल । | 2 - 0 = 2 |
| लगऽल कि उ | 3 - 0 = 3 | से ओकर आत्मा | 2 - 0 = 2 |
| कि माहेर अपने | 3 - 0 = 3 | कष्ट देब । | 2 - 0 = 2 |

### 4-gram

| PD based Dep-Tree-to-Str wins | | UD based Dep-Tree-to-Str wins | |
|---|---|---|---|
| क सम्मान करेंऽल , | 5 - 0 = 5 | देख चुकल हउअ ? | 3 - 0 = 3 |
| माहेर के लगऽल कि | 5 - 0 = 5 | तोहके कवन पसन्द हउअ | 3 - 0 = 3 |
| टीचर के पसंद करऽल | 4 - 0 = 4 | खुद कष्ट देब । | 2 - 0 = 2 |
| तु आपन झोला भर | 4 - 0 = 4 | तु खुद कष्ट देब | 2 - 0 = 2 |
| के लगऽल कि उ | 3 - 0 = 3 | ख्याल से ओकर आत्मा | 2 - 0 = 2 |
| अपने टीचर के पसंद | 4 - 1 = 3 | चिल्लात रहत हउअ ? | 1 - 0 = 1 |
| लगऽल कि उ ताकतवर | 2 - 0 = 2 | तु नौकरी काहे छोडल | 1 - 0 = 1 |
| कि उ ताकतवर ह | 2 - 0 = 2 | तोहके जरुर हमके बतावे | 1 - 0 = 1 |
| उमर के लगऽल कि | 2 - 0 = 2 | कपडा तोहार वहा हउए | 1 - 0 = 1 |
| उ ताकतवर ह , | 2 - 0 = 2 | केतना कपडा तोहार वढा | 1 - 0 = 1 |

Figure 5. 6: Details of PD and UD at the Confirmed N-grams Level

## Figure 5.7: Details of PD and UD at the Un-confirmed N-grams level

### 1-gram

| PD based Dep-Tree-to-Str loses | | UD based Dep-Tree-to-Str loses | |
|---|---|---|---|
| तु | 36 - 3 = 33 | तू | 325 - 215 = 110 |
| सोचत | 34 - 16 = 18 | के | 284 - 221 = 63 |
| सकत | 25 - 13 = 12 | का | 210 - 179 = 31 |
| तोहार | 21 - 9 = 12 | बा | 50 - 19 = 31 |
| पसंद | 19 - 9 = 10 | ऊ | 21 - 7 = 14 |
| लगत | 9 - 2 = 7 | रहे | 24 - 10 = 14 |
| क | 27 - 22 = 5 | से | 25 - 13 = 12 |
| अउर | 37 - 32 = 5 | ना | 111 - 99 = 12 |
| कहवाँ | 9 - 4 = 5 | लगे | 63 - 52 = 11 |
| हमके | 10 - 5 = 5 | एगो | 51 - 40 = 11 |

### 2-gram

| PD based Dep-Tree-to-Str loses | | UD based Dep-Tree-to-Str loses | |
|---|---|---|---|
| का तु | 30 - 6 = 24 | का तू | 75 - 38 = 37 |
| हम सोचत | 27 - 8 = 19 | बा ? | 17 - 1 = 16 |
| तु का | 18 - 6 = 12 | हम ख्याल | 14 - 1 = 13 |
| का ह | 14 - 4 = 10 | ना । | 27 - 15 = 12 |
| सोचत कि | 13 - 3 = 10 | का ? | 51 - 40 = 11 |
| ह ? | 14 - 5 = 9 | का बा | 11 - 0 = 11 |
| हम पसंद | 9 - 1 = 8 | तु आपन | 18 - 8 = 10 |
| हम आश्चर्य | 14 - 6 = 8 | के लगे | 23 - 13 = 10 |
| पर । | 13 - 7 = 6 | रहे । | 15 - 5 = 10 |
| उ का | 6 - 0 = 6 | हम का | 16 - 7 = 9 |

### 3-gram

| PD based Dep-Tree-to-Str loses | | UD based Dep-Tree-to-Str loses | |
|---|---|---|---|
| तु का ? | 13 - 0 = 13 | हम ख्याल से | 14 - 1 = 13 |
| का ह ? | 12 - 2 = 10 | का बा ? | 11 - 0 = 11 |
| हम सोचत कि | 10 - 1 = 9 | का तू का | 9 - 3 = 6 |
| करेंऽल , लेकिन | 3 - 0 = 3 | के लगे । | 8 - 2 = 6 |
| बा लेकिन ओकर | 3 - 0 = 3 | तु आपन झोला | 6 - 0 = 6 |
| तु काहे ? | 4 - 1 = 3 | के चाही । | 22 - 17 = 5 |
| मजबूत बा लेकिन | 3 - 0 = 3 | का करबा ? | 5 - 0 = 5 |
| , मजबूत बा | 3 - 0 = 3 | बा कि उ | 5 - 0 = 5 |
| का तु ? | 3 - 0 = 3 | सोचत बा कि | 5 - 0 = 5 |
| सोचत कि जॉन | 3 - 0 = 3 | जाए के पड़ी | 5 - 0 = 5 |

### 4-gram

| PD based Dep-Tree-to-Str loses | | UD based Dep-Tree-to-Str loses | |
|---|---|---|---|
| हम सोचत कि जॉन | 3 - 0 = 3 | सोचत बा कि उ | 5 - 0 = 5 |
| मजबूत बा लेकिन ओकर | 3 - 0 = 3 | जाए के पड़ी । | 4 - 0 = 4 |
| , मजबूत बा लेकिन | 3 - 0 = 3 | तु आपन झोला भर | 4 - 0 = 4 |
| सम्मान करेंऽल , लेकिन | 3 - 0 = 3 | के जाए के पड़ी | 3 - 0 = 3 |
| बा लेकिन ओकर पिता | 3 - 0 = 3 | उमर सोचलस , लेकिन | 3 - 0 = 3 |
| कहलस कि माहेर अपने | 3 - 0 = 3 | ओकर पिता का ख्याल | 3 - 0 = 3 |
| माहेर कहलस कि माहेर | 3 - 0 = 3 | हम आश्चर्य के बात | 3 - 0 = 3 |
| कुत्ता के तु कहले | 3 - 0 = 3 | आश्चर्य के बात कि | 3 - 0 = 3 |
| अउर कहलस कि गुरुजी | 2 - 0 = 2 | हम ख्याल से जॉन | 3 - 0 = 3 |
| हम ना जानत के | 2 - 0 = 2 | होवे के चाही । | 3 - 0 = 3 |

Figure 5. 7: Details of PD and UD at the Un-confirmed N-grams level

## 5.3 Human Evaluation

Along with the automatic evaluation metrics, the human evaluation metrics are also considered while evaluating MT outputs. But this strategy consumes more time and incurs a higher cost as compared to automatic evaluation. Human evaluation is mostly done on sentence-by-sentence basis which makes it cumbersome to issue a judgement on the entire discourse. This anomaly is resolved in automatic evaluation as the measures here are invaluable tools for regular development of MT systems. These measures are only imperfect substitutions for human assessment of translation quality which reveals interesting clues about the properties of automatic and manual scoring. Most MT researchers/evaluators follow metrics of adequacy and fluency in order to judge the MT output. It is very difficult to maintain a consistent standard for fluency and adequacy scale for different annotators. The judgement of humans on several systems also contrasts significantly with the quality of the different systems. But this method is the best strategy to improve any MT system's accuracy, especially for Indian languages.

### 5.3.1 Fluency and Adequacy

Human evaluation is a preferred methodology when more than one translation outputs are available. Most human evaluators were assigned following five point scale (shown in the Table 5.1 and 5.2) to evaluate at the level of fluency and adequacy of MT systems output (taken from the Koehn, 2010; Ojha et al., 2014).

| Fluency | |
|---|---|
| 5 | Flawless of Bhojpuri sentence |
| 4 | Good Bhojpuri sentence |
| 3 | Non-native sentence (like Hindi) |
| 2 | Disfluent |
| 1 | Incomprehensible |

Table 5. 1: Fluency Marking Scale

At this level, evaluator can evaluate the output of MT systems correctly on the given scales. Generally, researchers use a quality scale of 1-5 for fluency and adequacy i.e. 1 for incomprehensible/none, while 5 for flawless/all meaning. After fluency, adequacy is evaluated. For adequacy, the translated output is compared with the reference translation in order to know how natural is the output translation based on the quality scale of 1-5.

Therefore, this strategy has been adopted to evaluate the PD and UD-based EB-SMT systems output.

| Adequacy | |
|---|---|
| 5 | All meaning |
| 4 | Most meaning |
| 3 | Much meaning |
| 2 | Little meaning |
| 1 | None |

Table 5. 2: Adequacy Marking Scale

**(i)     Suggestion for Evaluators to evaluate the Systems outputs**:

The following instructions are given to evaluator for evaluation of the EB-SMT systems output:

- Look at the MT translated output first.
- Evaluate each sentence for its fluency.
- Mark it on the scale 1-5 (according to table 5.1 and 5.2).
- Look at the original source sentence only to verify the faithfulness of the translation (only for reference).
- If the marking needs revision, modify it to the new marking**.**
- After marking at the fluency level look at reference sentence and mark it on adequacy level at the scale of 1-5.

**(ii)    Methodology of EB-SMT Systems testing:**

The same test data[4] was used (which is used for automatic evaluation methods) to evaluate the PD and UD-based EB-SMT systems. Their outputs were then assigned to two evaluators who marked the PD and UD-based EB-SMT systems outputs based on adequacy and fluency levels. If marking is done for N sentences and each of the N sentences is given a mark based on the above scale, the two parameters (on the 5.6 and 5.7) are calculated as follows[5]:

$$\text{Adeuacy} = \frac{(Number\ of\ sentences\ with\ scores\ )}{N} \qquad (5.6)$$

$$\text{Fluency} = \sum_{i=1}^{N} Si/N \qquad (5.7)$$

---

4  See chapter 4 for details
5  To know more, see Ojha et al., 2014 "Evaluation of Hindi-English MT Systems".

### 5.3.2 PD and UD-based EB-SMT Systems: Human Evaluation Result

On the basis of human evaluation methodology described above, the PD and UD-based EB-SMT systems have been evaluated. Figure 5.8 demonstrates a comparative result of the PD and UD-based EB-SMT systems of the two evaluators. In this, PD-based system achieves highest fluency for both evaluators. At the adequacy level, on the other hand, the UD-based EB-SMT system received highest scores by evaluator-1 while evaluator-2 gave the highest score to PD-based EB-SMT system. When we take averages of the adequacy and fluency scores of both evaluators, we find that UD's adequacy score (approx. 80.46%) is higher as compared to the PD-based EB-SMT system's adequacy score (approx. 75.77%) while at the level of fluency, the PD's system performance is higher (approx. 62%) as compared that of the UD (57.63%).

When we closely observe the PD and UD-based EB-SMT systems' evaluation report of fluency (shown in figure 5.9) at each level, we find that PD has received a score above 50% for the '2' scale (except the evaluator-2 score). Another observation is that PD-based EB-SMT system received lowest score for '4' scale by all evaluators as compared to the UD-based EB-SMT system.

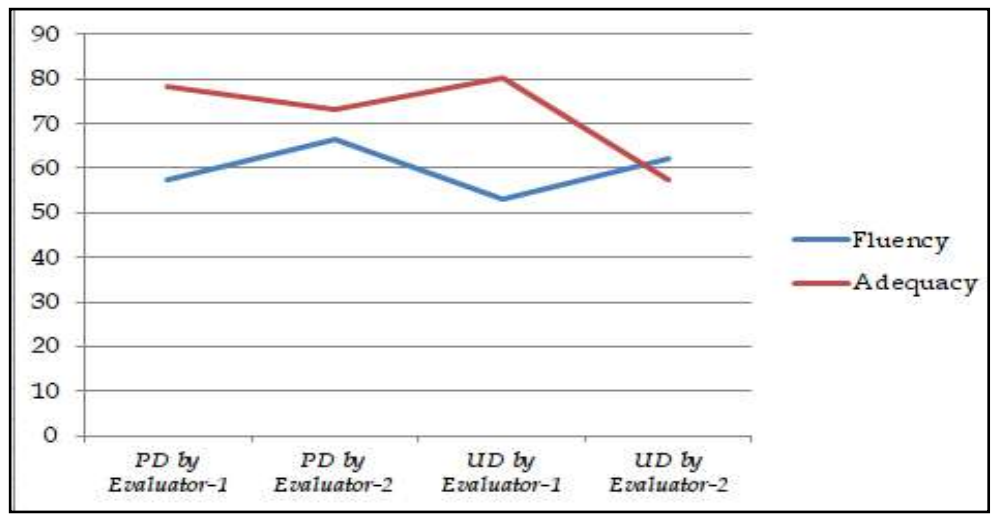

Figure 5. 8: A comparative Human Evaluation of PD and UD based EB-SMT Systems

On the other hand, the analysis of adequacy scores on each level shows that UD-based EB-SMT system received the highest score by all the evaluators for '2' scale as compared to that of the PD-based EB-SMT system (shown in figure 5.10). In another instance, the UD system did not receive any score for '3' and '1' scale by evaluator-2.

A similarity observed for both systems is that the overall score received from both the evaluators is not above 10% for '5' scale, for both systems.

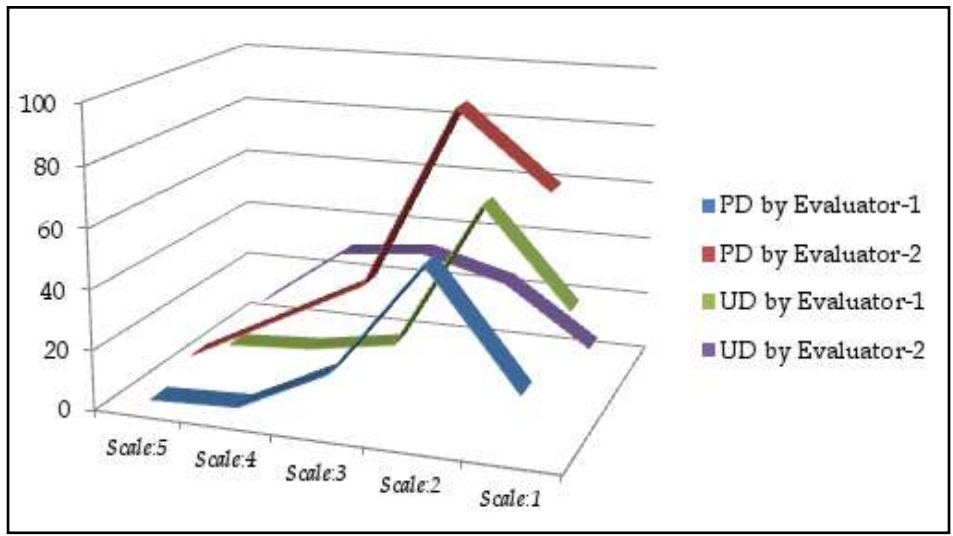

Figure 5. 9: PD and UD-based EB-SMT Systems at the Levels of Fluency

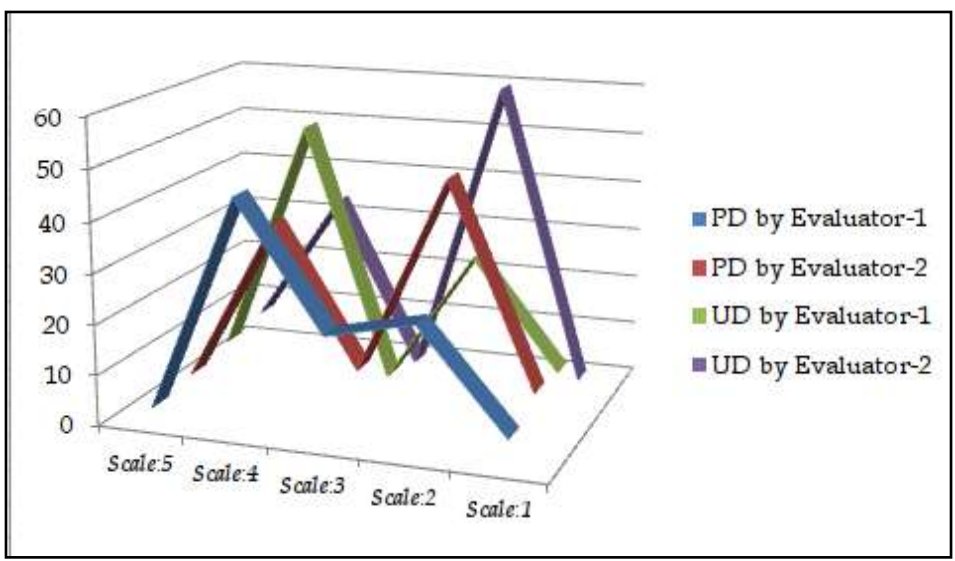

Figure 5. 10: PD and UD-based EB-SMT Systems at the Levels of Adequacy

**5.4 Error Analysis of the PD and UD based EB-SMT Systems**

In the translation process, the text of source language gets decoded, which is then encoded into the target language. One of the main challenges for translation is the linguistic divergences between the source and target languages at several levels. In case of MT, several others problems are also encountered. These problems occur at the level of decoding which raises several challenges for MT.

The errors in the translated MT outputs are categorised into the following three main levels (Chatzitheodorou et al., 2013). Each of these three main error categories is sub-divided into several sub-levels as explained below:

- **Style:** The level of 'Style' is concerned with style and formatting of the MT output. The stylistic issues are related to incorrect formats of addresses, dates, currency, errors of incorrect accents, misspelled words, incorrect punctuation, incorrect capitalisation and abbreviations, etc.
- **Words:** The level of 'Words' deals with vocabulary usage. This level is also divided into various sub-categories. For instance, single words error, the errors of wrong translation of idioms, un-translated words which are not found in the data, literal translation etc.
- **Linguistic**: The linguistic level focuses on the linguistic and grammatical aspects of the MT outputs. It consists of sub-levels of inflection errors (nominal, verbal and others), the error of wrong category selection, error in function words translation like articles and ad-positions, and the errors related to agreement.

The analysis of the EB-SMT outputs also results in several problems primarily related to agreement (like gender agreement), Named Entity Recognition (NER), structural mapping, and nominal inflection error (plural morphemes).

The following errors were encountered during the evaluation of PD and UD-based EB-SMT systems:

**(A) Errors on Style Level:**

**(a) Generation of Extra Word(s):**

In this category, an error is made because a few extra word(s) are generated in the MT output. For instance, in the example (I), if we compare UD and PD EB-SMT outputs with the reference translation, the word 'भी' is found to be an extra in both UD and PD-based EB-SMT outputs. Similarly, in the example (II), the UD EB-SMT output generates extra words 'नकार देहल'.

(I) I also have to go to the market.                (SL)

हमहू के भी बजार जाएके हऽ ।                          (PD-EB-SMT Output)

hamahU ke bhI bAzAra jAeska ha .                    (ITrans)

हम भी बाजार के जाए के पड़ी ।                         (UD-EB-SMT Output)

hama bhI bAjAra ke jAe ke pa.DI .                   (ITrans)

*Reference Translation of the TL:* हमहू के बजार जाएके हऽ ।

*ITrans of the reference Translation:* hamahU ke bajAra jAeska has.

(II) I regret but I am compelled to deny.           (SL)

हमके अफसोस हऽ लेकिन हम इनकार करे ----

के मजबूर हईं।                                        (PD-EB-SMT Output)

hamake aphasosa has lekina hama inakAra kare ----

ke majabUra haIM.                                   (ITrans)

हम लेकिन अफसोस इनकार करे के मजबूर हईं। नकार देहल।    (UD-EB-SMT Output)

hama lekina aphasosa inakAra kare ke majabUra haIM. ----

nakAra dehala .                                     (ITrans)

*Reference Translation of the TL*: हमके अफसोस बा लेकिन हम इनकार करे के मजबूर हईं ।

*ITrans of the reference Translation*: hamake aphasosa bA lekina hama inakAra kare ke majabUra haIM.

**(b) Spelling and Punctuation Errors:**

This error type classifies the errors related to the spelling and punctuation in the MT output. The punctuation errors can be illustrated with the help of examples in (II) and (III). In (II), the UD EB-SMT output has an extra punctuation mark 'ा' and in (III), both PD and UD EB-SMT systems outputs have generated an extra punctuation 'ा' at the end of the output sentence.

The spelling error is observed in the example (III), in which both UD and PD EB-SMT systems outputs have 'तू' in place of 'तु' of the reference translation.

(III) You are the chieftain of the central panchayat.        (SL)

 तू केन्दरिय पन्चायत के मुखिया हउअ ? ।        (PD-EB-SMT Output)

 tU kendariya pancAyata ke mukhiyA haua ?        (ITrans)

 तू केन्दरिय पन्चायत के मुखिया हउअ ? ।        (UD-EB-SMT Output)

 tU kendariya pancAyata ke mukhiyA haua ? .        (ITrans)

*Reference Translation of the TL*: तु केन्दरिय पन्चायत के मुखिया हउअ ?

*ITrans of the reference Translation*: tu kendariya pancAyata ke mukhiyA haua ?

**(B) Errors on Word Level:**

Two types of word level errors have been observed: WSD (word sense disambiguation) and Un-translated and Multi-word units error. These errors are further explained.

**(c)  WSD error:**

WSD error occurs when the EB-SMT output is unable to disambiguate the source text and therefore provide wrong word in the EB-SMT output. This is illustrated clearly in (IV), in which 'कपार' of the reference translation is wrongly translated as 'मुखिया' in both UD and PD-based outputs. An explanation of this error could be the inability of both the EB-SMT systems to disambiguate two different senses of source language word 'head'.

(IV) I feel a severe pain in my head.        (SL)

 हमार कपार भयानक पिराऽत मुखिया ।        (PD-EB-SMT Output)

 hamAra kapAra bhayAnaka pirAsta mukhiyA .        (ITrans)

 हमार  भयानक पिराऽत मुखिया ।        (UD-EB-SMT Output)

 hamAra mukhiyA bhayAnaka pirAsta.        (ITrans)

*Reference Translation of the TL*: हमार कपार भयानक पिराऽत।

*ITrans of the reference Translation*: hamAra kapAra bhayAnaka pirASta.

**(d) Un-translated and Multi-word units error**

In the example (V), 'becoming' and 'impudent' are not translated by both EB-SMT systems. Error in multi-word units also occurs in the same example: 'अधिक से अधिक' of PD-EB-SMT output has occurred instead of 'बहुते' of the reference translation.

(V) You are becoming more and more impudent day by day.     (SL)

का तू <u>becoming</u> अधिक से अधिक <u>impudent</u> अउर दिन दिन ।     (PD-EB-SMT Output)

kA tU becoming adhika se adhika impudent aura dina dina . (ITrans)

तु <u>becoming</u> द्वारा और ढेर दिन दिन <u>impudent</u> दिहल।     (UD-EB-SMT Output)

tu becoming dvArA aura Dhera dina dina impudent dihala.  (ITrans)

*Reference Translation of the TL*: तु दिन ब दिन बहुते बेशरम होत रहत हउअ ।

*ITrans of the reference Translation*: tu dina ba dina bahute besharama hota rahata haua .

**(C) Error on Linguistic Levels:**

At the linguistic level, we have found errors in inflection and structure which are described below.

**(e) Issues with Inflections**

In the example (VI), 'हमरे' of the reference translation is translated as 'हम' by both PD and UD EB-SMT system outputs. These inflectional errors occur in the output because the target language 'Bhojpuri' is morphological richer than the source language 'English'. One of the reasons of such errors could be the fact that there is no linguistic information provided for the target language (generation part).

(VI)  I think Ram is going to marry Sita.     (SL)

<u>हम</u> सोचत राम सीता से बियाह करे जाऽत हऽ ।     (PD-EB-SMT Output)

hama socata rAma sitA se biyAha kare jAऽta haऽ .     (ITrans)

<u>हम</u> ख्याल से राम सीता से बियाह करे जाऽत हऽ ।     (UD-EB-SMT Output)

hama khyAla se rAma sitA se biyAha kare jAऽta haऽ.     (ITrans)

*Reference Translation of the TL*: हमरे ख्याल से राम सीता से बियाह करे जाऽत हऽ ।

*ITrans of the reference Translation:* hamare khyAla se rAma sitA se biyAha kare

jAsta has.

**(f) Structural Issues**

In example (VII), in the UD-EB-SMT output 'भी' should occur after 'हम' instead of occurring at the end of the sentence.

(VII) I also don't know.   (SL)

    हम भी पता ना ।   (PD-EB-SMT Output)

    hama bhI patA nA .   (ITrans)

    हम ना जानत भी ।   (UD-EB-SMT Output)

    hama nA jAnata bhI .   (ITrans)

*Reference Translation of the TL*: हमहू नाई जानीऽल।

*ITrans of the reference Translation:* hamahU nAI jAnIsla.

**5.4.1 Error-Rate of the PD and UD-based EB-SMT Systems**

The Tables 5.3, 5.4 and 5.5 describe all the error sub-levels (of main error levels: Style, Word and Linguistic) for the PD and UD-based EB-SMT systems.

At the style error level (Table 5.3), both PD and UD systems produce maximum errors in the sub-level of 'Generation of Extra Words' while the lowest number of errors occur in the sub-levels of 'Spelling errors and Country standards' (for PD EB-SMT system) and in the sublevel of 'Punctuation' (for UD EB-SMT system).

| Sub-Levels of Style | PD-EB-SMT System | UD-EB-SMT System |
|---|---|---|
| Acronyms and Abbreviations | 34 | 108 |
| Generation of Extra words | 81 | 156 |
| Country standards | 8 | 25 |
| Spelling errors | 8 | 54 |
| Issues with Accent level | 15 | 39 |
| Punctuation | 17 | 7 |

Table 5. 3: Statistics of Error-Rate of the PD and UD based EB-SMT Systems at the Style Level

The Table 5.4 provides the error details of both PD and UD-based EB-SMT systems at the 'Word' error level. The error sub-level 'Single words' has received maximum errors for both PD and UD systems while the lowest number of errors occurs in the sub-level of 'Conjunctions' for both PD and UD EB-SMT systems.

|  | **PD-EB-SMT System** | **UD-EB-SMT System** |
|---|---|---|
| **Single words** | 213 | 348 |
| **Multi-word units** | 126 | 72 |
| **Terminology** | 4 | 51 |
| **Un-translated words** | 118 | 136 |
| **OOV (Out of Vocabulary)** | 30 | 23 |
| **Ambiguous translation** | 54 | 58 |
| **Literal translation** | 178 | 111 |
| **Conjunctions** | 1 | 3 |

Table 5. 4: Statistics of Error-Rate of the PD and UD based EB-SMT Systems at the Word Level

At the linguistic error level (Table 5.5), both PD and UD-based EB-SMT systems produce maximum errors in the sub-level of 'Verb inflection' while the lowest number of errors occur in the sub-levels of 'Article and Agreement' (for PD EBSMT system) and in the sub-level of 'Agreement' (for UD EBSMT system).

|  | **PD-EB-SMT System** | **UD-EB-SMT System** |
|---|---|---|
| **Verb inflection** | 123 | 216 |
| **Noun inflection** | 21 | 3 |
| **Other inflection** | 148 | 201 |
| **Wrong category** | 49 | 51 |
| **Article** | 1 | 165 |
| **Preposition** | 108 | 89 |
| **Agreement** | 1 | 2 |

Table 5. 5: Statistics of Error-Rate of the PD and UD based EB-SMT Systems at the Linguistic Level

**5.5 Conclusion**

In this chapter, out of the 24 (12 PBSMT, 2 HPBSMT, 8 FBSMT and 2 Dep-Tree-to-Str) EB-SMT systems developed, top two - PD and UD-based on Dep-Tree-to-Str EB-SMT systems have been evaluated and compared using various automatic and human evaluation methods. In these evaluations, we can analyze that the performance of PD's system is slightly better than UD's system. Out of automatic and human evaluations, the systems generated output were deeply anlaysed on three broader translation error levels, namely, style, word and linguistic level, which are explained with outputs of both systems. Finally the chapter has reported statistics of error-rate on the error levels of three translations.

# Chapter 6

# Conclusion and Future Work

MT has always been the focus of research and development in NLP. New ideas and methodologies have shaped and contributed towards the advancement of the field of MT. One of the important MT types is the SMT which has become very influential in the last two decades. The present PhD research concerns itself with the development of twenty four SMT systems for the English-Bhojpuri language pair. This research is quite significant as there is no SMT system available for this language pair. Another significance of the research is the creation of LT resources for the target language 'Bhojpuri' which is a low-resourced language. Various issues and challenges faced and tackled during the course of the research work have also been discussed. A brief explanation to the thesis has been provided below by providing a brief summary of the chapters.

Chapter 1 introduces this PhD research by detailing the motivation of the study, the methodology used for the study and the literature review of the existing MT related work in Indian languages.

Chapter 2 explores the feasibility of the Kāraka model (Pāṇinian Dependency) for enhancing the linguistic information which would help in better performance of the SMT system. The chapter also takes into account the usefulness of another popular dependency framework 'Universal dependency' which has been recently used for several cross-lingual tasks.

Chapter 3 gives details of various LT resources created for Bhojpuri. It should be mentioned here that although Bhojpuri is spoken by a large population, there is no publically available corpus for monolingual Bhojpuri or parallel English-Bhojpuri language pair. The initial resources created for the present study were a monolingual Bhojpuri corpus and a parallel English-Bhojpuri corpus. Both of these corpora were annotated with BIS-based POS tag-set. For the parallel English-Bhojpuri corpus, the source language English was also annotated at the dependency level. Two dependency frameworks were used for annotation: the Kāraka model based dependency (Pāṇinian

Dependency) and Universal Dependency (UD). The tools and methodology employed in this chapter have proven to be quite useful. Since the LT resources created for Bhojpuri and English are initial efforts, further enrichment of these resources for further research work is planned.

Chapter 4 discusses in detail the various experiments conducted for the development of 24 EB-SMT systems. These systems were developed using various translation models which are explained in this chapter. The statistical information of the data used for training, tuning and testing has also been provided. This chapter also reports the results of the 24 EB-SMT systems at the Precision, Recall, F-Measure and BLEU metrics.

Chapter 5 focuses on the evaluation task of the top two EB- SMT systems (PD based Dep-Tree-to-Str and UD based Dep-Tree-to-Str EB-SMT systems) which have given the best performance. For the purpose of evaluation, both Automatic and Human evaluation methods were used. One of the important result that came out this study is the better performance of the PD based EB-SMT system which performed better than all other EB-SMT systems.

**Future scope or extension of the study**:
1. One of the future goals is to improve the accuracy of the existing EB-SMT system by using transliteration, NER and hybrid methods (such as Placeholder method etc.).
2. In order to keep the scope of this study restricted, I have provided dependency annotation only for the source language, English. The next task would be to annotate target language, Bhojpuri, of parallel English-Bhojpuri corpus with Pāṇinian Dependency framework and to develop SMT system using tree-to-tree and string-to-tree models.
3. As mentioned earlier, the corpus created in this study is first of its kind which will be made available publically. The date size of parallel corpus will be further increased to 100K sentences to enhance accuracy and capture variation in Bhojpuri.
4. The present study focuses on the SMT methods. A future goal is to conduct the experiment using the Neural/Deep learning methods.

5. An extension of the present research to develop the SMT system for the Bhojpuri – English language pair will contribute to develop the bi-directional SMT system of this language pair.
6. There are many Indian languages which are low-resourced. One of the proposed future works is to explore the contribution of the present research for conducting a similar work in other Indian languages.

This PhD work has been a fruitful research endeavour in which SMT system for English-Bhojpuri language pair has been developed. Experiments with various methods and approaches were conducted to explore their feasibility for this work. The results and output of the research have been satisfying, however, there is still scope of improvement and further research work.

# Appendix 1

## Home Page of EB-SMT System with Input Sentence

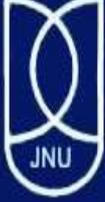

Online interface of the EB-SMT system is accessible at http://sanskrit.jnu.ac.in/eb-smt/index.jsp

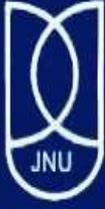

**Output of the EB-SMT system**

**Appendix 2**

**Sample Set of Bhojpuri Monolingual Corpus**

| ID | Sentences |
|---|---|
| BHSD01 | एतना सुनते हमार मतारी कहलस - " हे महारानी जी , जदि रउआ सचमुच अपना बचन के पक्का बानी त हमार एगो प्रार्थना बा कि अब आगे से हमरा एह प्रदेश के सब काम - काज , शिक्षा - दीक्षा खाली फ्रेंच में होखो , जर्मन में ना । |
| BHSD02 | एतना सुनके महारानी अचम्भा में पड़ गइली । |
| BHSD03 | खीसे काँपै लगली । |
| BHSD04 | उनकर आँख लाल हो गइल । |
| BHSD05 | बाकिर ऊ करस त का करस , बचन हार चुकल रहस । |
| BHSD06 | उनका कहे के पड़ल – ऐ लइकी ! नेपोलियन के सेनो जर्मनी पर कबो इतना कठोर प्रहार ना कइले रहे , जतना तें शक्तिशाली जर्मनी साम्राज्य पर कर देले बाड़े । |
| BHSD07 | जर्मनी के महारानी तोरा अइसन छोट बच्ची से बचन देके हार गइल । |
| BHSD08 | ई हम जिनगीभर ना भुला सकीं । |
| BHSD09 | जर्मनी जवन अपना बाहुबल से जीतले रहे , ओकरा के तूँ अपना बिबेक आ बानी से फेरु जीत लिहले । |
| BHSD10 | अब हम भलीभाँति जान गइनी कि लॉरेन प्रदेश अब अधिका दिन जर्मनी के अधीन ना रह सके । |
| BHSD11 | एतना कहके महारानी तेजी से उल्टे पाँव लौट गइली । |
| BHSD12 | डॉ० रघुवीर , एह घटना से रउआ आभास हो गइल होई कि हम कइसन मतारी के बेटी हईं । |
| BHSD13 | हम फ्रेंच भाषा - भाषी संसार में सबसे अधिक मान - सम्मान आ गौरव अपना मातृभाषा के दिहिले , काहेकि हमनी खातिर राष्ट्र प्रेम आ मातृभाषा प्रेम में कवनो अंतर नइखे । |
| BHSD14 | हमार आपन भाषा मिल गइल , त आगे चलके हमनी के जर्मनी से आजादियो मिल गइल । |
| BHSD15 | डॉ० रघुवीर ! रउआ अब जरूर समुझ गइल होखब जे हम का कहे के चाहत बानी । |
| BHSD16 | डॉ० रघुवीर लमहर सांस लेत मुट्ठी भींच लेलन । |
| BHSD17 | जइसे कुछ मन ही मन संकल्प कइले होखस । |
| BHSD18 | एकरा के कहल जाला भाषिक गौरवबोध आ मातृभाषिक अस्मिताबोध। |
| BHSD19 | बिहार विधानसभा के चुनाव अबहीं टटका मुद्दा बा बाकिर घूमा फिरा के ई मुद्दा हमेशा बनल रहेला कि पत्रकार के राय कवना गोल के बा। |
| BHSD20 | कहे खातिर त सगरी पत्रकार हमेशा तटस्थता आ ईमानदारी के चोला भा बुरका पहिरले रहेलें बाकिर तनिका धियान दे दीं त ओह चोला भा बुरका का पाछा झलकत उनकर असली चेहरो लउक जाई। |
| BHSD21 | ईमानदारी के हाल त ई बा कि अधिकतर स्थानीय पत्रकारन के लोग ब्लैकमेलर जइसन बुझेला। |
| BHSD22 | हमेशा ओकरा से डेराइल सहमल रहेला कि पता ना कब कवन बात ऊ अपना मीडिया में केकरा खिलाफ लिख पड़ दी। |
| BHSD23 | एहसे ऊपर उठीं त देखब कि करीब करीब सगरी बड़का चैनल के एंकर मोदी भंजक लउकेलें। |
| BHSD24 | ओह लोग के हमेशा कोशिश रहेला कि कवनो बात , कवनो खबर के घुमा फिरा के मोदी का खिलाफ इस्तेमाल क लीहल जाव । |
| BHSD25 | अब आगे बढ़े से पहिले हम रउरा सभ के बता दीहल जरुरी समुझत बानी कि हम बचपन से संघी हईं आ भाजपा से सहानुभूति राखीले। |
| BHSD26 | संघ के शाखा में गइला युग से अधिका हो गइल बाकिर संघ के विचार से निष्ठा जवन एक बेर बन गइल तवन आजु ले बनल बा। |
| BHSD27 | बाकिर का ई सगरी पत्रकारन ला जरूरी ना होखे के चाहीं कि ऊ खुलेआम बतावसु कि ऊ कवना राजनीतिक गोल के तरफदारी करेलें भा कवना गोल के सदस्य हउवें। |
| BHSD28 | ढेर दिन ना भइल जब एगो मीडिया एंकर खुलेआम अपना प्रोग्राम में आआपा के समर्कथन कइल करसु आ बाद में बाकायदा ओकरा में शामिल हो गइलन। |
| BHSD29 | बाकिर जब ले उनुका के चैनल से निकाल बाहर ना कइल गइल तबले ऊ अपना के निष्पक्ष पत्रकारिता के झंडाबरदार बतावे से बाज ना आवत रहसु। |
| BHSD30 | नाम बतावल जरुरी नइखे बाकिर हम आशुतोष के बात करत बानी। |
| BHSD31 | अइसने एगो बहुते क्रांतिकारी पत्रकार आ चैनल एंकर हउवन पुण्य प्रसुन बाजपेयी जे लाख बेइज्जति भइला का बादो आजु ले आपन राजनीतिक विचारधारा के खुलासा नइखन कइले। |
| BHSD32 | अब अइसन पत्रकारन के एंकर कइल प्रोग्राम कतना ईमानदार होखी से सभे जानत बा। |
| BHSD33 | केकर केकर लीहीं नाम , कमरी ओढ़ले सगरी गाँव। |
| BHSD34 | पिछला सरकार का बेरा ले मीडिया के लोग के हरामखोरी के भरपूर मौका मिलत रहुवे। |
| BHSD35 | सरकारी खरचा पर देश विदेश घूमे के मौको मिल जात रहुवे आ ऊ लोग आजु ले अपना मलकिनी के चरण वंदना करत चारण बनल बाड़ें। |
| BHSD36 | एह पत्रकारन के बेंवत नइखे जे ई राहुल गाँधी भा सोनिया गाँधी से सवाल कर सकसु। |
| BHSD37 | एक तरह से देखीं त लाख दुर्गुण होखला का बावजूद एह लोगन में स्वामिनभक्ति के गुण अतना बा कि कुकुरो लजा जइहें। |
| BHSD38 | बिहार विधानसभा चुनाव का मौका पर फेरू एह बिकाऊ आ भाँड़ मीडिया के नौटंकी खुलेआम चालू बा बाकिर कवनो पत्रकार में अतना ईमानदारी नइखे जे ऊ पहिले बता देव कि ओकर सहानुभूति कवना गोल भा गठबन्हन से बा। |
| BHSD39 | हमरा समुझ से हर पत्रकार आ चैनल एंकर खातिर जरूरी होखे के चाहीं कि हर प्रोग्राम से पहिले ओकर राजनीतिक विचारधारा खुलेआम बता देसु। |
| BHSD40 | साथ ही इहो जरूरी होखे के चाहीं कि चैनल भा अखबार छोड़ला का बाद तीन बरीस ले एह लोग के कवनो सरकारी पद सकारे भा चुनाव लड़े के मौका मत दीहल जाव। |
| BHSD41 | जइसे न्यायपालिका के शुचिता बनावल राखे खातिर व्यवस्था कइल जाला वइसहीं पत्रकारिता के शुचिता बनवले राखे खातिर पत्रकारो लोग पर समुचित लगाम लगावे के व्यवस्था होखे के चाहीं। |
| BHSD42 | आज दुनिया के बहुते भाषा मर - बिला रहल बाड़ी सऽ। |
| BHSD43 | एकर मतलब ई ना भइल कि जवन भाषा कवनो क्षेत्र - विशेष आ उहाँ के जन - जीवन में जियतो बाड़ी सऽ उन्हनियो के मुवल - बिलाइल मान लिहल जाव, जवन आजुओ अपना सांस्कृतिक समाजिक खासियत का साथ अपना 'निजता' के सुरक्षित रखले बाड़ी सऽ। |
| BHSD44 | सोच - संवेदन आ अनुभूतियन के सटीक, प्रासंगिक अभिव्यक्ति देबे में सक्षम आ शब्द - संपन्न अइसन भाषा - सब के सम्मान सहित सकारे में बहुतन के अजुओ हिचक बा। |
| BHSD45 | इन्हन कऽ स्वाभाविकता गँवारू आ पछुवाइल लागत बा आ भद्रजनन (?) का अंगरेजियत का आगा भाषा के संपन्नतो उपेक्षित हो जातिया। |
| BHSD46 | अइसनका 'शिष्ट - विशिष्ट' नागर लोगन का जमात में लुगरी वाला 'लोक' के अर्थ पिछड़ापन रहल बा। |
| BHSD47 | ऊ लोग अपना बनावटी आडंबर आ बड़प्पन में लोकभाषा के सुभाविक सिरजनशीलता के क्षणिक - मनरंजन के चीजु मानला। |
| BHSD48 | ई अंगरेजी दाँ अभिजात हिन्दीवादी लोग 'अनेकता में एकता' के गान करत ना अघाला, बाकि आपन भाषाई वर्चस्व आ दबंगई बनवला रखला खातिर, अनेकता के 'अस्तित्व' आ 'निजता' के नकरले में आपन शान - गुमान बूझेला। |
| BHSD49 | ई ऊहे हिन्दी - प्रेमी (?) हउवे लोग जेके खुदे हिन्दी का दुर्दशा के परवाह नइखे रहल। |

| BHSD50 | ई अंगरेजी के बनावटी आदर देई लोग बाकि अपना देश के लोकभषन के सम्मान पर हाय - तोबा मचावे लागी । |
|---|---|
| BHSD51 | भाषा का दिसाई गंभीर लउके वाला ई लोग तब तनिको विचलित ना होला , जब उनहीं के गाँव - शहर में उनहीं के आत्मीय लोग अपना भाषा का प्रति लगातार अगंभीरता के प्रदर्शन करेला । |
| BHSD52 | संवाद आ जनसंचार माध्यमन में भाषा के जवना तरह से लिहल जाये के चाहीं , वइसन ना भइला के कुछ वजह बाड़ी स । |
| BHSD53 | हमनी का खुद अपना बात - व्यवहार आ अभिव्यक्ति में भाषा के चलताऊ बनावट जा रहल बानी जा । |
| BHSD54 | अपना सुविधा आ सहूलियत का मोताबिक भाषा के अचार , मुरब्बा आ चटनी बना लिहल प्रचलन में आ गइल बा । |
| BHSD55 | ववाट्स ऐप आ फेसबुकिया 'शार्ट कट ' के मनमानी के त बाते अलगा बा । |
| BHSD56 | हिन्दी के अंगरेजी में लिखे के फैशन भा मजबूरी बा । |
| BHSD57 | 'एस एम एस ' आ 'मेल ' भाषा का दिसाई हमन का लापरवाही आ हड़बड़ी के अलगे दरसावत बा । |
| BHSD58 | कवनो नियम , कवनो अनुशासन नइखे । |
| BHSD59 | भाषा के भितरी सांस्कृतिक चेतना नष्ट हो रहल बिया । |
| BHSD60 | ओकरा सुभाविक प्रवाहे के बदल दिहल जाता । |
| BHSD61 | लोकभाषा त ओह दूब नियर हवे , जवन अपना माटी के कस के पकड़ले जकड़ले रहेले । |
| BHSD62 | ऊ बोले वालन के जीवन - संस्कृति आ चेतना के आखिरी दम तक बचवले रहेले आ ज्यों तनी खाद - पानी - हवा मिलल ऊ आपन हरियरी आ ताजगी लिहले जीवंत हो उठेले । |
| BHSD63 | भाषा के इहे गुन - धर्म लोक का गतिशीलता आ विशेषता के पहिचान देला । |
| BHSD64 | अपना देश में कतने अइसन लोकभाषा बाडी सɽ जवना में लोक के अंतरंग खुलेला आ खिलेला । |
| BHSD65 | जवना में अनुभव , बोध आ ज्ञान के अपार क्षमता - बल - बेंवत बा । |
| BHSD66 | ई परस्पर एक दुसरा से घुललो - मिलल बाड़ी सɽ , जइसे भोजपुरी - अवधी , भोजपुरी मैथिली , भोजपुरी छत्तीसगढ़ी , राजस्थानी , भोजपुरी अंगिका । |
| BHSD67 | ई कूल्हि लोकभाषा मिलिये - जुल के हिन्दी के जातीय विकास में सहायक भइल बाड़ी स आ हमन का राष्ट्रीय एकता के अउर पोढ़ बनवले बाड़ी सɽ । |
| BHSD68 | तबो का जने काहें जब एह भाषा - सब के निजता , सम्मान आ समानता क बात आवेला त कुछ कथित हिन्दीवादी राष्ट्रभक्तन ( ? ) के खुजली होखे लागेला । |
| BHSD69 | मातृभाषा से विलग होखे के सबक ऊ लोग सिखावेला , जेके राजभाषा के तिजारती , दफ्तरी आ बाजारू घालमेल में बेपटरी भइला के तनिको चिन्ता नइखे , जे अंग्रेजी में हिन्दी - डे , 'सेलिब्रेट ' करऽता । |
| BHSD70 | मीडिया चैनल , फिल्मकार आ नवकलाकार अपना कमाई खातिर , लोगन के चटोरपन का तुष्टि खातिर भाषा के जब जइसन चाहत बा तोड़ - मरोड़ लेता । |
| BHSD71 | विज्ञापन से लगायत साहित्य तक भाषा के देसी - विदेसी खिचड़ी , चटनी , अचार सिरका बनावल जा रहल बा । |
| BHSD72 | एने हिन्दी - भाषा में 'अराजकता ' खुल के अठखेली करत लउक रहल बिया । |
| BHSD73 | गली - सड़क चौराहा के लंपट - लखैरा , भांड़ आ मसखरा भाषा - साहित्य में सदवचन आ सत्साहित्य उगिल रहल बाड़न सऽ आ हिन्दिये के कथित महान साहित्यिक लोग उन्हनी का उगिलला के साहित्य बना देत बा । |
| BHSD74 | लोकप्रियता खातिर भाषा में चमत्कारी घालमेल आ उटपटांग हरकत के पूरा छूट बा । |
| BHSD75 | 'कचरा ' फइलावे वालन के महिमामंडनो होत बा , बस ऊ अपना 'गोल ' भा 'दल ' क होखे । |
| BHSD76 | एह अराजक - अभियान के हिन्दी भाषा - साहित्य के समझदारो लोगन क नेह - छोह - शह मिल रहल बा । |
| BHSD77 | कुछ अपवाद छोड़ के भाषा के जिये आ प्यार करे वाला लोग पहिलहूँ कुछ करे का स्थिति में ना रहे आ अजुओ नइखे । |
| BHSD78 | भाषा के ठीकेदार , व्यापारी आ महन्थ पहिलहूँ तिरछा मुसुकी काटत रहलन सऽ आ आजुओ काटत बाड़े सऽ । |
| BHSD79 | हिन्दी में , तकनीकी आ व्यवहारिक शब्दन का आड़ में कतने देसी - विदेशी शब्द आ मोहावरा , पढ़े वालन खातिर अजब - गजब अर्थ उगिल रहल बाडन सऽ । |
| BHSD80 | भाषा के हिंग्लिसियावत जवन नव बाजारू हिन्दी बिया ओमें बोलियन क तड़को बा । |
| BHSD81 | कुछ विद्वान त एके हिन्दी के बढ़त - 'ग्राफ ' आ ओकरा 'वैश्विक विस्तार ' से जोड़त छाती फुला रहल बा लोग । |
| BHSD82 | माने हिन्दी में रउवा आपन मातृभाषा लेके घुर्सि आईं आ एकर ऐसी - तैसी करत अपना के ओही में विलय क दीं , बाकि खबरदार अपना मातृभाषा के मान्यता के नाँव मत लीं । |
| BHSD83 | 'सत्ता ' आ 'राजनीति ' दूनो – शिक्षा , संस्कृति आ भाषा के ठेका कुछ खास संस्था , अकादमी आ संस्थानन के देके आपन पल्ला झार चुकल बाड़ी सऽ । |
| | |
| BHSD84 | कुछ विश्वविद्यालयी विभागो एह दायित्व के पेशेवर ढंग से निभा रहल बाड़न स ऽ । |
| BHSD85 | मीडिया चैनल आ भाषा - साहित्य के महोत्सव करे वालन क दखल अलग बा । |
| BHSD86 | करोड़न का कमाई खातिर कुछ फिल्मकारो ' भाषा ' के ऐसी - तैसी करे क अधिकार पवले बाड़न स ऽ । |
| BHSD87 | अब एमे भाषा खातिर गंभीरता देखावे , ओके संस्कारित - अनुशासित करे , के आगा आई ? केहू का लगे अतना फालतू टैम नइखे । |
| BHSD88 | अगर बटलो बा त ऊ चिचियाइल करस । |
| BHSD89 | प्रलाप - रूदन आ विलाप - मिलाप खातिर साल में एगो दिन तय क दिहल बा । |
| BHSD90 | ओह दिन , याने 'हिन्दी - डे ' पर रउवों आई - कुछ कहीं , कुछ सुनीं - फेरू घरे जाईं । |
| BHSD91 | ई सब कहला - गवला क मतलब केहू क विरोध नइखे । |
| BHSD92 | दरसल भाषा - संस्थान आ अकादमियन क आपन विवशता भा आंतरिक - राजनीति आ कर्म - धर्म हो सकेला । |
| BHSD93 | एही तरे राजनीतिक दलन के आपन निजी स्वारथ आ फायदा - नोकसान क चिन्ता बा । |
| BHSD94 | 'सत्ता ' , सुविधा , पद , सम्मान आ पुरस्कार केके ना लोभावे ? हमनी का खुदे अपना निजी स्वार्थ , जाति - धरम , क्षेत्र आ गोलबन्दी में बाझल बानी जा । |
| BHSD95 | आपुसी जलन , विखराव - विघटन आ हमनी क निजी पूर्वाग्रह सबसे बड़ रोड़ा बा । |
| BHSD96 | भाषा के कुछ पुरान - नया पैरोकारो चुप बाड़न स ऽ । |
| BHSD97 | ऊ उहाँ जरूर बोलिहन स ऽ , जब कवनो लोकभाषा के भारतीय भाषा - श्रेणी में - आठवीं अनुसूची में मान्यता देबे के बात आई । |
| BHSD98 | ओघरी हिन्दीवादियन के हिन्दी के बाँहि कटत लउके लागी आ राष्टं भाषा का रूप में राष्ट्रीय ता खतरा में नजर आवे लागी । |
| BHSD99 | ऊ ए सत्य के बेल्कुले भुला जइहें स ऽ कि हिन्दी के एही लोक - भषन से संजीवनी आ शक्ति मिलल बा । |
| BHSD100 | ब्रजी , अवधी , मैथिली , भोजपुरी , राजस्थानी जन - भाषा के साहित्य से हिन्दी - साहित्य संपन्न भइल बा । |
| BHSD101 | आपन क्षेत्रीय भाषा - संस्कृति आ पहिचान भइलो पर , हिन्दी पट्टी के राज्य राष्ट्रीय एकता आ राष्ट्रीय संकल्पना के आकार देले बाड़न स ऽ । |
| BHSD102 | राष्ट्रीय भाषा का रूप में , आठवीं अनुसूची में जनभाषा भोजपुरी के शामिल भइला पर कुछ हिन्दीवादी लोगन द्वारा बार - बार विरोध समझ से परे बा । |
| BHSD103 | अनजाने अब ई विरोध भोजपुरी खातिर आवाज उठावे वालन का भीतर , हिन्दी का प्रति असहिष्णुता के जन्म दे रहल बा । |
| BHSD104 | ई शुभ नइखे । |
| BHSD105 | अगर हिन्दी के समृद्ध आ संपन्न होत देखे के बा त हिन्दी पट्टी के प्रतिष्ठित आ संपन्न भोजपुरी भाषा के आठवीं अनुसूची में ना शामिल भइल , अन्याय - पूर्ण बा । |
| BHSD106 | 'भोजपुरी ' करोड़न हिन्दी - प्रेमियन के मातृभाषा हऽ । |
| BHSD107 | ओके समानता आ सम्मान के दरकार बा । |
| BHSD108 | करोड़ो भोजपुरिहा समानता आ स्वाभिमान का साथ , हिन्दी का भाषाई राष्टंवाद में शामिल रहल बाड़न सऽ । |
| BHSD109 | सामाजिक - सांस्कृतिक समरसता में ऊ बँगला , मराठी , गुजराती , तमिल , तेलगू, मलयाली , असमिया सबका साथ खड़ा बाड़न सऽ , फेर उन्हनी के आपन समृद्ध मातृभाषा काहें उपेक्षित रही ? |
| BHSD110 | सभे जानत बा कि हिन्दी अपना बोलियन आ जनपदीय - भाषा सब से तत्व ग्रहण करत आइल बिया । |
| BHSD111 | कूल्हि हिन्दी - राज्य अपना भाषाई खासियत आ विविधता का साथहीं संगठित भइल बाड़न सऽ त साहित्यिक रूप से समृद्ध आ शब्द संपन्न लोकभाषा भोजपुरी के छल - पाखंड से उपेक्षित क के , भा धौंस जमाइ के , भा अपना वर्चस्व से , ओके रोके क प्रयास कइल करोडन भोजपुरिहन में क्षोभ आ आक्रोश के जनम देई । |
| BHSD112 | सब अपना भाषा में पलल - बढ़ल बा आ ओकरा अपना मातृभाषा से लगाव बा । |
| BHSD113 | एकर मतलब ई थोरे भइल कि ओके हिन्दी भा दोसरा राष्ट्रीय भाषा से विरोध बा । |
| BHSD114 | जेकरा भोजपुरी भावेला ओके हिन्दियो भावेला । |
| BHSD115 | एगो ओकर मातृभाषा हऽ दुसरकी राजभाषा । |
| BHSD116 | 'अस्मिताबोध ' ओह सग्यानी - स्वाभिमान का होला , जेकरा दिल - दिमाग आ मन - मिजाज में अपना आ अपना पुरखन का उपलब्धियन के लेके मान - गुमान के भाव होखे । |
| BHSD117 | एकरा खातिर देश - काल - परिस्थिति के अनुकूल - अनुरूप देशज संस्कार - संस्कृति से ओह मनई के नाभि - नाल संबंध आ सरोकार के दरकार होला । |
| BHSD118 | ई सभका ला सहज - सुलभ ना हो सके । |
| | |
| BHSD119 | ई खाली किताबी ग्यान से ना उमगे । |
| BHSD120 | ई त अपना परिष्कृत परम्परा के जीवन्तता से जीअला से पोढ़ाला । |
| BHSD121 | ई अस्तित्वबोध से बहुत आगे के चीज ह । |
| BHSD122 | एकर सम्बन्ध जीवन आ जीवन्तता से होला आ अस्तित्व अउर जीवन में फरक होला । |
| BHSD123 | उत्पत्ति से बिनास तक जड़ , जानवर आ मनुष्य के बनल रहल अस्तित्व में रहल होला । |
| BHSD124 | जानवर जनम लेके मरन तक अस्तित्व में होला । |
| BHSD125 | जीवन ओकरे पास होला अथवा जीवन ओकरे के कहल जाला जे अपना जनम आ मरन के बीच कुछ अइसन उपलब्धि हासिल कर लेता , जवन ओकरा आ ओकरा आगे वाली पीढ़ी ला उपयोगिए ना , बल्कि जीवन खातिर अनिवार्य होले । |
| BHSD126 | एकरे 'के जिनगी का बाद के जिनगी ' कहल जाला आ अबहीं ले आदमिए ई जिनगी जीअत आइल बा , जड़ आ जानवर ना । |
| BHSD127 | जे ओह तमाम उपलब्धियन के मान - सम्मान आ गरिमामय पहचान देत स्वाभिमान के जिनगी जीएला , ओकरे मन - मानस में अस्मिताबोध पैदा होते । |
| BHSD128 | ई ना त अस्तित्व में मौजूद जड़ - जानवर से जुड़ल चीज ह अउर ना ओह नीरस - संवेदनहीन मनई का समझ के बिषय , जेकरा ला सब धान बाइसे पसेरी होला चाहे जेकरा मुड़ला माथ पर पानी ठहरबे ना करे , टघर जाए । |
| BHSD129 | हमरा दिल - दिमाग में अपना पुरखा लोग के अरजल बिकासमान भाषा , समाज , संस्कार , संस्कृति , जीवन - दर्शन आ प्रगतिशील ग्यान - परम्परा खातिर अछोर - अथोड़ मान - सम्मान के भाव भरल बा । |
| BHSD130 | हमरा खातिर भोजपुरी मात्र भाषा ना , मातृभाषा हिअ । |
| BHSD131 | हमरा के जनम देवेवाली माई के बोली हिअ । |
| BHSD132 | हमरा खातिर माइए हिअ । |
| BHSD133 | हमरा लोक जीवन में सात माई के महिमा गावल बा । |
| BHSD134 | ई सातो माई हमरा पालन - पोषण करेला लोग । |
| BHSD135 | हम एक एक करके रउओ सभे अपना सातो मतारी से मिलावे के चाहत बानी – जनम देवे वाली माई , बोली बनके कंठ में विराजे वाली माई ( माई भाषा ) मतलब माई के दिहल भाषा चाहे जीवन में जेकरा जरिए आदमी दुनिया के आउर भाषा अरज लेला – एह से अरजल सब भाषा के जननीभाषा , मतारी के गोदी आवते आ आँखि खोलते जेकर दरसन भइल , ऊ प्रकृति माई , मतारी के गोदी से उतरके जेकरा गोदी में |
| | गोड़ धराइल से धरती माई , माई द्वारा कुछ दिन दूध पिआके छोड़ देला के बाद जीवन भर मतारिए जइसन दूध देवे वाली गऊ माई , जल रूप में जीवन देववाली नदी माई , आहार बनके क्षुधा तृप्त करेवाली अन्नपूर्णा माई । |
| BHSD136 | एह सातो मतारी में बोली भा बानी भा भाषा रूपी माई के महिमा अपार बा । |
| BHSD137 | ई समस्त लोक - परलोक खातिर ध्वनि भा शब्द रूप में आपन विस्तार बढ़वले बाड़ी । |
| BHSD138 | मनुष्य का भाव - बिचार के लौकिक भा अलौकिक रूप में अभिव्यक्त करे खातिर ई बानी माता बीज , पौध , फूल आ फल जइसन क्रम से परा , पश्यन्ति आ मध्यमा के बाद बैखरी के अवस्था में आके मनुष्य के व्यक्त वाक् में प्राप्त होली । |
| BHSD139 | इनकरा बैखरी अवस्था के पावते आपन भाव बिचार के व्यक्त करे खातिर मनुष्य के भीतर बैखरा नधा जाला । |
| BHSD140 | ओकरे बदौलत मनुष्य व्यक्ति कहाए के पंक्ति में खड़ा हो पावेला । |
| BHSD141 | महाकवि दण्डी अपना काव्यादर्श में ओकरे के खूब फरिआ के कहले बाड़न कि ई सउँसे दुनिया घनघोर कूप अन्हरिया में रहित – जानवर जइसन जिनगी जिहित – जदि भाषा का शब्दात्मक अँजोर के उदय ना होखित – " इदमन्धतम : कृत्स्नं जायते भुवनत्रयम् । |
| BHSD142 | यदि शब्दाह्वयं ज्योतिरासंसारं न दीप्यते । । " - ( काव्यादर्श 1 / 4 ) |
| BHSD143 | हर व्यक्ति के जननीभाषा भा मातृ भाषा के गरिमा गावत भारत के सांस्कृतिक भाषा संस्कृत में साफ - साफ खोल के कहल बा कि जे अपना मातृभाषा के छोड़ के अनकर भाषा अपनावेला अथवा ओकरे उपासना में लाग जाला , ऊ अन्धकार का गँहिरा में गिर जाला । |
| BHSD144 | ऊ उहँवा पहुँच जाला , जहँवा सूरज के अँजोतो ना होखे । |
| BHSD145 | हमरा अपना मातृभाषा भोजपुरी के गौरवशाली अतीत आ संघर्षशील वर्तमान पर गुमान बा । |
| BHSD146 | एकर सम्बन्ध वैदिक ऋषि विश्वामित्र के जसी जजमान भोजगण , पौराणिक भोज पदधारी राज परिवार आ ऐतिहासिक काल के उज्जयिनी अउर कन्नौजी राजपूत भोजवंशी राजालोग से रहल बा । |
| BHSD147 | एही वैदिक भोजगण का कर्मभूमि वेदगर्भ मतलब सिद्धाश्रम बक्सर में विश्वामित्र , याज्ञवलक्य आदि ऋषि लोग वेदन का मंत्रन के दरसन कइल । |
| BHSD148 | पूरब से पच्छिम गइल एकरे अठारह भोज राज परिवारन का जस के गौरव गान युधिष्ठिर के राजसुय यज्ञ में भगवान कृष्ण गवले रहले । |
| BHSD149 | फेर उज्जयिनी आ कन्नौजी राजपूत राजवंशी के रूप में पच्छिम से पूरब अपना पुरखन के जमीन बक्सर में नयका आ पुरनका भोजपुर बसावे वाला लोग के भाषा भोजपुरी हमार मातृभाषा हई । |
| BHSD150 | हमरा एह बात के अस्मिताबोध बा । |
| BHSD151 | अपना मातृभाषा भोजपुरी के लेके हमरा भीतर मान - सम्मान आ स्वाभिमान के भाव एहू से बा , काहेकि एकर मूल वैदिक भाषा में प्रकट भइल बा । |
| BHSD152 | कारुषी आ पालि एकरे प्राचीन नाम रूप ह । |
| BHSD153 | बारहवीं सदी में कोसली से दूगो भाषा छिनगली स त पच्छमी रूप अवधी आ पूरबी रूप भोजपुरी कहाइल । |
| BHSD154 | काशिका , मल्लिका आ बज्जिका एकरे क्षेत्रीय नाम ह । |
| BHSD155 | भाषा वैज्ञानिक लोग मनले बा कि हमार मातृभाषा भोजपुरी तद्भव संस्कृति के कम से कम डेढ़ हजार साल पुरान भाषा हिअ । |
| BHSD156 | भोजपुरी के प्राचीन रूप में बौद्ध साहित्य , सिद्ध साहित्य आ नाथ साहित्य सहित लोक साहित्य - संस्कृति के मौजूदी एकरा जीवन्तता के अमर काव्य बा । |
| BHSD157 | मध्य काल के संत साहित्य से लेके आधुनिक साहित्य यात्रा तक एकरा बिकास परम्परा के गौरव शाली प्रमाण बा । |
| BHSD158 | हमार भोजपुरी आर्यावर्त ( 'आर्य ' मतलब , श्रेष्ठ आ 'आवर्त ' मतलब , निवास स्थान ) के मातृभाषा हिअ , मनुस्मृति , उपनिषद आदि में मूल आर्य भूमि बतावल गइल बा अर्थात् हिमालय आ विन्ध पर्वतमाला अउर गंगा - जमुना के बीच काशी - करुष - मल्ल आ बज्जि क्षेत्र । |
| BHSD159 | जवना क्षेत्र में विश्वामित्र , भोगगण , बुद्ध , महावीर , पुष्पमित्र , चन्द्रगुप्त मौर्य , सम्राट अशोक , स्कन्दगुप्त , शेरशाह , फतेह बहादुर शाही , मंगल पांडे , बाबू कुँवर सिंह , जयप्रकाश नारायण , डॉ० राजेन्द्र प्रसाद जइसन सपूत लोग जनम लेके भारते के ना , बल्कि सउँसे दुनिया के जीवनदर्शन आ अपना अधिकार खातिर संघर्ष करेके प्रेरणा दिहलें । |
| BHSD160 | जवना खातिर प्रसिद्ध नारायण सिंह का लिखेके पड़ल |
| BHSD161 | जवना भाषा में सरहप्पा , शबरप्पा , कुकुरिप्पा , भुसुकप्पा , गोरखनाथ , जालंधरनाथ , चौरंगीनाथ , भरथरि , गोपीचन्द , कबीर , धरमदास , दरिया , धरनी , शिवनारायण , पलटू , भीखम , टेकमन , लछमी सखी जइसन अनगिनत सिद्ध , नाथ जोगी , संत , महात्मा साहित्य सिरजल लोग । |
| BHSD162 | जवना भाषा के लोकसाहित्य , लिखित आधुनिक साहित्य में ओकरा संस्कृति , जीवनदर्शन , लोक ग्यान - बिग्यान के सहज - स्वाभाविक दरसन होला । |
| BHSD163 | अपना मातृभाषा भोजपुरी के ताकत के बदौलत हमार पुरखा लोग मारिसस , फिजी , गुआना , नेपाल जइसन कई देशन के बनावे आ बिकास के ऊँचाई पर पहुँचावे में आपन ऐतिहासिक जोगदान दिहल । |
| BHSD164 | ई लोग अपना मातृभाषा के ताकत के सहारे हिन्दी आ हिन्दुस्तान के मान - सम्मान - पहचान - उत्थान के डंका सउँसे दुनिया में बजावल । |
| BHSD165 | भारत के सम्पर्क भाषा हिन्दी के सिरजावे - सजावे - सम्पन्न बनावे आ देश - बिदेस में पहुँचावे में हमरा |
| | मातृभाषा भोजपुरी आ भोजपुरी भाषी हिन्दीसेवियन का अवदान के के नकार सकेला । |
| BHSD166 | हिन्दी के राजभाषा आ राष्ट्रभाषा के अधिकार दिलावे खातिर हमार भोजपुरी अपने हिन्दीसेवी सपूतन आ कपूतन के केतना उपेक्षा आ उपहास सहले बिआ आ सहत बिआ , ई केहू से छुपल बा । |
| BHSD167 | आजुओ तथाकथित हिन्दी के हिमायती भा भोजपुरी के सामग्री हिन्दी में हेला के बड़का - बड़का साहित्यिक पुरस्कार पावेवाला अपने नासमझ सपूतन , राजनेतन , कलाकारन , पत्रकारन , बुद्धिजीवियन के विरोध भा उदासीनता के चलते ओह भोजपुरी का अपने देश में संवैधानिक मान्यता नइखे मिल सकल । |
| BHSD168 | जवन एकरा बहुत पहिले मिल जाए के चाहत रहे । |
| BHSD169 | अबहीं तक एह बहुभाषी देश के 1651 - 52 बोली / भाषा में से बाइस भाषा के संवैधानिक मान्यता मिल चुकल बा । |
| BHSD170 | शुरु - शुरु में जवना चउदह भाषा के ई अधिकार मिलल , ऊ चउदह भाषा रहल – असमिया , बंगला , उड़िया , हिन्दी , पंजाबी , संस्कृत , गुजराती , कन्नड , कश्मीरी , मलयालम , मराठी , तमिल । |
| BHSD171 | तेलुगु आ उर्दू। |
| BHSD172 | सन् 1967 में 21 वाँ संविधान संशोधन करके सिन्धी , सन् 1992 में 71 वाँ संशोधन से कोंकणी , नेपाली आ मणिपुरी अउर सन् 2003 में 92 वाँ संशोधन से डोगरी , बोडो , संथाली आ मैथिली भाषा के संवैधानिक मान्यता दिहल गइल आ संवैधानिक मान्यता पावे के हर मापदंड आ शर्त के पूरा करेवाली हमरा मातृभाषा भोजपुरी के साथ हर राजनीतिक दल , राजनेता , सरकार आ विपक्ष हमेशा आश्वासन के झुनझुना थम्हावत धोखा देत आइल बा । |
| BHSD173 | संवैधानिक मान्यता पा चुकल भाषा सब से क्षेत्र , आबादी , राष्ट्र आ राष्ट्रभाषा हिन्दी के बिकास में अवदान , भाषिक आ साहित्यिक सम्पन्नता के आधार पर हमरा मातृभाषा भोजपुरी के साथ तुलनात्मक अध्ययन - विश्लेषण करीं त एकरा साथ भइल अन्याय के अंदाज लाग सकेला । |
| BHSD174 | हमार बाबा , ढेर पढ़ुवा लोगन का बचकाना गलती आ अज्ञान पर हँसत झट से कहसु, "पढ़ लिखि भइले लखनचन पाड़ा ! " |
| BHSD175 | पाड़ा माने मूरख ; सामाजिक अनुभव - ज्ञान से शून्य । |
| BHSD176 | आजकल तऽ लखनचंद क जमात अउर बढ़ले चलल जाता । |
| BHSD177 | ए जमात में अधूरा ज्ञान क अहंकार आ छेछड़पन दूनों बा । |
| BHSD178 | कुतरक त पुछहीं के नइखे । |
| BHSD179 | ई लोग कबो "आधुनिकता आ बदलाव" का नाँव पर त कबो देखावटी "स्त्री मुक्ति" का नाँव पर आ कबो कबो समता - समाजवाद का नाँव प पगुरी ( वैचारिक जुगाली ) करत मिलिए जाला । |

| BHSD180 | सावन का एही हरियरी तीज पर एगो बिद्वान भड़क गइले । |
|---|---|
| BHSD181 | उनका ए तीज बरत करे वाली लइकी आ मेहरारुवन में मानसिक गुलामी आ पिछड़ापन लउके लागल । |
| BHSD182 | कहले कि मेहरारुवन के ई बरत - परब बंद क देबे के चाही । |
| BHSD183 | एम्मे पुरुष - दासता क झलक मिलत बा । |
| BHSD184 | सुख सौभाग आ पति का दीरघ जीवन खातिर काहें खाली मेहरारुवे बरत करिहें सऽ ? फेसबुक पर उनका एह नव बिचार प "झलक दिखला जा" देखे वाली एगो बोल्ड मेहरारू उनकर टनकारे सपोट कइलस । |
| BHSD185 | फेर त फेसबुकिया भाई लोग के मन बहलाव क उद्दम भेंटाइ गइल । |
| BHSD186 | कुछ दिन से सावन का महीना में महादेव का भक्ति भाव में लरकल - लपटाइल लोगन क हर हर - बम बम सुनत हमहूँ रमल रहलीं । |
| BHSD187 | ओने झुलुवा झूलत , कजरी गावत लइकी मेहरारुन क उत्सव - परब हरियरी तीज आ गइल । |
| BHSD188 | प्रेम आ हुलसित उल्लास क अइसन सहज अभिव्यक्ति करे वाला परब जान लीं कि हमार सुखाइल मन अउरी हरियरा गइल । |
| BHSD189 | कजरी राग के रस - बरखा मन परान तृप्त क दिहलस । |
| BHSD190 | लोक - उत्सव इहे न हऽ ; हम सोचलीं । |
| BHSD191 | ओने हमार बिद्वान पढुवा मित्र पगुरी में बाझल बाड़े । |
| BHSD192 | उनके प्रेम आ हुलास से भरल ए मौसम में मानसिक गुलामी लउकत बा । |
| BHSD193 | ए लोक परब का बत में सुन्दर सुजोग वर ( जीवन साथी ) भा पति के आयु आ सौभाग क कामना में असमानता आ अनेत लउकत बा । |
| BHSD194 | ऊ समता खातिर अतना खखुवाइल बाड़े कि सुई का जगहा तलवार उठावे प आमादा बाड़े । |
| BHSD195 | अब हम उनके कइसे समझाईं कि "लोक" आ ओकरा सांस्कृतिक - अभिप्राय के रहस्य समझे खातिर लोक का भावभूमि पर उतरे के परेला । |
| BHSD196 | विरासत में मिलल आचार - बिचार ठीक से समझला का बादे न ओकरा के पुनरीक्षित भा नया रूप रंग देबे का बारे में सोचल जाई । |
| BHSD197 | तीज , चउथ आ भइया दूज आ जिउतिया ( जीवित पुत्रिका ) भूखे वाली मेहरारुवन क मनोभूमि प उतरऽ त बुझाई कि उनहन का एही अवदान पर परिवार आ समाज टिकल बा । |
| BHSD198 | उनहन का पवित्र मनोभाव आ कामना के मानसिक दासता क संज्ञा दे दिहल त्याग तप आ प्रार्थना के अपमाने न बा। |
| BHSD199 | अतीत क कवनो बात कइला भा ओकर परतोख देहला पर कुछ बुद्धजीवी कुटिल मुस्कान काटत हमके परंपरावादी कहि सकेला बाकि हम भोजपुरिया लोक आ जीवन संस्कृति के हईं। |
| BHSD200 | अतीत में कुल्हि माहुरे नइखे ; ओमे अमृतो छिपल बा जेवन हमन का जीवन के नया भावबोध आ अनुभव ज्ञान संपदा से जोरि सकेला। |
| BHSD201 | भौतिक तरक्की आ नव ज्ञान का जोम में हमन का पीढ़ी के संवेदनहीन आ मूल्यविहीन हो जाये क ज्यादा खतरा बा। |
| BHSD202 | लोकज्ञान पहिलहूँ शास्त्र ज्ञान ले कम ना मनात रहे आ शास्त्रमत 'लोकमत' से पछुवा जात रहे। |
| BHSD203 | लोक आ शास्त्र मिलिये के हमहन का समाज के संस्कारित कइलस। |
| BHSD204 | लोकधर्म मानवधर्म बनल आ मानव का स्वच्छन्दता आ निरंकुशता के अनुशासित कइलस ; ओके संस्कारित कइ के समाजिक बनवलस। |
| BHSD205 | हमहन क भोजपुरिया समाज एही आचार बिचार आ संस्कार से बन्हाई - मँजाई के चमकल। |
| BHSD206 | काल चक्र में अलग अलग समय सदर्भ में , एह लोक रीति में कुछ अच्छाई कुछ बुराई आइल होई ; बाकिर कुल मिलाई के हमहन का पारिवारिक संस्कृति क नेइं ( नींव ) एही तरे परल। |
| BHSD207 | अनेकता में एकता क उद्घोष करे वाला हमन के महान देश इहाँ क रहनिहार हर नागरिक के हऽ ; बाकिर अइसनो ना कि देश के हेठे दबाइ के सबकर अपने आजादी परमुख हो जाय। |
| BHSD208 | पहिले राजा जवन मरजी होखे करे बदे सुतंत्र आ निरंकुश रहे बाकि अब , लोगन के चुनल - बनावल "ओहदेदार" अपना निज के महत्वाकांक्षा आ तुष्टि खातिर लोगन के भेंड़ा बना के इस्तेमाल करे लागी त बतबढ़ ढेर आगा बढ़ जाई। |
| BHSD209 | ई कइसन आजादी कि "मोगैम्बो" के खुश करे खातिर लोग ओकर 'कारपेट' बन जाय ? हमहन क कइसन मानसिक गुलामी कि आपन नेता भइला का नाते ओकरा आगा पाछा पोछ डोलाई आ ओकरा अनेत के आड़ छाँह देईं जा ? |
| BHSD210 | स्वच्छ राजनीति का नाँव प' "नया आ अनोखा" करे निकलल जब अपना महत्वाकांक्षा आ खुशामद पसन्दी में सीमा लाँघे लागल त का कइल जाव ? हाय रे अधोगति , आज देश के स्वतंत्रता का महापरब पर "भारत / इन्डिया , ", "जयहिन्द" का बजाय अपने नाँव "अरविन्द केजरीवाल" ; ऊहो स्कूल का छोट लड़िकन के ऊपर , लाल ड्रेस पहिरा के शब्द का रूप में बइठा के ; आजादी खातिर मर मिटे वालन का जगहा अपने के महिमामंडित कइल जाता। |
| BHSD211 | काल्हु महामहिम राष्ट्रपतिजी हमनी के आ देश खातिर प्राण निछावर करे वाला सैनिकन क अभिनंदन क के कहलन कि हमन के अपना मानवी मूल्य आ विरासत बचावे क कोसिस करे के चाही। |
| BHSD212 | कहलन कि , "जवन देश अपना अतीत के मूल्य आ आदर्श भुला जाला ऊ अपना भविष्यो क खासियत |
| | गँवा देला। |
| BHSD213 | आज तमाशा देखीं। |
| BHSD214 | आज खबर मिलल कि हमनी का प्रदेश में कानपुर में कवनो लड़की के छेड़खानी करत मनचल / मनबढ़ुवन के रोकला पर , फर्ज निभावे वाला फौजी के पीट पीट के हत्या कर दिहल गइल। |
| BHSD215 | लोग कह सकेला कि अगर मोगैम्बो आजादी के अपना पक्ष में इस्तेमाल कर सकेला त मनबढ़ लइका काहे ना ? बाकिर ए दूनो घटनन से हमहन क मूल्यन क गिरावट आ पतनशीलता नइखे उजागर होत का ? राजधरम , लोक धरम आ मानव धरम सब सँगही चलेला। |
| BHSD216 | यथा राजा तथा प्रजा। |
| BHSD217 | आज एह हालत में के केकर अनुसरन करत बा ? साँच त ईहे बा कि हमन के आजादी त मिल गइल बाकि मानसिक गुलामी ना गइल। |
| BHSD218 | पोंछ डोलावल जब सोभाव बन जाला त नीक जबून ना बुझाला। |
| BHSD219 | आजादी मिलल , अधिकार मिलल बाकि ओकर इस्तेमाल करे क लूर सहूर ना आइल। |
| BHSD220 | अगर शक्ति आ सत्ताकेन्द्रित राजनीति भइल त ओकर फायदा उठावे वाला जमातो बढ़ल। |
| BHSD221 | स्वछंदता करे खातिर सरंक्षन मिलल ; बाकि एकर तिरस्कार आ बहिस्कार कइला का साथ साथ हमनियो के ई ठीक से रट लेबे के चाहीं कि स्वतंत्रता क माने स्वच्छंद भइल ना ह , आ आजादी क मतलब अराजकता त कत्तई ना हऽ। |
| BHSD222 | हमनी के आजादी के 68 वां सालिगरह के पहिले वाला साँझ हम रउरा सभ के आ दुनिया भर के भारतवासी लोग के हार्दिक अभिनंदन करत बानी। |
| BHSD223 | हम अपना सशस्त्र सेना , अर्ध - सैनिक बल आ आंतरिक सुरक्षा बल के सदस्यनो के खास अभिनंदन करत बानी। |
| BHSD224 | हम अपना ओह सगरी खिलाड़ियनो के बधाई देत बानी जे भारत आ विदेशो में आयोजित भइल प्रतियोगितन में शामिल भइले आ पुरस्कार जीतले। |
| BHSD225 | हम , 2014 के नोबेल शांति पुरस्कार विजेता श्री कैलाश सत्यार्थीओ के बधाई देत बानी , जे देश के नाम रोशन कइलन। |
| BHSD226 | मित्रलोग ; |
| BHSD227 | 15 अगस्त , 1947 के हमनी का राजनीतिक आजादी हासिल कइनी। |
| BHSD228 | आधुनिक भारत के उदय एगो ऐतिहासिक खुशी के मौका रहल , बाकिर ई देश के एह छोर से दोसरा छोर ले बहुते पीड़ा के खून से नहाइलो रहुवे। |
| BHSD229 | अंगरेजी शासन का खिलाफ भइल महान संघर्ष के पूरा दौर में जवन आदर्श आ भरोसा बनल रहुवे तबन ओह घरी दबाव में रहुवे। |
| BHSD230 | महानायकन के एगो महान पीढ़ी एह विकट चुनौती के सामना कइलसि। |
| BHSD231 | ओह पीढ़ी के दूरदर्शिता आ पोढ़पन हमहन के एह आदर्शन के , खीस आ तकलीफ के दबाव में बिछिलाए से भा गिर जाए से बचवलसि। |
| BHSD232 | ऊ असाधारण लोग हमनी के संविधान के सिद्धांतन में , सभ्यता से आइल दूरदर्शिता से जनमल भारत के गर्व , स्वाभिमान आ आत्मसम्मान के मिलवले , जवन पुनर्जागरण के प्रेरणा दिहलसि आ हमनी के आजादी मिलल। |
| BHSD233 | हमहन के सौभाग्य बा कि हमनी के अइसन संविधान मिलल जवन महानता क तरफ भारत के यात्रा के शुरुआत करवलसि। |
| BHSD234 | एह दस्तावेज क सबले मूल्यवान उपहार रहल लोकतंत्र , जवन हमनी के पुरातन मूल्यन के नयका संदर्भ में आधुनिक रूप दिहलसि आ तरह तरह के आजादी के संस्थागत कइलसि , ई आजादी के शोषितन आ वंचितन ला एगो जानदार मौका में बदल दिहलसि आ ओह लाखो लोग के बरोबरी का संगही सकारात्मक पक्षपात के उपहार दिहलसि जे सामाजिक अन्याय से पीड़ित रहले। |
| BHSD235 | ई एगो अइसन लैंगिक क्रांति के शुरुआत कइलसि जवन हमनी के देश के बढन्ती के उदाहरण बना दिहलसि। |
| BHSD236 | हमनी का ना चलत परंपरन आ कानूनन के खतम कइनी जा आ शिक्षा अउर रोजगार के जरिए औरतन ला बदलाव तय करा दिहनी। |
| BHSD237 | हमनी के संस्था सभ एही आदर्शवाद के बुनियादी ढांचा हई सँ। |
| BHSD238 | हे देशवासी ; |
| BHSD239 | निमनो से निमन विरासत के बचवले राखे ला ओकर लगातार देखभाल जरूरी होले। |
| BHSD240 | लोकतंत्र के हमहन के संस्था दबाव में बाड़ी सँ। |
| BHSD241 | संसद बातचीत का बजाय टकराव के अखाड़ा बन गइल बाड़ी सँ। |
| BHSD242 | एह समय , संविधान के प्रारूप समिति के अध्यक्ष डॉ। |
| BHSD243 | बी . आर . अम्बेडकर के ओह कहना के दोहरावल सही होखी , जे उहाँ के नवंबर , 1949 में संविधान सभा में अपना समापन व्याख्यान में दिहले रहीं : |
| BHSD244 | 'कवनो संविधान के चलावल पूरा से संविधाने के सुभाव पर निर्भर ना होखे। |

| BHSD245 | संविधान त राज्य के विधायिका , कार्यपालिका आ न्यायपालिका जइसन अंगे भर दे सकेला। |
|---|---|
| BHSD246 | ओह अंगन के चलावल जिनका पर निर्भर करेला ऊ ह जनता आ ओकर मनसा अउर राजनीति के साकार करे ला ओकर बनावल राजनीतिक गोल। |
| BHSD247 | ई के बता सकेला कि भारत के जनता आ ओकर बनावल गोल कवना तरह से काम करीहें ? ' |
| BHSD248 | अगर लोकतंत्र के संस्थाच दबाव में बाड़ी सँ त समय आ गइल बा कि जनता आ ओकर गोल एह पर गंभीर चिंतन करेसु। |
| BHSD249 | सुधार के उपाय भीतर से आवे के चाहीं। |
| BHSD250 | हमहन के देश के बढ़न्ती के आकलन हमनी के मूल्यन के बेंवत से होखी , बाकिर साथही ई आर्थिक प्रगति अउर देश के संसाधनन के बरोबरी वाला बाँटो से तय होखी। |
| BHSD251 | हमनी के अर्थव्यवस्था भविष्य ला बहुत आशा बंधावत बिया। |
| BHSD252 | 'भारत गाथा ' के नया अध्याय अबहीं लिखल बाकी बा। |
| BHSD253 | 'आर्थिक सुधार ' पर काम चलत बा। |
| BHSD254 | पिछला दस बरिस का दौरान हमनी के बढ़न्ती आ् हासिल सराहे जोग रहल , आ बहुते खुशी के बात बा कि कुछ गिरावट का बाद हमनी का साल 2014 - 15 में 7 । |
| BHSD255 | 3 प्रतिशत के विकास दर फेरु पा लिहले बानी। |
| BHSD256 | बाकिर एहसे पहिले कि एह बढ़न्ती के फायदा सबले धनी लोग के बैंक खातन में पहुंचे , ओकरा गरीब से गरीब आदमी ले चहुँपे के चाहीं। |
| BHSD257 | हमनी के एगो समहरी लोकतंत्र आ अर्थव्यवस्था ह। |
| BHSD258 | धन - दौलत के एह इन्तजाम में सभका ला जगहा बा। |
| BHSD259 | बाकिर सबले पहिले ओकरा मिले के चाहीं जे अभाव आ किल्लत के कष्ट झेलत बाड़े। |
| BHSD260 | हमनी के नीतियन के आगे चल के 'भूख से मुक्ति ' के चुनौती के सामना करे में बेंवतगर होखे के चाहीं। |
| BHSD261 | हे देशवासी ; मनुष्य अउर प्रकृति के बीच के पारस्परिक संबंधनन के बचावे राखे पड़ी। |
| BHSD262 | उदारमन वाली प्रकृति अपवित्र कइला पर आपदा बरपावे आ बरबाद करे वाली शक्ति में बदल सकेले , जवना चलते जानमाल के बड़हन नुकसान होला। |
| BHSD263 | एह घरी , जब हम रउरा सभे के संबोधित करत बानी देश के बहुते हिस्सा बडहन कठिनाई झेलत बाड़ के |
| | बरबादी से उबरे में लागल बाड़े । |
| BHSD264 | हमनी के ओह पीड़ितन ला फौरी राहत का साथही पानी के कमी आ अधिकी दुनू के काबू करे के लमहर दिन चले लायक उपाय खोजे के पड़ी । |
| BHSD265 | हे देशवासी ; जवन देश अपना अतीत के आदर्शवाद भुला देला ऊ भविष्य से बहुते कुछ खासो गँवा देला । |
| BHSD266 | कई पीढ़ियन के आकांक्षा आपूर्ति से कहीं अधिका बढ़ला का चलते हमनी के विद्यादायी संस्थान के गिनती लगातार बढ़ल जात बा बकिर नीचे से ऊपर ले गुणवत्ता के का हाल बा ? हमनी का गुरु शिष्य परंपरा के बहुते तर्कसंगत गर्व से याद करीलें , त फेर हमनी का एह संबंधन का जड़ में समाइल नेह , समर्पण आ प्रतिबद्धता के काहे छोड़ दिहनी जा ? गुरु कवनो कुम्हार के मुलायम आ कुशल हाथे जइसन अपना शिष्य के भविष्य बनावेला । |
| BHSD267 | विद्यार्थीओ श्रद्धा आ विनम्रता से शिक्षक के ऋण सकारेला । |
| BHSD268 | समाज , शिक्षक के गुण आ उनका विद्वता के मान सम्मान देला । |
| BHSD269 | का आजु हमनी के शिक्षा प्रणाली में अइसन होखत बा ? विद्यार्थियन , शिक्षकन आ अधिकारियन के रुक के आपन आत्मनिरीक्षण करे के चाहीं । |
| BHSD270 | हमनी के लोकतंत्र रचनात्मक ह काहे कि ई बहुलवादी ह , बाकिर एह विविधता के सहनशक्ति आ धीरज से पोसे के चाहीं । |
| BHSD271 | मतलबी लोग सदियन पुरान एह पंथनिरपेक्षता के नष्ट करे का कोशिश में आपसी भाईचारा के चोट पहुंचावेले । |
| BHSD272 | लगातार बेहतर होखत जात प्रौद्योगिकी के तुरते फुरत संप्रेषण के एह युग में हमहनके ई तय करे ला सतर्क रहे क चाहीं कि कुछ इनल - गिनल लोग के कुटिल चाल हमहन के जनता के बुनियादी एकता पर कबो हावी मत हो पावे । |
| BHSD273 | सरकार अउर जनता , दुनू ला कानून के शासन परम पावन होला बाकिर समाज के रक्षा कानूनो से बड़ एगो शक्तिओ करेले आ ऊ ह मानवता । |
| BHSD274 | महात्मा गांधी कहले रहीं , 'रउरा सभके मानवता पर भरोसा ना छोड़े के चाहीं । |
| BHSD275 | मानवता एगो समुद्र ह आ अगर समुद्र के कुछ बूंद गन्दा हो जाव त समुद्र गंदा ना हो जाले । |
| BHSD276 | शांति , मैत्री आ सहयोग अलग अलग देशन आ लोग के आपस में जोड़ेला । |
| BHSD277 | भारतीय उपमहाद्वीप के साझा भविष्य के पहिचानत , हमनी के आपसी संबंध मजगर करे के होखी , संस्थागत क्षमता बढ़ावे के होखी आ क्षेत्रीय सहयोग के विस्तार ला आपसी भरोसा के बढ़ावे होखी । |
| BHSD278 | जहां हमनी का दुनिया भर में आपन हित आगे बढ़ावे का दिसाई प्रगति करत बानी , ओहिजे भारत अपना नियरे के पड़ोस में सद्भावना आ समृद्धि बढ़ावहुं ला बढ़ - चढ़ के काम करत बा । |
| BHSD279 | ई खुशी के बात बा कि बांग्लादेश का साथे लमहर दिन से लटकल आवत सिवान के विवाद आखिर में सलटा लीहल गइल बा। |
| BHSD280 | जहाँ हमनी का दोस्ती में आपन हाथ अपना मन से आगे बढ़ावेनी जा ओहिजे हमनी का जानबूझ के कइल जात उकसावे वाली हरकतन के आ बिगड़त सुरक्षा माहौल से आंख ना मूंद सकीं जा। |
| BHSD281 | सीमा पार से चलावल जात शातिर आतंकवादी समूहन के निशाना भारत बन गइल बा। |
| BHSD282 | हिंसा के भाषा आ बुराई के राह के अलावा एह आतंकवादियन के ना त कवनो मजहब बा ना ई कवनो विचारधारा मानेलें। |
| BHSD283 | हमनी के पड़ोसियन के ई तय क लेबे के चाहीं कि ओकरा जमीन के इस्तेमाल भारत से दुश्मनी राखे वाला ताकत ना कर पावसु। |
| BHSD284 | हमहन के नीति आतंकवाद के इचिको बर्दाश्त ना करे वाला बनल रही। |
| BHSD285 | राज्य के नीति के एगो औजार का तरह आतंकवाद के इस्तेमाल के कवनो कोशिश के हमनी का खारिज करत बानी। |
| BHSD286 | हमहन के सिवान का भीतर घुसपैठ आ अशांति फइलावे के कोशिशन से कड़ाई से निपटल जाई। |
| BHSD287 | हम ओह शहीदन के श्रद्धांजलि देत बानी जे भारत के रक्षा में अपना जीवन के सर्वोच्च बलिदान दिहले। |
| BHSD288 | हम अपना सुरक्षा बलन के साहस आ वीरता के नमन करत बानी जे हमनी के देश के क्षेत्रीय अखंडता के रक्षा आ हमनी के जनता के हिफाजत ला निरंतर चौकस रहत बाड़ें। |
| BHSD289 | हम खास क के ओह बहादुर नागरिकनो के सराहत बानी जे अपना जान के जोखिम के परवाह कइले बिना बहादुरी देखावत एगो दुर्दांत आतंकवादी के पकड़ लिहलें। |
| BHSD290 | हे देशवासी ; भारत 130 करोड़ नागरिक , 122 गो भाषा , 1600 बोलियन आ 7 गो मजहबन के एगो जटिल देश ह। |
| BHSD291 | एकर ताकत , सोझा लउकत विरोधाभासन ले रचनात्मक सहमति का साथ मिलावे के आपन अनोखा बेंवत में समाइल बा। |
| BHSD292 | पंडित जवाहरलाल नेहरू के शब्दन में ई एगो अइसन देह ह जवन 'मजबूत बाकिर ना लउके वाला धागन' से एके सूत्र में बान्हल गइल बा आ 'ओकरा ईर्द - गिर्द एगो प्राचीन गाथा के मायावी विशेषता पसरल बा , जइसे कवनो सम्मोहन ओकरा मस्तिष्क के अपना वश में क लिहले होखे। |
| BHSD293 | ई एगो मिथक आ एगो विचार ह , एगो सपना आ एगो सोच ह , बाकिर साथही ई एकदमे असली , साकार आ सर्वव्यापी हवे।' |
| BHSD294 | हमहन के संविधान से मिलल उपजाऊ जमीन पर , भारत एगो जीवंत लोकतंत्र बन के बढ़ल बा। |
| BHSD295 | एकर जड़ गहिर ले गइल बाड़ी सँ बाकिर पतई मुरझाए लागल बाड़ी सँ । |
| BHSD296 | अब एकरा के नया करे के समय आइल बा । |
| BHSD297 | अगर हमनी का अबहियें डेग ना उठवनी जा त का सत्तर बरीस बाद हमहन के वारिस हमनी के ओतने सम्मान आ प्रशंसा साथे याद कर पइहें जइसे हमनी का 1947 में भारतवासियन के सपना साकार करे वालन के करीलें । |
| BHSD298 | जबाब भलही सहज ना होखे बाकिर सवाल त पूछही के पड़ी । |
| BHSD299 | धन्यवाद , जय हिंद ! |
| BHSD300 | ( स्वतंत्रता दिवस के 68 वां सालगिरह का पहिले वाला साँझ राष्ट्र के दिहल राष्ट्रपति प्रणव मुखर्जी के हिन्दी संबोधन के अनधिकृत अनुवाद । |
| BHSD301 | कवनो तरह के कमी भा गलती ला क्षमायाचना का साथे दीहल गइल बा । |
| BHSD302 | बाकिर अगर भोजपुरी के संविधान से मान्यता दिआवे के बा त एह तरह के कोशिश लगातार होखे के चाहीं । |
| BHSD303 | पिछला 11 अगस्त का दिने इंडिया इंटरनेशनल सेंटर में भोजपुरी समाज दिल्ली भारत में मारीशस के उच्चायुक्त जगदीश्वर गोवर्धन के अभिनंदन समारोह आयोजित कइलसि । |
| BHSD304 | एह समारोह में केन्द्रीय मंत्री आ राज्यपाल रह चुकल डाँ । |
| BHSD305 | भीष्मनारायण सिंह के अध्यक्षता में सांसद जगदम्बिका पाल , अर्जुन मेघवाल , आर . के . सिन्हा , मनोज तिवारी , विधायक आदर्श शास्त्री आ सीआईएसएफ के डी जी रहल के . एम . सिन्हा मौजूद रहनी जा । |
| BHSD306 | एह समारोह में भोजपुरी के संवैधानिक मान्यता के मुद्दा पर खूब खुलके बात भइल आ सरकार से एह मुद्दा के जल्दी से सलटावे के मांग कइल गइल । |
| BHSD307 | एही मौका पर भोजपुरी समाज के स्मारिका के विमोचनो भइल । |
| BHSD308 | भोजपुरी समाज दिल्ली के अध्यक्ष अजीत दुबे कहलन कि भारत में मारीशस के उच्चायुक्त के रूप में जगदीश्वर गोवर्धन के नियुक्ति से दुनु देश के संबंध अउरी पोढ़ होखी । |
| BHSD309 | भोजपुरी के बारे में पिछलका यूपीए सरकार पर वादा - खिलाफी के आरोप लगावत दुबे जी कहलन कि ऊ सरकार देश विदेश के करोड़न भोजपुरियन के महज खयाली पोलाव परोसलसि । |
| BHSD310 | आशा जतवलन कि अब देश में भारतीय भाषावन के पक्षधर सरकार आइल बिया त आशा लागल बा कि ई सरकार भोजपुरी के ओकर हक दे दी । |

| BHSD311 | बतवलन कि आजुए सांसदन के एगो प्रतिनिधिमंडल देश के प्रधानमंत्री नरेन्द्र मोदी से मिलल आ भोजपुरी , राजस्थानी आ भोटी भाषा के संवैधानिक मान्यता देबे के निहोरा कइलसि । |
|---|---|
| BHSD312 | अपना सम्मान में बोलावल एह सभा में बोलत मारीशस के उच्चायुक्त जगदीश्वर गोवर्धन कहलन कि मारीशस के छोट भारत कहाला आ हमहन के भाषा , संस्कृति , परम्परा पूरा तरह से भारतीयता के रंग में रंगाइल बावे । |
| BHSD313 | उहो आस जतवले कि भारतो में भोजपुरी के संवैधानिक मान्यता मिल जाई । |
| BHSD314 | सांसद जगदम्बिका पाल आ मनोज तिवारी भोजपुरी के एह आन्दोलन में हर तरह से शामिल रहे आ कवनो कोर कसर ना छोड़े के वादा कइलन । |
| BHSD315 | कार्यक्रम का संचालन समाज के वरिष्ठ उपाध्यक्ष प्रभुनाथ पाण्डेय कइलन आ वरिष्ठ उपाध्यक्ष गरीबदास , महामंत्री एल . एस . प्रसाद , संयोजक विनयमणि तिरपाठी , कृपालु टरस्ट के ट्रस्टी रामपुरी , देशबंधु के वरिष्ठ सलाहकार अरूण कुमार सिंह वगैरह अनेके कवि , लेखक , वकील , अध्यापक , समाजसेवी , पत्रकार आ अउरिओ बुद्धिजीवी उपस्थित रहलन । |
| BHSD316 | बहुत पहिले एक बेर क्रिकेट देखत खा , भारत के 'माही ' मिस्टर धोनी का उड़त छक्का के कमेन्टरी वाला 'हेलीकाप्टर शाट ' का कहलस , ओके नकलियावे क फैशन चल निकलल । |
| BHSD317 | नीचे से झटके म़े ऊपर उठावे वाला ई हेलिकाप्टर स्टाइल विज्ञापन वाला ले जाके लेहना काटे वाला मशीन से जोड़ दिहले सऽ । |
| BHSD318 | एही तरे राजनीति का नूराकुश्ती में आपुसी घात - प्रतिघात क स्टाइल बदल गइल बा । |
| BHSD319 | एहू में नया नया प्रयोग , आ नया रिसर्च हो रहल बा । |
| BHSD320 | दक्षिणपंथी , वामपंथी , नरमपंथी , चरमपंथी , मध्यममार्गी दल नया नया स्टाइल अपना रहल बाड़े सऽ । |
| BHSD321 | अखाड़ा म़ें उतरला का पहिलहीं पैंतरा चालू आ टेम्पो हाई करे खातिर बोलबाजी हावी हो जाता । |
| BHSD322 | "धुरिये म़ें जेंवर बरल ( रसरी बरल ) " गाँव क गपोड़ियन के मनोविनोद अकल्पनिय आनंद के विषय होला । |
| BHSD323 | राजनीतियो एकरा के अपनाई के "बेबात बवंडर"से काम रोकले में आपन हित देख ऽ तिया । |
| BHSD324 | पछिला साल दिल्ली का चुनाव में "बाल के खाल निकाले" वाला 'केजरीवार ' स्टाइल अस चमकल कि ओकर विरोधो करे वाला बड़ बड़ स्टाइलिस राजनीतिक ओही स्टाइल के अपनाई के बाल के खाल निकाले आ लीपे पोते शुरू दिहले स । |
| BHSD325 | काम न धाम सबकर नीद हराम । |
| BHSD326 | पहिले एगो अलीबाबा का पाछा चालीस गो चोर रहले सऽ । |
| BHSD327 | अब त चालीस का सँगे चउदह गो अउर जुटि गइल बाड़े स । |
| BHSD328 | सहयोगी भा पछलगुवन के आपन आपन दाल गलावे खातिर सझिया क हाँड़ी चाही । |
| BHSD329 | चालीस चोरन का गोदाम में हँड़िया क कवन कमी ? उहवाँ त सोना चानी क केतने हँड़िया बटलोही गाँजल रहे । |
| BHSD330 | सहजोगिया सोचले स कि दाल त दाल रिन्हाई ; हँड़िया फिरी भेंटाई ! चोरन क सरदार चिचियाव , एकरा पहिलही मय पछलगुवा एक - ए - गो हँड़िया लेके जोर जोर से चोकरे चिचियाये शुरू क दिहले सऽ । |
| BHSD331 | बूझि लीं कि केजरीवार स्टाइल फेर हिट हो गइल । |
| BHSD332 | बलुक एबेर त अतना हिट भइल , अतना हिट कि संसदे ठप्प हो गइल । |
| BHSD333 | चुनाव - चरचा में गुणा - गणित के हिसाब किताब मीडिया चैनल ढेर करेले सन । |
| BHSD334 | जइसे मान लीं कि लगले एगो प्रदेश म़े चुनाव होखे वाला बा ; त मये खुर्राट गणितज्ञ जात - पात आ धरम - करम से थाकि के दिलिये वाला बालू पर आपन भीति उठावे शुरू क दिहले स आ ओने जनता ए लोगन के "घुघुवा मन्ना "खेलावे लागल । |
| BHSD335 | घुघुवा मन्ना खेलावत खा एगो गीत गवाला । |
| BHSD336 | 'घुघुवा मन्ना , उपजे धन्ना । |
| BHSD337 | नई भीत उठेले , पुरानी भीत गिरेले । |
| BHSD338 | " त पुरनकी भीत बचावे का चक्कर म़े गठबन्धन क टोटरम होखे लागल । |
| BHSD339 | एह टोटरम में सइ गो टोटका ! त कबो लउवा लाठी चन्नन आवऽ ता त कबो 'ओका - बोका तीन तड़ोका ' । |
| BHSD340 | इजइल , बिजइल का करिहें ; ढोंढ़िया पचकबे करी । |
| BHSD341 | ई बालो ससुर ना सोचले होंइहें कि अइसन अफदरा परी ; कि उनकर अइसे आ एतना खाल निकालल जाई ? हमरा सोझिया मति से त ईहे बुझाता कि अब सुधार आ बिकास के डहर बहुते कठिन बा , काहे कि बाल के खाल निकाले वाला खेल में जनतो के मजा आवे लागल बा । |
| BHSD342 | मजा लेबे का चक्कर में , कहीं ऊ गफलत में मत परि जाव ? गुलामी , गरीबी , बदहाली , असमानता आ नियति के थपेड़ा खात - खात पीढ़ी गुजरि गइल । |
| BHSD343 | कतने 'तंत्र ' का बाद "लोकतंत्र" आइल आ अइबो कइल त 'बानर का हाथ क खेलवना ' हो गइल । |
| BHSD344 | हाय रे करम ! एके बनरा कब ले छोड़िहें सऽ ? |
| BHSD345 | पसेना से नहाइल रामचेला हाँफत हाँफत बाबा लस्टमानंद के लगे चहुँपले । |
| BHSD346 | सस्टांग दंडवर कइके आशीर्वाद लिहला के बाद रामचेला कहलन , गुरू आजकाल्हु गउवों में बदलाव के बयार बड़ा तेजी से बहऽता । |
| BHSD347 | लस्टमानंद पुछलन , काहे ? का हो गइल ? समय के साथे हर चीज बदल जाला । |
| BHSD348 | बदलाव संसार के नियम हवे , एहमें परेशान होखे ला का बा ? |
| BHSD349 | रामचेला कहलें , गुरू पहिले बिआह होत रहे बड़ा मजा आवो । |
| BHSD350 | लौंडा के नच से ले के मरजाद ले बिआह में रहे में बड़ा मजा आवो । |
| BHSD351 | एतने ले ना , बेटिहा किहां जब बारात चहुँपे आ जनवासा में लउन्डा के नाच होखे त बुढ़उवो लोग मजा लेबे में पीछे ना रहत रहुवे । |
| BHSD352 | लउन्डवो सब नाचते नाचत बुढ़ऊ लोग के अंचरा ओढ़ावे लागे आ बुढ़ऊ बाबा ईया के पइसा देत होखो चाहे ना , लउन्डवन के जरूर दे देस । |
| BHSD353 | तिलक आ बिआह में पात में भूईंया बइठ के खइला के आपने आनंद रहुवे । |
| BHSD354 | जेकर दुआर जेतना बड़ ओकरे ओतने बड़ हैसियत रहे । |
| BHSD355 | बड़का लोग किहाँ त एगो पांत में हजार आदमी ले खा लेत रहलन बाकिर अब सभे व्यस्त हो गइल बा । |
| BHSD356 | बिआह में मरजाद त दूर अब साँझी खा बारात जा ता आ राते में खा के लवद आव ता । |
| BHSD357 | भूईंया बइठ के पांत में खाए वाला सिस्टम के साथे पूड़ी तरकारी आ दही चिउरो खतम हो गइल । |
| BHSD358 | अब त का कहता लोग बुफे सिस्टम में खिआवे के । |
| BHSD359 | पता ना ई बुफे सिस्टम कहवां से आइल बा लेकिन एह सिस्टम में खाए वाला लोग लाज हया घोर के पी गइल बा । |
| BHSD360 | ससुर होखे चाहे भसुर , ओ लोग के सामनहीं कनिया लोग प्लेट में चम्मच हिलावत चपर चपर मुंह चलावऽता लोग । |
| BHSD361 | एह बुफे में बड़ छोट के लिहाज नइखे रह गइल । |
| BHSD362 | केहु केहू के सामने कवनो तरह खा लेता । |
| BHSD363 | दही चिउरा के जगहा अब चाउमीन , पनिर , डोसा , इडली अउरी का का कचकच खा ता लोग । |
| BHSD364 | अब रउए बताईं ई नवका रहन आ पेट अउरी चालढाल दुनू के निगलत जात बा , कि नां ? |
| BHSD365 | बुढ़ापा आदमी के अवसान का पहिले क आखिरी पड़ाव ( लास्ट स्टेज ) ह । |
| BHSD366 | शरीर के कमजोरी आ अक्षमता त बढ़िये जाला , ऊपर से परिवार आ समाज के उपेक्षा आ अपनन के अमानवी तिरस्कार बूढ़ - ठेल अदिमी के भितरियो से तूरि देला । |
| BHSD367 | बूढ़वन के अनुभव आ ज्ञान से लाभ उठवला का बजाय ओके चीझु बतुस लेखा एन . पी . ए . ( नान प्रोडक्टिव ) मानि के छोड़ि दिहल हमनी का सभ्य समाज खातिर लज्जाजनक बा ! |
| BHSD368 | अभी दुइये महीना पहिले साइत जून में एगो समाचार सुने मे़ आइल रहे कि महिपालपुर ( दिल्ली ) का "गुरुकुल बृद्धाश्रम" में पानी का कमी आ तेज गरमी से बूढ़ मरि गइलन सँ । |
| BHSD369 | कवनो किसान भा परिस्थिति के मारल अदिमी का आत्महत्या पर सबक लेके ओके रोके क उतजोग होखे भा ना होखे बाकि विधान सभा , संसद आ मीडिया चैनलन म़े हंगामा क सीन त अकसरे लउकेला । |
| BHSD370 | बाकि देश का नगर - महानगरन में वृद्धाश्रमन में फेंकल - ढकेलल गइल अनगिन बुजुर्गन के दुर्भाग , बेबसी मे रिरिक - रिरिक के मुवला पर ना त समाज , ना मानवाधिकारी , ना जनप्रतिनिधि ( विधायक ) आ ना बात - बेबात लाइव टेलिकास्ट देखावे वाला मीडिया में से केहू गंभीरता से ना ले ला । |
| BHSD371 | कुछ करे भा ना करे हालो ना पूछे । |
| BHSD372 | अबहीं 8 अगस्त के "हिन्दुस्तान" हिन्दी दैनिक में एगो वृद्धाश्रम के हाल पढ़े के मिलल । |
| BHSD373 | निहाल बिहार ( पच्छिम दिल्ली ) में किराया पर चले वाला "साईं वृद्धाश्रम" किहाँ जगह का कमी का चलते बूढ़ - बुजुर्ग फुटपाथ का टीन शेड में सूते के मजबूर बाड़न सँ । |
| BHSD374 | सर्दी , गरमी , बरसात का विपरीत हालातो में माछी मच्छर का बीच बेमार भइल आ मुवल सुभाविक बा , बाकि ई सब देख सुन के मन घवाहिल हो जा ता । |
| BHSD375 | सुने में ईहो आइल कि मँगनी आ दान से चले वाला एह बृद्धाश्रम में अकसर पुलिसो वाला फुटपाथ प दिन काटे वाला बुढ़वन के छोड़ जालन सँ । |
| BHSD376 | जगह क ठेकाने ना आ अँड़सा में ठेला ठेली ! बाहर भीतर से टूटल छितराइल वृद्धन के जवने ठेकान मिल जाव ऊहे बहुत । |
| BHSD377 | कम से कम एह खस्ताहाल आश्रमन म़े असरा आ दाना पानी त मिल जा ता । |
| BHSD378 | इहवाँ त आम आदमी क सरकार बिया आ राजधानी गुने केन्द्रो क सरकार सटले बिया । |
| BHSD379 | ऊहो मानवतावादिये हिय s । |
| BHSD380 | बाकि लब्बो लुआब ई कि इहाँ इन्सानी प्रेम आ संवेदना का बजाय इन्सानी फितरत आ कुतरक बहुत बा । |

| BHSD381 | लागत बा कि सँचहूं घोर कलिजुग आ गइल बा। |
|---|---|
| BHSD382 | आखिर एह धन्नासेठन आ बीस हजारी थाली खियावे वालन के कवनो मानवी जिम्मेदारी बा कि ना ? भला होखे "हिन्दुस्तान" अखबार का हिन्दी पत्रकार रोहित के , जे कम से कम राजनीतिए के पत्रकारिता माने वाला चरफर पत्रकारन में कवनो जरत धधकत समाजिक आ राज्य का मसला के उठावे आ छापे के साहस त कइलस। |
| BHSD383 | सरकार केहू क होखे , बनावल हमनिये के ह। |
| BHSD384 | त हमनी के ओसे पुछहूं क अथिकार लोकतत्रे देले बा। |
| BHSD385 | त हमन के अपनो से आ सरकारो से पूछे के चाहीं नू कि हमनी का व्यवस्था में अइसन अमानवीयता काहें बा ? एकरा निदान खातिर का होत बा ? |
| BHSD386 | गाँवो कस्बा में जगहे जगह कुपूत आ कुबहू बाड़ी सँ जवन बूढ़ माई बाप के डाहे झहियावे में अगसर रहेले सँ आ शहर त खैर मतलबपरस्ती खातिर बदनामे बा। |
| BHSD387 | अपना बूढ़ बुजुर्ग आ अशक्य पुरनियन का दिसाई बढ़त संवेदनहीनता आ अगंभीरता प हमार मन बहुत आहत बा आखिर ई कवन विकसित सभ्य समाज हवे जेकर इकाई "परिवार" काम से रिटायर अक्षम वृद्धन के उपयोग कइला का बाद अंतिम घरी असहाय छोड़ देता ? ई सरकार क कवन व्यवस्था ह s , कि ओकरा आँखी का सामने ओकरे राज्य के नागरिक दाना पानी इलाज आ आश्रय का अभाव में रिरिक रिरिक के मर जा ताड़न स ? मानवाधिकार एजेन्सियन के आँख काहें मुनाइल रहत बा ? एह बूढ़ पुरनियन के आह एह समाज आ देश पर न पड़ी ! आईं सभे अपना घर परिवार समाज से ई अमानवी संवेदनहीनता खतम करीं जा ! |
| BHSD388 | कबो ट्रेन के टाइम त कबो हवाई जहाज के टाइम का चलते अनसोहाते मौका मिल जाला बतकुच्चन से आराम करे के। |
| BHSD389 | सोहाव त ना बाकिर कुछ देर ला सोहावन जरूर लागेला। |
| BHSD390 | आ आज एहीसे अनसोहाते पर बतकुच्चन करे के मन बनवले बानी। |
| BHSD391 | अब रउरा एहसे अनस आवे , रउरा अनसा जाईं , राउर मन अनसाइल हो जाव त अलग बात बा बाकिर हमार मकसद रउरा के अनसावल इचिको नइखे। |
| BHSD392 | हँ कुछ अनसाह लोग हमेशा बहाना खोजत रहेला अनसाए के , त ओह लोग के जरूर मौका मिल जाई। |
| BHSD393 | रउरा पूछ सकीलें कि अनसोहाते से अनस के का संबंध , त बतावल जरूरी हो जाई कि जब कुछ अनसोहाते होखे लागेला तबे अनस बड़ेला , मन खिसियाला। |
| BHSD394 | एह अनस आ अनसोहाते के संबंध बहुते घनिष्ठ होला। |
| BHSD395 | काहे कि अनसोहाते होखे वाला बात , मन मरजी का खिलाफ होखे वाला बात , केहू के सोहाव ना। |
| BHSD396 | अचके भा अचानके होखे वाला बात जरूरी नइखे कि अनसोहातो होखे। |
| BHSD397 | अनसोहात के अचानक होखलो जरुरी ना होखे। |
| BHSD398 | ऊ त बस अनसोहात होला जवना से मन के अनसोहाती महसूस होला। |
| BHSD399 | मोहन आ सोहन नीक दुनू लागेला बाकिर मोहन अपना मोहिनी से मोहेला बाकिर सोहन अपना सोहाइल भइला का चलते। |
| BHSD400 | खेत भा बाग से खर पतवार के छाँट के निकाले के काम निराई भा सोहनी कहल जाला। |
| BHSD401 | का बताईं। |
| BHSD402 | घरे रहनी हँऽ तऽ काम पार नाहीं लागत रहल हऽ। |
| BHSD403 | खेते में जाए में छाती फाटत रहल हऽ। |
| BHSD404 | गोबर-गोहथारी तऽ दूर के बाती बा, चउवनों कुली के नादे पर बँधले में हाथे में सानी-पानी लगले के डर बनल रहत रहल हऽ। |
| BHSD405 | बिहाने-बिहाने अगर केहू जगा दी त मोन करी की गोली मार दीं। |
| BHSD406 | जब बाबा दुआर बहारे के कहिहें त कवनो बहाना बना के घर में घुस जात रहनी हँ। |
| BHSD407 | अरे एतने ना, पढ़ाइयो में एकदम्मे मन नाहीं लागत रहल हऽ। |
| BHSD408 | दुपहरिया में स्कूले के देवालि फानि के भगले में एगो अलगहीं मजा आवत रहल ह। |
| BHSD409 | खाली होत का रहल हऽ की बनी-ठनी के लुहेरवन कुली के संघे लुहेराबाजी। |
| BHSD410 | गाँव-जवार घुमाई अउर सांझीखान चउरहा पर पकउड़ी के कटाई। |
| BHSD411 | दस-पाँच रुपिया अगर ना कहीं से मिलल त 4-5 किलो धाने भा गोहुएँ ले के सइकिल दनदनावत बाजारी में निकल जात रहनी हँ। |
| BHSD412 | खूब घुमाई होत रहल ह। |
| BHSD413 | जवार-जिला के एक्को मेला छुटत ना रहल ह। |
| BHSD414 | जब घर के केहु पुरनिया कुछ कही, घोंघिया के परि जात रहनी हँ। |
| BHSD415 | अब नु बुझाता जी। |
| BHSD416 | बाबूजी, बाबा, माई, ईया, भईया, भउजी, काका, काकी, फूफा, फूआ, मामा, मामी सब केहु समझावत रहल हऽ की ए बाबू, निमन से पढ़ाई करऽ। |
| BHSD417 | घर की कामे में हाथ बँटावऽ। |
| BHSD418 | जीवन हमेसा सुखदाई रही। |
| BHSD419 | लेकिन इ सब बाती हमरा जहर बुझात रहल हऽ। |
| BHSD420 | हमरा इ बुझात रहल हऽ की जिनगिया हमेसा एहींगा रही। |
| BHSD421 | जिनगीभर काका, बाबूजी कमा के खिआई लोग अउरी हमार हुरदंगई चालू रही। |
| BHSD422 | अब नु बुझाता जी। |
| BHSD423 | खैर रउआँ का बुझाई, जेकरी पैर न फटे बिवाई, उ का जाने पीर पराई। |
| BHSD424 | पढ़ाइयो-ओढ़ाई छोड़ले की बादो बाबूजी केतना समझवने की बाबू, पढ़ाई-लिखाई में मन नइखे लागत तऽ कवनो बाती नाहीं। |
| BHSD425 | अपनी घर के काम करऽ, जवन दू-चारी कट्ठा खेत बा ओमें मेहनती करऽ। |
| BHSD426 | दू-चारी पइसा के आदमी हो जइबऽ। |
| BHSD427 | कवनो छोट-मोट कामो-धंधा खोजी के कऽ लेहल करऽ, अरे कुछु पार नइखे लागत तऽ एगो छोट-मोट दोकनिए खोली के बइठी जा। |
| BHSD428 | पर बाबुजी के इ सब बाति जहर बुझाई अउर चाहबि की केतना जलदी बाबुजी की सामने से दूर हो जाई। |
| BHSD429 | हमरा घूमले से, चउरहा पर बइठी के राजनीती कइले अउरी दूसरे के टाँगखिंचाई कइले से, चाय पीयले से, पान चभुरवले से, दू-चारी जाने की संघे लिहो-लिहो कइले से, कबो भाजपा अउरी कबो कांग्रेस के झंडा ढोवले से अउरी कबो-कबो चुनाव की समय झगड़ा अउरी मारा-मारी कइले से फुरसत काहाँ मिलत रहल हऽ। |
| BHSD430 | कबो-कबो निमनो बाती पर घर की बड़-बुजुरुगन पर घोंघिया के चढ़ी बइठत रहनी हँऽ। |
| BHSD431 | केहू केतनो निमन समझाई, हमरा जहरे बुझाई। |
| BHSD432 | एकदिन माई-बाप की समझवले से आजिज हो के बंबई चली अइनी। |
| BHSD433 | इहवाँ आ के जब सड़की पर सूते के परल, गारी-गुप्ता खाए के परल, हगले खातिर लाइन लगावे के परल, पानी की चलते कई-कई महीना बिना नहाए रहे के परल तऽ दिमाग ठिकाने आ गइल। |

| BHSD434 | रहले के परेसानी, खइले के परेसानी अउरी इहवाँ तक की दूसरे के सलामी बजवले के परेसानी। |
|---|---|
| BHSD435 | कास माई-बाप के बाती मान लेले रहतीं। |
| BHSD436 | कास, तनियो एसा पढ़ि-लिखि ले ले रहतीं। |
| BHSD437 | आजु केतना बुझाताऽ। |
| BHSD438 | अरे अगर गउँओ में हम मेहनत करतीं तऽ दू पइसा जरूर कमा लेतीं अउरी उहो इजती से। |
| BHSD439 | घरे माई की हाथे के बनावल खाए के त मिलित। |
| BHSD440 | घर की लोग के साथ मिलितऽ। |
| BHSD441 | मेहनती करेवाला गउँओ में आराम से बा। |
| BHSD442 | दु पइसा के आदमी बा। |
| BHSD443 | मेहनती करे वाला कवनोंगा दु जुनि के जोगाड़ कइए लेला। |
| BHSD444 | घरे गोबर काढ़त में छाती फाटे अउरी इहवाँ उहे गोबरा कढ़ाता। |
| BHSD445 | घरे साग-भाजी उपजवले, बेंचले में आपन बेइजती महसूस होखे अउरी उहे कमवा इहवाँ कइल जाता। |
| BHSD446 | बड़ी अपसोस होता। |
| BHSD447 | खैर अब जवन भइल उ भइल पर अब हमार लोग-लइका घरवे रही। |
| BHSD448 | जवन मेहनत हम इहवाँ करतानी, उ मेहनती हमार बाल-बच्चा घरवे करी, अपनन में, अपनन खातिर, अपनन की बीच। |
| BHSD449 | अब एकदम बुझा गइल बा की "बंबइया से नीक घरवे बा...गउवें बा।" अब दिमाग एकदम्मे ठिकाने आ गइल बा। |
| BHSD450 | जेई दिन हम दादर में पिटइनी, हम गरीब के समोसा सड़की पर अपने भाई लोग छिंटी देहल। |
| BHSD451 | आरे हमार कालर पकड़ाइल तऽ कवनो बाती नाहीं पर हमरी बहिन-बेटी के इजती उतारि लेहल गइल। |
| BHSD452 | ओई दिन हमरा अपनी माई-बाप के उ बाती यादी आ गइल, "बाबू! घरवो रही के, मेहनती कऽ के, इजती से दु पइसा कमाइल जा सकेला। |
| BHSD453 | तूँ बाहर जा के जेतना मेहनती करऽतारऽ अगर ओकरी अधवा घरवो करतऽ तऽ केतना ठीक रहीत। |

| BHSD454 | आपन लोगन की बीचे में रहतऽ, अपनी मन के मालिक रहतऽ।" एतने ना अउर भी प्रदेसन में भी हमके दुरदुरा देहल जाता। |
| --- | --- |
| BHSD455 | नेता लोग कहत फिरता की पूरा भारत एक हऽ। |
| BHSD456 | कहले में अउरी होखले में केतना अंतर बा। |
| BHSD457 | का एक भइले के इहे मतलब बा की केहु के कमइले के अधिकार नइखे अउरी अपने कमाई खइले के अधिकार नइखे। |
| BHSD458 | देस में बेरोजगारी त बटले बा, प्रांतवाद, क्षेत्रवाद आदि भी पूरा तरे हाबी बा। |
| BHSD459 | हर जगहि गरीबे पिसाता। |
| BHSD460 | गरीब के सुने वाला केहू नइखे। |
| BHSD461 | कहे के त भारत एक बा, सबके समान अधिकार बा, पर इ कुल खाली कहहीं में ठीक लागता। |
| BHSD462 | जमीनी स्तर पर देखीं त बहुते असमानता बा। |
| BHSD463 | केहू खात-खात मुअता त केहू खइले बिना। |
| BHSD464 | अरे कुछ लोग के त कड़ी मेहनत की बादो दु जून के रोटी नसीब नइखे होत। |
| BHSD465 | मनई भी का करो, गाँव में परेसानी होता त सहर ध लेता अउर इहाँ गदहा, बैल की तर काम करता। |
| BHSD466 | कास, सब लोग अपनी बचवन के पढ़ा पाइत। |
| BHSD467 | कास, सब किसोर, जुबा पढ़ले पर धेयान देतें। |
| BHSD468 | कास, सरकारो सिछा पर सबके अधिकार जमीनी स्तर पर ले आइत, सब ठीक हो जाइत। |
| BHSD469 | सब एकदम्मे ठीक हो जाइत। |
| BHSD470 | बचपन में बाबा बार-बार एगो किस्सा सुनावें। |
| BHSD471 | एगो बहुत धनिक-मानिक अदमी रहे। |
| BHSD472 | ओकर घर धन-धान्य से भरल रहे। |
| BHSD473 | घोरसार, हथीसार, गउसाल सब आबाद रहे। |
| BHSD474 | ओकरी घर में बहुते अनुसासन रहे। |
| BHSD475 | छोट बड़ के मान दे और बड़ छोट के नेह दे। |
| BHSD476 | एक बेर उ अदमी अकेले अपनी बगीचा में बइठल रहे। |
| BHSD477 | तवलेकहीं एगो मेहरारू उहाँ रोवत आइल। |
| BHSD478 | अदमी ओ मेहरारू से ओकरी रोवे के कारन पूछलसि। |
| BHSD479 | मेहरारू कहलसि की हम बिपति हई, हम तोहरी पर पड़े आइल बानी। |
| BHSD480 | पर तोहरी घर के अनुसासन, आपस के नेह-दुलार देखि के हमरा रोवाई आवता की एतना नीमन परिवार अब तहस-नहस हो जाई। |
| BHSD481 | उ अदमी तनी हँसल अउर कहलसि की ठीक बा। |
| BHSD482 | तूँ पड़े आइल बाड़ू त परबे करबू। |
| BHSD483 | पर एइसन पड़ऽ की हमार धन-धान्य बरबाद न होखो। |
| BHSD484 | घोड़ा घोरसारे रहि जां, हाथी हथीसाले अउर गाइ गउसाले। |
| BHSD485 | ओ अदमी के इ बाति सुनि के उ बिपति कहलसि की फेर त हमार पड़ल का कहाई। |
| BHSD486 | फिर उ अदमी कहलसि की ठीक बा तूं अपनी हिसाब से पड़ऽ पर हमरी घर के अनुसासन पर आपन नजर मति दिहऽ। |
| BHSD487 | बिपति मानी गइल। |
| BHSD488 | बिपती की पड़ते उनके सबकुछ बरबाद हो गइल। |
| BHSD489 | घर में कुछु ना बचल। |
| BHSD490 | घोरा, हाथी, खेत-बारी सबकुछ तहस-नहस हो गइल। |
| BHSD491 | घर में एतनो अन्न ना बंचल की एक्को बेरा के खाना बनि सको। |
| BHSD492 | एइसन हालत होते, पूरा परिवार एकट्ठा होके, ओ अदमी से पूछल की अब बताई का कइल जाव? |
| BHSD493 | उ अदमी कहलसि की घबरइले के ताक नइखे, घर में जवन एक-आध गो बरतन बँचल बा ओके बाँधि ल जा, हमनी जान अब्बे आपन गाँव-घर छोड़ि के दूसरे राज की ओर चलल जाई अउर उहवें कुछ कमाइल-खाइल जाई। |

| | | |
|---|---|---|
| BHSD494 | ओकरी बाति के सुनते ही घर-परिवार के लोग बँचल बरतन आदी बाँधि के ओ अदमी की पीछे-पीछे दूसरे राज की ओर निकल गइल। | |
| BHSD495 | चलत-चलत साम हो गइल। | |
| BHSD496 | उ अदमी एगो पेड़े की नीचे रुकि के अपनी घरवालन से कहलसि की आजु के राति हमनी जान इहवें काटल जाई अउर फेर बिहने सबेरे आगे बढ़ल जाई। | |
| BHSD497 | देखत-ही देखत ओ अदमी के एगो परिवार के सदस्य उहवें बँचल समान ध देहल लोग। | |
| BHSD498 | एक-आधगो बोरा-चट्टी बिछा के घर की बड़-बुजुर्गन अउर बच्चन के बइठा देहल लोग। | |
| BHSD499 | एकरी बाद उ अदमी घर की मेहरारू कुल से कहलसि की अब तोह लोगन खाना बनवले के इंतजाम करऽ जा। | |
| BHSD500 | उ एक आदमी के आगि ले आवे के त एक आदमी के पानी ले आवे के भेजलसि। |

| | Appendix 3 |
| --- | --- |
| | **Sample Set of English-Bhojpuri Parallel Corpus** |

| ID | EN-BHO Sentences |
| --- | --- |
| EBHJNUD01 | Two thousand copies of the book were printed.\|\|\|दो हजार पन्ना किताब के छापल गईल रहे । |
| EBHJNUD02 | The book which is on the table is mine.\|\|\|मेज पर जवन किताब हऽ उ हमार हऽ । |
| EBHJNUD03 | The cover of the book is soft.\|\|\|किताब के आवरण बहुत कोमल हऽ । |
| EBHJNUD04 | The ninth chapter of the book is very interesting.\|\|\|किताब के नौऊवा अध्याय बहुत आनन्दायक हऽ । |
| EBHJNUD05 | The book is very interesting.\|\|\|किताब बहुत रोचक हऽ । |
| EBHJNUD06 | The book is new.\|\|\|किताब नया हऽ । |
| EBHJNUD07 | The boy reading a book is rich.\|\|\|लईका जवन किताब पढ रहल बा धनी हऽ । |
| EBHJNUD08 | I read the book.\|\|\|हम किताब पढनी । |
| EBHJNUD09 | The book is old.\|\|\|किताब पुरान हऽ । |
| EBHJNUD10 | The book lies in a heap on the floor.\|\|\|किताब छत के अम्बार पर पड़ल हऽ । |
| EBHJNUD11 | Even the book is good.\|\|\|जबकि किताब अच्छा हऽ । |
| EBHJNUD12 | The book does not belong to me.\|\|\|किताब के संबंध हमरा से ना हऽ । |
| EBHJNUD13 | The book belonged to me.\|\|\|किताब के संबंध हमरा से रहे । |
| EBHJNUD14 | The book fell from the table to the floor.\|\|\|किताब मेज से छत पर गिर गईल । |
| EBHJNUD15 | The book on the table.\|\|\|किताब मेज पर हऽ । |
| EBHJNUD16 | The book is on the desk.\|\|\|किताब डेस्क पर हऽ । |
| EBHJNUD17 | The book lies on the table.\|\|\|किताब मेज पर पड़ल हऽ । |
| EBHJNUD18 | The book is being sent.\|\|\|किताब के भेजल जा रहल बा । |
| EBHJNUD19 | The book is green.\|\|\|किताब हीया हऽ । |
| EBHJNUD20 | Here is the book.\|\|\|एईजा किताब हऽ । |
| EBHJNUD21 | Here is a book.\|\|\|एईजा एगो किताब हऽ । |
| EBHJNUD22 | A book is being written by him.\|\|\|एगो किताब उनका द्वारा लिखल जा रहल बिया । |
| EBHJNUD23 | Of where has the post come ?\|\|\|कहा के पद आईल हऽ । |

| EBHJNUD24 | Kipling was not born in London.|||किपलिंग लंदन में ना पैदा भईल रहन । |
|---|---|
| EBHJNUD25 | Kim seems never to be alone.|||किम कभी भी अकेला न दिखेला । |
| EBHJNUD26 | Kim had a coffee in the café while she waited for the post office to open.|||किम के एगो काफी एगो केफे में रहे जबकि उ लड़की डाकघर के खुले के इन्तजार करत रहे । |
| EBHJNUD27 | Kiri is said to be very rich.|||किरी के बहुत धनी कहल जाला । |
| EBHJNUD28 | Kim must not drink the wine on the table.|||किम के शराब मेज पर ना पिये के चाही । |
| EBHJNUD29 | Kim said he could have heard the news, but Lee said that he could not.|||किम कहलस कि उ समाचार सुन सकलस, लेकिन ली कहलस कि उ ना सुन सकलस । |
| EBHJNUD30 | Kim has danced, and Sandy has.|||किम नृत्य कईलस लेकिन सेंडी ना । |
| EBHJNUD31 | Kim can dance, and Sandy can.|||किम नृत्य कर सकता और सेंडी भी । |
| EBHJNUD32 | Kim was dancing, and Sandy was.|||किम नाचत रहे और सेंडी भी । |
| EBHJNUD33 | Kim may not drink the wine on the table.|||किम के शराब मेज पर ना पिये के चाही । |
| EBHJNUD34 | Don't move Kim !|||आगे मत बढ किम ! |
| EBHJNUD35 | Fall !|||गिरल ! |
| EBHJNUD36 | Kiran took out the car and went straight to her sister's office.|||किरन कार लिहलस और सीधे अपना बहिन के आफिस गईल । |
| EBHJNUD37 | The falling leaves made me think about the coming autumn.|||पत्तन के गिरल इ बतावे ला कि रात आवेवाला बा । |
| EBHJNUD38 | It is mean to crow over a fallen foe.|||कौवा के गिरल के मतलब शत्रु से होला । |
| EBHJNUD39 | In disgust he threw up his appointment.|||घृणा में उ आपन नियुक्ति के फेंक दिहलस । |
| EBHJNUD40 | A major feature of the fort and palaces is the superb quality of stone carvings.|||किला व महलन के मुख्य चित्र पत्थर नक्काशी के महानतम गुण हऽ । |
| EBHJNUD41 | The castle is well worth a visit.|||दुर्ग घुमे के अच्छा जगह हऽ । |
| EBHJNUD42 | Go then, said the ant, "and dance winter away."|||जा औरी चींटीं से कह |

| | |
|---|---|
| | कि नाच जाड़ा चल गईल । |
| EBHJNUD43 | An army of ants will attack large and ferocious animals.|||चींटीयन के समूह बड़ व उग्र जानवरन पर आक्रमण करी । |
| EBHJNUD44 | The ants fought the wasps.|||चींटीयन ने भीड़ से लड़ल रहे । |
| EBHJNUD45 | How many brothers do Kishan have ?|||किसन केतना भाई हऽ । |
| EBHJNUD46 | How many sisters do Kishan have ?|||किसन के केतना बहन हई । |
| EBHJNUD47 | Where is Kishan's brother ?|||किसन के भाई कहा हऽ । |
| EBHJNUD48 | Who is the guest of Kishan ?|||किसन के अतिथी के हऽ । |
| EBHJNUD49 | Kishan was aking a question to teacher.|||किसन एगो प्रश्न अध्यापक से पूंछत रहे । |
| EBHJNUD50 | Kishan is about to come and I'll go with him.|||किसन आवेवला हऽ और हम ओकरा संगे जाईव । |
| EBHJNUD51 | Kishore is thinking of going to Ajmer.|||किशोर अजमेर जायेके सोच रहल बा । |
| EBHJNUD52 | Kishore will come.|||किशोर आई । |
| EBHJNUD53 | Kishore will come, won't he ?|||किशोर आई, उ ना आई । |
| EBHJNUD54 | Kishan does not get angry with anyone without thinking.|||किसन कभी भी बिना सोचे कभी केहूं पर गुस्सा ना होला । |
| EBHJNUD55 | Adolescents do not need specific sets of dietary guidelines .||| एडोलसेन्ट के विशेष रास्ता मार्गदर्शन समूह के आवश्यक्ता न हऽ । |
| EBHJNUD56 | Prices start from $1.|||दाम शुरु होला डालर 1 से । |
| EBHJNUD57 | The price was ill.|||दाम बेकार रहे । |
| EBHJNUD58 | Prices ought to come down soon.|||दाम जल्दी ही निचे आवे अके चाही । |
| EBHJNUD59 | The cost is twelve rupees.|||दाम 12 रुपया हऽ । |
| EBHJNUD60 | A valuable ring was found yesterday.|||काल एको मूल्यवान अंगूठी मिलल रहे । |
| EBHJNUD61 | Who is the head of chemistry department ?|||रासायन विभाग के मुख्य के हऽ ? |
| EBHJNUD62 | There was a spy on the corner. |||ओइजा कोना में एगो जासूस रहे । |
| EBHJNUD63 | A spy was on the corner. |||एगो जासूस कोना में रहे । |
| EBHJNUD64 | The hooks pierced his mouth.|||हुक ओकरा मुँह के छेद देहलस । |

| EBHJNUD65 | With whatever luxuries a bachelor may be surrounded, he will always find his happiness incomplete, unless he has a wife and children.|||जवानी में भले ही उ आरामदेह वाला जिंदगी से घिरल होई लेकिन उ हमेशा आपन खुशी उ अधूरा पावेला नादि त ओका एगो पत्नी व बच्चा भी हऽ । |
|---|---|
| EBHJNUD66 | The rat came in when the cat is away.|||मूस तब उ ऊवे जब बिल्ली चल गईल रहे । |
| EBHJNUD67 | The mouse tried to get out of the basket.|||चूहा टोकरी से बाहर आवे के कोशिश करत रहे । |
| EBHJNUD68 | Taking pity on the mouse, the magician turned it into a cat.|||चूहा पर मोह खाकर जादूगर ओके बिल्ली बना देहुवे । |
| EBHJNUD69 | The mouse fed greedily on the corn.|||चूहा मक्का के लोभ से खाइवे । |
| EBHJNUD70 | The mouse rejoiced in his good fortune.|||चूहा अपना अच्छा भाग पर खुश रहे । |
| EBHJNUD71 | The mouse was killed by the cat.|||चूहा बिलाई के द्वारा मार दिहल गईल रहे । |
| EBHJNUD72 | The hen has laid an egg.|||मुर्गी अंडा पर लेटल रहे । |
| EBHJNUD73 | When does the cock crow?|||जब कौउवा काव काव करत हऽ । |
| EBHJNUD74 | Something unusual happened. |||कुछ अनावश्यक हो जाला । |
| EBHJNUD75 | Some such seekers would grope in the way and would be taken to a forest.|||कुछ जिज्ञासु के रास्ता में टटोलल जाता और उनके जंगल में ले जाईल जाई । |
| EBHJNUD76 | Some say one thing and others another.|||कुछ एगो चीज कह और दूसरा केहूँ दूसर । |
| EBHJNUD77 | Few historians have written in more interesting manner than Gibbon.|||गिब्बन से अधिक रोचक तरीके से कुछ इतिहासकार लिखले हऊवन । |
| EBHJNUD78 | A few Americans have their offices in Kolkata.|||कुछ अमेरिकन के कोलकाता में आफिस हऽ । |
| EBHJNUD79 | Very few boys are as industrious as Latif.|||बहुत कुछ बालक लोग कारोबारी हउवन लतीफ के तरह । |
| EBHJNUD80 | A few Parsees write Gujarati correctly.|||कुछ पारसी गुजराती सही तरीका से लिखेलन । |

| | | |
|---|---|---|
| EBHJNUD81 | In no time, he had fallen to his death.|||कवनो समय में उ अपना मौत के तरफ बढल रहे । | |
| EBHJNUD82 | Show some superior ones.|||प्रदर्शन सर्वश्रेष्ट में से एक रहे । | |
| EBHJNUD83 | Some praise the work, and some architect.|||केहूँ काम के प्रशंसा करुवे त केहूँ रचना के । | |
| EBHJNUD84 | Do you have any work ?|||तोहरा के कवनो काम बा । | |
| EBHJNUD85 | Something may be worth doing.|||कुछ मूल्य के लायक काम हऽ । | |
| EBHJNUD86 | There is nothing to do.|||एईजा कुछु करे के ना हऽ । | |
| EBHJNUD87 | Some poets are at least as great as Tennyson.|||कुछ कवि कमसे कम एतना प्रसिद्ध हऊवन जेतना तानसेन । | |
| EBHJNUD88 | Some poets are not less great than Tennyson.|||कुछ कवि तानसेन स कम महान ना हऊवन । | |
| EBHJNUD89 | Some ants fight very fiercely.|||कुछ चींटीयां बहुत हिंसक तरीके से लड़ाई करत रहे । | |
| EBHJNUD90 | Nothing special.|||कुछ विशेष न हऽ । | |
| EBHJNUD91 | Some milk was split.||कुछ दूध फाड़ल गईल रहे । | |
| EBHJNUD92 | Some were acquitted, and some punished.|||कुछ के रिहा क देहल गईल रहे और कुछ के दण्ड दिहल गईल रहे । | |
| EBHJNUD93 | Some say he is a sharper.|||कुछ कहेलन कि उ बहुत तेज हऽ । | |
| BHHJNUD94 | How many old ages people cannot cross the road?|||केतना बुढ आदमी सड़क पार न कर सकेलन । | |
| EBHJNUD95 | Some were born great.|||कुछ महान पैदा भईल रहलन। | |
| EBHJNUD96 | Few persons can keep a secret.|||बहुत कम आदमी रहस्य के बनाये रख सकतन । | |
| EBHJNUD97 | Some boys started singing.|||कुछ लड़का गावें शुरु कईल रहे । | |
| EBHJNUD98 | The wounded man was being helped by some boys.|||घायल आदमी के कुछ लईकन द्वारा सहायता कईल गईल रहे । | |
| EBHJNUD99 | Few boys are not amenable to discipline.|||कुछ बालक अनुशासन के लेके उत्तरदायी न हऊवन । | |
| EBHJNUD100 | At some places, I went for interview too, but it didn't work | |

| | anywhere.|||कुछ जगहन पर हम साक्षात्कार खातिर गऊवीं लेकिन कहीं इ काम न कईलस । |

Meteor Precision by sentence length

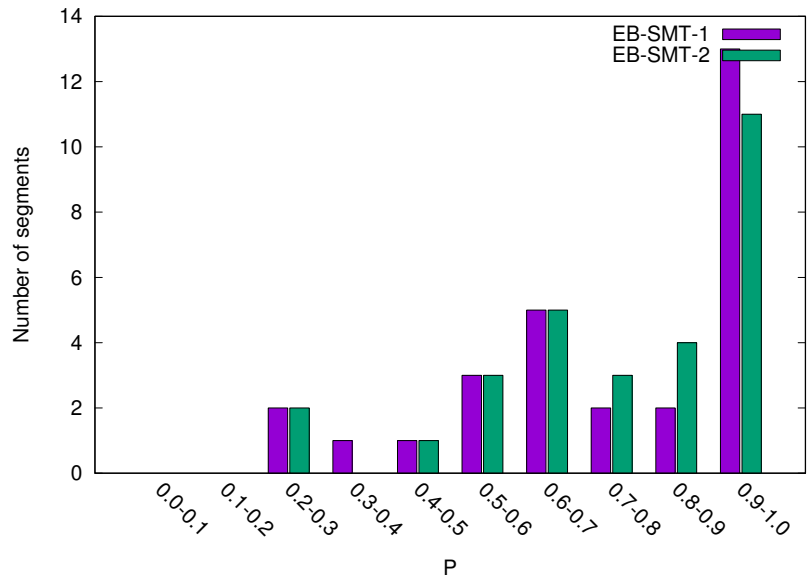

1–10 words

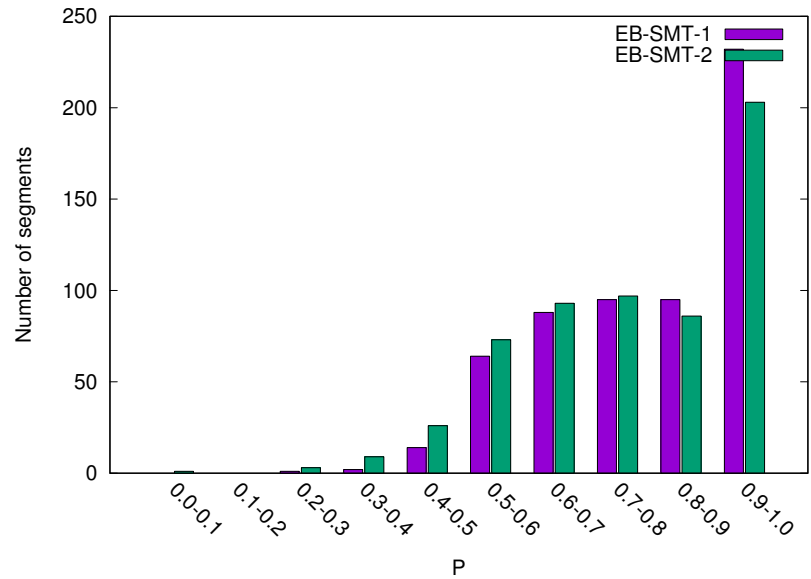

11-25 words

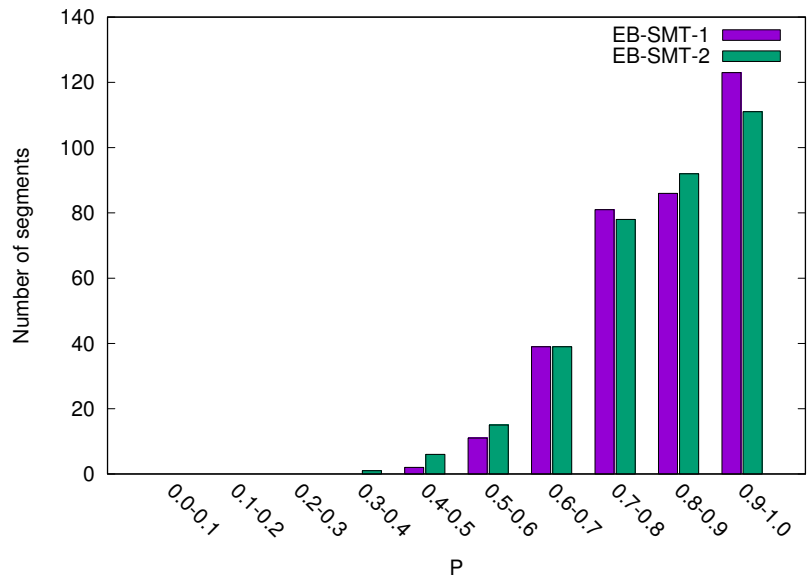

26–50 words

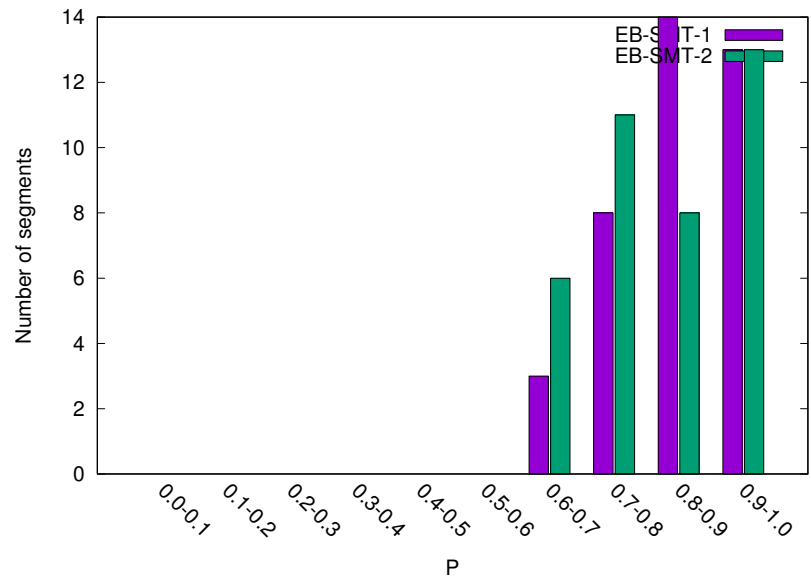

51+ words

**Appendix 4**

**Results of EB-SMT System 1 and 2 based on METEOR Metric**

## Meteor Recall by sentence length

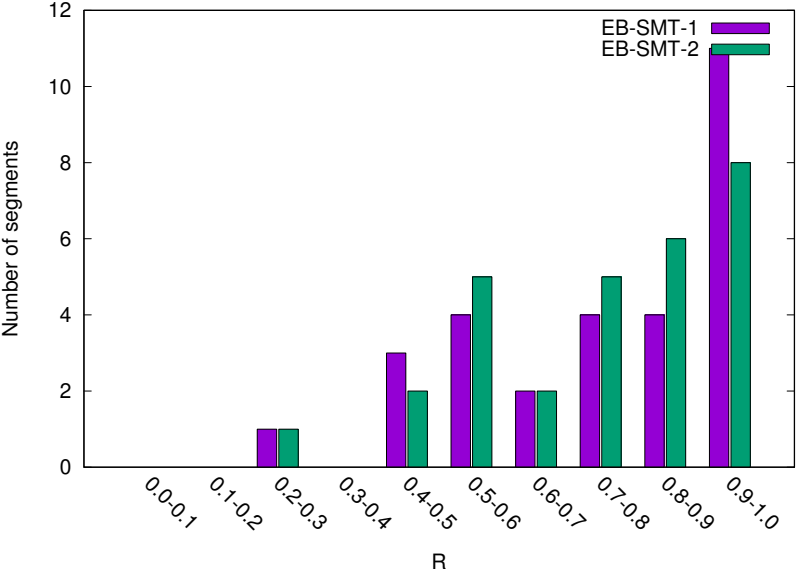

1–10 words

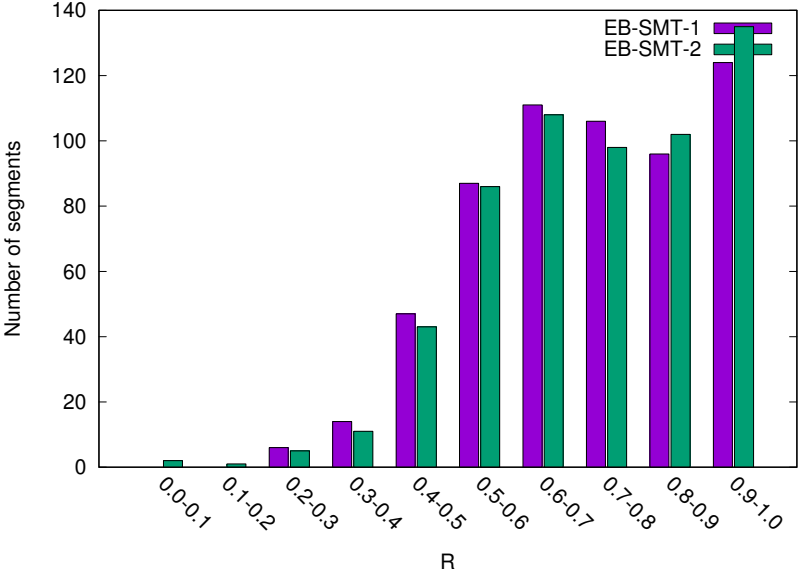

11-25 words

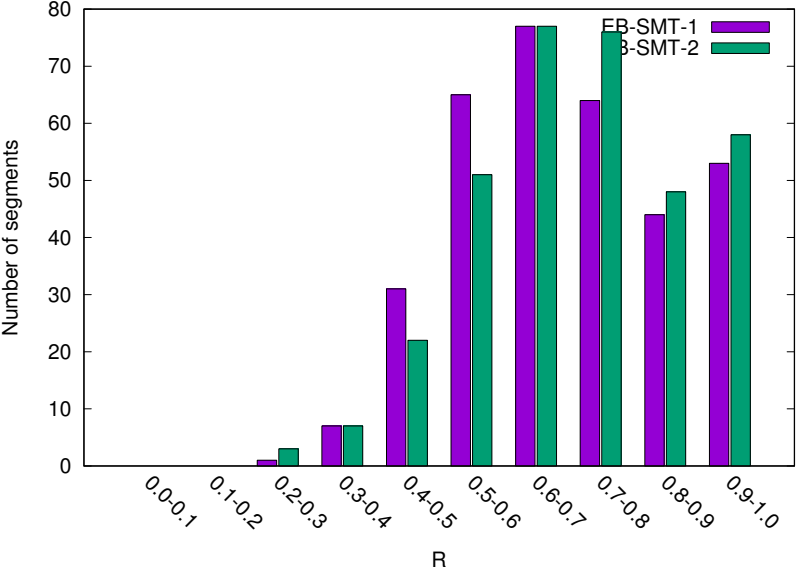

26–50 words

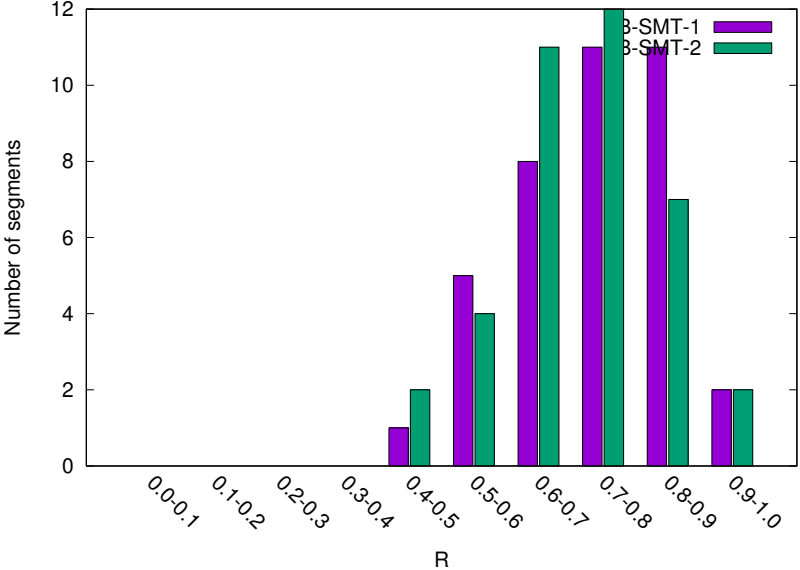

51+ words

## Meteor Fragmentation by sentence length

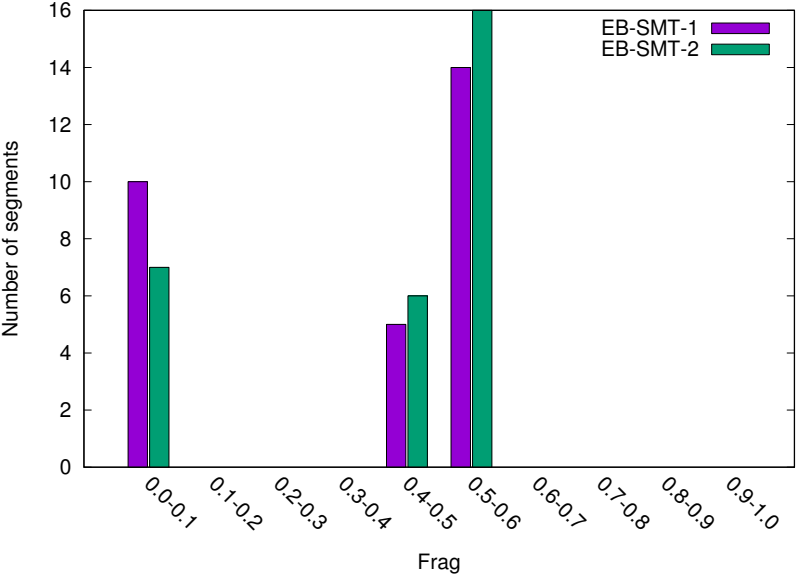
1–10 words

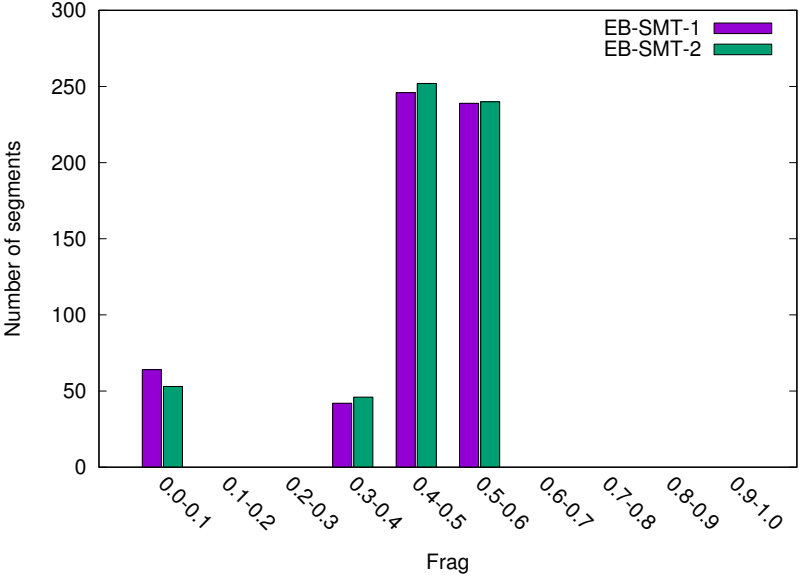
11-25 words

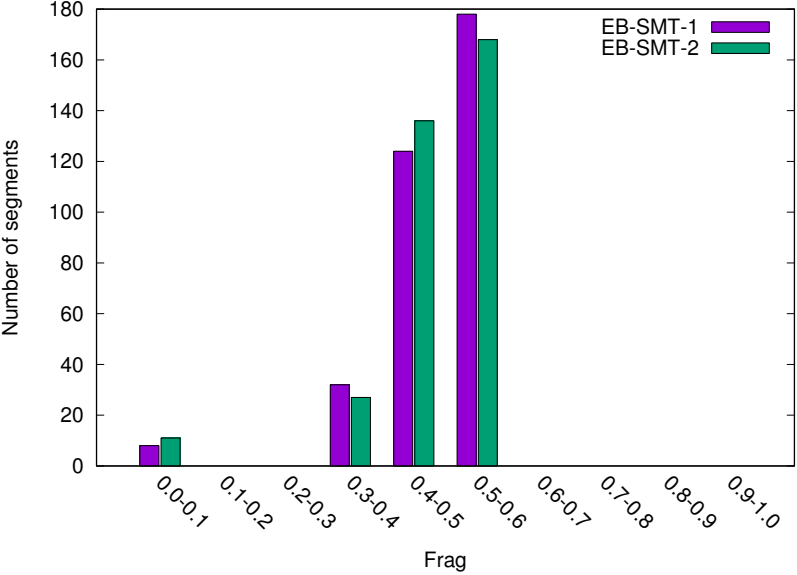
26–50 words

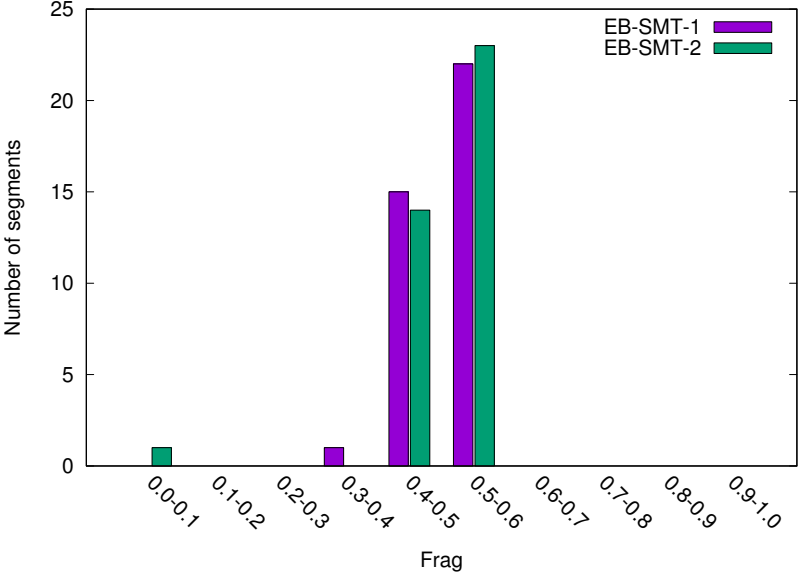
51+ words

# Appendix 5

**A Comparative Sample Set of Human Evaluation Scores of the PD based EB-SMT System**

| Sentence Id | Fluency:Eval-1 | Adequacy:Eval-1 | Fluency:Eval-2 | Adequacy:Eval-2 |
|---|---|---|---|---|
| 1 | 3 | 2 | 2 | 2 |
| 2 | 2 | 1 | 2 | 4 |
| 3 | 1 | 1 | 2 | 5 |
| 4 | 3 | 2 | 4 | 5 |
| 5 | 2 | 2 | 4 | 5 |
| 6 | 1 | 1 | 2 | 2 |
| 7 | 2 | 3 | 4 | 5 |
| 8 | 1 | 1 | 2 | 2 |
| 9 | 3 | 2 | 4 | 4 |
| 10 | 1 | 2 | 4 | 4 |
| 11 | 2 | 3 | 2 | 2 |
| 12 | 3 | 3 | 4 | 5 |
| 13 | 2 | 3 | 4 | 4 |
| 14 | 3 | 4 | 4 | 5 |
| 15 | 2 | 3 | 5 | 5 |
| 16 | 3 | 4 | 4 | 5 |
| 17 | 2 | 3 | 2 | 3 |
| 18 | 4 | 5 | 5 | 5 |
| 19 | 4 | 4 | 4 | 4 |
| 20 | 2 | 3 | 5 | 5 |
| 21 | 2 | 4 | 4 | 5 |
| 22 | 2 | 3 | 4 | 4 |
| 23 | 2 | 4 | 4 | 5 |
| 24 | 2 | 3 | 4 | 5 |
| 25 | 2 | 4 | 5 | 5 |
| 26 | 5 | 5 | 5 | 5 |
| 27 | 2 | 3 | 4 | 4 |
| 28 | 2 | 3 | 2 | 3 |
| 29 | 2 | 3 | 2 | 2 |
| 30 | 2 | 3 | 2 | 4 |
| 31 | 2 | 3 | 2 | 3 |
| 32 | 5 | 4 | 5 | 5 |
| 33 | 2 | 4 | 4 | 5 |
| 34 | 2 | 3 | 4 | 4 |
| 35 | 2 | 3 | 1 | 1 |
| 36 | 2 | 3 | 4 | 4 |
| 37 | 2 | 3 | 2 | 2 |
| 38 | 3 | 3 | 4 | 5 |
| 39 | 3 | 4 | 5 | 5 |
| 40 | 2 | 3 | 4 | 4 |
| 41 | 5 | 5 | 5 | 5 |
| 42 | 3 | 3 | 4 | 5 |
| 43 | 2 | 4 | 4 | 5 |
| 44 | 2 | 4 | 4 | 5 |
| 45 | 2 | 4 | 5 | 5 |
| 46 | 2 | 4 | 2 | 3 |
| 47 | 2 | 3 | 2 | 2 |
| 48 | 2 | 2 | 1 | 1 |
| 49 | 2 | 4 | 5 | 5 |
| 50 | 3 | 4 | 5 | 5 |
| 51 | 2 | 4 | 2 | 4 |
| 52 | 2 | 4 | 4 | 5 |
| 53 | 2 | 4 | 5 | 5 |
| 54 | 4 | 4 | 5 | 5 |
| 55 | 3 | 2 | 4 | 5 |
| 56 | 2 | 4 | 4 | 5 |
| 57 | 2 | 4 | 5 | 5 |
| 58 | 4 | 4 | 5 | 5 |
| 59 | 4 | 4 | 5 | 5 |
| 60 | 4 | 4 | 4 | 4 |
| 61 | 2 | 4 | 2 | 2 |
| 62 | 2 | 3 | 4 | 4 |
| 63 | 4 | 4 | 4 | 5 |
| 64 | 2 | 4 | 4 | 5 |
| 65 | 2 | 4 | 4 | 5 |
| 66 | 2 | 4 | 2 | 3 |
| 67 | 4 | 5 | 5 | 5 |
| 68 | 2 | 4 | 2 | 2 |
| 69 | 2 | 4 | 4 | 5 |
| 70 | 4 | 5 | 4 | 5 |
| 71 | 2 | 4 | 2 | 2 |
| 72 | 2 | 3 | 4 | 4 |
| 73 | 2 | 4 | 4 | 5 |
| 74 | 4 | 4 | 4 | 5 |
| 75 | 3 | 4 | 4 | 5 |
| 76 | 3 | 4 | 4 | 4 |
| 77 | 2 | 4 | 4 | 5 |
| 79 | 1 | 1 | 4 | 5 |
| 80 | 3 | 4 | 3 | 2 |
| 81 | 3 | 4 | 2 | 3 |
| 82 | 3 | 4 | 3 | 4 |
| 83 | 2 | 4 | 4 | 3 |
| 84 | 3 | 4 | 4 | 4 |
| 85 | 2 | 4 | 2 | 2 |
| 86 | 3 | 4 | 2 | 2 |
| 87 | 3 | 4 | 3 | 2 |
| 88 | 4 | 5 | 2 | 2 |
| 89 | 1 | 3 | 5 | 5 |
| 90 | 2 | 4 | 2 | 5 |
| 91 | 3 | 4 | 2 | 3 |
| 92 | 2 | 3 | 3 | 2 |
| 93 | 2 | 5 | 2 | 2 |
| 94 | 2 | 3 | 2 | 1 |
| 95 | 3 | 4 | 2 | 2 |
| 96 | 3 | 4 | 3 | 3 |
| 97 | 3 | 4 | 4 | 3 |
| 98 | 4 | 5 | 4 | 5 |
| 99 | 4 | 5 | 5 | 5 |
| 100 | 2 | 4 | 4 | 5 |

# Appendix 6

**A Comparative Sample Set of Human Evaluation Scores of the UD based EB-SMT System**

| Sentence Id | Fluency:Eval-1 | Adequacy:Eval-1 | Fluency:Eval-2 | Adequacy:Eval-2 |
|---|---|---|---|---|
| 1 | 2 | 2 | 4 | 4 |
| 2 | 4 | 2 | 4 | 4 |
| 3 | 4 | 5 | 2 | 2 |
| 4 | 2 | 4 | 2 | 3 |
| 5 | 2 | 4 | 4 | 4 |
| 6 | 2 | 5 | 2 | 2 |
| 7 | 4 | 4 | 4 | 4 |
| 8 | 2 | 5 | 2 | 2 |
| 9 | 1 | 3 | 4 | 2 |
| 10 | 1 | 4 | 4 | 2 |
| 11 | 1 | 4 | 2 | 2 |
| 12 | 2 | 4 | 4 | 4 |
| 13 | 2 | 5 | 2 | 2 |
| 14 | 2 | 4 | 4 | 4 |
| 15 | 2 | 4 | 4 | 5 |
| 16 | 1 | 3 | 4 | 4 |
| 17 | 1 | 2 | 2 | 2 |
| 18 | 2 | 4 | 4 | 4 |
| 19 | 2 | 4 | 2 | 3 |
| 20 | 3 | 4 | 4 | 4 |
| 21 | 1 | 4 | 4 | 3 |
| 22 | 1 | 4 | 1 | 1 |
| 23 | 2 | 4 | 4 | 3 |
| 24 | 1 | 3 | 4 | 3 |
| 25 | 2 | 5 | 4 | 4 |
| 26 | 4 | 4 | 4 | 4 |
| 27 | 2 | 4 | 4 | 3 |
| 28 | 2 | 4 | 4 | 2 |
| 29 | 2 | 4 | 2 | 2 |
| 30 | 1 | 4 | 2 | 2 |
| 31 | 1 | 4 | 2 | 2 |
| 32 | 4 | 4 | 4 | 4 |
| 33 | 2 | 3 | 2 | 2 |
| 34 | 2 | 4 | 2 | 2 |
| 35 | 2 | 2 | 1 | 1 |
| 36 | 1 | 3 | 4 | 2 |
| 37 | 2 | 4 | 1 | 1 |
| 38 | 2 | 4 | 4 | 4 |
| 39 | 2 | 4 | 4 | 4 |
| 40 | 2 | 2 | 4 | 3 |
| 41 | 2 | 4 | 4 | 4 |
| 42 | 2 | 3 | 4 | 4 |
| 43 | 2 | 4 | 4 | 3 |
| 44 | 2 | 4 | 4 | 3 |
| 45 | 2 | 4 | 4 | 4 |
| 46 | 3 | 3 | 2 | 2 |
| 47 | 4 | 4 | 4 | 4 |
| 48 | 1 | 4 | 1 | 1 |
| 49 | 2 | 3 | 4 | 4 |
| 50 | 3 | 4 | 2 | 3 |
| 51 | 2 | 4 | 2 | 3 |
| 52 | 2 | 5 | 4 | 3 |
| 53 | 1 | 4 | 4 | 4 |
| 54 | 2 | 4 | 2 | 3 |
| 55 | 2 | 2 | 1 | 1 |
| 56 | 1 | 4 | 4 | 3 |
| 57 | 4 | 3 | 4 | 5 |
| 58 | 4 | 4 | 4 | 5 |
| 59 | 2 | 3 | 4 | 4 |
| 60 | 3 | 4 | 4 | 4 |
| 61 | 3 | 3 | 2 | 2 |
| 62 | 1 | 2 | 4 | 3 |
| 63 | 2 | 4 | 4 | 3 |
| 64 | 3 | 4 | 2 | 3 |
| 65 | 2 | 2 | 4 | 4 |
| 66 | 2 | 2 | 4 | 4 |
| 67 | 4 | 4 | 4 | 5 |
| 68 | 4 | 4 | 4 | 4 |
| 69 | 3 | 3 | 4 | 1 |
| 70 | 2 | 2 | 4 | 5 |
| 71 | 4 | 4 | 2 | 3 |
| 72 | 4 | 5 | 4 | 5 |
| 73 | 5 | 5 | 4 | 4 |

| 74 | 2 | 4 | 2 | 3 |
| 75 | 4 | 5 | 4 | 3 |
| 76 | 2 | 4 | 4 | 4 |
| 77 | 2 | 5 | 4 | 5 |
| 78 | 4 | 5 | 4 | 4 |
| 79 | 1 | 1 | 4 | 2 |
| 80 | 2 | 4 | 4 | 2 |
| 81 | 2 | 4 | 2 | 2 |
| 82 | 2 | 4 | 1 | 1 |
| 83 | 2 | 4 | 2 | 3 |
| 84 | 2 | 4 | 4 | 3 |
| 85 | 2 | 4 | 4 | 3 |
| 86 | 2 | 4 | 4 | 4 |
| 87 | 2 | 4 | 4 | 3 |
| 88 | 2 | 4 | 4 | 4 |
| 89 | 2 | 4 | 4 | 4 |
| 90 | 2 | 5 | 2 | 2 |
| 91 | 2 | 5 | 2 | 3 |
| 92 | 2 | 4 | 4 | 2 |
| 93 | 2 | 4 | 2 | 2 |
| 94 | 2 | 4 | 4 | 3 |
| 95 | 2 | 4 | 4 | 4 |
| 96 | 2 | 4 | 4 | 3 |
| 97 | 2 | 4 | 2 | 2 |
| 98 | 2 | 4 | 4 | 4 |
| 99 | 2 | 5 | 4 | 4 |
| 100 | 5 | 5 | 4 | 4 |